\documentclass[11pt,oneside,a4paper]{article}
\usepackage[top=1 in, bottom=1 in, left=0.85 in, right=0.85 in]{geometry}
\usepackage{hyperref}
\usepackage{float}
\usepackage{amsmath,amssymb,graphicx,url}
\usepackage{amsthm}
\usepackage{thmtools,thm-restate,wrapfig,enumitem,mathabx}
\newtheorem{assumption}{Assumption}
\newtheorem{theorem}{Theorem}
\newtheorem{remark}{Remark}
\newtheorem{lemma}{Lemma}
\newtheorem{corollary}{Corollary}

\usepackage{color}
\newtheorem{claim}{Claim}

\usepackage[noend]{algorithmic}
\usepackage[noend,linesnumbered,ruled,vlined]{algorithm2e}
\usepackage{bbold}

\usepackage[utf8]{inputenc}
\usepackage[T1]{fontenc}
\usepackage{booktabs}
\usepackage{amsfonts}
\usepackage{nicefrac}
\usepackage{microtype}
\usepackage{xcolor}
\usepackage{subcaption}
\usepackage{multirow}
\usepackage{bbm}
\usepackage{mathtools}

\newcommand{\blue}[1]{\textcolor{blue}{#1}}

\DeclareMathOperator*{\argmax}{arg\,max}

\allowdisplaybreaks

\title{Learning While Scheduling in Multi-Server Systems\\with Unknown Statistics: MaxWeight with Discounted UCB}

\author{%
  Zixian Yang\\
  University of Michigan\\
  Ann Arbor, MI 48109, USA\\
  \texttt{zixian@umich.edu}\\
  \and
  R. Srikant\\
  University of Illinois at Urbana-Champaign\\
  Urbana, IL 61801, USA\\
  \texttt{rsrikant@illinois.edu} \\
  \and
  Lei Ying\\
  University of Michigan\\
  Ann Arbor, MI 48109, USA\\
  \texttt{leiying@umich.edu}\\
}
\date{}

\begin{document}

\maketitle

\begin{abstract}
Multi-server queueing systems are widely used models for job scheduling in machine learning, wireless networks, crowdsourcing, and healthcare systems.
This paper considers a multi-server system with multiple servers and multiple types of jobs, where different job types require different amounts of processing time at different servers. The goal is to schedule jobs on servers without knowing the statistics of the processing times. 
To fully utilize the processing power of the servers, it is known that one has to at least learn the service rates of different job types on different servers. Prior works on this topic decouple the learning and scheduling phases which leads to either excessive exploration or extremely large job delays. We propose a 
new algorithm, which combines the MaxWeight scheduling policy with discounted upper confidence bound (UCB), to simultaneously learn the statistics and schedule jobs to servers. 
We prove that under our algorithm the asymptotic average queue length is bounded by one divided by the traffic slackness, which is order-wise optimal.
We also obtain an exponentially decaying probability tail bound for any-time queue length.
These results hold for both stationary and nonstationary service rates.
Simulations confirm that the delay performance of our algorithm is several orders of magnitude better than previously proposed algorithms. 
\end{abstract}

\section{Introduction}
\label{sec:intro}

\begin{figure}[htb]
    \centering
    \includegraphics[width=0.3\linewidth]{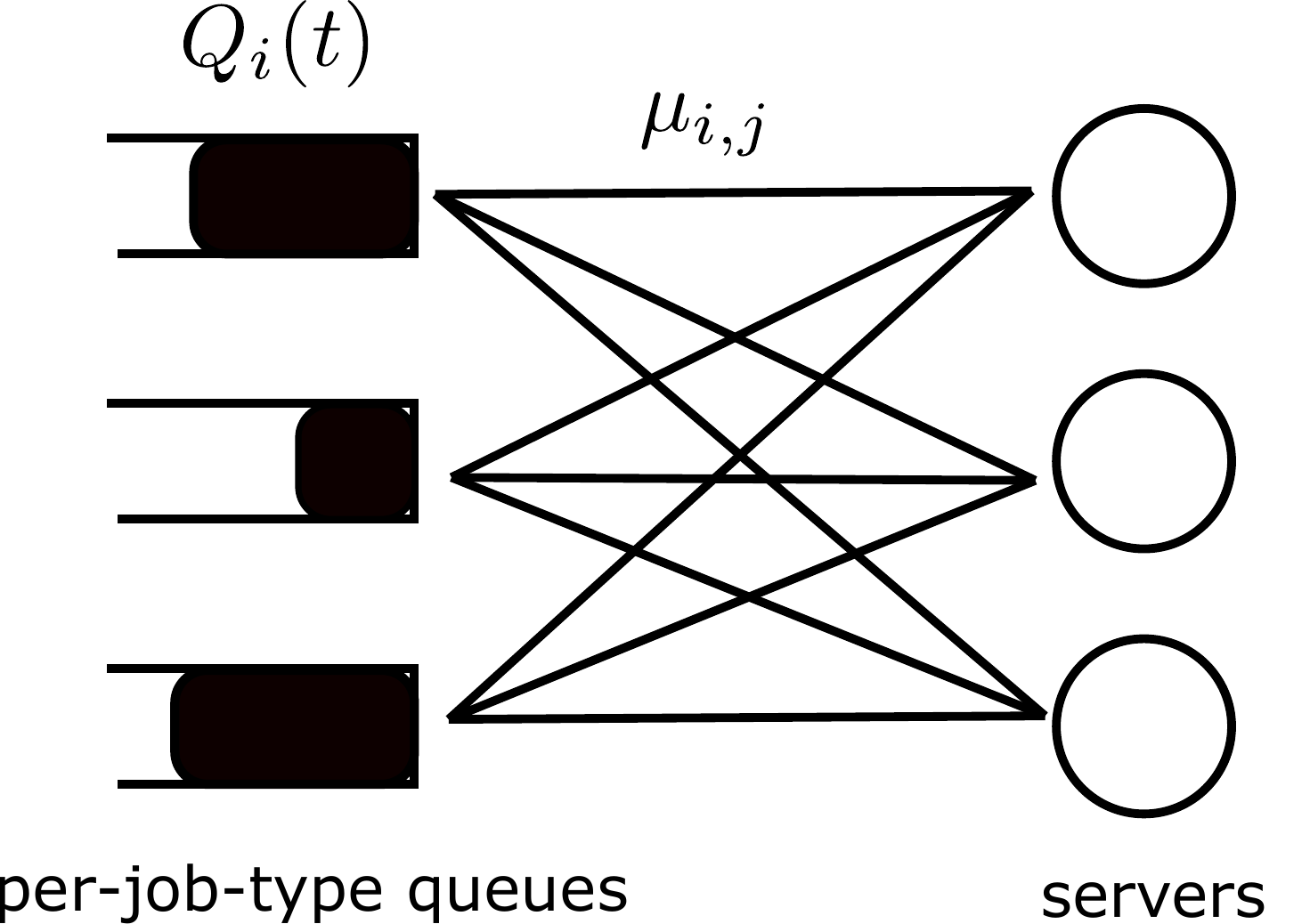}
    \caption{A Multi-Server System with Three Servers and Three Types of Jobs.}
    \label{fig:multi-server}
\end{figure}

A multi-server system is a system with multiple servers for serving jobs of different types as shown in Figure \ref{fig:multi-server}. An incoming job can be served by one of the servers and the service time depends on both the server and the job type. Multi-server systems have been used to model many real-world applications such as load balancing in a cloud-computing cluster, packet scheduling in multi-channel wireless networks, crowdsourcing, scheduling of doctors and patients in healthcare settings etc. In cloud-computing, a job may be a machine learning task and a server may be a virtual machine or a container, so the processing time of the machine learning task depends on the virtual machine's configuration. 
In crowdsourcing, jobs could be tagging of images and servers are workers, so the amount of the time a worker takes to tag the images depends on her familiarity of the images. Also, in healthcare systems, jobs could be patients and servers could be doctors and the time a doctor spends on a patient depends on both the patient's symptoms and doctor's experience and expertise.
In these cases, the scheduler may not know the statistics of processing times for a server before a sufficient number of jobs of the same type are processed at the server. Since scheduling decisions are based on learned processing times, and learning the processing times depends on the data samples collected through scheduling, poor scheduling and inadequate learning may reinforce each other, leading to instability (see the example in Appendix~\ref{app:sec:counter-example}). Therefore, it has been a problem of great interest to determine how to learn while scheduling in multi-server systems with unknown and potentially nonstationary environments.

In this paper, we consider a system as shown in Figure~\ref{fig:multi-server}, the incoming jobs form per-job-type queues on the left-hand side.
The scheduling problem in multi-server system is to schedule available servers to the per-job-type queues in order to maximize the throughput of the system and minimize the delay of jobs.
When the mean server times are known, the best known algorithm for scheduling in multi-server systems is the celebrated MaxWeight algorithm proposed by \cite{TasEph_93}. When a server is available, the MaxWeight algorithm picks the queue with the largest product of the queue length and the service rate.
Let $Q_i(t)$ denote the number of type-$i$ jobs waiting to be served and $1/\mu_{i,j}$ denote the mean service time of serving a type-$i$ job at server $j.$
When server $j$ is available, MaxWeight schedules a type $i^*_j$ job to server $j$ such that
$$i^*_j\in \arg\max Q_i(t)\mu_{i,j}.$$
A set of arrival rates is said to be supportable if there exists a scheduling algorithm such that, under this set of arrival rates, the queue lengths are bounded in an appropriate sense. The MaxWeight algorithm is provably throughput optimal \cite{TasEph_93}, i.e., it has the largest set of supportable arrival rates, also called the capacity region. Besides throughput optimality, MaxWeight has also near-optimal delay performance in various settings \cite{Sto_04,AndJunSto_07,ShaWis_07,KanWil_13,ErySri_12,MagSri_16}.

A key assumption behind the MaxWeight algorithm is that the scheduler knows the mean service rates $\mu_{i,j}$ for all $i$ and $j.$ This assumption is becoming increasingly problematic in emerging applications such as cloud computing and crowdsourcing due to either high variability of jobs (such as complex machine learning tasks) or servers (such as human experts in crowdsourcing). In these emerging applications, the mean service rates need to be learned while making scheduling decisions. Therefore, learning and scheduling are coupled and jointly determine the performance of the system because the scheduling decisions are based on estimated $\mu_{i,j}$ but the amount of samples the learner has for estimating $\mu_{i,j}$ depends on the number of times type-$i$ jobs are scheduled on server $j,$ i.e. depending on the scheduling decisions. 

A straightforward idea to learn the mean service rates is using the sample average, i.e., replacing $\mu_{i,j}$ with $1/\bar{s}_{i,j}$ where $\bar{s}_{i,j}$ is the empirical mean of the service time of type-$i$ jobs at server $j,$ based on the jobs completed at server $j$ so far. However, because of the coupling between learning and scheduling, this approach can be unstable. In this paper, we define stability as $\limsup_{t\rightarrow \infty}\frac{1}{t}\sum_{\tau=1}^{t} E[\sum_i Q_i(\tau)]<\infty$.
We provide a counter-example in Appendix~\ref{app:sec:counter-example}, which shows that such instability can occur. From the example, we observe that the problem of using empirical mean is that the initial bad samples led to a poor estimation of $\mu_{i,j},$ which led to poor scheduling decisions. They stop the scheduler from getting new samples from other queue-server pairs and therefore the system is ``locked in'' in a state with poor estimation and wrong scheduling decisions, which led to instability.

To overcome this problem, as in multi-armed bandit problems, we should encourage exploration: since the service rate of a server for a particular job can be estimated only by repeatedly scheduling jobs on all jobs, we should occasionally schedule jobs even on servers whose service rates are estimated to be small to overcome poor estimates due to randomness or nonstationary. For example, as in online learning, we can add an exploration bonus $b_{i,j},$ e.g., the upper confidence bound (UCB), to the empirical mean $\hat{\mu}_{i,j}.$
Indeed, there have been a sequence of recent studies that study job scheduling in multi-server systems as an online learning problem (multi-armed bandits or linear bandits) but a satisfactory solution has yet been developed. 
 We now review different categories of prior work and their limitations, and place our work in the context of the prior work:

\paragraph{\textbf{Queue-blind Algorithms:}} Queue blind algorithms do not take queue lengths into consideration at all when making scheduling decisions. In one line of work, the performance metric is the total reward received from serving jobs \cite{LiLiuJi_19,LiuLiShi_21}; however, for such algorithms, the queue lengths can potentially blow up to infinity asymptotically, which means that finite-time bounds for queue lengths can be excessively large and thus, such algorithms cannot be used in practice. Another line of work in the context of queue-blind scheduling algorithms addresses stability by assuming that the arrival rates of each type of job is known. They then use well-known scheduling algorithms such as $c\mu$-rule \cite{KriAraJoh_18} or weighted random routing \cite{ChoJosWan_21} or utility-based joint learning and scheduling \cite{HsuXuLin_22}. The drawback of such algorithms is that queue lengths can still be excessive large even if the queue lengths do not blow up to infinity asymptotically. The reason is the knowledge of queue lengths can encourage a phenomenon called \emph{resource pooling} which leads to greater efficiency. While we will not spend too much space explaining the concept of resource pooling, we hope that the following example clarifies the situation. Suppose you visit a grocery store and are not allowed to look at the queue lengths at each checkout lane before joining the checkout line. Then, some checkout lines can be excessively long, while others may even be totally empty. On the other hand, in practice, we look at the length of each checkout line and join the shortest one, which results in much better delay performance.

\paragraph{\textbf{Queue-Aware Algorithms:}} In early work on the problem \cite{NeeRagLa_12,KriAkhAra_18,KriSenJoh_21,yekkehkhany2020blind},
a fraction of time is allocated to probing the servers and the rest of the time is used to exploit this information. In the context of our problem, we would end up exploring all (job type, server) pairs the same number of times which is wasteful. On the other hand, exploration and exploitation are decoupled in such a forced exploration, which makes it easier to derive analytical derivation of performance bounds . If one uses  optimistic exploration such as UCB or related algorithms, the queue length information and the UCB-style estimation are coupled, which makes it difficult to analyze the system. Two approaches to decoupling UCB-style estimators have been studied prior to our paper: (a) In \cite{StaShrMod_19}, the algorithm proceeds in frames (a frame is a collection of contiguous time slots), where the queue length information is frozen at the beginning of each frame and UCB is used to estimate the service rates of the servers; additionally, UCB is reset at the end of each frame, and (b) In \cite{FreLykWen_22}, a schedule is fixed throughout each phase and thus, UCB is only executed for the jobs which are scheduled in that frame. The correlation between queues and UCB is more complicated here than in the algorithm of \cite{StaShrMod_19}, which requires more sophisticated analysis to conclude stability. 
Another challenge in the scheduling problem with unknown statistics is that the service rates may change over time and are nonstationary, so we need to carefully design algorithms.
The above two works \cite{StaShrMod_19,FreLykWen_22} do not have theoretical guarantees for the setting of nonstationary service rates.
In our paper, we propose an algorithm which does not explicitly decouple exploration and exploitation but continuously update the UCB bonuses and perform scheduling at each time instant, so the algorithm can quickly adapt to changes in stationary settings and is also able to adapt to nonstationary environments by using a discounted version of UCB~\cite{KocSze_06}.
On the other hand,
the fact that the schedule and discounted UCB are updated at every time step means that we require a new analysis of stability.
In particular, unlike prior work, our approach requires the use of concentration results for self-normalized means from \cite{GarMou_08}. In addition to differences in the algorithms and analysis, we also note other key differences between our paper and theirs \cite{StaShrMod_19,FreLykWen_22}:
The paper \cite{StaShrMod_19} considers scheduling in a general conflict graph, which includes our multi-server model as a special case. The paper \cite{FreLykWen_22} considers a general multi-agent setting that includes the centralized case as a special case. Both \cite{StaShrMod_19} and \cite{FreLykWen_22} assume the system is stationary but \cite{FreLykWen_22} allows dynamic arrivals and departures of queues while our paper studies a nonstationary, centralized setting that includes the stationary setting with a fixed set of queues as a special case. \cite{StaShrMod_19} and \cite{FreLykWen_22} consider Bernoulli services and Bernoulli arrivals while we consider general bounded arrivals and service times with nonpreemptive scheduling, i.e., once a job is scheduled, it cannot be stopped until completed.
In addition, in our model, multiple different servers are allowed to serve the same queue simultaneously, which is not allowed in the models of \cite{StaShrMod_19} and \cite{FreLykWen_22}.

This paper addresses the fundamental questions: how to learn and schedule without decoupling of the two and what is the fundamental impact of learning on queueing?  The main contributions of this paper are summarized below. 
\begin{itemize}[leftmargin=*]

\item {\textbf{Theoretical Results:}} We introduce the MaxWeight with discounted UCB algorithm. Discounted UCB was first proposed for nonstationary bandit problems \cite{KocSze_06}.
For our problem, with a revised discounted UCB, the discount factor allows us to handle the coupling between the queue lengths and the service rate estimators.
We establish the queue stability of MaxWeight with discounted UCB for nonstationary environments where the arrival rates and service rates may change over time. Given that the variation of service rates during the service time of a single job is bounded by $d~(d \le 1)$, we show that MaxWeight with discounted UCB can support any arrival rate vector $\boldsymbol{\lambda}$ such that $\boldsymbol{\lambda}+\delta \boldsymbol{1}$ is in the capacity region for some $\delta=\tilde{\Theta}(d)$, and the asymptotic time average of the expected queue length is bounded by $O(1/\delta_{\max})$, where $\delta_{\max}$ is the largest $\delta$ such that $\boldsymbol{\lambda}+\delta \boldsymbol{1}$ is in the capacity region.  This queue length bound holds for both stationary and nonstationary settings and improves the bound in~\cite{FreLykWen_22} significantly in the stationary setting, which is of order $O(1/\delta^3_{\max})$.
Note that the order $O(1/\delta_{\max})$ is {\em order-wise optimal} because even for Geo/Geo/1 queue with no need for learning or scheduling, the average queue length is $O(1/\delta_{\max})$ \cite{SriYin_14}. This result demonstrates that a carefully designed joint learning and scheduling algorithm can minimize the impact of learning and achieve the same order-wise queue length (in terms of $\delta_{\max}$) as in the case when the mean process times are known. This result is intuitive because, as learning continues, a well-designed learning algorithm should eventually be able to learn the mean processing times accurately. However, it is technically challenging due to the complex coupling between learning and scheduling, particularly in a nonstationary environment. Furthermore, we proved an upper bound on the moment generating function of the queue length, which implies an exponentially decaying probability tail in the distribution of any-time queue length. It is important to note that most existing results on learning and scheduling focus on the time-averaged queue length and do not provide any-time queue length bounds. In addition, we demonstrated that in the setting where service rates are time-invariant (stationary), the asymptotic average queue length under the MaxWeight with UCB algorithm (without discount) is bounded by $O(1/\delta_{\max})$ for arbitrarily small $\delta_{\max}$.

\item {\textbf{Methodology:}}
Our analysis is based on Lyapunov drift analysis. However, there are several difficulties due to joint scheduling and learning.
For the analysis of MaxWeight with discounted UCB, the estimated mean service time is the discounted sum of previous service times divided by the sum of the discount coefficients and the summation is taken over the time slots in which there is job completion, which themselves are random variables depending on the scheduling and learning algorithm.
To deal with this difficulty, we first transform the summation into a summation over the time slots in which a job starts, and then use a Hoeffding-type inequality for {\em self-normalized means} with a random number of summands~\cite[Theorem 18]{GarMou_08}\cite{GarMou_11} to obtain a concentration bound. Another difficulty is in bounding the discounted number of times server $j$ serves type-$i$ jobs.   Our method is to divide the interval into sub-intervals of carefully chosen lengths so that the discount coefficients can be lower bounded by a constant in each sub-interval. Moreover, for the analysis of the moment generating function of the queue length, our method is based on the idea in \cite{Haj_82}. However, the method in \cite{Haj_82} does not directly apply to our case because we need to carefully design the Lyapunov drift due to the coupling between the queue lengths and the service rate estimators. This different Lyapunov drift requires a more delicate analysis.
We believe these ideas may be useful for analyzing other joint learning and scheduling algorithms as well, in particular, providing a roadmap for tackling the coupling of learning and scheduling in the analysis. 

\begin{figure}[htb]
    \centering
    \includegraphics[width=0.45\linewidth]{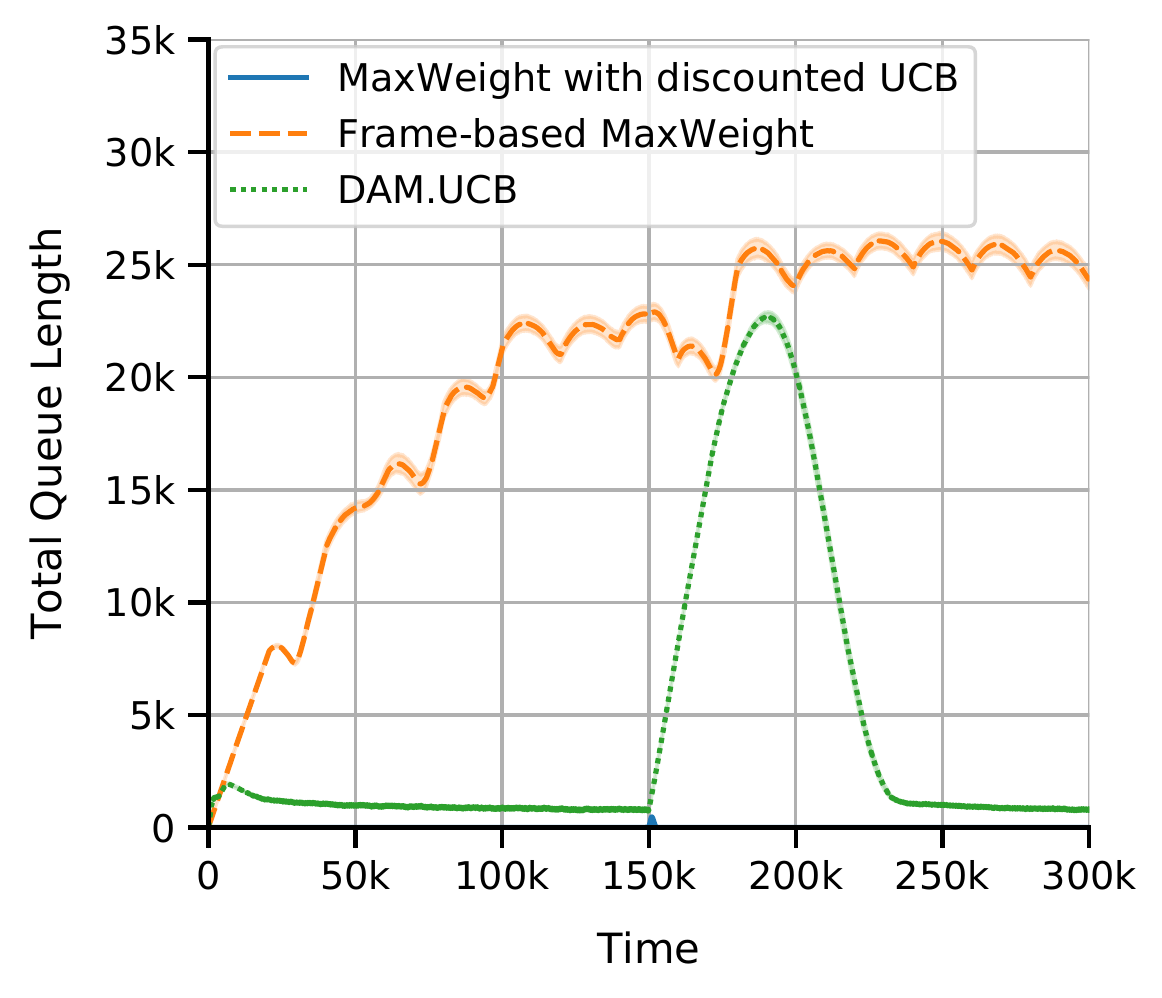}
    \caption{Comparison among MaxWeight with Discounted UCB (Proposed), Frame-Based MaxWeight~\cite{StaShrMod_19}, and DAM.UCB~\cite{FreLykWen_22} in the Nonstationary Setting where There is an Abrupt Change of Service Rates at Time Slot $150k$.}
    \label{fig:intro-simu-result}
\end{figure}

\item {\textbf{Numerical Studies:}} We compare the proposed algorithm with previously proposed algorithms in the literature. The results show that our algorithm achieves delays that are several orders of magnitude smaller than previously proposed algorithms. For example, in the setting where there is an abrupt change of service rates, our algorithm can quickly adapt to the change, as shown in Figure~\ref{fig:intro-simu-result}. 
Another noteworthy observation is that, although discounted UCB algorithm was originally designed for nonstationary environments, MaxWeight with discounted UCB still performs as well as MaxWeight with UCB in stationary environments.
\end{itemize}

\subsection{Extension Compared to the Proceedings Paper}
This paper is an extension of our proceedings paper~\cite{yang2023Learning}.
The new contributions made in this paper that extend the original proceedings paper are as follows:
\begin{itemize}[leftmargin=*]
    \item We have modified our original algorithm. The modifications include the calculation of the UCB of service rates, which gives us a better numerical performance. The design of the UCB bonus term unifies MaxWeight with discounted UCB algorithm and MaxWeight with UCB algorithm in one framework, where MaxWeight with UCB has a better theoretical guarantee in stationary settings.
    
    \item We have significantly improved the asymptotic average queue length bound from $O(1/\delta^3_{\max})$ to $O(1/\delta_{\max})$ under MaxWeight with discounted UCB algorithm. This improvement comes from a different proof, where we borrow the decoupling idea from \cite{FreLykWen_22} and combine it with the proof idea in our original proceedings paper.

    \item We have proved an upper bound on the moment generating function of the any-time queue length, which implies an exponentially decaying probability tail. Note that this bound considers queue length at any time while \cite{StaShrMod_19} and \cite{FreLykWen_22} only consider queue length averaged/summed over time. 

    \item We have proved that the asymptotic average queue length under MaxWeight with UCB algorithm is bounded by $O(1/\delta_{\max})$ in the setting where service rates are time-invariant, which holds for arbitrary small traffic slackness $\delta_{\max}$. The analysis extends the decoupling idea in \cite{FreLykWen_22} to our setting where multiple servers are allowed to serve the same queue simultaneously and the scheduling is nonpreemptive with general bounded service times.

    \item In the simulation, we consider more challenging settings where the service times follow truncated heavy-tailed Weibull distributions.
\end{itemize}

\section{Model}
\label{sec:model}
We consider a multi-server system with $J$ servers, indexed with $j\in\{1,2,\ldots,J\}$, and $I$ types of jobs, indexed with  $i\in\{1,2,\ldots,I\}$.  The system maintains a separate queue for each job type, as shown in Figure \ref{fig:multi-server}.

We consider a discrete-time system. The number of jobs that arrive at queue $i$ is denoted by $\left(A_i(t)\right)_{t\ge 0}$ where $t$ denotes the time slot. Assume that $\left(A_i(t)\right)_{t\ge 0}$ are independent with unknown mean $E[A_i(t)] = \lambda_i(t)$ and are bounded, i.e., $A_i(t)\le U_{\mathrm{A}}$ for all $i$ and $t$. We consider $U_{\mathrm{A}} \ge 1$ without loss of generality. Let $\boldsymbol{A}(t)\coloneqq(A_i(t))_{i=1}^{I}$ and $\boldsymbol{\lambda}(t)\coloneqq (\lambda_i(t))_{i=1}^{I}$.

We say a server is available in time slot $t$ if the server is not serving any job at the beginning of time slot $t$; otherwise, we say the server is busy.
At the beginning of each time slot, each available server picks a job from one of the queues. Note that each server can serve at most one job at a time and can start to serve another job only after finishing the current job, i.e., the job scheduling is nonpreemptive.
When a job from queue $i$ (job of type $i$) is picked by server $j$ in time slot $t$, it requires  ${S}_{i,j}(t)$ time slots to finish serving the job. For any $i,j$, $\left({S}_{i,j}(t)\right)_{t\ge 0}$ are independent random variables with unknown mean $E[{S}_{i,j}(t)] = \frac{1}{\mu_{i,j}(t)}$ and are bounded, i.e., ${S}_{i,j}(t) \le U_{\mathrm{S}}$ for all $i$, $j$ and $t$. $A_i(t)$ and ${S}_{i,j}(t)$ for different $i,j$ are also independent. Let $\boldsymbol{S}(t)\coloneqq(S_{i,j}(t))_{i=1, \ldots, I,j=1, \ldots, J}$ and $\boldsymbol{\mu}(t)\coloneqq(\mu_{i,j}(t))_{i=1, \ldots, I,j=1, \ldots, J}$.
Note that we allow $\boldsymbol{\lambda}(t)$ and $\boldsymbol{\mu}(t)$ to be time-varying to model nonstationary environments, and the value of $S_{i,j}(t)$ is generated at time slot $t$ and will not change after that.  

If server $j$ is available and picks queue $i$ in time slot $t$ or if server $j$ is busy serving queue $i$ in time slot $t$, we say server $j$ is scheduled to queue $i$ in time slot $t$. Let $I_j(t)$ denote the queue to which server $j$ is scheduled in time slot $t$. 
Define a waiting queue $\tilde{Q}_i(t)$ for each job type $i$. A job of type $i$ joins the waiting queue $\tilde{Q}_i(t)$ when it arrives, and leaves the waiting queue $\tilde{Q}_i(t)$ when it is picked by a server under the algorithm.
If an available server $j$ picks queue $i$ in time slot $t$ and there is no job in the waiting queue $i$, i.e., $\tilde{Q}_i(t) + A_i(t)=0$, we say server $j$ is idling in time slot $t$ and the server $j$ will be available in the next time slot.
Let $\eta_j(t)$ be an indicator function such that $\eta_j(t)=1$ if server $j$ is not idling in time slot $t$ and $\eta_j(t)=0$ otherwise.
Let $\mathbb{1}_{i,j}(t)$ be another indicator function such that $\mathbb{1}_{i,j}(t)=1$ if $I_j(t)=i$ and server $j$ finishes serving the job of type $i$ at the end of time slot $t$, or if $I_j(t)=i$ and server $j$ is idling. 

Let $Q_i(t)$ denote the actual queue length of jobs at queue $i$ at the beginning of time slot $t$ so $Q_i(t)$ is the total number of type-$i$ jobs in the system. Thus, $\tilde{Q}_i(t)$ is $Q_i(t)$ minus the number of type $i$ jobs that are in service. A job leaves the actual queue only when it is completed. Then we have the following queue dynamics:
\begin{align}\label{equ:queue-dynamics}
Q_i(t+1)=Q_i(t)+A_i(t)-\sum_j \mathbb{1}_{i,j}(t)\eta_j(t).
\end{align}
 
Our objective is to find an efficient learning and scheduling algorithm to stabilize $Q_i(t)$ for all $i$, i.e., preventing the queue lengths from going to infinity. In each time slot, the scheduling algorithm decides which queue to serve for each available server.

\section{Algorithm}
\label{sec:alg}
We propose MaxWeight with discounted UCB algorithm, which combines the MaxWeight scheduling algorithm~\cite{TasEph_92} with discounted UCB~\cite{KocSze_06} for learning the service statistics, as shown in Algorithm~\ref{alg:1} with $\gamma<1$.
\begin{algorithm}[ht]
\caption{MaxWeight with Discounted UCB / MaxWeight with UCB}\label{alg:1}
\begin{algorithmic}[1]
    \STATE {\textbf{Initialize:} $\mbox{Choose } \gamma\in(0,1]$; $\hat{N}_{i,j}(0)=0$, $\hat{\phi}_{i,j}(0)=0$, $M_{i,j}(0)=0$ for all $i,j$.}\label{alg:line:init}
    \STATE{If $t=0$, schedule each server to the queues uniformly at random.}
    \FOR{$t= 1$ to infinity}
            \FOR{$i=1,\ldots,I$ and $j=1,\ldots,J$}
                \IF{$I_j(t-1)=i$}
                \STATE {$M_{i,j}(t) = M_{i,j}(t-1) + 1$}
                \:\blue{// the number of time slots already served}
                \ENDIF
                \STATE{Update $\hat{N}_{i,j}(t)$ and $\hat{\phi}_{i,j}(t)$ according to~\eqref{equ:alg:update}.}
                \STATE{$\hat{\mu}_{i,j}(t) = \frac{\hat{N}_{i,j}(t)}{\hat{\phi}_{i,j}(t)}$} \:\blue{// estimate of the service rate}
                \label{alg:line:mu}
                \STATE {$b_{i,j}(t) = c_1 U_{\mathrm{S}}\sqrt{\frac{\log \left(\sum_{\tau=0}^{t-1} \gamma^{\tau} \right) }{\hat{N}_{i,j}(t)}}$} \label{alg:line:b}
                \:\blue{// UCB bonus term, where $c_1>0$ is a constant}
                \IF{$\mathbb{1}_{i,j}(t-1)=1$}
                \STATE {$M_{i,j}(t)=0$}
                \:\blue{// reset the counter if the server becomes available.}
                \ENDIF
            \ENDFOR
        \FOR{$j=1,\ldots,J$}
            \IF{server $j$ is available}
                \STATE{$\hat{i}^*_{j}(t) = \argmax_i \frac{Q_i(t)} {\max\left\{\frac{1}{\hat{\mu}_{i,j}(t)}-b_{i,j}(t), 1\right\}}$} \:\blue{// server $j$ picks $\hat{i}^*_{j}(t)$}\label{alg:line:max-weight}
            \ENDIF
        \ENDFOR
    \ENDFOR
\end{algorithmic}
\end{algorithm}

In Algorithm~\ref{alg:1}, we first fix the discount factor $\gamma$ beforehand and initialize the estimates $\hat{N}_{i,j}(0)$, $\hat{\phi}_{i,j}(0)$, and the counter $M_{i,j}(0)$, as shown in Line~\ref{alg:line:init}.
In the algorithm, $\hat{N}_{i,j}(t)$ is the discounted number of type-$i$ jobs served by server $j$ by time slot $t$ and $\hat{\phi}_{i,j}(t)$ is the discounted number of time slots used by server $j$ for serving type-$i$ jobs by time slot $t$.
If server $j$ is serving a type-$i$ job at time $t-1$, $M_{i,j}(t)$ is the service time the job has received by time slot $t$ (not including time slot $t$); otherwise, $M_{i,j}(t)=0.$
At time $t=0$, we schedule each server to the queues uniformly at random. If $t\ge 1$, we first update our estimates of service rates and the UCB bonuses and then do the scheduling using the MaxWeight algorithm with the true service rates replaced by the UCB. 
Specifically, at the beginning of each time slot $t$,
we update $\hat{N}_{i,j}(t)$ and $\hat{\phi}_{i,j}(t)$ as follows:
\begin{align}\label{equ:alg:update}
    \hat{N}_{i,j}(t) = & \gamma \hat{N}_{i,j}(t-1) + \gamma^{M_{i,j}(t-1)}\mathbb{1}_{i,j}(t-1)\eta_j(t-1)\nonumber\\
    \hat{\phi}_{i,j}(t) = & \gamma \hat{\phi}_{i,j}(t-1)
     + \gamma^{M_{i,j}(t-1)}\mathbb{1}_{i,j}(t-1)\eta_j(t-1)M_{i,j}(t).
\end{align}
That is, if the job has not yet finished or the server is idling, we simply multiply $\hat{N}_{i,j}(t-1)$ and $\hat{\phi}_{i,j}(t-1)$ by a discount factor $\gamma$; if the server is not idling and the job has finished, we update $\hat{N}_{i,j}(t-1)$ by multiplying $\gamma$ and adding a number $\gamma^{M_{i,j}(t-1)}$ and update $\hat{\phi}_{i,j}(t-1)$ by multiplying $\gamma$ and adding a discounted service time. The discount $\gamma^{M_{i,j}(t-1)}$ actually means that the service time is discounted starting from the time when the job starts. This update is slightly different from the discounted UCB in \cite{KocSze_06} and is needed for a technical reason. 
Then we obtain $\hat{\mu}_{i,j}(t)$, an estimate of the service rate, as shown in Line~\ref{alg:line:mu}, where we use the convention that $0/0=0$. For each available server, we pick the queue with the largest product of queue length and UCB of the service rate, as shown in Line~\ref{alg:line:max-weight}, where $\hat{i}^*_{j}(t)$ denotes the queue that server $j$ picks and ties are broken arbitrary. Note that $1/\max\{1/\hat{\mu}_{i,j}(t)-b_{i,j}(t), 1\}$ is the UCB of the service rate since $1/\hat{\mu}_{i,j}(t)=\hat{\phi}_{i,j}(t)/\hat{N}_{i,j}(t)$ is the estimate of the mean service time and $1/\hat{\mu}_{i,j}(t)-b_{i,j}(t)$ is the lower confidence bound (LCB) of the mean service time.

Note that Algorithm~\ref{alg:1} unifies the MaxWeight with discounted UCB algorithm and the MaxWeight with UCB algorithm in the same framework. When $\gamma<1$, Algorithm~\ref{alg:1} is MaxWeight with discounted UCB; when $\gamma=1$, Algorithm~\ref{alg:1} is Maxweight with UCB.

The use of discounted average instead of simple average reduces the influence of previous service times on the current estimate, and weakens the dependence between queue lengths and UCB. In nonstationary environments, it ensures  
that the estimation process can adapt to the nonstationary service rate since the discount factor reduces the influence of previous service times on the current estimate. UCB helps with the exploration of the service times for different servers and job types. The MaxWeight algorithm is known to be throughput optimal~\cite{SriYin_14}. These ideas are combined in the proposed MaxWeight with discounted UCB algorithm.

\section{Main Result}
\label{sec:result}

\subsection{MaxWeight with Discounted UCB}
\label{sec:result-d-ucb}

In this section, we will present our main result for the MaxWeight with discounted UCB algorithm with $\gamma<1$.
Define $g(\gamma)\coloneqq \frac{4}{1-\gamma}\log \frac{1}{1-\gamma}$.
We consider Algorithm~\ref{alg:1} with a sufficiently large $\gamma$ such that $\gamma \ge 1-\frac{1}{1-e^{1.5}}$ and $g(\gamma)\ge 8 U_{\mathrm{S}}$.
We make the following assumption on the time-varying mean service times and rates:
\begin{assumption}\label{assump:mu-1}
$\mu_{i,j}(t)$ satisfies the following two conditions:
\begin{enumerate}
    \item[(1)] For any $i, j$ and any $t_a, t_b$ such that $t_a\neq t_b$ and $\lvert t_a-t_b\rvert\le 2g(\gamma)$,
    \begin{align*}
    \left\lvert\frac{1}{\mu_{i,j}(t_a)}-\frac{1}{\mu_{i,j}(t_b)}\right\rvert \le \frac{1}{g(\gamma)}\left(\frac{1}{\gamma}\right)^{\left\lvert t_a-t_b\right\rvert-1};
    \end{align*}
    \item[(2)] There exists an absolute constant $p>0$ such that for any $i, j$ and any $t_a, t_b$ such that $\lvert t_a-t_b\rvert\le U_{\mathrm{S}}$,
    \begin{align*}
    \left\lvert\mu_{i,j}(t_a)-\mu_{i,j}(t_b)\right\rvert \le \frac{1}{\left[g(\gamma)\right]^p}.
    \end{align*}
\end{enumerate}
\end{assumption}
\begin{remark}\label{remark:1}
Note that in the first condition in Assumption~\ref{assump:mu-1}, $\frac{1}{\gamma}>1$, so the allowable change of the mean service time increases exponentially with respect to the time difference. Therefore, the second condition in Assumption~\ref{assump:mu-1} will be dominating for large $|t_a-t_b|$. Recall that $g(\gamma)\approx \frac{4}{1-\gamma},$ so the bound in condition (2) is roughly equivalent to that the maximum change that can occur when serving a job is 
$\frac{(1-\gamma)^p}{4^p}$ for some $p>0$ (note that  $U_{\mathrm{S}}$ is an upper bound on the service times). This bound increases as $\gamma$ decreases because the algorithm can quickly adapt by aggressively discounting the past samples. 
\end{remark}

For the nonstationary system considered in this paper, we introduce the following definition ${\mathcal C}(W)$ for the capacity region: 
\begin{align}\label{equ:def-capacity}
    {\mathcal C}(W) = & \biggl\{ (\boldsymbol{R}(t))_{t\ge 0}: \biggr. \mbox{ there exists } (\boldsymbol{\alpha}(t))_{t\ge 0} \mbox{ such that }\nonumber\\
    &\sum_{i}\alpha_{i,j}(t)\le 1 \mbox{ for all } j, t \mbox{ and for any } i, t, \mbox{ there exists } w(t) \mbox{ such that } 1\le w(t)\le W \mbox{ and } \nonumber\\
    &\biggl. \sum_{\tau=t}^{t+w(t)-1} R_i(\tau) \le \sum_{\tau=t}^{t+w(t)-1} \sum_{j} \alpha_{i,j}(\tau) \mu_{i,j}(\tau)  \biggr\},
\end{align}
where $\boldsymbol{\alpha}(t)\coloneqq(\alpha_{i,j}(t))_{i=1,\ldots,I,j=1,\ldots,J}$ and $W\ge 1$ is a constant.
$\boldsymbol{R}(t)$ can be interpreted as allocatable service rates for time $t.$
This capacity region means that for some $(\boldsymbol{R}(t))_{t\ge 0}$ in this region,
for any time $t$ and queue $i$, there exists a time window such that the sum of $R_i(t)$ over this time window is less than the sum of appropriately allocated service rates. If $(\boldsymbol{\alpha}(t))_{t\ge 0}$ is given, then a randomized scheduling algorithm using $(\boldsymbol{\alpha}(t))_{t\ge 0}$ guarantees that the service rate received by queue $i$ in a time window is at least as large as the sum of $R_i(t)$ in this time window. 
Note that ${\mathcal C}(W_1) \subseteq {\mathcal C}(W_2)$ if $W_1\le W_2$.
If $W=1$ and $\boldsymbol{R}(t)$ and $\boldsymbol{\mu}(t)$ are time-invariant, then this definition reduces to the capacity region definition for the stationary setting~\cite{SriYin_14}. Let $\boldsymbol{\lambda}\coloneqq (\boldsymbol{\lambda}(t))_{t\ge 0}$.
We assume that the arrival rates satisfy that ${\boldsymbol \lambda}+\delta {\boldsymbol 1}\in {\mathcal C}(W)$, where ${\boldsymbol 1}$ denotes an all-ones vector and we assume that $W\le \frac{g(\gamma)}{2}$. We present Theorem~\ref{theo:1} which shows that the MaxWeight with discounted UCB algorithm can stabilize the queues with such arrival rates.
Another interpretation is that our algorithm can stabilize any arrival rate that satisfies $\lambda_i(t)+\delta\le R_i(t)$ for all $i,t$ for some $(\boldsymbol{R}(t))_{t\ge 0}$ in the capacity region.

\begin{theorem}\label{theo:1}
Consider Algorithm~\ref{alg:1} with $c_1=2$, $1-\frac{1}{1+e^{1.5}} \le \gamma < 1$, and $g(\gamma) \ge 8U_{\mathrm{S}}$. Suppose $Q_i(0)=0$ for all $i$.
Under Assumption~\ref{assump:mu-1}, for arrival rates that satisfy ${\boldsymbol \lambda}+\delta {\boldsymbol 1}\in {\mathcal C}(W)$, where $W\le \frac{g(\gamma)}{2}$
and
\begin{align}\label{equ:condition-delta-1}
    \delta \ge 451 I J U^{2}_{\mathrm{S}} (1-\gamma)^{\min\{p, \frac{1}{3}\}} \log \frac{1}{1-\gamma},
\end{align}
we have
\begin{align}\label{equ:theo-1-finite-bound-2}
   \frac{1}{t}\sum_{\tau=1}^{t}E\left[\sum_i Q_i(\tau) \right] 
   \le \frac{I  U_{\mathrm{A}} g^2(\gamma) }{t} 
   + \left(1+\frac{W}{t}\right)\left(
    \frac{35322 I J^2 U^2_{\mathrm{S}} U^2_{\mathrm{A}} W}{\delta}
    +\frac{4I U^2_{\mathrm{A}} g^2(\gamma) }{\delta[t+1-g(\gamma)]}
    \right)
\end{align}
for all $t\ge g(\gamma)$, and thus
\begin{align}\label{equ:theo-1-asy-bound-2}
    \limsup_{t\rightarrow\infty}\frac{1}{t}\sum_{\tau=1}^{t} E\left[\sum_i Q_i(\tau)\right]\le
    \frac{35322 I J^2 U^2_{\mathrm{S}} U^2_{\mathrm{A}} W}{\delta}.
\end{align}
\end{theorem}
Proof of Theorem~\ref{theo:1} can be found in Appendix~\ref{app:sec:proof:theo:1}.

We will discuss Theorem~\ref{theo:1} in the stationary setting and the nonstationary setting in the following paragraphs. Note that the value of $\delta,$ the traffic slackness, measures the throughput loss, under MaxWeight with discounted UCB for given discount factor $\gamma.$

\paragraph{\textbf{Stationary Setting:}}
For the stationary setting, Theorem~\ref{theo:1} implies that if $\gamma$ is sufficiently close to 1, $\delta$ can be arbitrarily close to zero and hence the proposed algorithm can stabilize the queues with arrivals inside the capacity region, which means the throughput loss is close to zero. Given an arrival rate vector ${\boldsymbol \lambda}$ and letting $\delta_{\max}$ denote the largest $\delta$ such that ${\boldsymbol \lambda}+\delta {\boldsymbol 1}\in {\mathcal C}(W)$, Theorem~\ref{theo:1} implies that the asymptotic time average of expected queue length is bounded by $O(1/\delta_{\max})$, which is obtained by setting $\gamma \ge 1 - \tilde{\Theta}(\delta_{\max}^3)$ that satisfies the condition \eqref{equ:condition-delta-1}, where $p$ can be set to an arbitrary large value because Assumption~\ref{assump:mu-1} always holds in the stationary setting.

\paragraph{\textbf{Nonstationary Setting:}}
For the nonstationary setting, Assumption~\ref{assump:mu-1} comes into play because we need to consider the variation of service rates.
Suppose that the variation of service rates within the service time of a single job is bounded by $d$, $d\in(0,1]$. We want to obtain the smallest $\delta$ in Theorem~\ref{theo:1}, i.e., minimizing the throughput loss, while satisfying Assumption~\ref{assump:mu-1}.
We only consider the second condition in Assumption~\ref{assump:mu-1} since it is dominating as discussed in Remark~\ref{remark:1}. We consider the following two cases:
\begin{enumerate}[leftmargin=*]
    \item[(A)]  For any $p\ge 1/3$, we choose $g(\gamma)=1/d^{1/p}$. We can see that Assumption~\ref{assump:mu-1} (2) is satisfied with this $p$. Then by~\eqref{equ:condition-delta-1}, $\delta$ can be as small as $\delta=451 I J U^2_{\mathrm{S}} (1-\gamma)^\frac{1}{3} \log \frac{1}{1-\gamma}$. Note that $451 I J U^2_{\mathrm{S}} (1-\gamma)^\frac{1}{3} \log \frac{1}{1-\gamma}$ is decreasing in $\gamma$, $g(\gamma) = \frac{4}{1-\gamma}\log\frac{1}{1-\gamma}$ is increasing in $\gamma$, and $1/d^{1/p}$ is decreasing in $p$. Hence, we will choose $p=1/3$ in order to obtain the smallest $\delta$ since $p \ge 1/3$.
    Therefore, by setting $g(\gamma)=1/d^3$,
    $\delta$ can be as small as $\delta = \tilde{\Theta}(d)$.

    \item[(B)] For any $p<1/3$, we choose $g(\gamma)=1/d^{1/p}$. We can see that Assumption~\ref{assump:mu-1} (2) is satisfied. Then By~\eqref{equ:condition-delta-1} and the definition that $g(\gamma) = \frac{4}{1-\gamma}\log\frac{1}{1-\gamma}$, $\delta$ can be as small as $\delta = \tilde{\Theta}(d)$.
\end{enumerate}
Note that in each case although choosing $g(\gamma)<1/d^{1/p}$ also satisfies Assumption~\ref{assump:mu-1} (2), it will induce a larger throughput loss since the right-hand side of \eqref{equ:condition-delta-1} is decreasing in $\gamma$ and hence is also decreasing in $g(\gamma)$.
Combining these two cases, we conclude that the smallest possible $\delta$ in Theorem~\ref{theo:1} is of order $\tilde{\Theta}(d)$.
In other words, the throughput loss is almost linear in terms of the variation $d.$
Consider an arrival rate vector ${\boldsymbol \lambda}$ and let $\delta_{\max}$ denote the largest $\delta$ such that ${\boldsymbol \lambda}+\delta {\boldsymbol 1}\in {\mathcal C}(W)$. Suppose $\delta_{\max}$ is greater than the smallest possible $\delta$. Then $\delta_{\max}\ge \tilde{\Theta}(d)$. Hence, by setting
$g(\gamma)=\tilde{\Theta}(1/\delta_{\max}^3)$, Assumption~\ref{assump:mu-1} (2) is satisfied with $p=1/3$, and Theorem~\ref{theo:1} implies that the asymptotic time average of expected queue length is bounded by $O(1/\delta_{\max})$.
This order is optimal because even for Geo/Geo/1 queue with no need for learning or scheduling, the average queue length is $O(1/\delta_{\max})$ \cite{SriYin_14}.

In many networks of interest, the arrival rates of flows are controlled by an algorithm called the congestion control protocol~\cite{SriYin_14}. For congestion controlled flows, $\delta_{\max}$ is typically small; and for non-congestion controlled flows, called best-effort arrivals, $\delta_{\max}$ varies a lot.

We also want to point out that the assumption $W\le g(\gamma)/2$ in Theorem~\ref{theo:1} is reasonable.
In fact, $W$ captures the time-scale at which congestion controlled arrivals react to nonstationarity. 
Recall from Assumption~\ref{assump:mu-1} that $1/[g(\gamma)]^p$ can loosely quantify the amount of nonstationarity the proposed algorithm can handle.
Therefore, when the level of nonstationarity is high, the congestion controller needs to react faster, resulting in a small $W.$

Next, in order to derive an upper bound for the moment generating function of the queue length, we make an additional assumption on the time-varying mean service times as follows:
\begin{assumption}\label{assump:mu-2}
$\mu_{i,j}(t)$ satisfies the following condition. Let $c_2 \coloneqq 5(IU_{\mathrm{A}}+J)$. For any $i, j$ and any $t_a, t_b$ such that $t_a\neq t_b$ and $\lvert t_a-t_b\rvert\le \frac{(c_2+1) g(\gamma)}{\delta}$,
\begin{align*}
    \left\lvert\frac{1}{\mu_{i,j}(t_a)}-\frac{1}{\mu_{i,j}(t_b)}\right\rvert \le \frac{\delta}{(c_2+1)g(\gamma)}\left(\frac{1}{\gamma}\right)^{\left\lvert t_a-t_b\right\rvert-1}.
\end{align*}
\end{assumption}
Note that Assumption~\ref{assump:mu-2} has a similar form as Assumption~\ref{assump:mu-1} (1) but the condition is stronger. The allowable change of the mean service time increases exponentially with respect to the time difference. We present Theorem~\ref{theo:2} in the following which shows that the MaxWeight with discounted UCB algorithm also has theoretical guarantees on the any-time queue length.

\begin{theorem}\label{theo:2}
    Consider Algorithm~\ref{alg:1} with $c_1=2$, $1-\frac{1}{1+e^{1.5}} \le \gamma < 1$, and $g(\gamma) \ge 8U_{\mathrm{S}}$.
    Suppose $Q_i(0)=0$ for all $i$.
    Under Assumption~\ref{assump:mu-1} (2) and Assumption~\ref{assump:mu-2}, for arrival rates that satisfy ${\boldsymbol \lambda}+\delta {\boldsymbol 1}\in {\mathcal C}(W)$, where $W\le \frac{g(\gamma)}{2}$
    and 
    \begin{align}\label{equ:condition-delta-2}
        \delta \ge 153 I J U^2_{\mathrm{S}} U^{1/2}_{\mathrm{A}} W^{1/2} (1-\gamma)^{\min\{p, 1/2\}} \sqrt{\log \frac{1}{1-\gamma}},
    \end{align}
    we have
    \begin{align*}
        E\left[e^{\xi \|\boldsymbol{Q}(t)\|_2}\right]
        \le
        \frac{31 I J U_{\mathrm{A}}}{\delta},
    \end{align*}
    for all $\xi \le \frac{3\delta}{ g(\gamma) [5(I U_{\mathrm{A}} +J) +2] (I U_{\mathrm{A}} +J) \left[ 1 + \frac{15(I U_{\mathrm{A}} +J) +6}{\delta} \right] }=\Theta\left(\frac{\delta^2 }{I^3 J^3 U_{\mathrm{A}}^3 g(\gamma)}\right)$ and all $t$.
\end{theorem}
Proof of Theorem~\ref{theo:2} can be found in Appendix~\ref{app:sec:proof:theo:2}. Since $g(\gamma)=\tilde{\Theta}(\frac{1}{1-\gamma})$, the condition \eqref{equ:condition-delta-2} implies that $g(\gamma)\ge \tilde{\Theta}(\frac{1}{\delta^{\max\{\frac{1}{p},2\}}})$. Hence, the largest possible $\xi$ is $\tilde{\Theta}(\delta^{\max\{2+\frac{1}{p}, 4\}})$. 
Suppose that the variation of service rates within the service time of a single job is bounded by $d$. Following the same argument as that in the discussion of Theorem~\ref{theo:1}, we can show that the smallest $\delta$ in Theorem~\ref{theo:2} is also $\tilde{\Theta} (d)$, which can be obtained by setting $p=\frac{1}{2}$ and $g(\gamma) = \frac{1}{d^2}$. Hence, the largest possible $\xi$ is $\tilde{\Theta}(\delta_{\max}^4)$. 

A corollary to Theorem~\ref{theo:2} showing the probability tail bound for the any-time queue length is presented as follows:
\begin{corollary}\label{cor:to-theo-2-tail}
Let all the assumptions and conditions in Theorem~\ref{theo:2} hold. Let $x$ be any positive real number. Then we have
    \begin{align*}
        \Pr \left( \|\boldsymbol{Q}(t)\|_2 \ge x \right)
        \le \frac{31 I J U_{\mathrm{A}}}{\delta}\exp(-\xi x)
    \end{align*}
    for all $t$, where $\xi=\frac{3\delta}{ g(\gamma) [5(I U_{\mathrm{A}} +J) +2] (I U_{\mathrm{A}} +J) \left[ 1 + \frac{15(I U_{\mathrm{A}} +J) +6}{\delta} \right] }=\Theta\left(\frac{\delta^2 }{I^3 J^3 U_{\mathrm{A}}^3 g(\gamma)}\right)$.
\end{corollary}
Proof of Corollary~\ref{cor:to-theo-2-tail} can be found in Appendix~\ref{app:sec:proof:cor:to-theo-2-tail}. Corollary~\ref{cor:to-theo-2-tail} implies that the probability distribution of any-time queue length has an exponentially decaying tail.
From the discussion of Theorem~\ref{theo:2}, we know that the largest possible $\xi$ is $\tilde{\Theta}(\delta_{\max}^4)$. Note that $\xi$ is decreasing in $\gamma$, which is reasonable because if the discount factor increases, the algorithm adapts to the changing environment more slowly, resulting in a heavier tail on the queue length.

\subsection{MaxWeight with UCB for Time-Invariant Service Rates}
\label{sec:result-ucb}

In this section, we will present results for the MaxWeight with UCB algorithm without discount factor, i.e., Algorithm~\ref{alg:1} with $\gamma=1$. We want to derive stability guarantee of this algorithm for systems with time-invariant service rates, i.e., $\mu_{i,j}(t)=\mu_{i,j}(t')$ for any $t$ and $t'$. In this setting we will drop the time index of the service rate, denoted by $\mu_{i,j}$.

Theorem~\ref{theo:1} shows that a system with time-invariant service rates is stable if we use $\gamma<1$. However, Theorem~\ref{theo:1} does not apply if $\gamma=1$. Moreover, if we fix a $\gamma<1$, it is not clear whether the stability still holds for arbitrarily small traffic slackness $\delta$ because Theorem~\ref{theo:1} applies only when the condition \eqref{equ:condition-delta-1} holds. For MaxWeight with UCB algorithm ($\gamma=1$), we will show that it is stable under arbitrarily small $\delta$, i.e., under heavy-traffic regime.

Before we present the result, similar to ${\cal C}(W)$ in \eqref{equ:def-capacity}, we define the capacity region under the setting of time-invariant service rates, which is shown as follows:
\begin{align}\label{equ:def-capacity-stationary}
    {\mathcal C}'(W) = & \biggl\{ (\boldsymbol{R}(t))_{t\ge 0}: \biggr. \mbox{ there exists } (\boldsymbol{\alpha}'(t))_{t\ge 0} \mbox{ such that }\nonumber\\
    &\sum_{i}\alpha'_{i,j}(t)\le 1 \mbox{ for all } j, t \mbox{ and for all } i, t, 
    \mbox{ there exists } w(t) \mbox{ such that } 1\le w(t)\le W \mbox{ and } \nonumber\\
    &\biggl. \frac{1}{w(t)}\sum_{\tau=t}^{t+w(t)-1} R_i(\tau) \le \sum_{j} \alpha'_{i,j}(t) \mu_{i,j} \biggr\},
\end{align}
where $\boldsymbol{\alpha}'(t)\coloneqq(\alpha'_{i,j}(t))_{i=1,\ldots,I,j=1,\ldots,J}$ and $W\ge 1$ is a constant.
This capacity region means that for some $(\boldsymbol{R}(t))_{t\ge 0}$ in this region,
for any time $t$ and queue $i$, there exists a time window such that the average of $R_i(t)$ over this time window is less than appropriately allocated service rates.
This capacity region ${\cal C}'(W)$ actually includes the capacity region ${\cal C}(W)$ defined in \eqref{equ:def-capacity} if we consider the setting of time-invariant service rates. This can be easily verified by setting $\alpha'_{i,j}(t)=\frac{1}{w(t)}\sum_{\tau=t}^{t+w(t)-1} \alpha_{i,j}(\tau)$. This capacity region is also a more general case compared to the capacity region definition for the stationary setting~\cite{SriYin_14}~\cite{FreLykWen_22}, where both arrival rates and service rates are time-invariant.
We assume that the arrival rates satisfy that ${\boldsymbol \lambda}+\delta {\boldsymbol 1}\in {\mathcal C}'(W)$. Theorem~\ref{theo:3} shows that MaxWeight with UCB algorithm can stabilize the queues with such arrivals.

\begin{theorem}\label{theo:3}
    Consider Algorithm~\ref{alg:1} with $\gamma=1$ and $c_1=2$. Suppose $Q_i(0)=0$ for all $i$. Assume that the service rates are time-invariant. Then for arrival rates that satisfy ${\boldsymbol \lambda}+\delta {\boldsymbol 1}\in {\mathcal C}'(W)$, we have
    \begin{align*}
        \frac{1}{t}\sum_{\tau=1}^{t}E\left[\sum_i Q_i(\tau) \right]
        \le \left(1 + \frac{W}{t}\right) \left(\frac{903264 I J^4 U^6_{\mathrm{S}} U_{\mathrm{A}} \log^2 (t+1)}{\delta^4 (t+1)}
        + \frac{34 I J^2 U^2_{\mathrm{S}} U^2_{\mathrm{A}} W }{\delta}
        + \frac{604 I^2 J^2 U^2_{\mathrm{S}} U_{\mathrm{A}}}{\delta^2 (t+1)}\right)
    \end{align*}
    for all $t$, and thus
    \begin{align*}
        \limsup_{t\rightarrow \infty} \frac{1}{t} \sum_{\tau=1}^{t} E\left[ \sum_i Q_i(\tau) \right]
        \le \frac{34 I J^2 U^2_{\mathrm{S}} U^2_{\mathrm{A}} W }{\delta}.
    \end{align*}
\end{theorem}
Proof of Theorem~\ref{theo:3} can be found in Appendix~\ref{app:sec:proof:theo:3}. Theorem~\ref{theo:3} implies that in the setting where service rates are time-invariant, MaxWeight with UCB is stable under arbitrarily small traffic slackness and the asymptotic time average of expected queue length is bounded by $O(1/\delta_{\max})$.

\section{Proof Roadmaps}
\label{sec:proof-map}

In this section, we will present the proof ideas and roadmaps of Theorem~\ref{theo:1}, \ref{theo:2} under MaxWeight with discounted UCB algorithm, and Theorem~\ref{theo:3} under MaxWeight with UCB algorithm. The complete proof of these theorems and the proofs of all the lemmas can be found in the appendices.

\subsection{MaxWeight with Discounted UCB: Theorem~\ref{theo:1}}

In this subsection, we will present the proof ideas and roadmaps of Theorem~\ref{theo:1} under MaxWeight with discounted UCB algorithm.
Our proof of Theorem~\ref{theo:1} is based on Lyapunov drift analysis. Consider the Lyapunov function $L(t)\coloneqq \sum_i Q_i^2(t)$. 

\subsubsection{Decomposing the Lyapunov Drift}
\label{sec:proof-roadmap-decomp-drift}

First, we will divide the time horizon into intervals and later we can analyze the Lyapunov drift in each interval. Let $D_k$ denote the length of the $k^{\mathrm{th}}$ interval. The details of how we construct $D_k$ can be found in Appendix~\ref{app:sec:proof:theo:1}. 
The main idea is that we want to make sure that $D_k$ is approximately $g(\gamma)$ so that the estimates of the mean service times in the current interval will ``forget'' the old samples in previous intervals due to the discount factor $\gamma$.
Define $t_0=0$ and $t_k=t_{k-1}+D_{k-1}$ for $k\ge 1$. Then $[t_k,t_{k+1}]$ is the $k^{\mathrm{th}}$ interval.

Next, we analyze the Lyapunov drift in the $(k+1)^{\mathrm{th}}$ interval given the queue length $\boldsymbol{Q}(t_k)$ and $\boldsymbol{H}(t_k)$ at the beginning of the $k^{\mathrm{th}}$ interval, where $\boldsymbol{Q}(t)\coloneqq(Q_i(t))_{i=1, \ldots, I}$ and $\boldsymbol{H}(t)$ is defined as:
\begin{align*}
    \boldsymbol{H}(t)\coloneqq\left(\boldsymbol{\tilde{Q}}(t), \boldsymbol{M}(t), \hat{\boldsymbol{N}}(t), \hat{\boldsymbol{\phi}}(t)\right),
\end{align*}
where $\boldsymbol{\tilde{Q}}(t)\coloneqq(\tilde{Q}_i(t))_{i}$, $\boldsymbol{M}(t)\coloneqq(M_{i,j}(t))_{i,j}$, $\hat{\boldsymbol{N}}(t)\coloneqq(\hat{N}_{i,j}(t))_{i,j}$, and
$\hat{\boldsymbol{\phi}}(t)\coloneqq(\hat{\phi}_{i,j}(t))_{i,j}$. Utilizing the queue dynamics~\eqref{equ:queue-dynamics}, we can bound the Lyapunov drift by
\begin{align}
    & E\left[ L(t_{k+1}+D_{k+1}) - L(t_{k+1}) \left| \boldsymbol{Q}(t_k)=\boldsymbol{q}, \boldsymbol{H}(t_k)=\boldsymbol{h}\right.\right]\nonumber\\
    &\le \sum_{\tau=D_k}^{D_k+D_{k+1}-1} \hat{E}_{t_k}\left[ \sum_i  2 Q_i(t_k+\tau) A_i(t_k+\tau) \right] \label{equ:arrival-term-roadmap}\\
    & - \sum_{\tau=D_k}^{D_k+D_{k+1}-1} \hat{E}_{t_k}\left[ \sum_i  2 Q_i(t_k+\tau) \sum_j \mathbb{1}_{i,j}(t_k+\tau) \right] \label{equ:service-term-roadmap}\\
    & + O(g(\gamma)) \nonumber,
\end{align}
where $\hat{E}_{t_k}$ is a shorthand for expectation conditioned on $\{\boldsymbol{Q}(t_k)=\boldsymbol{q}, \boldsymbol{H}(t_k)=\boldsymbol{h}\}$. In order to obtain a negative Lyapunov drift, we analyze the above two terms, the arrival term \eqref{equ:arrival-term-roadmap} and the service term \eqref{equ:service-term-roadmap}. By writing the summation \eqref{equ:arrival-term-roadmap} in the form of the time windows defined in ${\cal C}(W)$ and using the inequality in~\eqref{equ:def-capacity} in each window, the arrival term~\eqref{equ:arrival-term-roadmap} can be upper bounded by
\begin{align}
    \eqref{equ:arrival-term-roadmap} \le & \hat{E}_{t_k} \left[2\sum_{j}\sum_{\tau=D_k}^{D_k+D_{k+1}-1} 
    \max_i Q_i(f_j(t_k+\tau))
    \mu_{i,j}(t_k+\tau) \right]  \label{equ:cancel-out-term-roadmap}\\
    & + O(g(\gamma)) - 2 \delta  \sum_{\tau=D_k}^{D_k+D_{k+1}-1} \hat{E}_{t_k} \left[ \sum_i   Q_i(t_k+\tau) \right], \label{equ:negative-drift-roadmap}
\end{align}
where $f_j(t)$ denotes the starting time of the job that is being served at server $j$ in time slot $t$.
We hope that the term~\eqref{equ:cancel-out-term-roadmap} can be later canceled out by the bound of the service term~\eqref{equ:service-term-roadmap}. Next, we analyze the service term~\eqref{equ:service-term-roadmap}.

\subsubsection{Bounding the Service Term}
\label{sec:bound-service}
Notice that the service term~\eqref{equ:service-term-roadmap} is a sum over all servers $j$. Let us first fix one $j$ and analyze the per-server service term:
\begin{align}\label{equ:service-term-roadmap-j}
    \sum_{\tau=D_k}^{D_k+D_{k+1}-1} \hat{E}_{t_k}\left[ \sum_i Q_i(t_k+\tau) \mathbb{1}_{i,j}(t_k+\tau) \right].
\end{align}
Bounding the per-server service term~\eqref{equ:service-term-roadmap-j} takes several steps. 
\paragraph{\textbf{Concentration of Service Times:}}
The first step is to prove a concentration result regarding the deviation of the estimates of the mean service times $1/\hat{\mu}_{i,j}(t)$ from the true mean service times $1/\mu_{i,j}(t)$. 
Consider a concentration event as follows:
\begin{align}\label{equ:concentration-event}
    {\mathcal E}_{t_k,j}& \coloneqq \biggl\{\mbox{for all }
    i,\tau\in\biggl[D_k-\frac{g(\gamma)}{8},D_k+D_{k+1}-1\biggr],
    \left|1/\hat{\mu}_{i,j}(t_k+\tau)-1/\mu_{i,j}(t_k+\tau)\right|\le b_{i,j}(t_k+\tau)\biggr\}.
\end{align}
\begin{lemma}\label{lemma:concentration}
Let Assumption~\ref{assump:mu-1} (1) holds. Suppose $c_1=2$, $1-\frac{1}{1+e^{1.5}} \le \gamma < 1$, and $g(\gamma) \ge 8U_{\mathrm{S}}$. For any $k\ge 0$, and any $j$, 
\[
\Pr ({\mathcal E}_{t_k,j}^{\mathrm c} \left| \mathbf{Q}(t_k)={\mathbf q}, \mathbf{H}(t_k)={\mathbf h}  \right. ) \le 186I(1-\gamma)^{1.5}.
\]
\end{lemma}
Lemma~\ref{lemma:concentration} shows that the deviation of the estimated mean service time from the true mean service time is bounded by the UCB bonus with high probability conditioned on the queue length $\boldsymbol{Q}(t_k)$ and $\boldsymbol{H}(t_k)$.
Proving Lemma~\ref{lemma:concentration} is the most challenging part of our proof of Theorem~\ref{theo:1}. Lemma~\ref{lemma:concentration} cannot be proved by simply using the Hoeffding inequality and the union bound like in the traditional analysis of UCB algorithms. There are three main difficulties.
First, the probability is conditioned on the queue length in the previous interval, which is related to the service times before the previous interval.
Thanks to the relation between the discount factor $\gamma$ and the length $g(\gamma)$ of each interval,
the contribution of the service times before the previous interval to the current estimate $\hat{\phi}_{i,j}(t)$ is negligible and can be bounded.
Another difficulty is that $\hat{\phi}_{i,j}(t)$ is the discounted sum of previous service times and the summation is taken over the time slots in which there is job completion. Those time slots are random variables, which implies that the discount coefficients of those service times are also random.
Also, $\hat{N}_{i,j}(t)$ is the sum of some discount coefficients, which is a random variable that takes values in the real line while in the standard MAB problem this is just a random integer.
Therefore, taking union bound over $\hat{N}_{i,j}(t)$ like in the standard MAB analysis does not work in our setting.
To deal with this difficulty, we first transform the summation into a summation over the time slots in which there is a job starting, and then use a Hoeffding-type inequality for self-normalized means with a random number of summands~\cite[Theorem 22]{GarMou_08}\cite{GarMou_11} to obtain a concentration bound.
Another issue is that the mean service times are time-varying and the estimate of the mean service time in the current time slot is based on the actual service times in previous time slots. We use the first condition in Assumption~\ref{assump:mu-1} to solve this time-varying issue.

Adding this high probability event ${\mathcal E}_{t_k,j}$ into \eqref{equ:service-term-roadmap-j} and multiplying and diving the same term $\mu_{I_j(t_k+\tau),j}(f_j(t_k+\tau))$, we obtain
\begin{align}\label{equ:service-term-roadmap-j-bound-2-variant}
    \eqref{equ:service-term-roadmap-j}
    \ge \hat{E}_{t_k}\Biggl[ \sum_{\tau=D_k}^{D_k+D_{k+1}-1}  Q_{I_j(t_k+\tau)}(t_k+\tau) \mu_{I_j(t_k+\tau),j}(f_j(t_k+\tau)) \frac{\mathbb{1}_{I_j(t_k+\tau),j}(t_k+\tau)}{\mu_{I_j(t_k+\tau),j}(f_j(t_k+\tau))} \mathbb{1}_{{\mathcal E}_{t_k,j}} \Biggr].
\end{align}

\paragraph{\textbf{Bounding the Product of Queue Length and Service Rate:}} Next, we want to bound the product of queue length and service rate, i.e., $Q_{I_j(t_k+\tau)}(t_k+\tau) \mu_{I_j(t_k+\tau),j}(f_j(t_k+\tau))$ in \eqref{equ:service-term-roadmap-j-bound-2-variant}.
Since the algorithm picks the largest product of queue length and UCB of the service rate, this term can be lower bounded by $\max_i Q_i(f_j(t_k+\tau)) \mu_{i,j}(f_j(t_k+\tau))$ minus some term containing the UCB bonuses. Substituting this lower bound back to \eqref{equ:service-term-roadmap-j-bound-2-variant}, we obtain
\begin{align}\label{equ:sum-ucb-roadmap}
    \eqref{equ:service-term-roadmap-j} \ge & \hat{E}_{t_k}\biggl[ \sum_{\tau=D_k}^{D_k+D_{k+1}-1}  \max_i Q_i(f_j(t_k+\tau)) \mu_{i,j}(f_j(t_k+\tau)) \frac{\mathbb{1}_{I_j(t_k+\tau),j}(t_k+\tau)}{\mu_{I_j(t_k+\tau),j}(f_j(t_k+\tau))} \mathbb{1}_{{\mathcal E}_{t_k,j}} \biggr]\nonumber\\
    & - U_{\mathrm{S}} S_{\mathrm{Q}{\text-}\mathrm{UCB}} - O(g(\gamma)),
\end{align}
where $S_{\mathrm{Q}{\text-}\mathrm{UCB}}$ is the sum of queue-length-weighted UCB bonuses defined by
\begin{align*}
    S_{\mathrm{Q}{\text-}\mathrm{UCB}} \coloneqq &  \hat{E}_{t_k}\Biggl[ \sum_{\tau=D_k}^{D_k+D_{k+1}-1} Q_{I_j(t_k+\tau)}(t_k+\tau) \min\{2b_{I_j(t_k+\tau),j}(f_j(t_k+\tau)), 1\} \mathbb{1}_{I_j(t_k+\tau),j}(t_k+\tau) \Biggr].
\end{align*}

\paragraph{\textbf{Bounding the Sum of Queue-Length-Weighted UCB Bonuses:}}
In order to bound the sum of queue-length-weighted UCB bonuses $S_{\mathrm{Q}{\text-}\mathrm{UCB}}$, we consider two cases. If $\min\{2b_{I_j(t_k+\tau),j}(f_j(t_k+\tau)), 1\} = O(\delta)$, then $S_{\mathrm{Q}{\text-}\mathrm{UCB}}$ is also small and is negligible compared to the negative term in \eqref{equ:negative-drift-roadmap}.
Otherwise,
there are two difficulties of bounding $S_{\mathrm{Q}{\text-}\mathrm{UCB}}$. First, we need to decouple the queue length and the the UCB bonus. We borrow the idea from the method of proving Lemma 5.4 and Lemma 5.5 in~\cite{FreLykWen_22}, where they spread the queue length over previous time slots. However, the difference between our analysis and that in~\cite{FreLykWen_22} is that we need to spread the queue length over the interval $[t_k+D_k, t_k+D_k+D_{k+1}-1]$ rather than the whole time horizon, where we need an additional condition on $g(\gamma)$. Another difficulty is that we need a lower bound for $\hat{N}_{i,j}(t)$ so that we can bound the UCB bonus. Our method is to divide the interval into approximately $(1-\gamma)g(\gamma)$ sub-intervals with each sub-interval containing approximately $\frac{1}{1-\gamma}$ samples so that the discount coefficients can be lower bounded by a constant in each sub-interval. Combining the above ideas, we can obtain that
\begin{align*}
    S_{\mathrm{Q}{\text-}\mathrm{UCB}} \le \frac{3\delta}{4J U_{\mathrm{S}}} \hat{E}_{t_k}\Biggl[ \sum_{\tau=D_k}^{D_k+D_{k+1}-1} \sum_i Q_{i}(t_k+\tau)
     \Biggr] + O \left(\frac{1}{\delta^3}\log^4 \frac{1}{1-\gamma}\right) + O(g(\gamma)).
\end{align*}
Substituting the above bound into \eqref{equ:sum-ucb-roadmap}, we have
\begin{align}
    \eqref{equ:service-term-roadmap-j}  \ge & \hat{E}_{t_k}\biggl[ \sum_{\tau=D_k}^{D_k+D_{k+1}-1}  \max_i Q_i(f_j(t_k+\tau)) \mu_{i,j}(f_j(t_k+\tau)) \frac{\mathbb{1}_{I_j(t_k+\tau),j}(t_k+\tau)}{\mu_{I_j(t_k+\tau),j}(f_j(t_k+\tau))} \mathbb{1}_{{\mathcal E}_{t_k,j}} \biggr]\label{equ:job-comp-ind-roadmap}\\
    - & \frac{3\delta}{4J} \hat{E}_{t_k}\Biggl[ \sum_{\tau=D_k}^{D_k+D_{k+1}-1} \sum_i Q_{i}(t_k+\tau)
     \Biggr] - O \left(\frac{1}{\delta^3}\log^4 \frac{1}{1-\gamma}\right) - O(g(\gamma)),\label{equ:sum-ucb-roadmap-result}
\end{align}
where the first two terms in \eqref{equ:sum-ucb-roadmap-result} are negligible compared to the negative term in \eqref{equ:negative-drift-roadmap} if $g(\gamma)$ is sufficiently large. 

\paragraph{\textbf{Bounding the Weighted Sum of Job Completion Indicators:}}
The next step is to bound the weighted sum of job completion indicators \eqref{equ:job-comp-ind-roadmap}. First, removing the indicator $\mathbb{1}_{{\mathcal E}_{t_k,j}}$ does not change the value too much because the event ${\mathcal E}_{t_k,j}$ holds with high probability by Lemma~\ref{lemma:concentration}. Note that $1/\mu_{I_j(t_k+\tau),j}(f_j(t_k+\tau))$ are mean service times. 
In the expectation, we can replace the mean service times with actual service times. Intuitively, the weighted sum of the actual service times is close to the sum of the weights over the time slots, i.e., $\sum_{\tau=D_k}^{D_k+D_{k+1}-1} \max_i Q_i(f_j(t_k+\tau)) \mu_{i,j}(t_k+\tau)$, if $\mu_{i,j}$ does not change too much within the duration of each service (Assumption~\ref{assump:mu-1} (2)). That is, $\eqref{equ:job-comp-ind-roadmap} \gtrapprox \hat{E}_{t_k} [\sum_{\tau=D_k}^{D_k+D_{k+1}-1} \max_i Q_i(f_j(t_k+\tau)) \mu_{i,j}(t_k+\tau) ]$, where ``$\gtrapprox$'' means that we drop some negligible terms.
Substituting the above bound into \eqref{equ:job-comp-ind-roadmap} and then summing over all servers $j$, we have the following bound for the service term:
\begin{align}\label{equ:service-term-final-roadmap}
    \eqref{equ:service-term-roadmap} \lessapprox & -2\sum_j\hat{E}_{t_k}\left[\sum_{\tau=D_k}^{D_k+D_{k+1}-1} \max_i Q_i(f_j(t_k+\tau)) \mu_{i,j}(t_k+\tau)\right] + O(g(\gamma)).
\end{align}
Substituting \eqref{equ:service-term-final-roadmap} into \eqref{equ:service-term-roadmap} and then substituting \eqref{equ:cancel-out-term-roadmap} and \eqref{equ:negative-drift-roadmap} into \eqref{equ:arrival-term-roadmap}, we have
\begin{align*}
    & E\left[ L(t_{k+1}+D_{k+1}) - L(t_{k+1}) \left| \boldsymbol{Q}(t_k)=\boldsymbol{q}, \boldsymbol{H}(t_k)=\boldsymbol{h}\right.\right]\nonumber\\
    \lessapprox & -\frac{\delta}{4} \hat{E}_{t_k} \left[ \sum_{\tau=D_k}^{D_k+D_{k+1}-1} \sum_i Q_i(t_k+\tau)\right] 
    + O(g(\gamma)).
\end{align*}
Finally, by doing a telescoping sum over all the intervals,
we obtain the result in Theorem~\ref{theo:1}.

\subsection{MaxWeight with Discounted UCB: Theorem~\ref{theo:2}}

In this subsection, we will present the proof ideas and roadmaps of Theorem~\ref{theo:2} under MaxWeight with Discounted UCB algorithm. Our proof of Theorem~\ref{theo:2} is based on a different Lyapunov function $L'(t)\coloneqq \sqrt{\sum_i Q_i^2(t)}=\|\boldsymbol{Q}(t)\|_2$ and the idea in \cite{Haj_82}.

Different from Theorem~\ref{theo:1}, we consider a different Lyapunov drift for any $t$:
\begin{align*}
    E\left[ L'(t+\frac{g(\gamma)}{2}+G_t) - L'(t+\frac{g(\gamma)}{2}) \left| \boldsymbol{Q}(t)=\boldsymbol{q}, \boldsymbol{H}(t)=\boldsymbol{h}\right.\right]
\end{align*}
where $G_t = \Theta(g(\gamma)/\delta)$. The exact definition of $G_t$ can be found in Appendix~\ref{app:sec:proof:theo:2}. The value of $G_t$ makes sure that the negative drift is small enough while $G_t$ is not too large. Following the idea of proving Theorem~\ref{theo:1}, we can obtain a negative Lyapunov drift as follows:
\begin{align}\label{equ:negative-drift-theo-2-roadmap}
    E\left[ L'(t+\frac{g(\gamma)}{2}+G_t) - L'(t+\frac{g(\gamma)}{2}) \left| \boldsymbol{Q}(t)=\boldsymbol{q}, \boldsymbol{H}(t)=\boldsymbol{h}\right.\right] \le - (2+\frac{1}{16}) (I U_{\mathrm{A}} + J) g(\gamma),
\end{align}
which holds for any $t$ when $\sum_i Q_i(t) = \sum_i q_i \ge \frac{8(I U_{\mathrm{A}} + J) J g(\gamma)}{\delta}$.

Define $t'$ such that $t' \approx t-\frac{g(\gamma)}{2}-G_{t'}$.
Let $\Delta_n(\tau)\coloneqq \|\boldsymbol{Q}(\tau+n)\|_2 - \|\boldsymbol{Q}(\tau)\|_2$.
Then
\begin{align}\label{equ:exp-term-roadmap}
    \exp (\xi \|\boldsymbol{Q}(t)\|_2) \approx \exp \left(\xi \|\boldsymbol{Q}(t')\|_2 \right) \exp \left(\xi \Delta_{\frac{g(\gamma)}{2}}(t') \right)\exp \left(\xi \Delta_{G_{t'}}(t'+ \frac{g(\gamma)}{2}) \right).
\end{align}
Using the fact that the total queue length can increase by at most $I U_{\mathrm{A}}$ or decrease by at most $J$ in one time slot, we can bound the middle term $\exp (\xi \Delta_{\frac{g(\gamma)}{2}}(t'))$ in \eqref{equ:exp-term-roadmap} by $\exp(\xi \frac{(I U_{\mathrm{A}} + J)g(\gamma)}{2})$. Hence, by setting a small enough $\xi$ and using Taylor expansion, we can obtain
\begin{align}\label{equ:taylor-roadmap}
    \exp (\xi \|\boldsymbol{Q}(t)\|_2) \le \exp \left(\xi \|\boldsymbol{Q}(t')\|_2 \right) \left(1 + \xi \left[\Delta_{G_{t'}}\left(t'+\frac{g(\gamma)}{2}\right)+ \left(\frac{3g(\gamma)}{2} + 1\right) (IU_{\mathrm{A}} + J)\right]\right).
\end{align}
Note that the term $\Delta_{G_{t'}}(t'+ \frac{g(\gamma)}{2})$ in \eqref{equ:taylor-roadmap} is exactly the Lyapunov drift $L'(t'+\frac{g(\gamma)}{2}+G_{t'}) - L'(t'+\frac{g(\gamma)}{2})$. Consider the case where $\sum_i Q_i(t') > \frac{8(I U_{\mathrm{A}} + J) J g(\gamma)}{\delta}$. Then we can use the negative Lyapunov drift \eqref{equ:negative-drift-theo-2-roadmap}. Combining the negative drift with \eqref{equ:taylor-roadmap} and dealing with the other case where $\sum_i Q_i(t') \le \frac{8(I U_{\mathrm{A}} + J) J g(\gamma)}{\delta}$ (details are omitted and can be found in Appendix~\ref{app:sec:proof:theo:2}), we can obtain
\begin{align}\label{equ:exp-bound-7-roadmap}
    E\left[ \exp\left(\xi \|\boldsymbol{Q}(t)\|_2\right) \right]    
    \le \left(
    1 - \frac{\xi g(\gamma) (I U_{\mathrm{A}} + J)}{2}
    \right)
     E \left[ \exp\left(\xi \|\boldsymbol{Q}(t')\|_2 \right) \right]
    + O\left(\frac{\xi g(\gamma)}{\delta}\right).
\end{align}
Let $\rho \coloneqq 1 - \frac{\xi g(\gamma) (I U_{\mathrm{A}} + J)}{2}$.
Then we can recursively applying \eqref{equ:exp-bound-7-roadmap} to obtain
\begin{align*}
    E\left[ \exp\left(\xi \|\boldsymbol{Q}(t)\|_2\right) \right] 
    \le O\left(\frac{\xi g(\gamma)}{(1-\rho)\delta}\right) = O\left(\frac{1}{\delta}\right).
\end{align*}

\subsection{MaxWeight with UCB: Theorem~\ref{theo:3}}
Our proof of Theorem~\ref{theo:3} under MaxWeight with UCB is based on the decoupling idea in \cite{FreLykWen_22}, which spreads the queue length over previous time slots. In our setting, multiple servers are allowed to serve the same queue simultaneously and the scheduling is nonpreemptive with general bounded service times, which is different compared with the setting in \cite{FreLykWen_22}. This requires additional effort in the proof.

Although the decoupling idea in \cite{FreLykWen_22} can be used to prove the stability for MaxWeight with UCB, 
without using our Lemma~\ref{lemma:concentration}, this approach is not able to prove stability for MaxWeight with discounted UCB algorithm (Theorem~\ref{theo:1}) because in the discounting case, the probability of error is always lower bounded by a constant no matter how long the horizon is due to the discount factor.

\section{Simulation Results}
\label{sec:simu}

In this section, we evaluate the proposed algorithms numerically through simulation. We compare the proposed \emph{MaxWeight with discounted UCB} and \emph{MaxWeight with UCB} with several baselines, including the \emph{frame-based MaxWeight} algorithm~\cite{StaShrMod_19} and \emph{DAM.UCB} algorithm~\cite{FreLykWen_22}.

\begin{figure}[ht]
    \centering
    \subfloat[Zoom Out View]{%
    \includegraphics[width=0.45\textwidth]{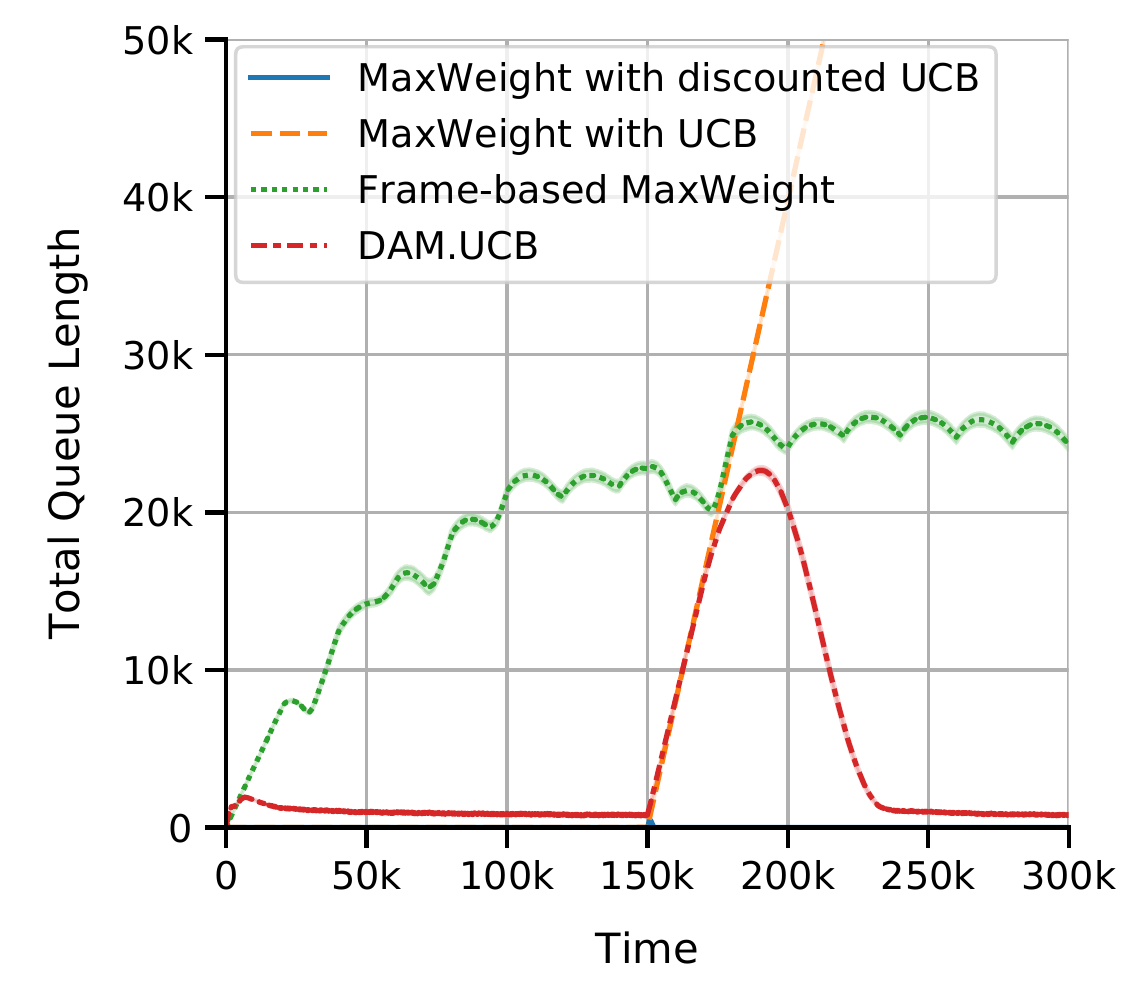}%
    }%
    \hspace{4ex}
    \subfloat[Zoom In View]{%
    \includegraphics[width=0.45\textwidth]{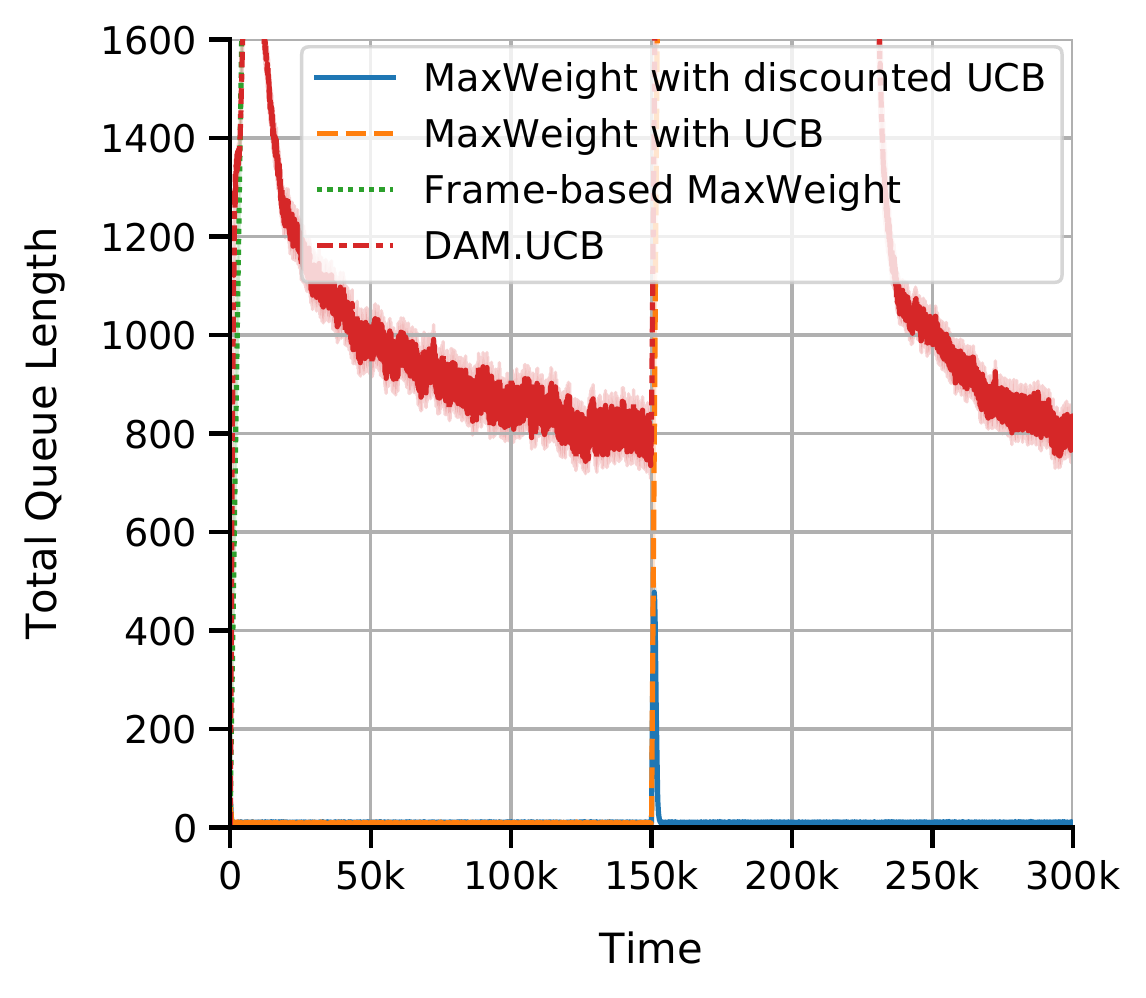}%
    }
    \caption{Nonstationary Service Rates.}
    \label{fig:simu-nonstationary-10-users}
\end{figure}

We consider a system with $10$ job types and $10$ servers.
The arrival $A_i(t)$ follows a Bernoulli distribution. The service time $S_{i,j}(t)$ takes value in $\{1, 2, \ldots, 100\}$ ($U_{\mathrm{S}}=100$) and follows a truncated heavy-tailed Weibull distribution with a time-invariant or time-varying mean.
We compare the algorithms in the stationary and nonstationary settings. In the stationary setting, the arrival rates and service rates are time-invariant. In the nonstationary setting, the service rates change to a completely different set of rates at time slot 150k and remain unchanged after that.
The simulation results are averaged over 100 runs. More details about the settings and parameters can be found in Appendix~\ref{app:simu}. The results are shown in Figure~\ref{fig:simu-stationary-10-users} and Figure~\ref{fig:simu-nonstationary-10-users}.

\begin{figure}[ht]
    \centering
    \subfloat[Zoom Out View]{%
    \includegraphics[width=0.45\textwidth]{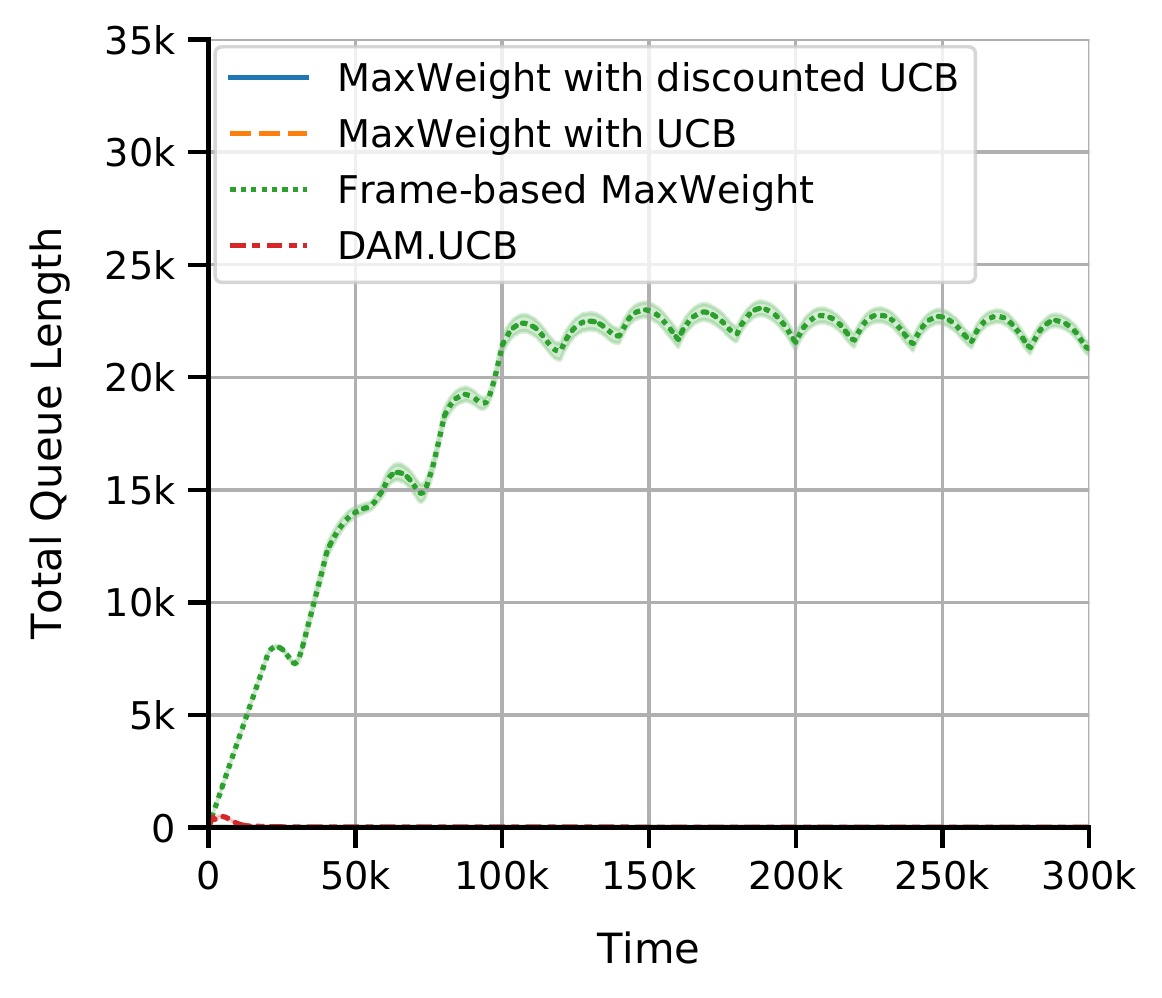}%
    }%
    \hspace{4ex}
    \subfloat[Zoom In View]{%
    \includegraphics[width=0.45\textwidth]{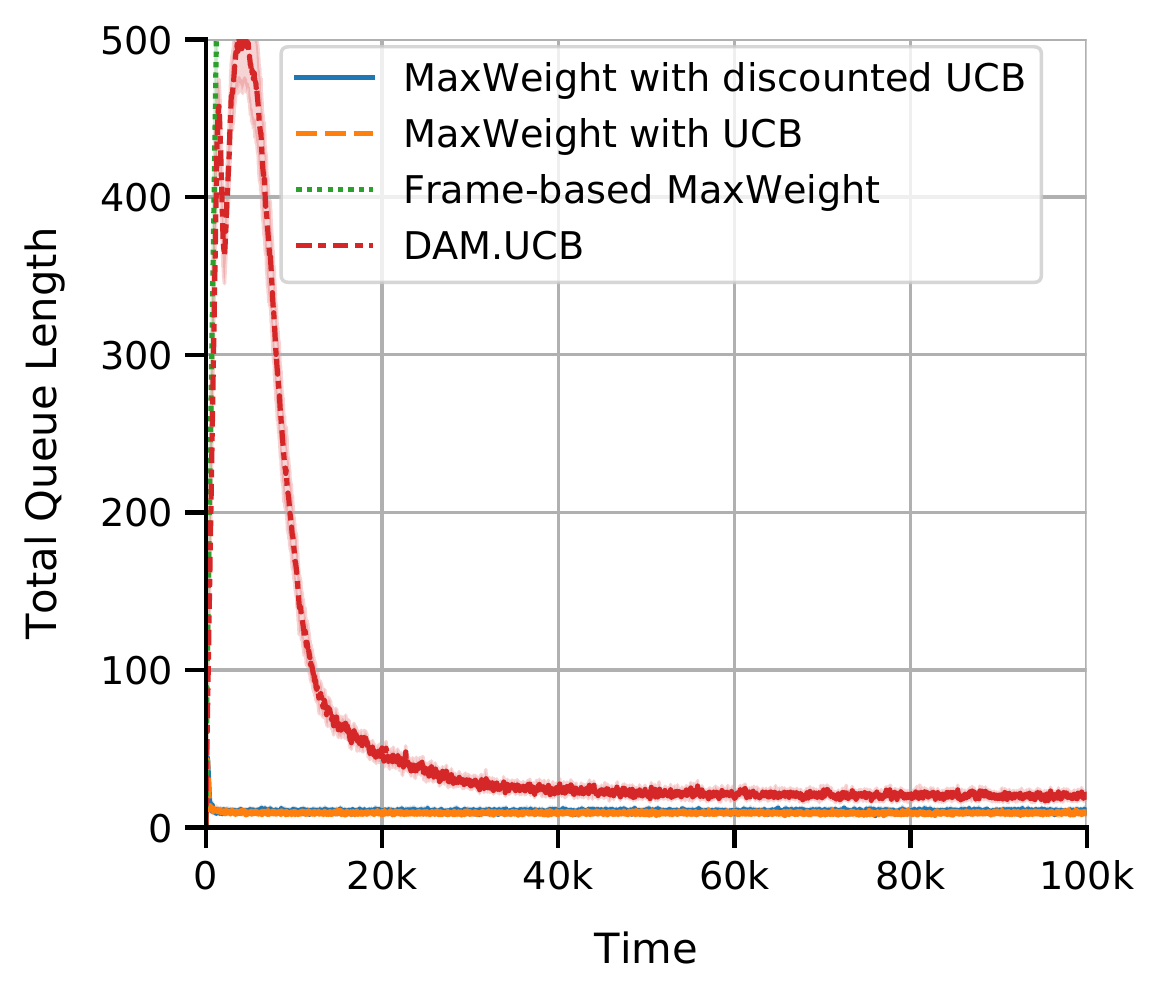}%
    \label{fig:simu-stationary-10-users-zoom-out}
    }
    \caption{Stationary Arrival Rates and Service Rates.}
    \label{fig:simu-stationary-10-users}
\end{figure}

Figure~\ref{fig:simu-stationary-10-users} shows the results in the stationary setting. As seen in the figure, all four algorithms are 
stable. However, the queue length of \emph{frame-based MaxWeight} is several orders of magnitude larger than the other three algorithms including our algorithms because \emph{frame-based MaxWeight} restarts the estimation and UCB of service rates at the beginning of every frame, which causes poor estimation. Another reason is that \emph{frame-based MaxWeight} uses the queue length at the beginning of each frame to make decisions, which leads to wrong decisions in the frame because the queue length information becomes outdated. The queue length of \emph{DAM.UCB} is worse than that of \emph{MaxWeight with discounted UCB} and \emph{MaxWeight with UCB} especially in the beginning. Note that in Figure~\ref{fig:simu-stationary-10-users} we set the size of each frame (called epoch in~\cite{FreLykWen_22}) of \emph{DAM.UCB} to be $1$. If a larger epoch size is used, the performance of \emph{DAM.UCB} becomes even worse, which can be found in Appendix~\ref{app:simu}, because \emph{DAM.UCB} uses the same schedule in each epoch, which causes wrong decisions due to outdated information. The difference of performance between \emph{DAM.UCB} with epoch size $1$ and our algorithms comes from different design of UCB.
We also notice that \emph{MaxWeight with discounted UCB} performs as well as \emph{MaxWeight with UCB} although discounted UCB was originally designed for nonstationary environments.

Figure~\ref{fig:simu-nonstationary-10-users} shows the results in the nonstationary setting. Both \emph{DAM.UCB} and \emph{MaxWeight with UCB} do not perform well because when the service rates change, these algorithms are still learning service rates using outdated samples.
Although \emph{frame-based MaxWeight} looks stable, its queue length is several orders of magnitude larger than the proposed \emph{MaxWeight with discounted UCB}. We believe that the key reason why our algorithm performs the best is that our algorithm can quickly adapt to the changing statistics thanks to the discount factor and the continuous update of both learning and scheduling decisions.

\section{Conclusions}
\label{sec:conc}

This paper considered scheduling in multi-server queueing systems with unknown arrival and service statistics, and proposed a new scheduling algorithm, MaxWeight with discounted UCB. Based on the Lyapunov drift analysis, concentration inequalities of self-normalized means, and the idea of spreading queue length over an interval, we proved that MaxWeight with discounted UCB guarantees queue stability (in the mean) and the asymptotic average queue length is bounded by $O(1/\delta_{\max})$ when the arrival rates are strictly within the service capacity region with traffic slackness $\delta_{\max}$. We also proved that the distribution of any-time queue length has an exponentially decaying probability tail. These results hold for both stationary systems and nonstationary systems.


\clearpage
\newpage
\bibliographystyle{plain}
\bibliography{inlab-refs,refs}

\begin{thebibliography}{10}

\bibitem{AndJunSto_07}
Matthew Andrews, Kyomin Jung, and Alexander Stolyar.
\newblock Stability of the max-weight routing and scheduling protocol in
  dynamic networks and at critical loads.
\newblock In {\em Proc. Ann. ACM Symp. Theory of Computing (STOC)}, pages
  145--154, 2007.

\bibitem{ChoJosWan_21}
Tuhinangshu Choudhury, Gauri Joshi, Weina Wang, and Sanjay Shakkottai.
\newblock Job dispatching policies for queueing systems with unknown service
  rates.
\newblock In {\em Proc. ACM Int. Symp. Mobile Ad Hoc Networking and Computing
  (MobiHoc)}, pages 181--190, 2021.

\bibitem{ChuLu_06}
Fan Chung and Linyuan Lu.
\newblock Concentration inequalities and martingale inequalities: A survey.
\newblock {\em Internet Mathematics}, 3:127 -- 79, 2006.

\bibitem{devroye1996probabilistic}
Luc Devroye, L{\'a}szl{\'o} Gy{\"o}rfi, and G{\'a}bor Lugosi.
\newblock {\em A probabilistic theory of pattern recognition}.
\newblock Springer-Verlag, 1996.

\bibitem{ErySri_12}
Atilla Eryilmaz and R.~Srikant.
\newblock Asymptotically tight steady-state queue length bounds implied by
  drift conditions.
\newblock {\em Queueing Syst.}, 72(3-4):311--359, December 2012.

\bibitem{FreLykWen_22}
Daniel Freund, Thodoris Lykouris, and Wentao Weng.
\newblock Efficient decentralized multi-agent learning in asymmetric queuing
  systems.
\newblock In {\em Proc. Conf. Learning Theory (COLT)}, volume 178, pages
  4080--4084, 02--05 Jul 2022.

\bibitem{GarMou_08}
Aur{\'e}lien Garivier and Eric Moulines.
\newblock On upper-confidence bound policies for non-stationary bandit
  problems.
\newblock {\em arXiv preprint arXiv:0805.3415}, 2008.

\bibitem{GarMou_11}
Aur{\'e}lien Garivier and Eric Moulines.
\newblock On upper-confidence bound policies for switching bandit problems.
\newblock In {\em {Int. Conf. Algorithmic Learning Theory (ALT)}}, pages
  174--188. Springer, 2011.

\bibitem{Haj_82}
B.~Hajek.
\newblock Hitting-time and occupation-time bounds implied by drift analysis
  with applications.
\newblock {\em Ann. Appl. Prob.}, pages 502--525, 1982.

\bibitem{HsuXuLin_22}
Wei-Kang Hsu, Jiaming Xu, Xiaojun Lin, and Mark~R Bell.
\newblock Integrated online learning and adaptive control in queueing systems
  with uncertain payoffs.
\newblock {\em Operations Research}, 70(2):1166--1181, 2022.

\bibitem{KanWil_13}
WN~Kang and RJ~Williams.
\newblock Diffusion approximation for an input-queued switch operating under a
  maximum weight matching policy.
\newblock {\em Stoch. Syst.}, 2(2):277--321, 2013.

\bibitem{KocSze_06}
Levente Kocsis and Csaba Szepesv{\'a}ri.
\newblock Discounted {UCB}.
\newblock In {\em {2nd PASCAL Challenges Workshop}}, volume~2, 2006.

\bibitem{KriAkhAra_18}
Subhashini Krishnasamy, P.~T. Akhil, Ari Arapostathis, Rajesh Sundaresan, and
  Sanjay Shakkottai.
\newblock Augmenting max-weight with explicit learning for wireless scheduling
  with switching costs.
\newblock {\em IEEE/ACM Trans. Netw.}, 26(6):2501--2514, 2018.

\bibitem{KriAraJoh_18}
Subhashini Krishnasamy, Ari Arapostathis, Ramesh Johari, and Sanjay Shakkottai.
\newblock On learning the c$\mu$ rule in single and parallel server networks.
\newblock In {\em Proc. Annu. Allerton Conf. Communication, Control and
  Computing}, pages 153--154. IEEE, 2018.

\bibitem{KriSenJoh_21}
Subhashini Krishnasamy, Rajat Sen, Ramesh Johari, and Sanjay Shakkottai.
\newblock Learning unknown service rates in queues: {A} multiarmed bandit
  approach.
\newblock {\em Operations Research}, 69(1):315--330, 2021.
\newblock The conference version appeared in NeurIPS 2016.

\bibitem{LiLiuJi_19}
F.~{Li}, J.~{Liu}, and B.~{Ji}.
\newblock Combinatorial sleeping bandits with fairness constraints.
\newblock In {\em Proc. IEEE Int. Conf. Computer Communications (INFOCOM)},
  pages 1702--1710, 2019.

\bibitem{LiuLiShi_21}
Xin Liu, Bin Li, Pengyi Shi, and Lei Ying.
\newblock An efficient pessimistic-optimistic algorithm for stochastic linear
  bandits with general constraints.
\newblock In {\em Advances Neural Information Processing Systems (NeurIPS)},
  2021.

\bibitem{MagSri_16}
Siva~Theja Maguluri and R~Srikant.
\newblock Heavy traffic queue length behavior in a switch under the maxweight
  algorithm.
\newblock {\em Stochastic Systems}, 6(1):211--250, 2016.

\bibitem{NeeRagLa_12}
Michael~J. Neely, Scott~T. Rager, and Thomas~F. La~Porta.
\newblock Max weight learning algorithms for scheduling in unknown
  environments.
\newblock {\em IEEE Trans. Autom. Control}, 57(5):1179--1191, 2012.

\bibitem{ShaWis_07}
D.~Shah and D.~Wischik.
\newblock Heavy traffic analysis of optimal scheduling algorithms for switched
  networks.
\newblock 2007.
\newblock Submitted to Annals of Applied Probability.

\bibitem{SriYin_14}
R.~Srikant and Lei Ying.
\newblock {\em Communication Networks: {A}n Optimization, Control and
  Stochastic Networks Perspective}.
\newblock Cambridge University Press, 2014.

\bibitem{StaShrMod_19}
Thomas Stahlbuhk, Brooke Shrader, and Eytan Modiano.
\newblock Learning algorithms for scheduling in wireless networks with unknown
  channel statistics.
\newblock {\em Ad Hoc Networks}, 85:131--144, 2019.

\bibitem{Sto_04}
A.~L. Stolyar.
\newblock Max{W}eight scheduling in a generalized switch: {S}tate space
  collapse and workload minimization in heavy traffic.
\newblock {\em Adv. in Appl. Probab.}, 14(1), 2004.

\bibitem{TasEph_93}
L.~Tassiulas and A.~Ephremides.
\newblock Dynamic server allocation to parallel queues with randomly varying
  connectivity.
\newblock {\em IEEE Trans. Inf. Theory}, 39:466--478, Mar. 1993.

\bibitem{TasEph_92}
Leandros Tassiulas and Anthony Ephremides.
\newblock Stability properties of constrained queueing systems and scheduling
  policies for maximum throughput in multihop radio networks.
\newblock {\em IEEE Trans. Autom. Control}, 37:1936--1948, December 1992.

\bibitem{yang2023Learning}
Zixian Yang, R.~Srikant, and Lei Ying.
\newblock Learning while scheduling in multi-server systems with unknown
  statistics: Maxweight with discounted ucb.
\newblock In {\em Proceedings of The 26th International Conference on
  Artificial Intelligence and Statistics (AISTATS)}, volume 206, pages
  4275--4312. PMLR, 25--27 Apr 2023.

\bibitem{yekkehkhany2020blind}
Ali Yekkehkhany and Rakesh Nagi.
\newblock Blind gb-pandas: A blind throughput-optimal load balancing algorithm
  for affinity scheduling.
\newblock {\em IEEE/ACM Transactions on Networking}, 28(3):1199--1212, 2020.

\end{thebibliography}


\clearpage
\newpage
\appendix

\section*{Appendices: Table of Contents}

In the appendices, we provide a counter-example of MaxWeight with empirical mean algorithm, complete proofs of Theorem~\ref{theo:1}, Theorem~\ref{theo:2}, Corollary~\ref{cor:to-theo-2-tail}, Theorem~\ref{theo:3}, proofs of all the lemmas, and additional details of the simulations. The contents are listed as follows:
\begin{itemize}
    \item Section~\ref{app:sec:counter-example} is a counter-example of MaxWeight with empirical mean algorithm, which was mentioned in Section~\ref{sec:intro}.
    \item Section~\ref{app:sec:proof:theo:1} contains the proof of Theorem~\ref{theo:1}.
    \item Section~\ref{app:sec:proof:theo:2} contains the proof of Theorem~\ref{theo:2}.
    \item Section~\ref{app:sec:proof:cor:to-theo-2-tail} contains the proof of Corollary~\ref{cor:to-theo-2-tail}.
    \item Section~\ref{app:sec:proof:theo:3} contains the proof of Theorem~\ref{theo:3}.
    \item Section~\ref{app:sec:proof:lemmas} contains the proofs of all the lemmas.
    \begin{itemize}
        \item Section~\ref{app:proof-lemma-concentration}: Proof of Lemma~\ref{lemma:concentration}.
        \item Section~\ref{app:proof-lemma-prop-D}: Proof of Lemma~\ref{lemma:prop-D}.
        \item Section~\ref{app:proof-lemma-queue-bound}: Proof of Lemma~\ref{lemma:queue-bound}.
        \item Section~\ref{app:proof-lemma-bounds-on-diff-queue-lens}: Proof of Lemma~\ref{lemma:bounds-on-diff-queue-lens}.
        \item Section~\ref{app:proof-lemma-decouple-queue-and-ucb-bonus}: Proof of Lemma~\ref{lemma:decouple-queue-and-ucb-bonus}.
        \item Section~\ref{app:proof-lemma-eij}: Proof of Lemma~\ref{lemma:eij}.
        \item Section~\ref{app:proof-lemma-bound-on-sum-queue}: Proof of Lemma~\ref{lemma:bound-on-sum-queue}.
        \item Section~\ref{app:proof-lemma-prop-G}: Proof of Lemma~\ref{lemma:prop-G}.
        \item Section~\ref{app:proof-lemma-concentration-theo-2}: Proof of Lemma~\ref{lemma:concentration-theo-2}.
        \item Section~\ref{app:proof-lemma-taylor}: Proof of Lemma~\ref{lemma:taylor}.
        \item Section~\ref{app:proof-lemma-concentration-3}: Proof of Lemma~\ref{lemma:concentration-3}.
        \item Section~\ref{app:proof-lemma-sum-error-queue}: Proof of Lemma~\ref{lemma:sum-error-queue}.
        \item Section~\ref{app:proof-lemma-decouple-queue-len-ucb}: Proof of Lemma~\ref{lemma:decouple-queue-len-ucb}.
    \end{itemize}
    \item Section~\ref{app:simu} contains additional details of the simulations, including the settings and the parameters we use, and the zoom-out views of the figures in Section~\ref{sec:simu}.
\end{itemize}

\clearpage

\section{A Counter-Example of MaxWeight with Empirical Mean Algorithm}
\label{app:sec:counter-example}
In this section, we will present an example showing that the MaxWeight with empirical mean algorithm is unstable.

Consider a multi-server system with two servers and two job types with the following statistics: 
$$\Pr\left(S_{i,j}=1\right)=0.99,\quad \Pr\left(S_{i,j}=100\right)=0.01\quad \hbox{for} \quad i=j$$ and  
$$\Pr\left(S_{i,j}=10\right)=1\quad \hbox{for} \quad i\not=j,$$
where $S_{i,j}$ is the service time of type $i$ jobs at server $j$.
We further assume the following job arrival process:  $A_i(t)=1$ for any $i$ and any $t=1, 3, \ldots$ and $A_i(t)=0$ for any $i$ and $t=2, 4, \ldots,$ where $A_i(t)$ is the number of type $i$ jobs that arrive at time slot $t$. We next consider the queue lengths over time under MaxWeight with empirical mean. Let $\hat{\mu}_{i,j}$ denote the empirical mean of service rates. Assume the algorithm uses $\hat{\mu}_{i,j}=1$ as a default value for initial empirical mean if there is no data sample for $S_{i,j}.$ 
\begin{itemize}
    \item Time slot 1: A type-$i$ job is scheduled at server $i$ and $S_{i,i}=100$ for $i=1,2$ which occurs with probability 0.01. 
    \item Time slot 101: Both queues have 49 jobs. We have estimated $\hat{\mu}_{i,i}=0.01$ and $\hat{\mu}_{i,j}=1$ ($i\not=j$) as the default value. The algorithm now schedules a type-$i$ job to server $j$ for $j\not=i.$
    \item Time slot $ 111:$ Both queues have 53 jobs. The estimated service rates are $\hat{\mu}_{i,i}=0.01$ and $\hat{\mu}_{i,j}=0.1$ for $i\not=j.$ Based on MaxWeight with mean-service-rate, the scheduler schedules type-$i$ jobs to server $j$ for $i\not=j.$
    \item Time slot $> 111:$ Since $S_{i,j}$ is a constant for $i\not=j,$ the estimated service rates do not change after the jobs are completed. Since the estimated service rates do not change as long as type-$i$ jobs are scheduled on server $j$ such that $i\not=j$, the schedule decisions also remain the same such that type-$i$ jobs are continuously scheduled to server $j$ for $i\not=j$. Since it takes 10 time slots to finish a job and there is a job arrival every two slots, both queues go to infinity. 

\end{itemize}
Note that if we schedule type-$i$ jobs to server $i,$ the mean queue lengths are bounded because in this case, the mean service time is $1.99$ time slots and the arrival rate is one job every two time slots. 

From the example above, we can see that the problem of using empirical mean is that the initial bad samples led to a poor estimation of $\mu_{i,j},$ which led to poor scheduling decisions.  Since the scheduler only gets new samples from the served jobs, it was not able to correct the wrong estimate of $\hat{\mu}_{i,i}=0.01$ when type-$i$ jobs are no long routed to server $i$ after time slot 101.  Therefore, the system was ``locked in'' in a state with poor estimation and wrong scheduling decisions, which led to instability. 

\section{Proof of Theorem~\ref{theo:1}}
\label{app:sec:proof:theo:1}

\begin{figure}[htb]
    \centering
    \includegraphics[width=0.9\linewidth]{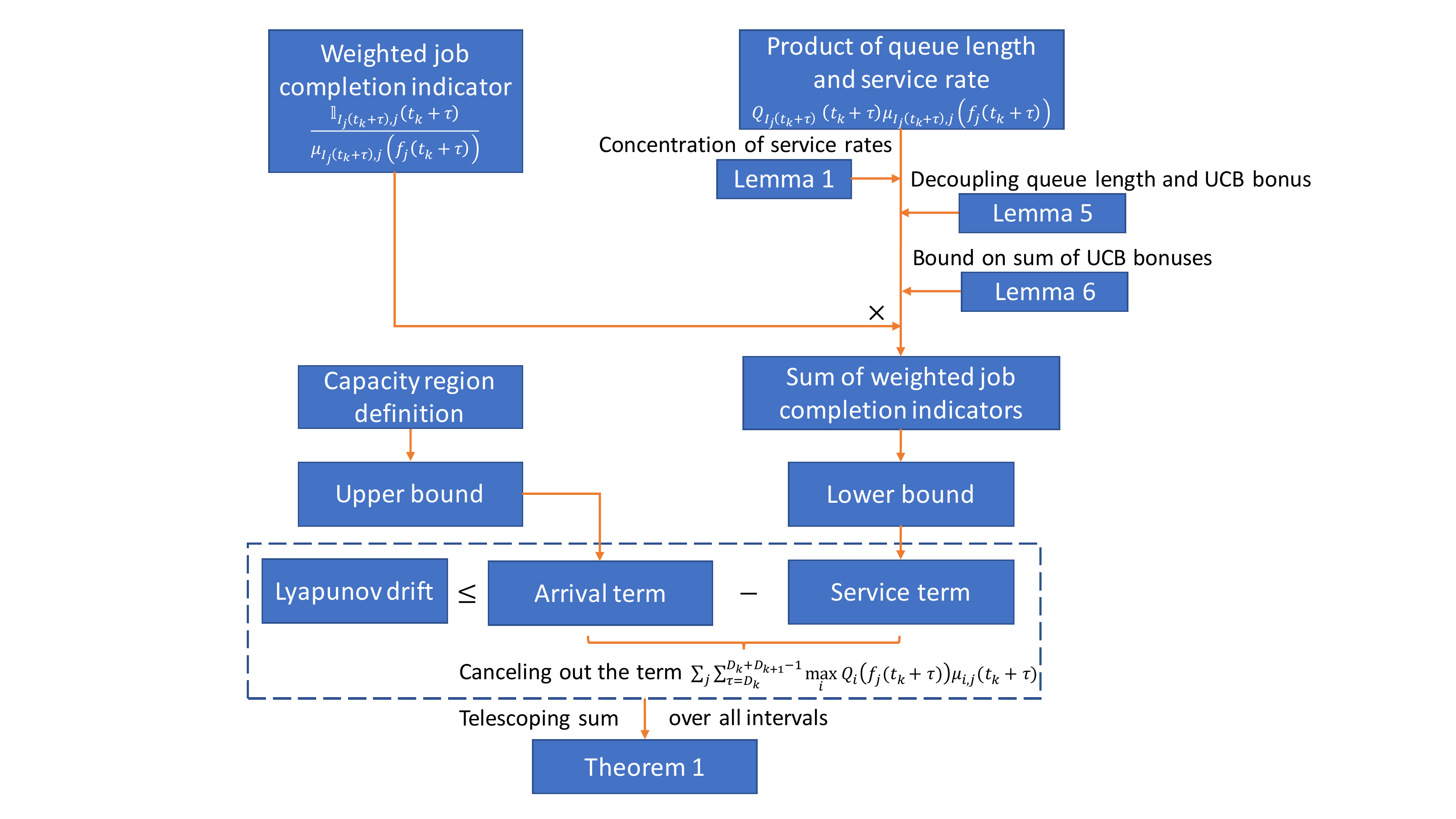}
    \caption{The Proof Roadmap of Theorem~\ref{theo:1}.}
    \label{fig:proof-roadmap}
\end{figure}

In this section, we will present the complete proof of Theorem~\ref{theo:1}.
Figure~\ref{fig:proof-roadmap} shows the proof raodmap of Theorem~\ref{theo:1}.
Before presenting the proof, we define a few additional notations.
In the proof, if server $j$ is not available at the beginning of time slot $t$, i.e., $\sum_i M_{i,j}(t) > 0$, we let $\hat{i}^*_{j}(t)=0$. Let $T\coloneqq g(\gamma)=\frac{4}{1-\gamma}\log \frac{1}{1-\gamma}$ for ease of notation. Denote by $\hat{P}_{t_k}(\cdot)$ the conditional probability $\Pr\left(\cdot|\mathbf{Q}(t_k)={\mathbf q}, \mathbf{H}(t_k)={\mathbf h}\right)$.
Denote by
$\hat{E}_{t_k}[\cdot]$ the conditional expectation $E[\cdot \left| \boldsymbol{Q}(t_k)=\boldsymbol{q}, \boldsymbol{H}(t_k)=\boldsymbol{h}\right.]$.

We now present the proof of Theorem~\ref{theo:1} in the following subsections.

\subsection{Dividing the Time Horizon}

Firstly, we want to divide the time horizon into intervals.
We assume $\frac{T}{8}$ is an integer without loss of generality.
Since ${\boldsymbol \lambda}+\delta {\boldsymbol 1}\in {\mathcal C}(W)$,
for any time slot $\tau$, there exists a $w(\tau)$ that satisfies the inequality in the capacity region definition~\eqref{equ:def-capacity}. Define 
\begin{align}\label{equ:def:tau-l-t}
  \tau_0(t)\coloneqq t,\qquad \tau_l(t)\coloneqq\tau_{l-1}(t)+w(\tau_{l-1}(t)) \mbox{ for $l\ge 1$}.
\end{align}
Define $D(t)$ such that
\begin{align*}
    D(t)=& \min_{n} \sum_{l=0}^n w(\tau_l(t)) \qquad \mbox{s.t. } \sum_{l=0}^{n} w(\tau_l(t)) \ge \frac{T}{2}.
\end{align*}
Denote by $n^*(t)$ the optimal solution to the above optimization problem. Note that $n^*(t)$ and $D(t)$ are fixed numbers rather than random variables for a given $t$. We have the following upper and lower bounds for $D(t)$:
\begin{lemma}\label{lemma:prop-D}
    Suppose $W\le \frac{T}{2}$. Then $\frac{T}{2}\le D(t) \le \frac{T}{2}+W \le T$ for any $t$.
\end{lemma}
Proof of this lemma can be found in Section~\ref{app:proof-lemma-prop-D}.
Let $t_0=0$ and $t_k=t_{k-1}+D(t_{k-1})$ for $k\ge 1$.
Let $D_k\coloneqq D(t_k)$ for simplicity. Then the time horizon can be divided into intervals with length $D_0,D_1,\ldots,D_k,\ldots$, where the $k^{\mathrm{th}}$ interval is $[t_k, t_{k+1}]$. We remark that this partition of the time horizon into time intervals is for the analysis only. The proposed algorithm does not need to know this partition and does not use the time interval information for scheduling and learning. 

In the next subsection, we will analyze and decompose the Lyapunov Drift in each interval.

\subsection{Decomposing the Lyapunov Drift}

Consider the Lyapunov function $L(t)\coloneqq \sum_i Q_i^2(t)$. 
We first consider the Lyapunov drift for the interval $[t_{k+1}, t_{k+1}+D_{k+1}]$  given the queue length $\boldsymbol{Q}(t_k)$ and $\boldsymbol{H}(t_k)$. We analyze the drift conditioned on $\boldsymbol{Q}(t_k)$ and $\boldsymbol{H}(t_k)$ instead of $\boldsymbol{Q}(t_{k+1})$ and $\boldsymbol{H}(t_{k+1})$ to weaken the dependence of the UCB bonuses and the estimated service rates on the conditional values. 
We have
\begin{align}\label{equ:drift-1}
    & E\left[ L(t_{k+1}+D_{k+1}) - L(t_{k+1}) \left| \boldsymbol{Q}(t_k)=\boldsymbol{q}, \boldsymbol{H}(t_k)=\boldsymbol{h}\right.\right]\nonumber\\
    = & E\left[ L(t_k+D_k+D_{k+1}) - L(t_k + D_k) \left| \boldsymbol{Q}(t_k)=\boldsymbol{q}, \boldsymbol{H}(t_k)=\boldsymbol{h}\right.\right]\nonumber\\
    = & \sum_{\tau=D_k}^{D_k+D_{k+1}-1} E\left[ L(t_k+\tau+1) - L(t_k + \tau) \left| \boldsymbol{Q}(t_k)=\boldsymbol{q}, \boldsymbol{H}(t_k)=\boldsymbol{h}\right.\right].
\end{align}
We first look at each term in the summation above. Note that by the queue dynamic~\eqref{equ:queue-dynamics} we can obtain the following upper bound for $Q_i(t+1)$:
\begin{lemma}\label{lemma:queue-bound}
For any $i,t$,
$Q_i(t+1) \le \max \left\{ J, Q_i(t)+A_i(t)-\sum_j \mathbb{1}_{i,j}(t)\right\}$.
\end{lemma}
Proof of this lemma can be found in Section~\ref{app:proof-lemma-queue-bound}.
Recall the definition of $\hat{E}_{t_k}[\cdot]$.
By Lemma~\ref{lemma:queue-bound}, we have
\begin{align}\label{equ:drift-2}
    &\hat{E}_{t_k}\left[ L(t_k+\tau+1) - L(t_k + \tau) \right]= \hat{E}_{t_k}\left[ \sum_i \left(Q_i^2(t_k+\tau+1) - Q_i^2(t_k+\tau)\right) \right]\nonumber\\
    \le & \hat{E}_{t_k}\left[ \sum_i \biggl[
    \max\biggl\{J^2, \bigl(Q_i(t_k+\tau)+A_i(t_k+\tau)-\sum_j \mathbb{1}_{i,j}(t_k+\tau) \bigr)^2 \biggr\}
    - Q_i^2(t_k+\tau)\biggr] \right]\nonumber\\
    \le & \hat{E}_{t_k}\left[ \sum_i \biggl[
    J^2 + \bigl(Q_i(t_k+\tau)+A_i(t_k+\tau)-\sum_j \mathbb{1}_{i,j}(t_k+\tau) \bigr)^2
    - Q_i^2(t_k+\tau)\biggr] \right]\nonumber\\
    = & \hat{E}_{t_k}\left[ \sum_i  2 Q_i(t_k+\tau) \bigl( A_i(t_k+\tau)-\sum_j \mathbb{1}_{i,j}(t_k+\tau)\bigr) \right]
    + \hat{E}_{t_k}\left[ \sum_i \bigl( A_i(t_k+\tau)-\sum_j \mathbb{1}_{i,j}(t_k+\tau) \bigr)^2 \right] + IJ^2
\end{align}
where the second inequality is due to the fact that $\max\{x^2,y^2\}\le x^2 + y^2$, and the second term in the last line can be bounded as follows:
\begin{align}\label{equ:drift-variance-term}
    & \sum_i \bigl( A_i(t_k+\tau)-\sum_j \mathbb{1}_{i,j}(t_k+\tau) \bigr)^2 \le  \sum_i \bigl( \max\bigl\{A_i(t_k+\tau),\sum_j \mathbb{1}_{i,j}(t_k+\tau)\bigr\} \bigr)^2 \nonumber\\
    \le & \sum_i (A_i(t_k+\tau))^2 + \sum_i \bigl( \sum_j \mathbb{1}_{i,j}(t_k+\tau)\bigr)^2
    \le IU_{\mathrm{A}}^2 + \bigl[ \sum_i\sum_j \mathbb{1}_{i,j}(t_k+\tau) \bigr]^2
    \le IU_{\mathrm{A}}^2 + J^2,
\end{align}
where the last two steps are due to the fact that $A_i(t_k+\tau) \le U_{\mathrm{A}}$ and $\sum_i\sum_j \mathbb{1}_{i,j}(t_k+\tau)\le J$. Hence, from~\eqref{equ:drift-2} and \eqref{equ:drift-variance-term}, we have
\begin{align*}
    & \hat{E}_{t_k}\left[ L(t_k+\tau+1) - L(t_k + \tau) \right]\\
    \le & \hat{E}_{t_k}\left[ \sum_i  2 Q_i(t_k+\tau) A_i(t_k+\tau) \right] - \hat{E}_{t_k}\left[ \sum_i  2 Q_i(t_k+\tau) \sum_j \mathbb{1}_{i,j}(t_k+\tau) \right] + IU_{\mathrm{A}}^2 + J^2 + IJ^2.
\end{align*}
Substituting the above inequality into~\eqref{equ:drift-1}, we have
\begin{align}
    & E\left[ L(t_k+D_k+D_{k+1}) - L(t_k + D_k) \left| \boldsymbol{Q}(t_k)=\boldsymbol{q}, \boldsymbol{H}(t_k)=\boldsymbol{h}\right.\right]\nonumber\\
    \le & \sum_{\tau=D_k}^{D_k+D_{k+1}-1} \hat{E}_{t_k}\left[ \sum_i  2 Q_i(t_k+\tau) A_i(t_k+\tau) \right] \label{equ:arrival-term}\\
    & - \sum_{\tau=D_k}^{D_k+D_{k+1}-1} \hat{E}_{t_k}\left[ \sum_i  2 Q_i(t_k+\tau) \sum_j \mathbb{1}_{i,j}(t_k+\tau) \right] + ( IU_{\mathrm{A}}^2 + J^2 + IJ^2 ) T \label{equ:service-term}
\end{align}
where the inequality uses the the upper bound on $D_{k+1}$ in Lemma~\ref{lemma:prop-D}. We will next find the bounds for the arrival term~\eqref{equ:arrival-term} and the service term~\eqref{equ:service-term}.

In the next subsection, we will bound the arrival term~\eqref{equ:arrival-term}.

\subsection{Bounding the Arrival Term}
\label{app:proof-theo-1-bound-arrival}
We first analyze the arrival term~\eqref{equ:arrival-term}. We have
\begin{align*}
    \eqref{equ:arrival-term}
    = & \sum_{\tau=D_k}^{D_k+D_{k+1}-1} \hat{E}_{t_k} \left[ E \left[\sum_i  2 Q_i(t_k+\tau) A_i(t_k+\tau) \left| \boldsymbol{Q}(t_k+\tau), \boldsymbol{Q}(t_k)=\boldsymbol{q}, \boldsymbol{H}(t_k)=\boldsymbol{h} \right.\right] \right]\\
    = & \sum_{\tau=D_k}^{D_k+D_{k+1}-1} \hat{E}_{t_k} \left[ \sum_i  2 Q_i(t_k+\tau) E \left[ A_i(t_k+\tau) \left| \boldsymbol{Q}(t_k+\tau), \boldsymbol{Q}(t_k)=\boldsymbol{q}, \boldsymbol{H}(t_k)=\boldsymbol{h} \right.\right] \right]\\
    = & \sum_{\tau=D_k}^{D_k+D_{k+1}-1} \hat{E}_{t_k} \left[ \sum_i  2 Q_i(t_k+\tau)  \lambda_i(t_k+\tau) \right],
\end{align*}
where the first equality is by the law of iterated expectation and the last equality is due to the fact that $A_i(t_k+\tau)$ is independent of $\boldsymbol{Q}(t_k+\tau)$, $\boldsymbol{Q}(t_k)$, and $\boldsymbol{H}(t_k)$. By adding and subtracting $\delta$, we have
\begin{align}\label{equ:arrival-term-1}
    \eqref{equ:arrival-term} = \sum_{\tau=D_k}^{D_k+D_{k+1}-1} \hat{E}_{t_k} \left[ \sum_i  2 Q_i(t_k+\tau)  (\lambda_i(t_k+\tau) + \delta) \right] - 2 \delta  \sum_{\tau=D_k}^{D_k+D_{k+1}-1} \hat{E}_{t_k} \left[ \sum_i   Q_i(t_k+\tau) \right].
\end{align}
Since ${\boldsymbol \lambda}+\delta {\boldsymbol 1}\in {\mathcal C}(W)$, by the definitions of $t_{k+1}$ and $D_{k+1}$, we have
\begin{align}\label{equ:transform-arrival-1}
    & \sum_{\tau=D_k}^{D_k+D_{k+1}-1} \sum_i 2 Q_i(t_k+\tau) (\lambda_i(t_k+\tau) + \delta) = \sum_{\tau=t_k+D_k}^{t_k+D_k+D_{k+1}-1} \sum_i 2 Q_i(\tau) (\lambda_i(\tau) + \delta) \nonumber\\
    = & \sum_{\tau=t_{k+1}}^{t_{k+1}+D_{k+1}-1} \sum_i 2 Q_i(\tau) (\lambda_i(\tau) + \delta) =  \sum_{\tau=t_{k+1}}^{t_{k+1}+\sum_{l=0}^{n^*(t_{k+1})} w(\tau_l(t_{k+1}))-1} \sum_i 2 Q_i(\tau) (\lambda_i(\tau) + \delta) \nonumber\\
    = & \sum_{l=0}^{n^*(t_{k+1})} ~ \sum_{\tau = t_{k+1} + \sum_{l'=0}^{l-1} w(\tau_{l'}(t_{k+1}))}^{t_{k+1} + \sum_{l'=0}^{l} w(\tau_{l'}(t_{k+1}))-1} \sum_i 2 Q_i(\tau)  (\lambda_i(\tau) + \delta).
\end{align}
By the queue dynamics~\eqref{equ:queue-dynamics} and the bounds on the arrival rate and service rate, we have the following bounds on the difference between queue lengths in two different time slots:
\begin{lemma}\label{lemma:bounds-on-diff-queue-lens}
    For any $t,i$, $\tau \ge 0$, we have
    \begin{enumerate}
        \item[(1)] $Q_i(t) - J\tau \le Q_i(t+\tau) \le Q_i(t) + \tau U_{\mathrm{A}}$;
        \item[(2)] $\sum_{i\in{\cal I}} Q_i(t+\tau) \ge \sum_{i\in {\cal I}} Q_i(t) - J\tau$ ~for any subset of queues ${\cal I}\subseteq \{1,2,\ldots,I\}$.
    \end{enumerate}
\end{lemma}
Proof of this lemma can be found in Section~\ref{app:proof-lemma-bounds-on-diff-queue-lens}. Recall the definition of $\tau_l(t)$ in~\eqref{equ:def:tau-l-t}. By Lemma~\ref{lemma:bounds-on-diff-queue-lens}, it holds that for any $\tau\in[\tau_l(t_{k+1}), \tau_{l+1}(t_{k+1})-1]$,
\begin{align}\label{equ:bound-Q_i_t_k_plus_tau}
    Q_i(\tau) \le Q_i(\tau_l(t_{k+1})) + (\tau - \tau_l(t_{k+1})) U_{\mathrm{A}} \le Q_i(\tau_l(t_{k+1})) + W U_{\mathrm{A}},
\end{align}
where the last inequality holds since $\tau - \tau_l(t_{k+1}) < \tau_{l+1}(t_{k+1}) - \tau_l(t_{k+1}) \le W$ by the bound of $w(t)$ in the definition of ${\mathcal C}(W)$. Then, substituting~\eqref{equ:bound-Q_i_t_k_plus_tau} into~\eqref{equ:transform-arrival-1},  we have
\begin{align}\label{equ:transform-arrival-2}
    & \sum_{\tau=D_k}^{D_k+D_{k+1}-1} \sum_i 2 Q_i(t_k+\tau) (\lambda_i(t_k+\tau) + \delta)\nonumber\\
    \le & \sum_{l=0}^{n^*(t_{k+1})} \sum_i 2 Q_i(\tau_l(t_{k+1})) \sum_{\tau = \tau_l(t_{k+1})}^{\tau_{l+1}(t_{k+1})-1}  (\lambda_i(\tau) + \delta) 
    + \sum_{l=0}^{n^*(t_{k+1})} \sum_{\tau = \tau_l(t_{k+1})}^{\tau_{l+1}(t_{k+1})-1} 2 I WU_{\mathrm{A}} (U_{\mathrm{A}} + 1),
\end{align}
where we also use the fact that $\lambda_i(\tau)+\delta\le U_{\mathrm{A}} + 1$. Since ${\boldsymbol \lambda}+\delta {\boldsymbol 1}\in {\mathcal C}(W)$, by the definitions of $\tau_{l+1}(t_{k+1})$ and ${\mathcal C}(W)$, we can bound the first term in~\eqref{equ:transform-arrival-2} as follows:
\begin{align*}
    & \sum_{l=0}^{n^*(t_{k+1})} \sum_i 2 Q_i(\tau_l(t_{k+1})) \sum_{\tau = \tau_l(t_{k+1})}^{\tau_{l+1}(t_{k+1})-1}  (\lambda_i(\tau) + \delta) \nonumber\\
    = &  \sum_{l=0}^{n^*(t_{k+1})} \sum_i 2 Q_i(\tau_l(t_{k+1})) \sum_{\tau = \tau_l (t_{k+1})}^{\tau_l (t_{k+1}) + w(\tau_{l}(t_{k+1}) ) -1} (\lambda_i(\tau) + \delta)\nonumber\\
    \le & \sum_{l=0}^{n^*(t_{k+1})} \sum_i 2 Q_i(\tau_l(t_{k+1})) \sum_{\tau = \tau_l (t_{k+1})}^{\tau_l (t_{k+1}) + w(\tau_{l}(t_{k+1}) ) -1} \sum_{j} \alpha_{i,j}(\tau)\mu_{i,j}(\tau).
\end{align*}
Substituting the above bound back into~\eqref{equ:transform-arrival-2}, we obtain
\begin{align*}
    & \sum_{\tau=D_k}^{D_k+D_{k+1}-1} \sum_i 2 Q_i(t_k+\tau) (\lambda_i(t_k+\tau) + \delta) \nonumber\\
    \le & \sum_{l=0}^{n^*(t_{k+1})} \sum_i 2 Q_i(\tau_l(t_{k+1})) \sum_{\tau = \tau_l (t_{k+1})}^{\tau_l (t_{k+1}) + w(\tau_{l}(t_{k+1}) ) -1} \sum_{j} \alpha_{i,j}(\tau)\mu_{i,j}(\tau) 
    + \sum_{l=0}^{n^*(t_{k+1})} \sum_{\tau = \tau_l(t_{k+1})}^{\tau_{l+1}(t_{k+1})-1} 2 I WU_{\mathrm{A}} (U_{\mathrm{A}} + 1),
\end{align*}
In the same way, we can transforming the double summations regarding $l$ and $\tau$ back to a single summation to obtain
\begin{align}\label{equ:arrival-term-2}
    & \sum_{\tau=D_k}^{D_k+D_{k+1}-1} \sum_i 2 Q_i(t_k+\tau) (\lambda_i(t_k+\tau) + \delta) \nonumber\\
    \le & \sum_{\tau=t_{k+1}}^{t_{k+1}+D_{k+1}-1} 
    \sum_i 2 Q_i(\tau_{l(\tau)}(t_{k+1}))  
    \sum_{j} \alpha_{i,j}(\tau)\mu_{i,j}(\tau) 
    +  2 D_{k+1} I WU_{\mathrm{A}} (U_{\mathrm{A}} + 1),
\end{align}
where $\tau_{l(\tau)}(t_{k+1})$ is starting time of the window which $\tau$ is in, i.e., 
\begin{align*}
    \tau_{l(\tau)}(t_{k+1})\coloneqq \tau_{l}(t_{k+1}) \mbox{ where $l$ is such that } \tau\in[\tau_{l}(t_{k+1}), \tau_{l+1}(t_{k+1})-1].
\end{align*}
Since $\sum_i \alpha_{i,j}(\tau) \le 1$, we can further bound~\eqref{equ:arrival-term-2} as follows:
\begin{align}\label{equ:arrival-term-3}
    & \sum_{\tau=D_k}^{D_k+D_{k+1}-1} \sum_i 2 Q_i(t_k+\tau) (\lambda_i(t_k+\tau) + \delta) \nonumber\\
    \le & \sum_{\tau=t_{k+1}}^{t_{k+1}+D_{k+1}-1} 
    \sum_{j} 2 \max_i Q_i(\tau_{l(\tau)}(t_{k+1}))  
    \mu_{i,j}(\tau) \left(\sum_{i'}\alpha_{i',j}(\tau)\right)
    +  2 D_{k+1} I WU_{\mathrm{A}} (U_{\mathrm{A}} + 1)\nonumber\\
    \le & \sum_{\tau=t_{k+1}}^{t_{k+1}+D_{k+1}-1} 
    \sum_{j} 2 \max_i Q_i(\tau_{l(\tau)}(t_{k+1}))  
    \mu_{i,j}(\tau)
    +  2 D_{k+1} I WU_{\mathrm{A}} (U_{\mathrm{A}} + 1)\nonumber\\
    = & 2\sum_{j} \sum_{\tau=D_k}^{D_k+D_{k+1}-1} 
    \max_i Q_i(\tau_{l(t_k+\tau)}(t_{k+1}))  
    \mu_{i,j}(t_k+\tau)
    + 2 D_{k+1} I WU_{\mathrm{A}} (U_{\mathrm{A}} + 1)\nonumber\\
    \le & 2\sum_{j} \sum_{\tau=D_k}^{D_k+D_{k+1}-1} 
    \max_i Q_i(\tau_{l(t_k+\tau)}(t_{k+1}))  
    \mu_{i,j}(t_k+\tau)
    +  2 I WU_{\mathrm{A}} (U_{\mathrm{A}} + 1) T,
\end{align}
where the last inequality is by Lemma~\ref{lemma:prop-D}.
Define a mapping $f_j$ that maps a time slot to another time slot such that if $y = f_j(x)$ then $y$ is the time slot when server $j$ picked the job that was being served at server $j$ in time slot $x$. If server $j$ was idling in time slot $x$, then let $f_j(x)=x$. That is,
$
    f_j(x)\coloneqq\max\{ t: t\le x, \hat{i}^*_{j}(t) = I_j(x) \}.
$
Note that for any $\tau\in[D_k, D_k+D_{k+1}-1]$, we have
\begin{align*}
    \tau_{l(t_k+\tau)}(t_{k+1}) - f_j(t_k+\tau) = & \left[\tau_{l(t_k+\tau)}(t_{k+1}) - (t_k+\tau)\right] + \left[(t_k+\tau) - f_j(t_k+\tau)\right]\nonumber\\
    \le & \left[(t_k+\tau) - f_j(t_k+\tau)\right] \le U_{\mathrm{S}},
\end{align*}
where the first inequality is due to the fact that $\tau_{l(t_k+\tau)}(t_{k+1}) \le (t_k+\tau)$ according to the definition of $\tau_{l(t_k+\tau)}(t_{k+1})$ and the last inequality is by the definition of $f_j(t_k+\tau)$ and the service time bound $U_{\mathrm{S}}$. Similarly,
\begin{align*}
    \tau_{l(t_k+\tau)}(t_{k+1}) - f_j(t_k+\tau) = & \left[\tau_{l(t_k+\tau)}(t_{k+1}) - (t_k+\tau)\right] + \left[(t_k+\tau) - f_j(t_k+\tau)\right]\nonumber\\
    \ge & \tau_{l(t_k+\tau)}(t_{k+1}) - (t_k+\tau) \ge - W,
\end{align*}
where the first inequality is due to the fact that $t_k+\tau \ge f_j(t_k+\tau)$ and the last inequality is by the bound of each window. Hence, $\tau_{l(t_k+\tau)}(t_{k+1}) - f_j(t_k+\tau)\in[-W, U_{\mathrm{S}}]$. Then by Lemma~\ref{lemma:bounds-on-diff-queue-lens}, we have
\begin{itemize}
    \item If $\tau_{l(t_k+\tau)}(t_{k+1}) \ge f_j(t_k+\tau)$, then 
    \begin{align*}
    Q_i(\tau_{l(t_k+\tau)}(t_{k+1})) \le Q_i(f_j(t_k+\tau)) + (\tau_{l(t_k+\tau)}(t_{k+1}) - f_j(t_k+\tau)) U_{\mathrm{A}} \le Q_i(f_j(t_k+\tau)) + U_\mathrm{S} U_{\mathrm{A}}.
    \end{align*}
    \item If $\tau_{l(t_k+\tau)}(t_{k+1}) \le f_j(t_k+\tau)$, then
    \begin{align*}
    Q_i(\tau_{l(t_k+\tau)}(t_{k+1})) \le Q_i(f_j(t_k+\tau)) + (f_j(t_k+\tau) - \tau_{l(t_k+\tau)}(t_{k+1})) J \le Q_i(f_j(t_k+\tau)) + JW.
    \end{align*}
\end{itemize}
Therefore, we have
\begin{align*}
    Q_i(\tau_{l(t_k+\tau)}(t_{k+1})) \le Q_i(f_j(t_k+\tau)) + \max\{U_\mathrm{S} U_{\mathrm{A}}, JW\}.
\end{align*}
Substituting the above inequality back into \eqref{equ:arrival-term-3} and using the fact that $\mu_{i,j}(t_k+\tau)\le 1$, we obtain
\begin{align*}
    & \sum_{\tau=D_k}^{D_k+D_{k+1}-1} \sum_i 2 Q_i(t_k+\tau) (\lambda_i(t_k+\tau) + \delta) \nonumber\\
    \le & 2\sum_{j} \sum_{\tau=D_k}^{D_k+D_{k+1}-1} 
    \max_i Q_i(f_j(t_k+\tau))
    \mu_{i,j}(t_k+\tau) + 2JD_{k+1}\max\{U_\mathrm{S} U_{\mathrm{A}}, JW\}
    +  2 I WU_{\mathrm{A}} (U_{\mathrm{A}} + 1) T\nonumber\\
    \le & 2\sum_{j} \sum_{\tau=D_k}^{D_k+D_{k+1}-1} 
    \max_i Q_i(f_j(t_k+\tau))
    \mu_{i,j}(t_k+\tau) + 2JT\max\{U_\mathrm{S} U_{\mathrm{A}}, JW\}
    +  2 I WU_{\mathrm{A}} (U_{\mathrm{A}} + 1) T,
\end{align*}
where the last inequality is by Lemma~\ref{lemma:prop-D}.
Substituting the above inequality back into \eqref{equ:arrival-term-1}, we obtain
\begin{align}\label{equ:arrival-term-final}
    \eqref{equ:arrival-term} 
    \le & 2\sum_{j} \hat{E}_{t_k} \left[\sum_{\tau=D_k}^{D_k+D_{k+1}-1} 
    \max_i Q_i(f_j(t_k+\tau))
    \mu_{i,j}(t_k+\tau) \right] - 2 \delta  \sum_{\tau=D_k}^{D_k+D_{k+1}-1} \hat{E}_{t_k} \left[ \sum_i   Q_i(t_k+\tau) \right]\nonumber\\
    & + 2JT\max\{U_\mathrm{S} U_{\mathrm{A}}, JW\}
    +  2 I WU_{\mathrm{A}} (U_{\mathrm{A}} + 1) T.
\end{align}
In the next subsection, we will bound the the service term~\eqref{equ:service-term}.

\subsection{Bounding the Service Term}
Now we analyze the service term~\eqref{equ:service-term}.
Let us first fix a server $j$. We want to lower bound the following per-server service term:
\begin{align*}
    \sum_{\tau=D_k}^{D_k+D_{k+1}-1} \hat{E}_{t_k}\left[ \sum_i  Q_i(t_k+\tau) \mathbb{1}_{i,j}(t_k+\tau) \right].
\end{align*}
The process takes several steps, which are shown in the following.

\subsubsection{Step 1: Adding the Concentration Event}

Recall the high probability concentration event ${\mathcal E}_{t_k,j}$ which is defined in~\eqref{equ:concentration-event}. We have
\begin{align*}
     \sum_{\tau=D_k}^{D_k+D_{k+1}-1} \hat{E}_{t_k}\left[ \sum_i  Q_i(t_k+\tau) \mathbb{1}_{i,j}(t_k+\tau) \right]
    \ge \hat{E}_{t_k} \left[ \sum_{\tau=D_k}^{D_k+D_{k+1}-1}  \sum_i  Q_i(t_k+\tau) \mathbb{1}_{i,j}(t_k+\tau) \mathbb{1}_{{\mathcal E}_{t_k,j}}\right].
\end{align*}
Note that $\mathbb{1}_{i,j}(t_k+\tau)=1$ can happen only on the queue to which server $j$ is scheduled in time slot $t_k+\tau$, i.e., the queue $I_j(t_k+\tau)$.
Hence, we have
\begin{align*}
    & \sum_{\tau=D_k}^{D_k+D_{k+1}-1} \hat{E}_{t_k}\left[ \sum_i  Q_i(t_k+\tau) \mathbb{1}_{i,j}(t_k+\tau) \right]\nonumber\\
    \ge & \hat{E}_{t_k} \left[ \sum_{\tau=D_k}^{D_k+D_{k+1}-1}  Q_{I_j(t_k+\tau)}(t_k+\tau) \mathbb{1}_{I_j(t_k+\tau),j}(t_k+\tau) \mathbb{1}_{{\mathcal E}_{t_k,j}}\right].
\end{align*}
Recall the definition of $f_j$. By multiplying and dividing the same term, we have
\begin{align}\label{equ:service-algorithm-term-1}
    & \sum_{\tau=D_k}^{D_k+D_{k+1}-1} \hat{E}_{t_k}\left[ \sum_i  Q_i(t_k+\tau) \mathbb{1}_{i,j}(t_k+\tau) \right]\nonumber\\
    \ge & \hat{E}_{t_k}\Biggl[ \sum_{\tau=D_k}^{D_k+D_{k+1}-1}  Q_{I_j(t_k+\tau)}(t_k+\tau) \mu_{I_j(t_k+\tau),j}(f_j(t_k+\tau)) \frac{\mathbb{1}_{I_j(t_k+\tau),j}(t_k+\tau)}{\mu_{I_j(t_k+\tau),j}(f_j(t_k+\tau))} \mathbb{1}_{{\mathcal E}_{t_k,j}} \Biggr].
\end{align}

\subsubsection{Step 2: Bounding the Product of Queue Length and Service Rate}
We next want to lower bound the term $Q_{I_j(t_k+\tau)}(t_k+\tau) \mu_{I_j(t_k+\tau),j}(f_j(t_k+\tau))$ in~\eqref{equ:service-algorithm-term-1}. The following analysis in this subsection is under the concentration event ${\mathcal E}_{t_k,j}$. 
Since $U_{\mathrm{S}}\le \frac{T}{8}$, we have
$f_j(t_k+\tau)\ge t_k+\tau -U_{\mathrm{S}}\ge t_k+D_k -U_{\mathrm{S}} \ge t_k+D_k - \frac{T}{8}.$
Also note that
$f_j(t_k+\tau)\le t_k+\tau \le t_k + D_k+D_{k+1}-1.$
Hence, we have
\begin{align}\label{equ:range-of-f_j}
    f_j(t_k+\tau)\in \left[t_k+D_k - \frac{T}{8}, t_k + D_k+D_{k+1}-1\right].
\end{align}
Define
\begin{align*}
    \bar{\mu}_{I_j(t_k+\tau),j}(f_j(t_k+\tau))
    \coloneqq &
    \frac{1}{\max\left\{\frac{1}{\hat{\mu}_{I_j(t_k+\tau),j}(f_j(t_k+\tau))}-b_{I_j(t_k+\tau),j}(f_j(t_k+\tau)), 1\right\}}\nonumber\\
    \underline{\mu}_{I_j(t_k+\tau),j}(f_j(t_k+\tau))
    \coloneqq &
    \frac{1}{\frac{1}{\hat{\mu}_{I_j(t_k+\tau),j}(f_j(t_k+\tau))}+b_{I_j(t_k+\tau),j}(f_j(t_k+\tau))}
\end{align*}
By~\eqref{equ:range-of-f_j} and the definition of the concentration event ${\mathcal E}_{t_k,j}$ in~\eqref{equ:concentration-event}, we have
\begin{align*}
    \frac{1}{\hat{\mu}_{I_j(t_k+\tau),j}(f_j(t_k+\tau))}-b_{I_j(t_k+\tau),j}(f_j(t_k+\tau)) \le & \frac{1}{\mu_{I_j(t_k+\tau),j}(f_j(t_k+\tau))}\nonumber\\
    \frac{1}{\hat{\mu}_{I_j(t_k+\tau),j}(f_j(t_k+\tau))} + b_{I_j(t_k+\tau),j}(f_j(t_k+\tau)) \ge & \frac{1}{\mu_{I_j(t_k+\tau),j}(f_j(t_k+\tau))}.
\end{align*}
Therefore, combining the above inequalities and the fact that $\frac{1}{\mu_{I_j(t_k+\tau),j}(f_j(t_k+\tau))}\ge 1$, we have
\begin{align*}
    \underline{\mu}_{I_j(t_k+\tau),j}(f_j(t_k+\tau)) 
    \le \mu_{I_j(t_k+\tau),j}(f_j(t_k+\tau)) 
    \le \bar{\mu}_{I_j(t_k+\tau),j}(f_j(t_k+\tau)).
\end{align*}
Then we have
\begin{align}\label{equ:q-mu-term-lower-bound}
    & Q_{I_j(t_k+\tau)}(t_k+\tau) \mu_{I_j(t_k+\tau),j}(f_j(t_k+\tau)) \nonumber\\
    = &
    Q_{I_j(t_k+\tau)}(t_k+\tau) \bar{\mu}_{I_j(t_k+\tau),j}(f_j(t_k+\tau)) \nonumber\\
    & + Q_{I_j(t_k+\tau)}(t_k+\tau) \left( \mu_{I_j(t_k+\tau),j}(f_j(t_k+\tau)) - \bar{\mu}_{I_j(t_k+\tau),j}(f_j(t_k+\tau))\right)\nonumber\\
    \ge & Q_{I_j(t_k+\tau)}(t_k+\tau) \bar{\mu}_{I_j(t_k+\tau),j}(f_j(t_k+\tau)) \nonumber\\
    & + Q_{I_j(t_k+\tau)}(t_k+\tau) \left( \underline{\mu}_{I_j(t_k+\tau),j}(f_j(t_k+\tau)) - \bar{\mu}_{I_j(t_k+\tau),j}(f_j(t_k+\tau))\right).
\end{align}
Note that
\begin{align}\label{equ:mu-difference-bound}
    & \underline{\mu}_{I_j(t_k+\tau),j}(f_j(t_k+\tau)) - \bar{\mu}_{I_j(t_k+\tau),j}(f_j(t_k+\tau))\nonumber\\
    = & \frac{1}{\frac{1}{\hat{\mu}_{I_j(t_k+\tau),j}(f_j(t_k+\tau))}+b_{I_j(t_k+\tau),j}(f_j(t_k+\tau))} - \frac{1}{\max\left\{\frac{1}{\hat{\mu}_{I_j(t_k+\tau),j}(f_j(t_k+\tau))}-b_{I_j(t_k+\tau),j}(f_j(t_k+\tau)), 1\right\}}\nonumber\\
    = & \frac{\max\left\{\frac{1}{\hat{\mu}_{I_j(t_k+\tau),j}(f_j(t_k+\tau))}-b_{I_j(t_k+\tau),j}(f_j(t_k+\tau)), 1\right\} - \left(\frac{1}{\hat{\mu}_{I_j(t_k+\tau),j}(f_j(t_k+\tau))}+b_{I_j(t_k+\tau),j}(f_j(t_k+\tau))\right)}{\left(\frac{1}{\hat{\mu}_{I_j(t_k+\tau),j}(f_j(t_k+\tau))}+b_{I_j(t_k+\tau),j}(f_j(t_k+\tau))\right) \max\left\{\frac{1}{\hat{\mu}_{I_j(t_k+\tau),j}(f_j(t_k+\tau))}-b_{I_j(t_k+\tau),j}(f_j(t_k+\tau)), 1\right\}}\nonumber\\
    \ge & \frac{-2b_{I_j(t_k+\tau),j}(f_j(t_k+\tau))}{\left(\frac{1}{\hat{\mu}_{I_j(t_k+\tau),j}(f_j(t_k+\tau))}+b_{I_j(t_k+\tau),j}(f_j(t_k+\tau))\right) \max\left\{\frac{1}{\hat{\mu}_{I_j(t_k+\tau),j}(f_j(t_k+\tau))}-b_{I_j(t_k+\tau),j}(f_j(t_k+\tau)), 1\right\}}\nonumber\\
    \ge & \frac{-2b_{I_j(t_k+\tau),j}(f_j(t_k+\tau))}{\frac{1}{\hat{\mu}_{I_j(t_k+\tau),j}(f_j(t_k+\tau))}+b_{I_j(t_k+\tau),j}(f_j(t_k+\tau)) }\nonumber\\
    \ge & -2b_{I_j(t_k+\tau),j}(f_j(t_k+\tau)),
\end{align}
where the last inequality uses the fact that $\frac{1}{\hat{\mu}_{i,j}(t)} \ge 1$ for any $i,j,t$.
Also note that $\underline{\mu}_{I_j(t_k+\tau),j}(f_j(t_k+\tau)) - \bar{\mu}_{I_j(t_k+\tau),j}(f_j(t_k+\tau))\ge -1$.
Hence, combining~\eqref{equ:mu-difference-bound} and \eqref{equ:q-mu-term-lower-bound}, we have
\begin{align}\label{equ:q-mu-term-2}
     & Q_{I_j(t_k+\tau)}(t_k+\tau) \mu_{I_j(t_k+\tau),j}(f_j(t_k+\tau)) \nonumber\\
     \ge & Q_{I_j(t_k+\tau)}(t_k+\tau) \bar{\mu}_{I_j(t_k+\tau),j}(f_j(t_k+\tau)) 
     - Q_{I_j(t_k+\tau)}(t_k+\tau) \min\{2b_{I_j(t_k+\tau),j}(f_j(t_k+\tau)), 1\}.
\end{align}
Note that by Lemma~\ref{lemma:bounds-on-diff-queue-lens} and the fact that $t_k+\tau - f_j(t_k+\tau)\le U_{\mathrm{S}}$, we have
$Q_{I_j(t_k+\tau)}(t_k+\tau) \ge Q_{I_j(t_k+\tau)}(f_j(t_k+\tau)) - JU_{\mathrm{S}}.$ Then we have
\begin{align}\label{equ:q-mu-term-1}
    Q_{I_j(t_k+\tau)}(t_k+\tau) \bar{\mu}_{I_j(t_k+\tau),j}(f_j(t_k+\tau)) \ge Q_{I_j(t_k+\tau)}(f_j(t_k+\tau)) \bar{\mu}_{I_j(t_k+\tau),j}(f_j(t_k+\tau))  - JU_{\mathrm{S}},
\end{align}
where we use the fact that $\bar{\mu}_{I_j(t_k+\tau),j}(f_j(t_k+\tau))\le 1$.
By Line~\ref{alg:line:max-weight} in Algorithm~\ref{alg:1} and the definition of $\bar{\mu}_{I_j(t_k+\tau),j}(f_j(t_k+\tau))$, we have
\begin{align}\label{equ:Q-mu-term-4}
    Q_{I_j(t_k+\tau)}(f_j(t_k+\tau)) \bar{\mu}_{I_j(t_k+\tau),j}(f_j(t_k+\tau))
    = \max_i  \frac{Q_{i}(f_j(t_k+\tau))}{\max\left\{\frac{1}{\hat{\mu}_{i,j}(f_j(t_k+\tau))}-b_{i,j}(f_j(t_k+\tau)), 1\right\}}.
\end{align}
Combining~\eqref{equ:q-mu-term-2},~\eqref{equ:q-mu-term-1}, and~\eqref{equ:Q-mu-term-4}, we have
\begin{align}\label{equ:Q-mu-term-5}
    &Q_{I_j(t_k+\tau)}(t_k+\tau) \mu_{I_j(t_k+\tau),j}(f_j(t_k+\tau))\nonumber\\
    \ge &  \max_i  \frac{Q_{i}(f_j(t_k+\tau))}{\max\left\{\frac{1}{\hat{\mu}_{i,j}(f_j(t_k+\tau))}-b_{i,j}(f_j(t_k+\tau)), 1\right\}} \nonumber\\
    & - Q_{I_j(t_k+\tau)}(t_k+\tau) \min\{2b_{I_j(t_k+\tau),j}(f_j(t_k+\tau)), 1\} - JU_{\mathrm{S}} \nonumber\\
    \ge &\max_i  Q_{i}(f_j(t_k+\tau)) \mu_{i,j}(f_j(t_k+\tau)) - Q_{I_j(t_k+\tau)}(t_k+\tau) \min\{2b_{I_j(t_k+\tau),j}(f_j(t_k+\tau)), 1\} - JU_{\mathrm{S}},
\end{align}
where the last inequality uses the fact that $\max\left\{\frac{1}{\hat{\mu}_{i,j}(f_j(t_k+\tau))}-b_{i,j}(f_j(t_k+\tau)), 1\right\} \le \frac{1}{\mu_{i,j}(f_j(t_k+\tau))}$, which is based on the concentration event ${\mathcal E}_{t_k,j}$ and the fact that $\frac{1}{\mu_{i,j}(f_j(t_k+\tau))}\ge 1$.
Substituting~\eqref{equ:Q-mu-term-5} into~\eqref{equ:service-algorithm-term-1}, we have
\begin{align}
    & \sum_{\tau=D_k}^{D_k+D_{k+1}-1} \hat{E}_{t_k}\left[ \sum_i  Q_i(t_k+\tau) \mathbb{1}_{i,j}(t_k+\tau) \right]\nonumber\\
    \ge & \hat{E}_{t_k}\Biggl[ \sum_{\tau=D_k}^{D_k+D_{k+1}-1}  \biggl( \max_i  Q_{i}(f_j(t_k+\tau)) \mu_{i,j}(f_j(t_k+\tau)) - JU_{\mathrm{S}} \nonumber\\
    &\qquad \quad \qquad \qquad  - Q_{I_j(t_k+\tau)}(t_k+\tau) \min\{2b_{I_j(t_k+\tau),j}(f_j(t_k+\tau)), 1\} \biggr)  \frac{\mathbb{1}_{I_j(t_k+\tau),j}(t_k+\tau)}{\mu_{I_j(t_k+\tau),j}(f_j(t_k+\tau))} \mathbb{1}_{{\mathcal E}_{t_k,j}}\Biggr].
    \nonumber\\
    \ge & 
    \hat{E}_{t_k}\left[ \sum_{\tau=D_k}^{D_k+D_{k+1}-1}  \max_i  Q_{i}(f_j(t_k+\tau)) \mu_{i,j}(f_j(t_k+\tau))
    \frac{\mathbb{1}_{I_j(t_k+\tau),j}(t_k+\tau)}{\mu_{I_j(t_k+\tau),j}(f_j(t_k+\tau))} \mathbb{1}_{{\mathcal E}_{t_k,j}}\right] - JU^2_{\mathrm{S}}T\label{equ:sum-service-time-term-1}\\
    & - U_{\mathrm{S}} \hat{E}_{t_k}\Biggl[ \sum_{\tau=D_k}^{D_k+D_{k+1}-1} Q_{I_j(t_k+\tau)}(t_k+\tau) \min\{2b_{I_j(t_k+\tau),j}(f_j(t_k+\tau)), 1\} \mathbb{1}_{I_j(t_k+\tau),j}(t_k+\tau) \Biggr] \label{equ:ucb-summation-term-1},
\end{align}
where the last inequality is by Lemma~\ref{lemma:prop-D} and the fact that
$\frac{1}{\mu_{I_j(t_k+\tau),j}(f_j(t_k+\tau))} \le U_{\mathrm{S}}$.

\subsubsection{Step 3: Bounding the Sum of Queue-Length-Weighted UCB Bonuses}

We first look at the term~\eqref{equ:ucb-summation-term-1}.
Define for any $i,j,t,$
\begin{align*}
    \tilde{b}_{i,j}(t) \coloneqq \min\{2b_{i,j}(t), 1\}
\end{align*}
for ease of notation.
Recall the definition of the waiting queue.
Note that if server $j$ is idling in time slot $f_j(t_k+\tau)$, then $t_k+\tau = f_j(t_k+\tau)$ and the waiting queue $\tilde{Q}_{I_j(t_k+\tau)}(t_k+\tau)=0$. Hence, we have
$\tilde{Q}_{I_j(t_k+\tau)}(t_k+\tau) = \tilde{Q}_{I_j(t_k+\tau)}(t_k+\tau)\eta_j(f_j(t_k+\tau)).$
Also note that
$0\le Q_{I_j(t_k+\tau)}(t_k+\tau) - \tilde{Q}_{I_j(t_k+\tau)}(t_k+\tau) \le J$
by definition. Hence, we have
\begin{align}\label{equ:idling-queue-3}
    Q_{I_j(t_k+\tau)}(t_k+\tau) \le & \tilde{Q}_{I_j(t_k+\tau)}(t_k+\tau) + J\nonumber\\
    = & \tilde{Q}_{I_j(t_k+\tau)}(t_k+\tau)\eta_j(f_j(t_k+\tau)) + J\nonumber\\
    \le & Q_{I_j(t_k+\tau)}(t_k+\tau)\eta_j(f_j(t_k+\tau)) + J.
\end{align}
Hence, by~\eqref{equ:idling-queue-3}, Lemma~\ref{lemma:prop-D}, and the fact that $\tilde{b}_{I_j(t_k+\tau),j}(f_j(t_k+\tau)) \le 1$, we have
\begin{align}\label{equ:ucb-summation-term-3}
    & \hat{E}_{t_k}\Biggl[ \sum_{\tau=D_k}^{D_k+D_{k+1}-1} Q_{I_j(t_k+\tau)}(t_k+\tau) \tilde{b}_{I_j(t_k+\tau),j}(f_j(t_k+\tau))
    \mathbb{1}_{I_j(t_k+\tau),j}(t_k+\tau)  \Biggr]\nonumber\\
    \le & \hat{E}_{t_k}\Biggl[ \sum_{\tau=D_k}^{D_k+D_{k+1}-1} Q_{I_j(t_k+\tau)}(t_k+\tau) \tilde{b}_{I_j(t_k+\tau),j}(f_j(t_k+\tau)) \eta_j(f_j(t_k+\tau))
    \mathbb{1}_{I_j(t_k+\tau),j}(t_k+\tau)
     \Biggr] + JT\nonumber\\
    = & \sum_i \hat{E}_{t_k}\Biggl[ \sum_{\tau=D_k}^{D_k+D_{k+1}-1} Q_{i}(t_k+\tau) \tilde{b}_{i,j}(f_j(t_k+\tau)) \eta_j(f_j(t_k+\tau))
    \mathbb{1}_{i,j}(t_k+\tau) \mathbb{1}_{I_j(t_k+\tau)=i}
     \Biggr] + JT.
\end{align}
Considering the event $\{\tilde{b}_{i,j}(f_j(t_k+\tau)) \le \frac{\delta}{2JU_{\mathrm{S}}}\}$, we further have
\begin{align}\label{equ:ucb-summation-term-4}
    & \hat{E}_{t_k}\Biggl[ \sum_{\tau=D_k}^{D_k+D_{k+1}-1} Q_{i}(t_k+\tau) \tilde{b}_{i,j}(f_j(t_k+\tau)) \eta_j(f_j(t_k+\tau))
    \mathbb{1}_{i,j}(t_k+\tau) \mathbb{1}_{I_j(t_k+\tau)=i}
     \Biggr]\nonumber\\
    = & \hat{E}_{t_k}\Biggl[ \sum_{\tau=D_k}^{D_k+D_{k+1}-1} Q_{i}(t_k+\tau) \tilde{b}_{i,j}(f_j(t_k+\tau)) \eta_j(f_j(t_k+\tau))
    \mathbb{1}_{i,j}(t_k+\tau) \mathbb{1}_{I_j(t_k+\tau)=i}
    \mathbb{1}_{\tilde{b}_{i,j}(f_j(t_k+\tau)) \le \frac{\delta}{2JU_{\mathrm{S}}}} 
     \Biggr]\nonumber\\
    & + \hat{E}_{t_k}\Biggl[ \sum_{\tau=D_k}^{D_k+D_{k+1}-1} Q_{i}(t_k+\tau) \tilde{b}_{i,j}(f_j(t_k+\tau)) \eta_j(f_j(t_k+\tau))
    \mathbb{1}_{i,j}(t_k+\tau) \mathbb{1}_{I_j(t_k+\tau)=i}
    \mathbb{1}_{\tilde{b}_{i,j}(f_j(t_k+\tau)) > \frac{\delta}{2JU_{\mathrm{S}}}} 
     \Biggr]\nonumber\\
    \le & \frac{\delta}{2JU_{\mathrm{S}}}  \hat{E}_{t_k}\Biggl[ \sum_{\tau=D_k}^{D_k+D_{k+1}-1} Q_{i}(t_k+\tau)
     \Biggr]\nonumber\\
     & + \hat{E}_{t_k}\Biggl[ \sum_{\tau=D_k}^{D_k+D_{k+1}-1} Q_{i}(t_k+\tau) \tilde{b}_{i,j}(f_j(t_k+\tau)) \eta_j(f_j(t_k+\tau))
    \mathbb{1}_{i,j}(t_k+\tau) \mathbb{1}_{I_j(t_k+\tau)=i}
    \mathbb{1}_{\tilde{b}_{i,j}(f_j(t_k+\tau)) > \frac{\delta}{2JU_{\mathrm{S}}}} 
     \Biggr]\nonumber\\
     \le & \frac{\delta}{2JU_{\mathrm{S}}}  \hat{E}_{t_k}\Biggl[ \sum_{\tau=D_k}^{D_k+D_{k+1}-1} Q_{i}(t_k+\tau)
     \Biggr]\nonumber\\
     & + \hat{E}_{t_k}\Biggl[ \sum_{\tau=D_k}^{D_k+D_{k+1}-1} Q_{i}(t_k+\tau) \tilde{b}_{i,j}(f_j(t_k+\tau)) \eta_j(f_j(t_k+\tau))
    \mathbb{1}_{i,j}(t_k+\tau) \mathbb{1}_{I_j(t_k+\tau)=i}\nonumber\\
    &\qquad \qquad \qquad \qquad \mathbb{1}_{\hat{N}_{i,j}(f_j(t_k+\tau)) < \frac{64 J^2 U_{\mathrm{S}}^4 \log \frac{1}{1-\gamma}}{\delta^2}} 
     \Biggr],
\end{align}
where the last inequality holds since
\begin{align}\label{equ:bound-b-tilde}
    \tilde{b}_{i,j}(t) = \min\{2b_{i,j}(t), 1\} = & \min\left\{4 U_{\mathrm{S}} \sqrt{\frac{\log \left(\sum_{\tau'=0}^{t-1} \gamma^{\tau'}\right)}{\hat{N}_{i,j}(t)}}
    , 1\right\}\nonumber\\
    \le & 4 U_{\mathrm{S}} \sqrt{\frac{\log \frac{1}{1-\gamma}}{\hat{N}_{i,j}(t)}}
\end{align}
for any $t$
according to Line~\ref{alg:line:b} in Algorithm~\ref{alg:1}.
For the second term in~\eqref{equ:ucb-summation-term-4}, we want to decouple the queue length and the UCB bonus so that we can bound the sum of the UCB bonuses.
Define
\begin{align*}
    e_{i,j}\coloneqq \sum_{\tau=D_k}^{D_k+D_{k+1}-1} \tilde{b}_{i,j}(f_j(t_k+\tau)) \eta_j(f_j(t_k+\tau))
    \mathbb{1}_{i,j}(t_k+\tau) \mathbb{1}_{I_j(t_k+\tau)=i}
    \mathbb{1}_{\hat{N}_{i,j}(f_j(t_k+\tau)) < \frac{64 J^2 U_{\mathrm{S}}^4 \log \frac{1}{1-\gamma}}{\delta^2}}.
\end{align*}
Inspired by the method of proving Lemma 5.4 and Lemma 5.5 in~\cite{FreLykWen_22}, we proved the following lemma:
\begin{lemma}\label{lemma:decouple-queue-and-ucb-bonus}
    If $T\ge 2 \left \lceil\frac{4 J U_{\mathrm{S}} e_{i,j}}{\delta}\right \rceil$, we have
    \begin{align*}
        & \sum_{\tau=D_k}^{D_k+D_{k+1}-1} Q_{i}(t_k+\tau) \tilde{b}_{i,j}(f_j(t_k+\tau)) \eta_j(f_j(t_k+\tau))
        \mathbb{1}_{i,j}(t_k+\tau) \mathbb{1}_{I_j(t_k+\tau)=i}
        \mathbb{1}_{\hat{N}_{i,j}(f_j(t_k+\tau)) < \frac{64 J^2 U_{\mathrm{S}}^4 \log \frac{1}{1-\gamma}}{\delta^2}} \nonumber\\
        \le & \frac{\delta}{4JU_{\mathrm{S}}} \sum_{\tau=D_k}^{D_k+D_{k+1}-1} Q_{i}(t_k+\tau) + \frac{4JU_{\mathrm{S}} \max\{U_{\mathrm{A}}, J\} e^2_{i,j}}{\delta}.
    \end{align*}
\end{lemma}
Proof of this lemma can be found in Section~\ref{app:proof-lemma-decouple-queue-and-ucb-bonus}. The difference between our proof and the proof in~\cite{FreLykWen_22} is that we need to spread the queue length over the interval $[t_k+D_k, t_k+D_k+D_{k+1}-1]$ rather than the whole time horizon, which is the reason why we need the additional condition $T\ge 2 \left \lceil\frac{4 J U_{\mathrm{S}} e_{i,j}}{\delta}\right \rceil$.
We next prove an upper bound for the sum of UCB bonuses $e_{i,j}$ using the following lemma:
\begin{lemma}\label{lemma:eij}
Consider $c_1=2$ and $\frac{1}{2} \le \gamma<1$. Let $L$, $U_N$ be two positive constants.
    For any $\tau_l$, $\tau_h$ such that $\tau_l \ge U_{\mathrm{S}}, \tau_h \le L - 1$, we have
    \begin{align*}
        S_\mathrm{UCB} \coloneqq & \sum_{\tau=\tau_l}^{\tau_h} \tilde{b}_{i,j}(f_j(t+\tau)) \eta_j(f_j(t+\tau))
        \mathbb{1}_{i,j}(t+\tau) \mathbb{1}_{I_j(t+\tau)=i}
        \mathbb{1}_{\hat{N}_{i,j}(f_j(t+\tau)) < U_{N}}\nonumber\\
        \le & \left\lceil 2L (1-\gamma) \right\rceil \left( 1 + 16 U_{\mathrm{S}} \sqrt{U_N \log \frac{1}{1-\gamma}} \right)
    \end{align*}
    for any $t$.
\end{lemma}
Proof of this lemma can be found in Section~\ref{app:proof-lemma-eij}.
The proof idea is to divide the interval of the summation into $\lceil 2L (1-\gamma) \rceil$ sub-intervals with each sub-interval containing $\lceil \frac{1}{2(1-\gamma)} \rceil$ samples so that the discount coefficients within each sub-interval can be lower bounded by a constant.
By Lemma~\ref{lemma:prop-D}, we know that $D_k\ge \frac{T}{2} > \frac{T}{8} \ge U_{\mathrm{S}}$ and $D_k + D_{k+1} - 1\le 2T - 1$. Hence, by applying Lemma~\ref{lemma:eij} with $\tau_l=D_k$, $\tau_h=D_k+D_{k+1}-1$, $L=2T$, $U_N=\frac{64 J^2 U_{\mathrm{S}}^4 \log \frac{1}{1-\gamma}}{\delta^2}$, we obtain an upper bound for $e_{i,j}$:
\begin{align}\label{equ:bound-eij}
    e_{i,j} \le \left\lceil 16\log \frac{1}{1-\gamma} \right\rceil \left( 1 + \frac{128 J U^3_{\mathrm{S}}\log \frac{1}{1-\gamma}}{\delta} \right) \le \frac{2059 J U^3_{\mathrm{S}} \log^2 \frac{1}{1-\gamma}}{\delta},
\end{align}
where the last inequality is by $\gamma \ge 1-\frac{1}{1+e^{1.5}}$.
Substituting \eqref{equ:bound-eij} into the result of Lemma~\ref{lemma:decouple-queue-and-ucb-bonus}, we have
\begin{align}\label{equ:bound-sum-ucb-N-small}
    & \sum_{\tau=D_k}^{D_k+D_{k+1}-1} Q_{i}(t_k+\tau) \tilde{b}_{i,j}(f_j(t_k+\tau)) \eta_j(f_j(t_k+\tau))
    \mathbb{1}_{i,j}(t_k+\tau) \mathbb{1}_{I_j(t_k+\tau)=i}
    \mathbb{1}_{\hat{N}_{i,j}(f_j(t_k+\tau)) < \frac{64 J^2 U_{\mathrm{S}}^4 \log \frac{1}{1-\gamma}}{\delta^2}} \nonumber\\
    \le & \frac{\delta}{4JU_{\mathrm{S}}} \sum_{\tau=D_k}^{D_k+D_{k+1}-1} Q_{i}(t_k+\tau) +  \frac{4 \times 2059^2 J^3 U^7_{\mathrm{S}} \max\{U_{\mathrm{A}},J\} \log^4 \frac{1}{1-\gamma}}{\delta^3},
\end{align}
which holds as long as $T \ge 2 \left \lceil\frac{4 J U_{\mathrm{S}} e_{i,j}}{\delta}\right \rceil$. From the bound \eqref{equ:bound-eij} on $e_{i,j}$, a sufficient condition for \eqref{equ:bound-sum-ucb-N-small} to hold is that
$T\ge \frac{16473 J^2 U^4_{\mathrm{S}} \log^2 \frac{1}{1-\gamma}}{\delta^2}$, since $T\ge \frac{16473 J^2 U^4_{\mathrm{S}} \log^2 \frac{1}{1-\gamma}}{\delta^2} \ge 2 \left \lceil\frac{4 J U_{\mathrm{S}} e_{i,j}}{\delta}\right \rceil$.
We can easily verify this sufficient condition using the condition~\eqref{equ:condition-delta-1} in Theorem~\ref{theo:1}.
Substituting~\eqref{equ:bound-sum-ucb-N-small} into~\eqref{equ:ucb-summation-term-4}, we have
\begin{align}\label{equ:ucb-summation-term-5}
    & \hat{E}_{t_k}\Biggl[ \sum_{\tau=D_k}^{D_k+D_{k+1}-1} Q_{i}(t_k+\tau) \tilde{b}_{i,j}(f_j(t_k+\tau)) \eta_j(f_j(t_k+\tau))
    \mathbb{1}_{i,j}(t_k+\tau) \mathbb{1}_{I_j(t_k+\tau)=i}
     \Biggr]\nonumber\\
     \le & \frac{3\delta}{4J U_{\mathrm{S}}} \hat{E}_{t_k}\Biggl[ \sum_{\tau=D_k}^{D_k+D_{k+1}-1} Q_{i}(t_k+\tau)
     \Biggr] + \frac{4 \times 2059^2 J^3 U^7_{\mathrm{S}} \max\{U_{\mathrm{A}},J\} \log^4 \frac{1}{1-\gamma}}{\delta^3}.
\end{align}
Substituting~\eqref{equ:ucb-summation-term-5} into~\eqref{equ:ucb-summation-term-3} and then into~\eqref{equ:ucb-summation-term-1}, we have
\begin{align}\label{equ:sum-ucb-final}
    \eqref{equ:ucb-summation-term-1} 
    \ge - \frac{3\delta}{4J} \hat{E}_{t_k}\Biggl[ \sum_{\tau=D_k}^{D_k+D_{k+1}-1} \sum_i Q_{i}(t_k+\tau)
     \Biggr] - \frac{4 \times 2059^2 I J^3 U^8_{\mathrm{S}} \max\{U_{\mathrm{A}},J\} \log^4 \frac{1}{1-\gamma}}{\delta^3} - JU_{\mathrm{S}}T.
\end{align}

\subsubsection{Step 4: Bounding the Weighted Sum of Job Completion Indicators}

We next look at the term
\begin{align*}
    \hat{E}_{t_k}\left[ \sum_{\tau=D_k}^{D_k+D_{k+1}-1}  \max_i  Q_{i}(f_j(t_k+\tau)) \mu_{i,j}(f_j(t_k+\tau))
    \frac{\mathbb{1}_{I_j(t_k+\tau),j}(t_k+\tau)}{\mu_{I_j(t_k+\tau),j}(f_j(t_k+\tau))} \mathbb{1}_{{\mathcal E}_{t_k,j}}\right]
\end{align*}
in~\eqref{equ:sum-service-time-term-1}.
Let $v_j(t)\coloneqq \max_i Q_{i}(t) \mu_{i,j}(t)$ for any time slot $t$. Since $\mathbb{1}_{{\mathcal E}_{t_k,j}} + \mathbb{1}_{{\mathcal E}^{\mathrm{c}}_{t_k,j}} = 1$, we have
\begin{align}\label{equ:sum-service-time-term-2}
    & \hat{E}_{t_k}\Biggl[ \sum_{\tau=D_k}^{D_k+D_{k+1}-1} v_j(f_j(t_k+\tau))  \frac{\mathbb{1}_{I_j(t_k+\tau),j}(t_k+\tau)}{\mu_{I_j(t_k+\tau),j}(f_j(t_k+\tau))}\mathbb{1}_{{\mathcal E}_{t_k,j}}\Biggr]\nonumber\\
    = &  \hat{E}_{t_k}\Biggl[ \sum_{\tau=D_k}^{D_k+D_{k+1}-1}  \frac{v_j(f_j(t_k+\tau))\mathbb{1}_{I_j(t_k+\tau),j}(t_k+\tau)}{\mu_{I_j(t_k+\tau),j}(f_j(t_k+\tau))}\Biggr] \nonumber\\
    & - \hat{E}_{t_k}\Biggl[ \sum_{\tau=D_k}^{D_k+D_{k+1}-1}  \frac{v_j(f_j(t_k+\tau))\mathbb{1}_{I_j(t_k+\tau),j}(t_k+\tau)}{\mu_{I_j(t_k+\tau),j}(f_j(t_k+\tau))}\mathbb{1}_{{\mathcal E}^{\mathrm{c}}_{t_k,j}}\Biggr].   
\end{align}
Note that
\begin{align*}
    \sum_{\tau=D_k}^{D_k+D_{k+1}-1}  \frac{v_j(f_j(t_k+\tau))\mathbb{1}_{I_j(t_k+\tau),j}(t_k+\tau)}{\mu_{I_j(t_k+\tau),j}(f_j(t_k+\tau))} 
    \le  U_{\mathrm{S}} \sum_{\tau=D_k}^{D_k+D_{k+1}-1} v_j(f_j(t_k+\tau)),
\end{align*}
since $1/{\mu_{i,j}(t)}\le U_{\mathrm{S}}$ for all $i,j,t$.
For all $t\in[t_k, t_k+2T-1]$, we have
\begin{align}\label{equ:bound-v-j}
    v_j(t) = \max_i Q_i(t)\mu_{i,j}(t) \le \max_i Q_i(t) \le \max_i \left(Q_i(t_k)+2TU_{\mathrm{A}}\right) \le \sum_i Q_i(t_k) + 2TU_{\mathrm{A}},
\end{align}
where the first inequality is by $\mu_{i,j}(f_j(t_k+\tau))\le 1$ and the second inequality is by Lemma~\ref{lemma:bounds-on-diff-queue-lens}.
From~\eqref{equ:range-of-f_j}, we have $f_j(t_k+\tau)\in[t_k, t_k+2T-1]$. Hence by~\eqref{equ:bound-v-j}, we have
$v_j(f_j(t_k+\tau)) \le \sum_i Q_i(t_k) + 2 T U_{\mathrm{A}}$. Hence, we have
\begin{align}\label{equ:upper-bound-sum-service-time}
    \sum_{\tau=D_k}^{D_k+D_{k+1}-1}  \frac{v_j(f_j(t_k+\tau))\mathbb{1}_{I_j(t_k+\tau),j}(t_k+\tau)}{\mu_{I_j(t_k+\tau),j}(f_j(t_k+\tau))} \le U_{\mathrm{S}} D_{k+1} \left(\sum_i Q_i(t_k) + 2 T U_{\mathrm{A}}\right).
\end{align}
Hence, combining~\eqref{equ:sum-service-time-term-2} and~\eqref{equ:upper-bound-sum-service-time} and using the notation $\hat{P}_{t_k}(\cdot)$, we have
\begin{align}\label{equ:sum-service-time-term-2-2}
    & \hat{E}_{t_k}\Biggl[ \sum_{\tau=D_k}^{D_k+D_{k+1}-1} v_j(f_j(t_k+\tau))  \frac{\mathbb{1}_{I_j(t_k+\tau),j}(t_k+\tau)}{\mu_{I_j(t_k+\tau),j}(f_j(t_k+\tau))}\mathbb{1}_{{\mathcal E}_{t_k,j}}\Biggr]\nonumber\\
    \ge & \hat{E}_{t_k}\Biggl[ \sum_{\tau=D_k}^{D_k+D_{k+1}-1}  \frac{v_j(f_j(t_k+\tau))\mathbb{1}_{I_j(t_k+\tau),j}(t_k+\tau)}{\mu_{I_j(t_k+\tau),j}(f_j(t_k+\tau))}\Biggr] 
    -\hat{P}_{t_k} ({\mathcal E}_{t_k,j}^{\mathrm c}) U_{\mathrm{S}} D_{k+1} \left(\sum_i q_i + 2 T U_{\mathrm{A}}\right)\nonumber\\
    \ge & \hat{E}_{t_k}\Biggl[ \sum_{\tau=D_k}^{D_k+D_{k+1}-1}  \frac{v_j(f_j(t_k+\tau))\mathbb{1}_{I_j(t_k+\tau),j}(t_k+\tau)}{\mu_{I_j(t_k+\tau),j}(f_j(t_k+\tau))}\Biggr] 
    - 186 I U_{\mathrm{S}}D_{k+1} (1-\gamma)^{1.5} \left(\sum_i q_i + 2 T U_{\mathrm{A}}\right),
\end{align}
where the second inequality is by Lemma~\ref{lemma:concentration}.
We can write the first term of~\eqref{equ:sum-service-time-term-2-2} in a different form by summing over the time slots in which the jobs start, i.e.,
\begin{align}\label{equ:sum-service-time-temp1}
    & \hat{E}_{t_k}\Biggl[ \sum_{\tau=D_k}^{D_k+D_{k+1}-1}  \frac{v_j(f_j(t_k+\tau))\mathbb{1}_{I_j(t_k+\tau),j}(t_k+\tau)}{\mu_{I_j(t_k+\tau),j}(f_j(t_k+\tau))}\Biggr] \nonumber\\
    \ge & \hat{E}_{t_k}\Biggl[ \sum_{\tau=D_k}^{D_k+D_{k+1}-1}  \frac{v_j(t_k+\tau)\mathbb{1}_{\hat{i}^*_j(t_k+\tau)\neq 0}}{\mu_{\hat{i}^*_j(t_k+\tau),j}(t_k+\tau)}\Biggr] - U_{\mathrm{S}} \left( \sum_i q_i + 2 T U_{\mathrm{A}}\right) ,
\end{align}
where the inequality holds since the last job starting before $t_k+D_k+D_{k+1}$ may not finish before $t_k+D_k+D_{k+1}$ and the first job finishing at or after $t_k+D_k$ may not start at or after $t_k+D_k$, and we also use~\eqref{equ:bound-v-j} and the fact that $1/\mu_{i,j}(t)\le U_{\mathrm{S}}$ for all $i,j,t$.
Next we look at the first term in~\eqref{equ:sum-service-time-temp1}.
Let $\mathbb{1}_{\mathrm{idling}}(j, t)\coloneqq 1- \eta_j(t)$, which is equal to $1$ when server $j$ is idling. Dividing the sum into two cases based on whether server $j$ is idling or non-idling,
we have
\begin{align}\label{equ:idling-non-idling}
    & \hat{E}_{t_k}\Biggl[ \sum_{\tau=D_k}^{D_k+D_{k+1}-1}  \frac{v_j(t_k+\tau)\mathbb{1}_{\hat{i}^*_j(t_k+\tau)\neq 0}}{\mu_{\hat{i}^*_j(t_k+\tau),j}(t_k+\tau)}\Biggr] \nonumber\\
    = &
     \hat{E}_{t_k}\Biggl[ \sum_{\tau=D_k}^{D_k+D_{k+1}-1}  \frac{v_j(t_k+\tau)\mathbb{1}_{\hat{i}^*_j(t_k+\tau)\neq 0}\eta_j(t_k+\tau)}{\mu_{\hat{i}^*_j(t_k+\tau),j}(t_k+\tau)}\Biggr]\nonumber\\
     & + \hat{E}_{t_k}\Biggl[ \sum_{\tau=D_k}^{D_k+D_{k+1}-1}  \frac{v_j(t_k+\tau)\mathbb{1}_{\hat{i}^*_j(t_k+\tau)\neq 0} \mathbb{1}_{\mathrm{idling}}(j, t_k+\tau)}{\mu_{\hat{i}^*_j(t_k+\tau),j}(t_k+\tau)}\Biggr] \nonumber\\
    \ge & 
    \hat{E}_{t_k}\Biggl[ \sum_{\tau=D_k}^{D_k+D_{k+1}-1}  \frac{v_j(t_k+\tau)\mathbb{1}_{\hat{i}^*_j(t_k+\tau)\neq 0}\eta_j(t_k+\tau)}{\mu_{\hat{i}^*_j(t_k+\tau),j}(t_k+\tau)}\Biggr]\nonumber\\
    & + \hat{E}_{t_k}\Biggl[ \sum_{\tau=D_k}^{D_k+D_{k+1}-1}  v_j(t_k+\tau)\mathbb{1}_{\hat{i}^*_j(t_k+\tau)\neq 0} \mathbb{1}_{\mathrm{idling}}(j, t_k+\tau) \Biggr],
\end{align}
where the last inequality is due to the fact that $\mu_{i,j}(t)\le 1$ for all $i,j,t$. Note that $E[S_{i,j}(t_k+\tau)] = 1/\mu_{i,j}(t_k+\tau)$ and $\hat{E}_{t_k}[S_{i,j}(t_k+\tau)] = E[S_{i,j}(t_k+\tau)]$ since $S_{i,j}(t_k+\tau)$ is independent of $\boldsymbol{Q}(t_k)$ and $\boldsymbol{H}(t_k)$. Hence, we have
\begin{align}\label{equ:equalities}
    & \hat{E}_{t_k}\Biggl[ \sum_{\tau=D_k}^{D_k+D_{k+1}-1}  \frac{v_j(t_k+\tau)\mathbb{1}_{\hat{i}^*_j(t_k+\tau)\neq 0}\eta_j(t_k+\tau)}{\mu_{\hat{i}^*_j(t_k+\tau),j}(t_k+\tau)}\Biggr] \nonumber\\
    = & \sum_{i=1}^{I} \hat{E}_{t_k}\Biggl[ \sum_{\tau=D_k}^{D_k+D_{k+1}-1}  \frac{v_j(t_k+\tau)\mathbb{1}_{\hat{i}^*_j(t_k+\tau)=i}\eta_j(t_k+\tau)}{\mu_{i,j}(t_k+\tau)}\Biggr]\nonumber\\
    = & \sum_{i=1}^{I} \hat{E}_{t_k}\Biggl[ \sum_{\tau=D_k}^{D_k+D_{k+1}-1}  v_j(t_k+\tau)\mathbb{1}_{\hat{i}^*_j(t_k+\tau)=i}\eta_j(t_k+\tau)
    \hat{E}_{t_k}[S_{i,j}(t_k+\tau)]\Biggr]\nonumber\\
    = & \sum_{i=1}^{I} \hat{E}_{t_k}\Biggl[ \sum_{\tau=D_k}^{D_k+D_{k+1}-1}  v_j(t_k+\tau)\mathbb{1}_{\hat{i}^*_j(t_k+\tau)=i}\eta_j(t_k+\tau)
    \hat{E}_{t_k} [S_{i,j}(t_k+\tau)\left| \boldsymbol{Q}(t_k+\tau), \boldsymbol{H}(t_k+\tau), \boldsymbol{A}(t_k+\tau) \right.] \Biggr]\nonumber\\
    = & \sum_{i=1}^{I} \hat{E}_{t_k}\Biggl[ \sum_{\tau=D_k}^{D_k+D_{k+1}-1}  
    \hat{E}_{t_k} [v_j(t_k+\tau)\mathbb{1}_{\hat{i}^*_j(t_k+\tau)=i}\eta_j(t_k+\tau) S_{i,j}(t_k+\tau)\left| \boldsymbol{Q}(t_k+\tau), \boldsymbol{H}(t_k+\tau), \boldsymbol{A}(t_k+\tau) \right.] \Biggr]\nonumber\\
    = & \sum_{i=1}^{I} \sum_{\tau=D_k}^{D_k+D_{k+1}-1}  
    \hat{E}_{t_k}\left[ \hat{E}_{t_k} [v_j(t_k+\tau)\mathbb{1}_{\hat{i}^*_j(t_k+\tau)=i}\eta_j(t_k+\tau) S_{i,j}(t_k+\tau)\left| \boldsymbol{Q}(t_k+\tau), \boldsymbol{H}(t_k+\tau), \boldsymbol{A}(t_k+\tau) \right.] \right]\nonumber\\
    = &  \sum_{i=1}^{I} \sum_{\tau=D_k}^{D_k+D_{k+1}-1} 
    \hat{E}_{t_k}[v_j(t_k+\tau)\mathbb{1}_{\hat{i}^*_j(t_k+\tau)=i}\eta_j(t_k+\tau) S_{i,j}(t_k+\tau)]\nonumber\\
    = & \sum_{\tau=D_k}^{D_k+D_{k+1}-1} 
    \hat{E}_{t_k}[v_j(t_k+\tau)\mathbb{1}_{\hat{i}^*_j(t_k+\tau)\neq 0}\eta_j(t_k+\tau) S_{\hat{i}^*_j(t_k+\tau),j}(t_k+\tau)],
\end{align}
where the third equality is due to the independence between $S_{i,j}(t_k+\tau)$ and $\boldsymbol{Q}(t_k+\tau), \boldsymbol{H}(t_k+\tau), \boldsymbol{A}(t_k+\tau)$, the fourth equality is due to the fact that $v_j(t_k+\tau),\mathbb{1}_{\hat{i}^*_j(t_k+\tau)=i},\eta_j(t_k+\tau)$ are fully determined by $\boldsymbol{Q}(t_k+\tau), \boldsymbol{H}(t_k+\tau), \boldsymbol{A}(t_k+\tau)$, and the sixth equality is by the law of iterated expectation. Note that in these derivations we view $\hat{E}_{t_k}$ as the expectation under the probability measure $\hat{P}_{t_k}$. Substituting~\eqref{equ:equalities} into~\eqref{equ:idling-non-idling}, we have
\begin{align}\label{equ:sum-service-time-temp3}
    & \hat{E}_{t_k}\Biggl[ \sum_{\tau=D_k}^{D_k+D_{k+1}-1}  \frac{v_j(t_k+\tau)\mathbb{1}_{\hat{i}^*_j(t_k+\tau)\neq 0}}{\mu_{\hat{i}^*_j(t_k+\tau),j}(t_k+\tau)}\Biggr] \nonumber\\
    \ge & \sum_{\tau=D_k}^{D_k+D_{k+1}-1} 
    \hat{E}_{t_k}[v_j(t_k+\tau)\mathbb{1}_{\hat{i}^*_j(t_k+\tau)\neq 0}\eta_j(t_k+\tau) S_{\hat{i}^*_j(t_k+\tau),j}(t_k+\tau)]\nonumber\\
    & + \hat{E}_{t_k}\Biggl[ \sum_{\tau=D_k}^{D_k+D_{k+1}-1}  v_j(t_k+\tau)\mathbb{1}_{\hat{i}^*_j(t_k+\tau)\neq 0} \mathbb{1}_{\mathrm{idling}}(j, t_k+\tau) \Biggr]\nonumber\\
    = & \hat{E}_{t_k}\Biggl[ \sum_{\tau=D_k}^{D_k+D_{k+1}-1}  
    v_j(t_k+\tau)\mathbb{1}_{\hat{i}^*_j(t_k+\tau)\neq 0} \left(\eta_j(t_k+\tau) S_{\hat{i}^*_j(t_k+\tau),j}(t_k+\tau) + \mathbb{1}_{\mathrm{idling}}(j, t_k+\tau)\right)\Biggr].
\end{align}
Note that the term $\eta_j(t_k+\tau) S_{\hat{i}^*_j(t_k+\tau),j}(t_k+\tau) + \mathbb{1}_{\mathrm{idling}}(j, t_k+\tau)$ is the actual time that server $j$ spends on the queue $\hat{i}^*_j(t_k+\tau)$. Hence, \eqref{equ:sum-service-time-temp3} can be rewritten using $f_j$ in the following way:
\begin{align}\label{equ:sum-service-time-temp4}
    & \hat{E}_{t_k}\Biggl[ \sum_{\tau=D_k}^{D_k+D_{k+1}-1}  
    v_j(t_k+\tau)\mathbb{1}_{\hat{i}^*_j(t_k+\tau)\neq 0} \left(\eta_j(t_k+\tau) S_{\hat{i}^*_j(t_k+\tau),j}(t_k+\tau) + \mathbb{1}_{\mathrm{idling}}(j, t_k+\tau)\right)\Biggr] \nonumber\\
    = & \hat{E}_{t_k}\left[ \sum_{\tau=\tau_{\mathrm{start}}}^{\tau_{\mathrm{end}}} v_j(f_j(t_k+\tau)) \right],
\end{align}
where $t_k + \tau_{\mathrm{start}}$ is the starting (or idling) time of the first schedule that starts at or after $t_k + D_k$ and $t_k + \tau_{\mathrm{end}}$ is the finishing (or idling) time of the last schedule that starts at or before $t_k+D_k+D_{k+1}-1$.
By~\eqref{equ:bound-v-j} and the facts that $t_k+\tau_{\mathrm{start}}<t_k + D_k + U_{\mathrm{S}}$ and $t_k+\tau_{\mathrm{end}}\ge t_k + D_k + D_{k+1} - 1$, we have
\begin{align}\label{equ:sum-service-time-temp5}
    & \hat{E}_{t_k}\left[ \sum_{\tau=t_{\mathrm{start}}}^{t_{\mathrm{end}}} v_j(f_j(t_k+\tau)) \right] \nonumber\\
    \ge & \hat{E}_{t_k}\left[ \sum_{\tau=D_k}^{D_k+D_{k+1}-1} v_j(f_j(t_k+\tau)) \right] - U_{\mathrm{S}} \left(\sum_i q_i + 2TU_{\mathrm{A}}\right)\nonumber\\
    \ge & \hat{E}_{t_k}\left[\sum_{\tau=D_k}^{D_k+D_{k+1}-1} \max_i Q_i(f_j(t_k+\tau)) \left(\mu_{i,j}(t_k+\tau)-\frac{1}{T^p}\right)\right]  - U_{\mathrm{S}} \left(\sum_i q_i + 2TU_{\mathrm{A}}\right)\nonumber\\
    = & \hat{E}_{t_k}\left[\sum_{\tau=D_k}^{D_k+D_{k+1}-1} \max_i \left(Q_i(f_j(t_k+\tau)) \mu_{i,j}(t_k+\tau)-\frac{1}{T^p} Q_i(f_j(t_k+\tau)) \right)\right]  - U_{\mathrm{S}} \left(\sum_i q_i + 2TU_{\mathrm{A}}\right)\nonumber\\
    \ge & \hat{E}_{t_k}\left[\sum_{\tau=D_k}^{D_k+D_{k+1}-1} \left( \max_i Q_i(f_j(t_k+\tau)) \mu_{i,j}(t_k+\tau)-\frac{1}{T^p} \biggl(\sum_{i'}Q_{i'}(t_k+\tau) + JU_{\mathrm{S}}\biggr)\right) \right] \nonumber\\
    & - U_{\mathrm{S}} \left(\sum_i q_i + 2TU_{\mathrm{A}}\right)\nonumber\\
    = &  \hat{E}_{t_k}\left[\sum_{\tau=D_k}^{D_k+D_{k+1}-1} \max_i Q_i(f_j(t_k+\tau)) \mu_{i,j}(t_k+\tau)\right] 
    - \frac{1}{T^p} \hat{E}_{t_k}\left[ \sum_{\tau=D_k}^{D_k+D_{k+1}-1} \sum_i Q_i(t_k+\tau) \right]\nonumber\\
    & - U_{\mathrm{S}}\left(\sum_i q_i + 2TU_{\mathrm{A}}\right) - \frac{J U_{\mathrm{S}} D_{k+1} }{T^p},
\end{align}
where the second inequality uses the fact that $t_k+\tau-f_j(t_k+\tau)\le U_{\mathrm{S}}$ and Assumption~\ref{assump:mu-1} (2), and the last inequality is by Lemma~\ref{lemma:bounds-on-diff-queue-lens} and the fact that $t_k+\tau-f_j(t_k+\tau)\le U_{\mathrm{S}}$.
Combining~\eqref{equ:sum-service-time-temp3},~\eqref{equ:sum-service-time-temp4}, and~\eqref{equ:sum-service-time-temp5}, we have
\begin{align}\label{equ:sum-service-time-temp6}
    & \hat{E}_{t_k}\Biggl[ \sum_{\tau=D_k}^{D_k+D_{k+1}-1}  \frac{v_j(t_k+\tau)\mathbb{1}_{\hat{i}^*_j(t_k+\tau)\neq 0}}{\mu_{\hat{i}^*_j(t_k+\tau),j}(t_k+\tau)}\Biggr] \nonumber\\
    \ge & \hat{E}_{t_k}\left[\sum_{\tau=D_k}^{D_k+D_{k+1}-1} \max_i Q_i(f_j(t_k+\tau)) \mu_{i,j}(t_k+\tau)\right] 
    - \frac{1}{T^p} \hat{E}_{t_k}\left[ \sum_{\tau=D_k}^{D_k+D_{k+1}-1} \sum_i Q_i(t_k+\tau) \right]\nonumber\\
    & - U_{\mathrm{S}}\left(\sum_i q_i + 2TU_{\mathrm{A}}\right) - \frac{J U_{\mathrm{S}} D_{k+1} }{T^p}.
\end{align}
Substituting~\eqref{equ:sum-service-time-temp6} into~\eqref{equ:sum-service-time-temp1} and then into~\eqref{equ:sum-service-time-term-2-2}, we have
\begin{align}\label{equ:sum-service-time-temp7}
    & \hat{E}_{t_k}\Biggl[ \sum_{\tau=D_k}^{D_k+D_{k+1}-1} v_j(f_j(t_k+\tau))  \frac{\mathbb{1}_{I_j(t_k+\tau),j}(t_k+\tau)}{\mu_{I_j(t_k+\tau),j}(f_j(t_k+\tau))}\mathbb{1}_{{\mathcal E}_{t_k,j}}\Biggr]\nonumber\\
    \ge & \hat{E}_{t_k}\left[\sum_{\tau=D_k}^{D_k+D_{k+1}-1} \max_i Q_i(f_j(t_k+\tau)) \mu_{i,j}(t_k+\tau)\right] 
    - \frac{1}{T^p} \hat{E}_{t_k}\left[ \sum_{\tau=D_k}^{D_k+D_{k+1}-1} \sum_i Q_i(t_k+\tau) \right]\nonumber\\
    & - \frac{J U_{\mathrm{S}} D_{k+1} }{T^p} - \left(2 U_{\mathrm{S}} + 186 I U_{\mathrm{S}}D_{k+1} (1-\gamma)^{1.5}\right) \left(\sum_i q_i + 2TU_{\mathrm{A}}\right).
\end{align}
For $\sum_i q_i$, we have the following lemma.
\begin{lemma}\label{lemma:bound-on-sum-queue}
$\sum_i  q_i \le  \frac{1}{D_{k+1}}\sum_{\tau=D_k}^{D_k+D_{k+1}-1} \hat{E}_{t_k}\left[ \sum_i Q_i(t_k+\tau) + 2JT \right]$.
\end{lemma}
Proof of this lemma can be found in Section~\ref{app:proof-lemma-bound-on-sum-queue}. By Lemma~\ref{lemma:bound-on-sum-queue},~\eqref{equ:sum-service-time-temp7}, and Lemma~\ref{lemma:prop-D}, we have
\begin{align}\label{equ:bound-sum-weighted-q}
    & \hat{E}_{t_k}\Biggl[ \sum_{\tau=D_k}^{D_k+D_{k+1}-1} v_j(f_j(t_k+\tau))  \frac{\mathbb{1}_{I_j(t_k+\tau),j}(t_k+\tau)}{\mu_{I_j(t_k+\tau),j}(f_j(t_k+\tau))}\mathbb{1}_{{\mathcal E}_{t_k,j}}\Biggr]\nonumber\\
    \ge & \hat{E}_{t_k}\left[\sum_{\tau=D_k}^{D_k+D_{k+1}-1} \max_i Q_i(f_j(t_k+\tau)) \mu_{i,j}(t_k+\tau)\right]
    - \frac{1}{T^p} \hat{E}_{t_k}\left[ \sum_{\tau=D_k}^{D_k+D_{k+1}-1} \sum_i Q_i(t_k+\tau) \right] 
    - \frac{J U_{\mathrm{S}} D_{k+1} }{T^p}\nonumber\\
    &  - \left(2 U_{\mathrm{S}} + 186 I U_{\mathrm{S}}D_{k+1} (1-\gamma)^{1.5}\right) \left(\frac{1}{D_{k+1}} \sum_{\tau=D_k}^{D_k+D_{k+1}-1} \hat{E}_{t_k} \biggl[\sum_i Q_i(t_k+\tau)\biggr] + 2JT + 2TU_{\mathrm{A}}\right)\nonumber\\
    \ge &  \hat{E}_{t_k}\left[\sum_{\tau=D_k}^{D_k+D_{k+1}-1} \max_i Q_i(f_j(t_k+\tau)) \mu_{i,j}(t_k+\tau)\right]
    - 5 U_{\mathrm{S}} (J + U_{\mathrm{A}}) T - 372 I U_{\mathrm{S}} (J + U_{\mathrm{A}}) T^2 (1-\gamma)^{1.5} \nonumber\\
    & - \left(\frac{1}{T^p}+\frac{4U_{\mathrm{S}}}{T}+186 I U_{\mathrm{S}} (1-\gamma)^{1.5}\right) \hat{E}_{t_k} \left[ \sum_{\tau=D_k}^{D_k+D_{k+1}-1} \sum_i Q_i(t_k+\tau)\right]. 
\end{align}
Note that
\begin{align*}
    T^2 (1-\gamma)^{1.5} = 4 T (1-\gamma)^{0.5}\log \frac{1}{1-\gamma} \le \frac{8T}{e},
\end{align*}
where the inequality is by the fact that $(1-\gamma)^{0.5}\log \frac{1}{1-\gamma} \le \frac{2}{e}$. Then from \eqref{equ:bound-sum-weighted-q} we have
\begin{align*}
    & \hat{E}_{t_k}\Biggl[ \sum_{\tau=D_k}^{D_k+D_{k+1}-1} v_j(f_j(t_k+\tau))  \frac{\mathbb{1}_{I_j(t_k+\tau),j}(t_k+\tau)}{\mu_{I_j(t_k+\tau),j}(f_j(t_k+\tau))}\mathbb{1}_{{\mathcal E}_{t_k,j}}\Biggr]\nonumber\\
    \ge & \hat{E}_{t_k}\left[\sum_{\tau=D_k}^{D_k+D_{k+1}-1} \max_i Q_i(f_j(t_k+\tau)) \mu_{i,j}(t_k+\tau)\right]
    - 5 U_{\mathrm{S}} (J + U_{\mathrm{A}}) T - 1095 I U_{\mathrm{S}} (J + U_{\mathrm{A}}) T \nonumber\\
    & - \left(\frac{1}{T^p}+\frac{4U_{\mathrm{S}}}{T}+186 I U_{\mathrm{S}} (1-\gamma)^{1.5}\right) \hat{E}_{t_k} \left[ \sum_{\tau=D_k}^{D_k+D_{k+1}-1} \sum_i Q_i(t_k+\tau)\right]. 
\end{align*}
Substituting the above inequality into~\eqref{equ:sum-service-time-term-1}, we have
\begin{align}\label{equ:sum-service-time-term-final}
    \eqref{equ:sum-service-time-term-1} \ge  & \hat{E}_{t_k}\left[\sum_{\tau=D_k}^{D_k+D_{k+1}-1} \max_i Q_i(f_j(t_k+\tau)) \mu_{i,j}(t_k+\tau)\right]\nonumber\\
    & - \left(\frac{1}{T^p}+\frac{4U_{\mathrm{S}}}{T}+186 I U_{\mathrm{S}} (1-\gamma)^{1.5}\right) \hat{E}_{t_k} \left[ \sum_{\tau=D_k}^{D_k+D_{k+1}-1} \sum_i Q_i(t_k+\tau)\right]\nonumber\\
    & - 5 U_{\mathrm{S}} (J + U_{\mathrm{A}}) T - 1095 I U_{\mathrm{S}} (J + U_{\mathrm{A}}) T - J U^2_{\mathrm{S}} T.
\end{align}

Combining~\eqref{equ:sum-service-time-term-1},~\eqref{equ:ucb-summation-term-1},~\eqref{equ:sum-ucb-final}, and~\eqref{equ:sum-service-time-term-final}, we have
\begin{align*}
    & \sum_{\tau=D_k}^{D_k+D_{k+1}-1} \hat{E}_{t_k}\left[ \sum_i  Q_i(t_k+\tau) \mathbb{1}_{i,j}(t_k+\tau) \right]\nonumber\\
    \ge & \hat{E}_{t_k}\left[\sum_{\tau=D_k}^{D_k+D_{k+1}-1} \max_i Q_i(f_j(t_k+\tau)) \mu_{i,j}(t_k+\tau)\right]\nonumber\\
    & - \left(\frac{1}{T^p}+\frac{4U_{\mathrm{S}}}{T}+186 I U_{\mathrm{S}} (1-\gamma)^{1.5}+\frac{3\delta}{4J}\right) \hat{E}_{t_k} \left[ \sum_{\tau=D_k}^{D_k+D_{k+1}-1} \sum_i Q_i(t_k+\tau)\right]\nonumber\\
    & - 5 U_{\mathrm{S}} (J + U_{\mathrm{A}}) T - 1095 I U_{\mathrm{S}} (J + U_{\mathrm{A}}) T - J U^2_{\mathrm{S}} T
    - \frac{4 \times 2059^2 I J^3 U^8_{\mathrm{S}} \max\{U_{\mathrm{A}},J\} \log^4 \frac{1}{1-\gamma}}{\delta^3} - JU_{\mathrm{S}}T.
\end{align*}
Substituting the above inequality into~\eqref{equ:service-term} ,we have
\begin{align}\label{equ:service-term-final}
    \eqref{equ:service-term} \le & -2\sum_j\hat{E}_{t_k}\left[\sum_{\tau=D_k}^{D_k+D_{k+1}-1} \max_i Q_i(f_j(t_k+\tau)) \mu_{i,j}(t_k+\tau)\right]\nonumber\\
    & + \left(\frac{2J}{T^p}+\frac{8JU_{\mathrm{S}}}{T}+372 I J U_{\mathrm{S}}(1-\gamma)^{1.5}+\frac{3\delta}{2}\right) \hat{E}_{t_k} \left[ \sum_{\tau=D_k}^{D_k+D_{k+1}-1} \sum_i Q_i(t_k+\tau)\right]\nonumber\\
    & + \frac{8 \times 2059^2 I J^4 U^8_{\mathrm{S}} \max\{U_{\mathrm{A}},J\} \log^4 \frac{1}{1-\gamma}}{\delta^3}\nonumber\\
    & + 10 J U_{\mathrm{S}} (J + U_{\mathrm{A}}) T + 2190 I J U_{\mathrm{S}} (J + U_{\mathrm{A}}) T + 2 J^2 U^2_{\mathrm{S}} T
    + 2J^2 U_{\mathrm{S}}T + (IU_{\mathrm{A}}^2 + J^2 + IJ^2 ) T\nonumber\\
    \le & -2\sum_j\hat{E}_{t_k}\left[\sum_{\tau=D_k}^{D_k+D_{k+1}-1} \max_i Q_i(f_j(t_k+\tau)) \mu_{i,j}(t_k+\tau)\right]\nonumber\\
    & + \left(169 I J U_{\mathrm{S}} (1-\gamma)^{\min\{p,1\}} +\frac{3\delta}{2}\right) \hat{E}_{t_k} \left[ \sum_{\tau=D_k}^{D_k+D_{k+1}-1} \sum_i Q_i(t_k+\tau)\right]\nonumber\\
    & + \frac{67831696 I J^5 U^8_{\mathrm{S}} U_{\mathrm{A}} \log^4 \frac{1}{1-\gamma}}{\delta^3} + 4407 I J^2 U^2_{\mathrm{S}} U^2_{\mathrm{A}} T,
\end{align}
where the last inequality is by the definition $T = \frac{4}{1-\gamma}\log \frac{1}{1-\gamma}$ and $\gamma \ge 1-\frac{1}{1+e^{1.5}}$.

In the next subsection, we will combine the bounds of the arrival term and the service term and then sum over all intervals.

\subsection{Telescoping Sum}

Combining~\eqref{equ:arrival-term},~\eqref{equ:service-term},~\eqref{equ:arrival-term-final}, and~\eqref{equ:service-term-final}, we have
\begin{align*}
    & E\left[ L(t_k+D_k+D_{k+1}) - L(t_k + D_k) \left| \boldsymbol{Q}(t_k)=\boldsymbol{q}, \boldsymbol{H}(t_k)=\boldsymbol{h}\right.\right]\nonumber\\
    \le & - \left(\frac{\delta}{2} - 169 I J U_{\mathrm{S}} (1-\gamma)^{\min\{p,1\}} \right) \hat{E}_{t_k} \left[ \sum_{\tau=D_k}^{D_k+D_{k+1}-1} \sum_i Q_i(t_k+\tau)\right]\nonumber\\
    & + \frac{67831696 I J^5 U^8_{\mathrm{S}} U_{\mathrm{A}} \log^4 \frac{1}{1-\gamma}}{\delta^3} + 4415 I J^2 U^2_{\mathrm{S}} U^2_{\mathrm{A}} W T \nonumber\\
    \le & -\frac{\delta}{4} \hat{E}_{t_k} \left[ \sum_{\tau=D_k}^{D_k+D_{k+1}-1} \sum_i Q_i(t_k+\tau)\right] 
    + \frac{67831696 I J^5 U^8_{\mathrm{S}} U_{\mathrm{A}} \log^4 \frac{1}{1-\gamma}}{\delta^3} + 4415 I J^2 U^2_{\mathrm{S}} U^2_{\mathrm{A}} W T,
\end{align*}
where the last inequality holds since
$\frac{\delta}{4} \ge \frac{451}{4} I J U^{2}_{\mathrm{S}} (1-\gamma)^{\min\{p, \frac{1}{3}\}} \log \frac{1}{1-\gamma} \ge 169 I J U_{\mathrm{S}} (1-\gamma)^{\min\{p,1\}}$
by the condition~\eqref{equ:condition-delta-1} and $\gamma \ge 1-\frac{1}{1+e^{1.5}}$.
Taking expectation on both sides, we have
\begin{align}\label{equ:negative-drift-final}
    & E\left[ L(t_k+D_k+D_{k+1}) - L(t_k + D_k) \right]\nonumber\\
    \le & -\frac{\delta}{4} E \left[ \sum_{\tau=D_k}^{D_k+D_{k+1}-1} \sum_i Q_i(t_k+\tau)\right] 
    + \frac{67831696 I J^5 U^8_{\mathrm{S}} U_{\mathrm{A}} \log^4 \frac{1}{1-\gamma}}{\delta^3} + 4415 I J^2 U^2_{\mathrm{S}} U^2_{\mathrm{A}} W T.
\end{align}
Let $t\ge T$. Since $D_0\le T \le t$ (by Lemma~\ref{lemma:prop-D}), we have
\begin{align*}
    \frac{1}{t}\sum_{\tau=1}^{t} E\left[ \sum_i Q_i(\tau) \right] 
    = & \frac{1}{t}\sum_{\tau=1}^{D_0-1} E\left[ \sum_i Q_i(\tau) \right] + \frac{1}{t}\sum_{\tau=D_0}^{t} E\left[ \sum_i Q_i(\tau) \right]. 
\end{align*}
Note that there exists an integer $K$ such that
$
    t\le \sum_{k=0}^{K}D_k - 1 < t+\frac{T}{2}+W
$
by Lemma~\ref{lemma:prop-D}. Then we have
\begin{align}\label{equ:finite-bound-K-ineq}
    \sum_{k=0}^{K}D_k - 1 \ge t \ge \sum_{k=0}^{K}D_k - \frac{T}{2} - W \ge \sum_{k=1}^{K}D_k - W,
\end{align}
where the last inequality is by Lemma~\ref{lemma:prop-D}. Hence, we have
\begin{align}\label{equ:finite-bound-1}
    \frac{1}{t}\sum_{\tau=1}^{t} E\left[ \sum_i Q_i(\tau) \right]
    = & \frac{1}{t}\sum_{\tau=1}^{D_0-1} E\left[ \sum_i Q_i(\tau) \right] + \frac{\sum_{k=1}^{K}D_k}{t}\frac{1}{\sum_{k=1}^{K}D_k}\sum_{\tau=D_0}^{t} E\left[ \sum_i Q_i(\tau) \right]\nonumber\\
    \le & \frac{1}{t}\sum_{\tau=1}^{D_0-1} E\left[ \sum_i Q_i(\tau) \right] + \frac{t+W}{t}\frac{1}{\sum_{k=1}^{K}D_k}\sum_{\tau=D_0}^{\sum_{k=0}^{K}D_k - 1} E\left[ \sum_i Q_i(\tau) \right].
\end{align}
Summing both sides of \eqref{equ:negative-drift-final} over $k=0,1,\ldots,K-1$, we have
\begin{align*}
    E\left[ L\left(\sum_{k=0}^{K}D_{k}\right) - L\left(D_0\right) \right]
    \le & - \frac{\delta}{4} \sum_{\tau=D_0}^{\sum_{k=0}^{K}D_{k}-1} E\left[ \sum_i Q_i(\tau)\right] \nonumber\\
    & + \frac{67831696 I J^5 U^8_{\mathrm{S}} U_{\mathrm{A}} K \log^4 \frac{1}{1-\gamma}}{\delta^3} + 4415 I J^2 U^2_{\mathrm{S}} U^2_{\mathrm{A}} W T K.
\end{align*}
Hence, we have
\begin{align*}
    \sum_{\tau=D_0}^{\sum_{k=0}^{K}D_{k}-1} E \left[ \sum_i Q_i(\tau)\right]  \le & \frac{4}{\delta}  E\left[ - L\left(\sum_{k=0}^{K}D_{k}\right) + L\left(D_0\right) \right] \nonumber\\
    & + \frac{271326784 I J^5 U^8_{\mathrm{S}} U_{\mathrm{A}} K \log^4 \frac{1}{1-\gamma}}{\delta^4} + \frac{17660 I J^2 U^2_{\mathrm{S}} U^2_{\mathrm{A}} W T K}{\delta}.
\end{align*}
Dividing both sides by $\sum_{k=1}^{K}D_{k}$, we have
\begin{align}\label{equ:finite-bound-2}
    & \frac{1}{\sum_{k=1}^{K}D_{k}}\sum_{\tau=D_0}^{\sum_{k=0}^{K}D_{k}-1} E \left[ \sum_i Q_i(\tau)\right]  \nonumber\\
    \le & \frac{4}{\delta\sum_{k=1}^{K}D_{k}}  E\left[ - L\left(\sum_{k=0}^{K}D_{k}\right) + L\left(D_0\right) \right] 
    + \frac{271326784 I J^5 U^8_{\mathrm{S}} U_{\mathrm{A}} K \log^4 \frac{1}{1-\gamma}}{\delta^4 \sum_{k=1}^{K}D_{k}} + \frac{17660 I J^2 U^2_{\mathrm{S}} U^2_{\mathrm{A}} W T K}{\delta \sum_{k=1}^{K}D_{k} }\nonumber\\
    \le & \frac{4}{\delta\sum_{k=1}^{K}D_{k}}  E\left[ L\left(D_0\right) \right]
    + \frac{542653568 I J^5 U^8_{\mathrm{S}} U_{\mathrm{A}} \log^4 \frac{1}{1-\gamma}}{\delta^4 T} + \frac{35320 I J^2 U^2_{\mathrm{S}} U^2_{\mathrm{A}} W}{\delta},
\end{align}
where the last inequality uses Lemma~\ref{lemma:prop-D}.
Substituting \eqref{equ:finite-bound-2} into \eqref{equ:finite-bound-1}, we have
\begin{align}\label{equ:finite-bound-3}
    & \frac{1}{t}\sum_{\tau=1}^{t} E\left[ \sum_i Q_i(\tau) \right]\nonumber\\
    \le & \frac{1}{t}\sum_{\tau=1}^{D_0-1} E\left[ \sum_i Q_i(\tau) \right]\nonumber\\
    & + \frac{t+W}{t}\left(\frac{4}{\delta\sum_{k=1}^{K}D_{k}}  E\left[ L\left(D_0\right) \right]
    + \frac{542653568 I J^5 U^8_{\mathrm{S}} U_{\mathrm{A}} \log^4 \frac{1}{1-\gamma}}{\delta^4 T} + \frac{35320 I J^2 U^2_{\mathrm{S}} U^2_{\mathrm{A}} W}{\delta}\right)\nonumber\\
    \le & \frac{I T^2 U_{\mathrm{A}}}{t}
    + \left(1+\frac{W}{t}\right)\left(\frac{4I U^2_{\mathrm{A}} T^2 }{\delta\sum_{k=1}^{K}D_{k}}
    + \frac{542653568 I J^5 U^8_{\mathrm{S}} U_{\mathrm{A}} \log^4 \frac{1}{1-\gamma}}{\delta^4 T} + \frac{35320 I J^2 U^2_{\mathrm{S}} U^2_{\mathrm{A}} W}{\delta}\right)\nonumber\\
    \le & \frac{I T^2 U_{\mathrm{A}}}{t}
    + \left(1+\frac{W}{t}\right)\left(\frac{4I U^2_{\mathrm{A}} T^2 }{\delta(t+1-T)}
    + \frac{542653568 I J^5 U^8_{\mathrm{S}} U_{\mathrm{A}} \log^4 \frac{1}{1-\gamma}}{\delta^4 T} + \frac{35320 I J^2 U^2_{\mathrm{S}} U^2_{\mathrm{A}} W}{\delta}\right),
\end{align}
where the second inequality is obtained by using Lemma~\ref{lemma:prop-D} and Lemma~\ref{lemma:bounds-on-diff-queue-lens} to bound $Q_i(\tau)$ and $L(D_0)$ with the initial condition $Q_i(0)=0$, and the last inequality holds since $\sum_{k=1}^{K}D_{k} = \sum_{k=0}^{K}D_{k} - D_0 \ge t+ 1 - D_0 \ge t+1-T$ by \eqref{equ:finite-bound-K-ineq} and Lemma~\ref{lemma:prop-D}.
Using the condition~\eqref{equ:condition-delta-1} in Theorem~\ref{theo:1} and the definition of $T$, from~\eqref{equ:finite-bound-3}, we obtain
\begin{align*}
    \frac{1}{t}\sum_{\tau=1}^{t} E\left[ \sum_i Q_i(\tau) \right]
    \le \frac{I T^2 U_{\mathrm{A}}}{t}
    + \left(1+\frac{W}{t}\right)\left(\frac{4I U^2_{\mathrm{A}} T^2 }{\delta(t+1-T)}
    +\frac{35322 I J^2 U^2_{\mathrm{S}} U^2_{\mathrm{A}} W}{\delta}\right).
\end{align*}
The finite-time bound~\eqref{equ:theo-1-finite-bound-2} in Theorem~\ref{theo:1} is proved.
Letting $t\rightarrow \infty$, we obtain 
\begin{align*}
    \limsup_{t\rightarrow\infty}\frac{1}{t}\sum_{\tau=1}^{t}  E\left[\sum_i Q_i(\tau)\right]\le
    \frac{35322 I J^2 U^2_{\mathrm{S}} U^2_{\mathrm{A}} W}{\delta}.
\end{align*}
The asymptotic bound~\eqref{equ:theo-1-asy-bound-2} in Theorem~\ref{theo:1} is proved.

\section{Proof of Theorem~\ref{theo:2}}
\label{app:sec:proof:theo:2}

We will present the complete proof of Theorem~\ref{theo:2} in the following subsections. In the proof, if server $j$ is not available at the beginning of time slot $t$, i.e., $\sum_i M_{i,j}(t) > 0$, we let $\hat{i}^*_{j}(t)=0$. Let $T\coloneqq g(\gamma)=\frac{4}{1-\gamma}\log \frac{1}{1-\gamma}$ for ease of notation. We assume $\frac{T}{8}$ is an integer without loss of generality. 
Denote by $\hat{P}_{t}(\cdot)$ the conditional probability $\Pr\left(\cdot|\mathbf{Q}(t)={\mathbf q}, \mathbf{H}(t)={\mathbf h}\right)$.
Denote by
$\hat{E}_{t}[\cdot]$ the conditional expectation $E[\cdot \left| \boldsymbol{Q}(t)=\boldsymbol{q}, \boldsymbol{H}(t)=\boldsymbol{h}\right.]$.

\subsection{Dividing the Time Horizon}
In order to bound $E\left[e^{\xi \|\boldsymbol{Q}(t)\|_2}\right]$ in Theorem~\ref{theo:2}, we divide $[0,t]$ into intervals. The approach is similar to that used in the proof of Theorem~\ref{theo:1}, but the length of each interval is different, which is important in this proof.
Since ${\boldsymbol \lambda}+\delta {\boldsymbol 1}\in {\mathcal C}(W)$,
for any time slot $\tau$, there exists a $w(\tau)$ that satisfies the inequality in the capacity region definition~\eqref{equ:def-capacity}.
For any time slot $t$, let
\begin{align}\label{equ:def-tau-l-t}
    \tau'_0(t)\coloneqq t+\frac{T}{2}, \qquad \tau'_l(t)\coloneqq \tau'_{l-1}(t) + w(\tau'_{l-1}(t)) ~~ \mbox{for $l\ge 1$}.
\end{align}
Let $G_t$ be an integer such that
\begin{align}\label{equ:def-gt}
    G_t=& \min_{n} \sum_{l=0}^n w(\tau'_l(t)) \qquad \mbox{s.t. } \sum_{l=0}^{n} w(\tau'_l(t)) \ge \frac{c_2T}{\delta},
\end{align}
where $c_2=5(IU_{\mathrm{A}}+J)$ is the constant defined in Assumption~\ref{assump:mu-2}. Then we have the following upper bound and lower bound for $G_t$:
\begin{lemma}\label{lemma:prop-G}
    Suppose $W\le \frac{T}{2}$. Then $\frac{c_2 T}{\delta}\le G_t \le \frac{c_2 T}{\delta} + W \le \frac{(c_2+\frac{1}{2})T}{\delta}$ for all $t$.
\end{lemma}
Proof of this lemma is similar to that of Lemma~\ref{lemma:prop-D} and can be found in Section~\ref{app:proof-lemma-prop-G}. 

In order to bound $E\left[e^{\xi \|\boldsymbol{Q}(t)\|_2}\right]$ in Theorem~\ref{theo:2}, our idea is to first divide $[0,t]$ into intervals $[t'_1,t]$, $[t'_2,t'_1]$, $[t'_3,t'_2]$, $\ldots$, $[0, t'_K]$ such that the length of $[t'_{k+1}, t'_{k}]$ is approximately $\frac{T}{2}+G_{t'_{k+1}}$ for $0\le k\le K-1$ with $t'_0\coloneqq t$. The exact length of each interval and the number of intervals $K$ will be specified later in the proof. And then we will bound the Lyapunov drift for $G_t$ time slots for all $t$.

In the next subsection, we will decompose the Lyapunov drift for $G_t$ time slots.

\subsection{Decomposing the Lyapunov Drift}
Fix any time slot $t\ge 0$.
Consider the Lyapunov function $L'(t)\coloneqq \sqrt{\sum_i Q_i^2(t)}=\|\boldsymbol{Q}(t)\|_2$. We consider the Lyapunov drift for the interval $[t+\frac{T}{2}, t+\frac{T}{2}+G_t]$ given the queue length $\boldsymbol{Q}(t)$ and $\boldsymbol{H}(t)$. We analyze the drift conditioned on $\boldsymbol{Q}(t)$ and $\boldsymbol{H}(t)$ instead of $\boldsymbol{Q}(t+\frac{T}{2})$ and $\boldsymbol{H}(t+\frac{T}{2})$ to weaken the dependence of the UCB bonuses and the estimated service rates on the conditional values. 
We have
\begin{align}\label{equ:theo-2-drift-1}
    & E\left[ L'(t+\frac{T}{2}+G_t) - L'(t+\frac{T}{2}) \left| \boldsymbol{Q}(t)=\boldsymbol{q}, \boldsymbol{H}(t)=\boldsymbol{h}\right.\right]\nonumber\\
    = & \sum_{\tau=\frac{T}{2}}^{\frac{T}{2}+G_t-1} E\left[ L'(t+\tau+1) - L'(t + \tau) \left| \boldsymbol{Q}(t)=\boldsymbol{q}, \boldsymbol{H}(t)=\boldsymbol{h}\right.\right].
\end{align}
We first look at each term in the summation above.
Recall the definition of $\hat{E}_{t}[\cdot]$. Suppose $\|\boldsymbol{Q}(t+\tau)\|_2>0$. Then we have
\begin{align}\label{equ:theo-2-drift-2}
    \hat{E}_{t}\left[ L'(t+\tau+1) - L'(t + \tau) \right] =& \hat{E}_{t}\left[ \|\boldsymbol{Q}(t+\tau+1)\|_2 - \|\boldsymbol{Q}(t + \tau)\|_2 \right]\nonumber\\
    \le & \hat{E}_{t}\left[\frac{1}{2\|\boldsymbol{Q}(t+\tau)\|_2}\left(\|\boldsymbol{Q}(t+\tau+1)\|_2^2 - \|\boldsymbol{Q}(t + \tau)\|_2^2\right)\right],
\end{align}
where the inequality is by the fact that $x-y\le \frac{x^2-y^2}{2y}$ for any $x$ and $y > 0$. Note that
\begin{align}\label{equ:lower-bound-q-t-tau}
    \|\boldsymbol{Q}(t+\tau)\|_2 = & \sqrt{\sum_i Q_i^2(t+\tau)}\nonumber\\
    \ge & \frac{1}{\sqrt{I}} \sum_i Q_i(t+\tau)\nonumber\\
    \ge & \frac{1}{\sqrt{I}} \left(\sum_i Q_i(t+\tau) - J \tau\right)\nonumber\\
    > & \frac{1}{\sqrt{I}} \left(\sum_i Q_i(t) - \frac{(c_2+1)JT}{\delta}  \right),
\end{align}
where the first inequality uses the Cauchy-Schwarz inequality, the second inequality is by Lemma~\ref{lemma:bounds-on-diff-queue-lens}, and the last inequality holds since $\tau \le \frac{T}{2}+G_t-1 < \frac{(c_2+1)T}{\delta}$ by Lemma~\ref{lemma:prop-G}. Therefore, from~\eqref{equ:lower-bound-q-t-tau} we know that \eqref{equ:theo-2-drift-2} holds when $\sum_i q_i = \sum_i Q_i(t) \ge \frac{(c_2+1)JT}{\delta}$.
Following the same argument as the proof of Theorem~\ref{theo:1}, we can further bound \eqref{equ:theo-2-drift-2} as follows:
\begin{align}\label{equ:theo-2-drift-3}
    & \hat{E}_{t}\left[ L'(t+\tau+1) - L'(t + \tau) \right]\nonumber\\
    \le & \hat{E}_{t}\left[\frac{1}{2\|\boldsymbol{Q}(t+\tau)\|_2}\left(
    \sum_i \biggl[
    \max\biggl\{J^2, \bigl(Q_i(t+\tau)+A_i(t+\tau)-\sum_j \mathbb{1}_{i,j}(t+\tau) \bigr)^2 \biggr\}
    - Q_i^2(t+\tau)\biggr] \right)\right]\nonumber\\
    \le & \hat{E}_{t}\left[\frac{1}{2\|\boldsymbol{Q}(t+\tau)\|_2} \left(\sum_i \biggl[
    J^2 + \bigl(Q_i(t+\tau)+A_i(t+\tau)-\sum_j \mathbb{1}_{i,j}(t+\tau) \bigr)^2
    - Q_i^2(t+\tau)\biggr]\right)\right]\nonumber\\
    = & \hat{E}_{t}\Biggl[\frac{1}{2\|\boldsymbol{Q}(t+\tau)\|_2} \biggl(
    \sum_i  2 Q_i(t+\tau) \bigl( A_i(t+\tau)-\sum_j \mathbb{1}_{i,j}(t+\tau)\bigr)
    + \sum_i \bigl( A_i(t+\tau)-\sum_j \mathbb{1}_{i,j}(t+\tau) \bigr)^2\nonumber\\
    & \qquad \qquad \qquad \qquad + IJ^2
    \biggr)\Biggr]\nonumber\\
    = & \hat{E}_{t}\left[\frac{1}{2\|\boldsymbol{Q}(t+\tau)\|_2}
    \sum_i  2 Q_i(t+\tau) \bigl( A_i(t+\tau)-\sum_j \mathbb{1}_{i,j}(t+\tau)\bigr)\right]  + \hat{E}_{t}\left[ \frac{IJ^2}{2\|\boldsymbol{Q}(t+\tau)\|_2} \right]\nonumber\\
    & + \hat{E}_{t}\left[ \frac{1}{2\|\boldsymbol{Q}(t+\tau)\|_2} \sum_i \bigl( A_i(t+\tau)-\sum_j \mathbb{1}_{i,j}(t+\tau) \bigr)^2 \right]\nonumber\\
    \le & \hat{E}_{t}\left[\frac{1}{2\|\boldsymbol{Q}(t+\tau)\|_2}
    \sum_i  2 Q_i(t+\tau) \bigl( A_i(t+\tau)-\sum_j \mathbb{1}_{i,j}(t+\tau)\bigr)\right] + \hat{E}_{t} \left[ \frac{IU_{\mathrm{A}}^2 + J^2 + IJ^2}{2\|\boldsymbol{Q}(t+\tau)\|_2} \right]
\end{align}
where the first inequality is by Lemma~\ref{lemma:queue-bound}, the second inequality is due to the fact that $\max\{x^2,y^2\}\le x^2 + y^2$, and the last inequality is by
\begin{align*}
    & \sum_i \bigl( A_i(t+\tau)-\sum_j \mathbb{1}_{i,j}(t+\tau) \bigr)^2 \le  \sum_i \bigl( \max\bigl\{A_i(t+\tau),\sum_j \mathbb{1}_{i,j}(t+\tau)\bigr\} \bigr)^2\\
    \le & \sum_i (A_i(t+\tau))^2 + \sum_i \bigl( \sum_j \mathbb{1}_{i,j}(t+\tau)\bigr)^2
    \le IU_{\mathrm{A}}^2 + \bigl[ \sum_i\sum_j \mathbb{1}_{i,j}(t+\tau) \bigr]^2
    \le IU_{\mathrm{A}}^2 + J^2,
\end{align*}
where the last two steps are due to the fact that $A_i(t+\tau) \le U_{\mathrm{A}}$ and $\sum_i\sum_j \mathbb{1}_{i,j}(t+\tau)\le J$.
Substituting \eqref{equ:theo-2-drift-3} into \eqref{equ:theo-2-drift-1}, we have
\begin{align}
    & E\left[ L'(t+\frac{T}{2}+G_t) - L'(t+\frac{T}{2}) \left| \boldsymbol{Q}(t)=\boldsymbol{q}, \boldsymbol{H}(t)=\boldsymbol{h}\right.\right]\nonumber\\
    \le & \sum_{\tau=\frac{T}{2}}^{\frac{T}{2}+G_t-1} 
    \hat{E}_{t}\left[\frac{1}{2\|\boldsymbol{Q}(t+\tau)\|_2}
    \sum_i  2 Q_i(t+\tau)  A_i(t+\tau)\right] \label{equ:theo-2-arrival-term}\\
    & - \sum_{\tau=\frac{T}{2}}^{\frac{T}{2}+G_t-1} \hat{E}_{t}\left[\frac{1}{2\|\boldsymbol{Q}(t+\tau)\|_2} 
    \sum_i  2 Q_i(t+\tau) \sum_j \mathbb{1}_{i,j}(t+\tau) \right]\label{equ:theo-2-service-term} \\
    & + \sum_{\tau=\frac{T}{2}}^{\frac{T}{2}+G_t-1} \hat{E}_{t} \left[ \frac{IU_{\mathrm{A}}^2 + J^2 + IJ^2}{2\|\boldsymbol{Q}(t+\tau)\|_2} \right], \label{equ:theo-2-constant-term}
\end{align}
which holds when $\sum_i q_i \ge \frac{(c_2+1)JT}{\delta}$.
We will next find the bounds for the arrival term~\eqref{equ:theo-2-arrival-term} and the service term~\eqref{equ:theo-2-service-term}.

In the next subsection, we will bound the arrival term~\eqref{equ:theo-2-arrival-term}.

\subsection{Bounding the Arrival Term}
We first analyze the arrival term~\eqref{equ:theo-2-arrival-term}. We have
\begin{align*}
    \eqref{equ:theo-2-arrival-term} = & \sum_{\tau=\frac{T}{2}}^{\frac{T}{2}+G_t-1} 
    \hat{E}_{t}\left[
    E \left[ \frac{1}{2\|\boldsymbol{Q}(t+\tau)\|_2}
    \sum_i  2 Q_i(t+\tau)  A_i(t+\tau) \left| \boldsymbol{Q}(t+\tau), \boldsymbol{Q}(t)=\boldsymbol{q}, \boldsymbol{H}(t)=\boldsymbol{h} \right. \right]
    \right]\nonumber\\
    = & \sum_{\tau=\frac{T}{2}}^{\frac{T}{2}+G_t-1} 
    \hat{E}_{t}\left[
    \frac{1}{2\|\boldsymbol{Q}(t+\tau)\|_2}
    \sum_i  2 Q_i(t+\tau)  E \left[  A_i(t+\tau) \left| \boldsymbol{Q}(t+\tau), \boldsymbol{Q}(t)=\boldsymbol{q}, \boldsymbol{H}(t)=\boldsymbol{h} \right. \right]
    \right]\nonumber\\
    = & \sum_{\tau=\frac{T}{2}}^{\frac{T}{2}+G_t-1} 
    \hat{E}_{t}\left[
    \frac{1}{2\|\boldsymbol{Q}(t+\tau)\|_2}
    \sum_i  2 Q_i(t+\tau)  \lambda_i(t+\tau)
    \right].
\end{align*}
Adding and Subtracting $\delta$, we have
\begin{align}\label{equ:theo-2-arrival-term-1}
    \eqref{equ:theo-2-arrival-term} = & 
    \sum_{\tau=\frac{T}{2}}^{\frac{T}{2}+G_t-1} 
    \hat{E}_{t}\left[
    \frac{1}{2\|\boldsymbol{Q}(t+\tau)\|_2}
    \sum_i  2 Q_i(t+\tau)  \left(\lambda_i(t+\tau) + \delta \right)
    \right]\nonumber\\ 
    & - \sum_{\tau=\frac{T}{2}}^{\frac{T}{2}+G_t-1} \delta \hat{E}_{t}\left[ \frac{1}{2\|\boldsymbol{Q}(t+\tau)\|_2}
    \sum_i  2 Q_i(t+\tau) \right]\nonumber\\
    \le & \sum_{\tau=\frac{T}{2}}^{\frac{T}{2}+G_t-1} 
    \hat{E}_{t}\left[
    \frac{1}{2\|\boldsymbol{Q}(t+\tau)\|_2}
    \sum_i  2 Q_i(t+\tau)  \left(\lambda_i(t+\tau) + \delta \right)
    \right] 
    - \delta G_t,
\end{align}
where the last inequality is due to the fact that $\|\boldsymbol{Q}(t+\tau)\|_2 \le \|\boldsymbol{Q}(t+\tau)\|_1 = \sum_i  Q_i(t+\tau)$.
Recall the definition of $G_t$, which is the optimal value of the optimization problem~\eqref{equ:def-gt}. Let $n_{G}^*(t)$ be the optimal solution to~\eqref{equ:def-gt}. Then we have
\begin{align}\label{equ:theo-2-arrival-term-2}
    & \sum_{\tau=\frac{T}{2}}^{\frac{T}{2}+G_t-1} 
    \hat{E}_{t}\left[
    \frac{1}{2\|\boldsymbol{Q}(t+\tau)\|_2}
    \sum_i  2 Q_i(t+\tau)  \left(\lambda_i(t+\tau) + \delta \right)
    \right]\nonumber\\
    = & \sum_{\tau=t + \frac{T}{2}}^{t + \frac{T}{2}+G_t-1} 
    \hat{E}_{t}\left[
    \frac{1}{2\|\boldsymbol{Q}(\tau)\|_2}
    \sum_i  2 Q_i(\tau)  \left(\lambda_i(\tau) + \delta \right)
    \right]\nonumber\\
    = & \sum_{\tau=t + \frac{T}{2}}^{t + \frac{T}{2}+\sum_{l=0}^{n_{G}^*(t)} w(\tau'_l(t)) -1} 
    \hat{E}_{t}\left[
    \frac{1}{2\|\boldsymbol{Q}(\tau)\|_2}
    \sum_i  2 Q_i(\tau)  \left(\lambda_i(\tau) + \delta \right)
    \right]\nonumber\\
    = & \sum_{l=0}^{n_{G}^*(t)} ~\sum_{\tau=t + \frac{T}{2} + \sum_{l'=0}^{l-1} w(\tau'_{l'}(t)) }^{t + \frac{T}{2}+\sum_{l'=0}^{l} w(\tau'_{l'}(t)) -1} 
    \hat{E}_{t}\left[
    \frac{1}{2\|\boldsymbol{Q}(\tau)\|_2}
    \sum_i  2 Q_i(\tau)  \left(\lambda_i(\tau) + \delta \right)
    \right]\nonumber\\
    = & \sum_{l=0}^{n_{G}^*(t)} 
    \sum_i ~\sum_{\tau=t + \frac{T}{2} + \sum_{l'=0}^{l-1} w(\tau'_{l'}(t)) }^{t + \frac{T}{2}+\sum_{l'=0}^{l} w(\tau'_{l'}(t)) -1} 
    \left(\lambda_i(\tau) + \delta \right)
    \hat{E}_{t}\left[
    \frac{Q_i(\tau)}{\|\boldsymbol{Q}(\tau)\|_2}
    \right]\nonumber\\
    = & \sum_{l=0}^{n_{G}^*(t)} 
    \sum_i \sum_{\tau=\tau'_l(t) }^{\tau'_{l+1}(t) -1} 
    \left(\lambda_i(\tau) + \delta \right)
    \hat{E}_{t}\left[
    \frac{Q_i(\tau)}{\|\boldsymbol{Q}(\tau)\|_2}
    \right],
\end{align}
where the last equality is by the definition of $\tau'_l(t)$ in~\eqref{equ:def-tau-l-t}. Note that by Lemma~\ref{lemma:bounds-on-diff-queue-lens} we have
\begin{align*}
    Q_i(\tau) \le Q_i(\tau'_l(t)) + (\tau - \tau'_l(t)) U_{\mathrm{A}} \le Q_i(\tau'_l(t)) + W U_{\mathrm{A}}
\end{align*}
and
\begin{align*}
    Q_i(\tau) \ge Q_i(\tau'_l(t)) -  (\tau - \tau'_l(t)) J \ge Q_i(\tau'_l(t)) - WJ,
\end{align*}
which implies that $Q_i(\tau) \ge \left(Q_i(\tau'_l(t)) - WJ\right)_{+}$, where $(x)_{+}$ denotes $\max\{x, 0\}$. Therefore, we have
\begin{align}\label{equ:upper-bound-q-by-q-norm}
    \hat{E}_{t}\left[
    \frac{Q_i(\tau)}{\|\boldsymbol{Q}(\tau)\|_2}
    \right] 
    \le \hat{E}_{t}\left[
    \frac{Q_i(\tau'_l(t)) + W U_{\mathrm{A}}}{\|\boldsymbol{Q}(\tau)\|_2} \right]
    \le \hat{E}_{t}\left[\frac{Q_i(\tau'_l(t))}{\|\left(\boldsymbol{Q}(\tau'_l(t))- WJ {\boldsymbol 1} \right)_{+}\|_2}\right]
    + \hat{E}_{t}\left[\frac{W U_{\mathrm{A}}}{\|\boldsymbol{Q}(\tau)\|_2}\right],
\end{align}
where ${\boldsymbol 1}$ denotes an all-ones vector and we extend the definition of $(\cdot)_{+}$ to vectors by taking the element-wise maximum with a zero vector. Note that 
\begin{align}\label{equ:lower-bound-q-tau-w-j}
    \|\left(\boldsymbol{Q}(\tau'_l(t))- WJ {\boldsymbol 1} \right)_{+}\|_2 
    \ge & \frac{1}{\sqrt{I}} \sum_i \left(Q_i(\tau'_l(t))- WJ \right)_{+}\nonumber\\
    \ge & \frac{1}{\sqrt{I}} \left(\sum_i Q_i(\tau'_l(t))- WIJ \right)_{+}\nonumber\\
    \ge & \frac{1}{\sqrt{I}} \left(\sum_i Q_i(t) - J\left(\frac{T}{2} + G_t\right) - WIJ \right)_{+}\nonumber\\
    \ge & \frac{1}{\sqrt{I}} \left(\sum_i Q_i(t) - \frac{(c_2+1)JT}{\delta} - WIJ \right)_{+},
\end{align}
where the first inequality is by the Cauchy Schwarz inequality, the third inequality is by Lemma~\ref{lemma:bounds-on-diff-queue-lens}, and the last inequality is by Lemma~\ref{lemma:prop-G}. Hence, from \eqref{equ:lower-bound-q-tau-w-j} we know that \eqref{equ:upper-bound-q-by-q-norm} holds when $\sum_i q_i = \sum_i Q_i > \frac{(c_2+1)JT}{\delta} + WIJ$ in order to make sure that the denominator $\|(\boldsymbol{Q}(\tau'_l(t))- WJ {\boldsymbol 1} )_{+}\|_2$ is not equal to zero.
Substituting \eqref{equ:upper-bound-q-by-q-norm} into \eqref{equ:theo-2-arrival-term-2}, we have
\begin{align*}
    & \sum_{\tau=\frac{T}{2}}^{\frac{T}{2}+G_t-1} 
    \hat{E}_{t}\left[
    \frac{1}{2\|\boldsymbol{Q}(t+\tau)\|_2}
    \sum_i  2 Q_i(t+\tau)  \left(\lambda_i(t+\tau) + \delta \right)
    \right]\nonumber\\
    \le & \sum_{l=0}^{n_{G}^*(t)} 
    \sum_i  \sum_{\tau=\tau'_l(t) }^{\tau'_{l+1}(t) -1} 
    \left(\lambda_i(\tau) + \delta \right) \left( \hat{E}_{t}\left[\frac{Q_i(\tau'_l(t))}{\|\left(\boldsymbol{Q}(\tau'_l(t))- WJ {\boldsymbol 1} \right)_{+}\|_2}\right]
    + \hat{E}_{t}\left[\frac{W U_{\mathrm{A}}}{\|\boldsymbol{Q}(\tau)\|_2}\right] \right)\nonumber\\
    = & \sum_{l=0}^{n_{G}^*(t)} 
    \sum_i  \sum_{\tau=\tau'_l(t) }^{\tau'_{l+1}(t) -1} 
    \left(\lambda_i(\tau) + \delta \right)  \hat{E}_{t}\left[\frac{Q_i(\tau'_l(t))}{\|\left(\boldsymbol{Q}(\tau'_l(t))- WJ {\boldsymbol 1} \right)_{+}\|_2}\right]\nonumber\\
    & + \sum_{l=0}^{n_{G}^*(t)} 
    \sum_i  \sum_{\tau=\tau'_l(t) }^{\tau'_{l+1}(t) -1} 
    \left(\lambda_i(\tau) + \delta \right)
    \hat{E}_{t}\left[\frac{W U_{\mathrm{A}}}{\|\boldsymbol{Q}(\tau)\|_2}\right] \nonumber\\
    = & \sum_{l=0}^{n_{G}^*(t)} 
    \sum_i \hat{E}_{t}\left[\frac{Q_i(\tau'_l(t))}{\|\left(\boldsymbol{Q}(\tau'_l(t))- WJ {\boldsymbol 1} \right)_{+}\|_2}\right] \sum_{\tau=\tau'_l(t) }^{\tau'_{l+1}(t) -1} 
    \left(\lambda_i(\tau) + \delta \right) \nonumber\\ 
    & + \sum_{\tau=\frac{T}{2}}^{\frac{T}{2}+G_t - 1} 
    \sum_i 
    \left(\lambda_i(t+\tau) + \delta \right)
    \hat{E}_{t}\left[\frac{W U_{\mathrm{A}}}{\|\boldsymbol{Q}(t + \tau)\|_2}\right],
\end{align*}
where the last equality is by transforming the double summations regarding $l$ and $\tau$ back to a single summation, which is similar to the derivation of \eqref{equ:theo-2-arrival-term-2}. Since $\lambda_i(t+\tau)\le U_{\mathrm{A}}$, we have
\begin{align}\label{equ:theo-2-arrival-term-3}
    & \sum_{\tau=\frac{T}{2}}^{\frac{T}{2}+G_t-1} 
    \hat{E}_{t}\left[
    \frac{1}{2\|\boldsymbol{Q}(t+\tau)\|_2}
    \sum_i  2 Q_i(t+\tau)  \left(\lambda_i(t+\tau) + \delta \right)
    \right]\nonumber\\
    \le & \sum_{l=0}^{n_{G}^*(t)} 
    \sum_i \hat{E}_{t}\left[\frac{Q_i(\tau'_l(t))}{\|\left(\boldsymbol{Q}(\tau'_l(t))- WJ {\boldsymbol 1} \right)_{+}\|_2}\right] \sum_{\tau=\tau'_l(t) }^{\tau'_{l+1}(t) -1} 
    \left(\lambda_i(\tau) + \delta \right)\nonumber\\
    & +  I W U_{\mathrm{A}} (U_{\mathrm{A}} + 1 )
    \sum_{\tau=\frac{T}{2}}^{\frac{T}{2}+G_t - 1} 
    \hat{E}_{t}\left[\frac{1}{\|\boldsymbol{Q}(t + \tau)\|_2}\right].
\end{align}
Since ${\boldsymbol \lambda}+\delta {\boldsymbol 1}\in {\mathcal C}(W)$, by the definitions of $\tau'_{l+1}(t)$ and ${\mathcal C}(W)$, we can bound the first term in~\eqref{equ:theo-2-arrival-term-3} as follows
\begin{align*}
    & \sum_{l=0}^{n_{G}^*(t)} 
    \sum_i \hat{E}_{t}\left[\frac{Q_i(\tau'_l(t))}{\|\left(\boldsymbol{Q}(\tau'_l(t))- WJ {\boldsymbol 1} \right)_{+}\|_2}\right] \sum_{\tau=\tau'_l(t) }^{\tau'_{l+1}(t) -1} 
    \left(\lambda_i(\tau) + \delta \right)\nonumber\\
    = & \sum_{l=0}^{n_{G}^*(t)} 
    \sum_i \hat{E}_{t}\left[\frac{Q_i(\tau'_l(t))}{\|\left(\boldsymbol{Q}(\tau'_l(t))- WJ {\boldsymbol 1} \right)_{+}\|_2}\right] \sum_{\tau=\tau'_l(t) }^{\tau'_l(t) + w(\tau'_l(t)) -1} 
    \left(\lambda_i(\tau) + \delta \right)\nonumber\\
    \le & \sum_{l=0}^{n_{G}^*(t)} 
    \sum_i \hat{E}_{t}\left[\frac{Q_i(\tau'_l(t))}{\|\left(\boldsymbol{Q}(\tau'_l(t))- WJ {\boldsymbol 1} \right)_{+}\|_2}\right] \sum_{\tau=\tau'_l(t) }^{\tau'_l(t) + w(\tau'_l(t)) -1} 
    \sum_j \alpha_{i,j}(\tau) \mu_{i,j}(\tau).
\end{align*}
Similarly, transforming the double summations regarding $l$ and $\tau$ back to a single summation, we have
\begin{align}\label{equ:theo-2-arrival-term-4}
    & \sum_{l=0}^{n_{G}^*(t)} 
    \sum_i \hat{E}_{t}\left[\frac{Q_i(\tau'_l(t))}{\|\left(\boldsymbol{Q}(\tau'_l(t))- WJ {\boldsymbol 1} \right)_{+}\|_2}\right] \sum_{\tau=\tau'_l(t) }^{\tau'_{l+1}(t) -1} 
    \left(\lambda_i(\tau) + \delta \right)\nonumber\\
    \le & \sum_{\tau=t+\frac{T}{2}}^{t+\frac{T}{2}+G_t - 1} 
    \hat{E}_{t}\left[\frac{1}{\left\|\left(\boldsymbol{Q}(\tau'_{l(\tau)}(t))- WJ {\boldsymbol 1} \right)_{+}\right\|_2} \sum_i Q_i(\tau'_{l(\tau)}(t))
    \sum_j \alpha_{i,j}(\tau) \mu_{i,j}(\tau) \right],
\end{align}
where $\tau'_{l(\tau)}(t)$ is starting time of the window which $\tau$ is in, i.e., 
\begin{align*}
    \tau'_{l(\tau)}(t)\coloneqq \tau'_{l}(t) \mbox{ where $l$ is such that } \tau\in[\tau'_{l}(t), \tau'_{l+1}(t)-1].
\end{align*}
Since $\sum_i \alpha_{i,j}(\tau) \le 1$, we can further bound~\eqref{equ:theo-2-arrival-term-4} as follows:
\begin{align}\label{equ:theo-2-arrival-term-5}
    & \sum_{l=0}^{n_{G}^*(t)} 
    \sum_i \hat{E}_{t}\left[\frac{Q_i(\tau'_l(t))}{\|\left(\boldsymbol{Q}(\tau'_l(t))- WJ {\boldsymbol 1} \right)_{+}\|_2}\right] \sum_{\tau=\tau'_l(t) }^{\tau'_{l+1}(t) -1} 
    \left(\lambda_i(\tau) + \delta \right)\nonumber\\
    \le & \sum_{\tau=t+\frac{T}{2}}^{t+\frac{T}{2}+G_t - 1} 
    \hat{E}_{t}\left[\frac{1}{\left\|\left(\boldsymbol{Q}(\tau'_{l(\tau)}(t))- WJ {\boldsymbol 1} \right)_{+}\right\|_2} \sum_j \max_i Q_i(\tau'_{l(\tau)}(t)) \mu_{i,j}(\tau)
     \sum_{i'} \alpha_{i',j}(\tau)  \right]\nonumber\\
     \le & \sum_{\tau=t+\frac{T}{2}}^{t+\frac{T}{2}+G_t - 1} 
    \hat{E}_{t}\left[\frac{1}{\left\|\left(\boldsymbol{Q}(\tau'_{l(\tau)}(t))- WJ {\boldsymbol 1} \right)_{+} \right\|_2} \sum_j \max_i Q_i(\tau'_{l(\tau)}(t)) \mu_{i,j}(\tau)  \right]\nonumber\\
    = & \sum_j \sum_{\tau=\frac{T}{2}}^{\frac{T}{2}+G_t - 1} 
    \hat{E}_{t}\left[\frac{1}{\bigl\|\bigl(\boldsymbol{Q}(\tau'_{l(t+\tau)}(t))- WJ {\boldsymbol 1} \bigr)_{+} \bigr\|_2}  \max_i Q_i(\tau'_{l(t+\tau)}(t)) \mu_{i,j}(t+\tau)  \right].
\end{align}
Recall the definition of the mapping $f_j$. $f_j$ maps a time slot to another time slot such that if $y = f_j(x)$ then $y$ is the time slot when server $j$ picked the job that was being served at server $j$ in time slot $x$. If server $j$ was idling in time slot $x$, then $f_j(x)=x$. That is,
$
    f_j(x)\coloneqq\max\{ \tau: \tau\le x, \hat{i}^*_{j}(\tau) = I_j(x) \}.
$
We will use $Q_i(f_j(t+\tau))$ to bound $Q_i(\tau'_{l(t+\tau)}(t))$ in \eqref{equ:theo-2-arrival-term-5}. This process is similar to that in Section~\ref{app:proof-theo-1-bound-arrival} in the proof of Theorem~\ref{theo:1}, but we present the details here for completeness.
Note that for any $\tau\in[\frac{T}{2}, \frac{T}{2} + G_t - 1]$, we have
\begin{align*}
    \tau'_{l(t+\tau)}(t) - f_j(t+\tau) = & \left[\tau'_{l(t+\tau)}(t) - (t+\tau)\right] + \left[(t+\tau) - f_j(t+\tau)\right]\nonumber\\
    \le & (t+\tau) - f_j(t+\tau) \le U_{\mathrm{S}},
\end{align*}
where the first inequality is due to the fact that $\tau'_{l(t+\tau)}(t) \le (t+\tau)$ according to the definition of $\tau'_{l(t+\tau)}(t)$ and the last inequality is by the definition of $f_j(t+\tau)$ and the service time bound $U_{\mathrm{S}}$. Similarly,
\begin{align*}
    \tau'_{l(t+\tau)}(t) - f_j(t+\tau) = & \left[\tau'_{l(t+\tau)}(t) - (t+\tau)\right] + \left[(t+\tau) - f_j(t+\tau)\right]\nonumber\\
    \ge & \tau'_{l(t+\tau)}(t) - (t+\tau) \ge - W,
\end{align*}
where the first inequality is due to the fact that $t+\tau \ge f_j(t+\tau)$ and the last inequality is by the bound of each window. Hence, we have
\begin{align}\label{equ:theo-2-easy-bound}
    \tau'_{l(t+\tau)}(t) - f_j(t+\tau)\in[-W, U_{\mathrm{S}}].
\end{align}
Then by Lemma~\ref{lemma:bounds-on-diff-queue-lens}, we have
\begin{itemize}
    \item If $\tau'_{l(t+\tau)}(t) \ge f_j(t+\tau)$, then 
    \begin{align*}
    Q_i(\tau'_{l(t+\tau)}(t)) \le Q_i(f_j(t+\tau)) + (\tau'_{l(t+\tau)}(t) - f_j(t+\tau)) U_{\mathrm{A}} \le Q_i(f_j(t+\tau)) + U_\mathrm{S} U_{\mathrm{A}}.
    \end{align*}
    \item If $\tau'_{l(t+\tau)}(t) \le f_j(t+\tau)$, then
    \begin{align*}
    Q_i(\tau'_{l(t+\tau)}(t)) \le Q_i(f_j(t+\tau)) + (f_j(t+\tau) - \tau'_{l(t+\tau)}(t)) J \le Q_i(f_j(t+\tau)) + JW.
    \end{align*}
\end{itemize}
Therefore, we have
\begin{align*}
    Q_i(\tau'_{l(t+\tau)}(t)) \le Q_i(f_j(t+\tau)) + \max\{U_\mathrm{S} U_{\mathrm{A}}, JW\}.
\end{align*}
Substituting the above bound into \eqref{equ:theo-2-arrival-term-5}, we have
\begin{align}\label{equ:theo-2-arrival-term-6}
    & \sum_{l=0}^{n_{G}^*(t)} 
    \sum_i \hat{E}_{t}\left[\frac{Q_i(\tau'_l(t))}{\|\left(\boldsymbol{Q}(\tau'_l(t))- WJ {\boldsymbol 1} \right)_{+}\|_2}\right] \sum_{\tau=\tau'_l(t) }^{\tau'_{l+1}(t) -1} 
    \left(\lambda_i(\tau) + \delta \right)\nonumber\\
    \le & \sum_j \sum_{\tau=\frac{T}{2}}^{\frac{T}{2}+G_t - 1} 
    \hat{E}_{t}\left[\frac{1}{\bigl\|\bigl(\boldsymbol{Q}(\tau'_{l(t+\tau)}(t))- WJ {\boldsymbol 1} \bigr)_{+} \bigr\|_2}  \max_i Q_i(f_j(t+\tau)) \mu_{i,j}(t+\tau)  \right]\nonumber\\
    & + J \max\{U_\mathrm{S} U_{\mathrm{A}}, JW\} \sum_{\tau=\frac{T}{2}}^{\frac{T}{2}+G_t - 1} 
    \hat{E}_{t}\left[\frac{1}{\bigl\|\bigl(\boldsymbol{Q}(\tau'_{l(t+\tau)}(t))- WJ {\boldsymbol 1} \bigr)_{+} \bigr\|_2} \right],
\end{align}
where the inequality also uses the fact that $\mu_{i,j}(t+\tau)\le 1$. Substituting \eqref{equ:theo-2-arrival-term-6} into \eqref{equ:theo-2-arrival-term-3} and then into \eqref{equ:theo-2-arrival-term-1}, we have
\begin{align}\label{equ:theo-2-arrival-term-final}
    \eqref{equ:theo-2-arrival-term} \le & \sum_j \sum_{\tau=\frac{T}{2}}^{\frac{T}{2}+G_t - 1} 
    \hat{E}_{t}\left[\frac{1}{\bigl\|\bigl(\boldsymbol{Q}(\tau'_{l(t+\tau)}(t))- WJ {\boldsymbol 1} \bigr)_{+} \bigr\|_2}  \max_i Q_i(f_j(t+\tau)) \mu_{i,j}(t+\tau)  \right]\nonumber\\
    & + J \max\{U_\mathrm{S} U_{\mathrm{A}}, JW\} \sum_{\tau=\frac{T}{2}}^{\frac{T}{2}+G_t - 1} 
    \hat{E}_{t}\left[\frac{1}{\bigl\|\bigl(\boldsymbol{Q}(\tau'_{l(t+\tau)}(t))- WJ {\boldsymbol 1} \bigr)_{+} \bigr\|_2} \right]\nonumber\\
    & + I W U_{\mathrm{A}} (U_{\mathrm{A}} + 1 )
    \sum_{\tau=\frac{T}{2}}^{\frac{T}{2}+G_t - 1} 
    \hat{E}_{t}\left[\frac{1}{\|\boldsymbol{Q}(t + \tau)\|_2}\right] - \delta G_t,
\end{align}
which holds when $\sum_i q_i > \frac{(c_2+1)JT}{\delta} + WIJ$.

\subsection{Bounding the Service Term}

Now we analyze the service term~\eqref{equ:theo-2-service-term}. Let us first fix a server j. We want to lower bound the following per-server service term:
\begin{align*}
    \sum_{\tau=\frac{T}{2}}^{\frac{T}{2}+G_t-1} \hat{E}_{t}\left[\frac{1}{2\|\boldsymbol{Q}(t+\tau)\|_2} 
    \sum_i  2 Q_i(t+\tau) \mathbb{1}_{i,j}(t+\tau) \right].
\end{align*}
The process takes several steps, which are shown in the following.

\subsubsection{Step 1: Adding the Concentration Event}
This step is similar to that in the proof of Theorem~\ref{theo:1}. We present all the details for completeness.

For any $t,j$, define an event ${\mathcal E}'_{t,j}$ as follows:
\begin{align}\label{equ:theo-2-def-conc-event}
    {\mathcal E}'_{t,j}\coloneqq \left\{ \mbox{for all } i, \tau\in\left[ \frac{3T}{8}, \frac{T}{2} + G_t - 1 \right], \left|1/\hat{\mu}_{i,j}(t+\tau)-1/\mu_{i,j}(t+\tau)\right|\le b_{i,j}(t+\tau) \right\}.
\end{align}
We will later prove that this concentration event ${\mathcal E}'_{t,j}$ holds with high probability. Adding an indicator function of the event, we have
\begin{align*}
    & \sum_{\tau=\frac{T}{2}}^{\frac{T}{2}+G_t-1} \hat{E}_{t}\left[\frac{1}{2\|\boldsymbol{Q}(t+\tau)\|_2} 
    \sum_i  2 Q_i(t+\tau) \mathbb{1}_{i,j}(t+\tau) \right]\nonumber\\
    \ge & \sum_{\tau=\frac{T}{2}}^{\frac{T}{2}+G_t-1} \hat{E}_{t}\left[\frac{1}{\|\boldsymbol{Q}(t+\tau)\|_2} 
    \sum_i  Q_i(t+\tau) \mathbb{1}_{i,j}(t+\tau) \mathbb{1}_{{\mathcal E}'_{t,j}} \right].
\end{align*}
Note that $\mathbb{1}_{i,j}(t+\tau)=1$ can happen only on the queue to which server $j$ is scheduled in time slot $t+\tau$, i.e., the queue $I_j(t+\tau)$.
Hence, we have
\begin{align*}
    & \sum_{\tau=\frac{T}{2}}^{\frac{T}{2}+G_t-1} \hat{E}_{t}\left[\frac{1}{2\|\boldsymbol{Q}(t+\tau)\|_2} 
    \sum_i  2 Q_i(t+\tau) \mathbb{1}_{i,j}(t+\tau) \right]\nonumber\\
    \ge & \sum_{\tau=\frac{T}{2}}^{\frac{T}{2}+G_t-1} \hat{E}_{t}\left[\frac{Q_{I_j(t+\tau)}(t+\tau) \mathbb{1}_{I_j(t+\tau),j}(t+\tau) \mathbb{1}_{{\mathcal E}'_{t,j}}}{\|\boldsymbol{Q}(t+\tau)\|_2} 
    \right].
\end{align*}
Recall the definition of $f_j$. By multiplying and dividing the same term, we have
\begin{align}\label{equ:theo-2-service-term-1}
    & \sum_{\tau=\frac{T}{2}}^{\frac{T}{2}+G_t-1} \hat{E}_{t}\left[\frac{1}{2\|\boldsymbol{Q}(t+\tau)\|_2} 
    \sum_i  2 Q_i(t+\tau) \mathbb{1}_{i,j}(t+\tau) \right]\nonumber\\
    \ge & \sum_{\tau=\frac{T}{2}}^{\frac{T}{2}+G_t-1} \hat{E}_{t}\left[\frac{1}{\|\boldsymbol{Q}(t+\tau)\|_2} 
    Q_{I_j(t+\tau)}(t+\tau) \mu_{I_j(t+\tau),j}(f_j(t+\tau))  \frac{\mathbb{1}_{I_j(t+\tau),j}(t+\tau)}{\mu_{I_j(t+\tau),j}(f_j(t+\tau))}
    \mathbb{1}_{{\mathcal E}'_{t,j}}
    \right].
\end{align}

\subsubsection{Step 2: Bounding the Product of Queue Length and Service Rate}
This step is similar to that in the proof of Theorem~\ref{theo:1}. We present all the details for completeness.

We want to lower bound the term $Q_{I_j(t+\tau)}(t+\tau) \mu_{I_j(t+\tau),j}(f_j(t+\tau))$ in~\eqref{equ:theo-2-service-term-1}. The following analysis in this subsection is under the concentration event ${\mathcal E}'_{t,j}$. 
Since $U_{\mathrm{S}}\le \frac{T}{8}$, we have
$f_j(t+\tau)\ge t+\tau -U_{\mathrm{S}}\ge t+\frac{T}{2} -U_{\mathrm{S}} \ge t+\frac{3T}{8}.$
Also note that
$f_j(t+\tau)\le t+\tau \le t + \frac{T}{2} + G_t -1.$
Hence, we have
\begin{align}\label{equ:theo-2-range-of-f_j}
    f_j(t+\tau)\in \left[t + \frac{3T}{8}, t + \frac{T}{2} + G_t -1\right].
\end{align}
Define
\begin{align*}
    \bar{\mu}_{I_j(t+\tau),j}(f_j(t+\tau))
    \coloneqq &
    \frac{1}{\max\left\{\frac{1}{\hat{\mu}_{I_j(t+\tau),j}(f_j(t+\tau))}-b_{I_j(t+\tau),j}(f_j(t+\tau)), 1\right\}}\nonumber\\
    \underline{\mu}_{I_j(t+\tau),j}(f_j(t+\tau))
    \coloneqq &
    \frac{1}{\frac{1}{\hat{\mu}_{I_j(t+\tau),j}(f_j(t+\tau))}+b_{I_j(t+\tau),j}(f_j(t+\tau))}
\end{align*}
By~\eqref{equ:theo-2-range-of-f_j} and the definition of the concentration event ${\mathcal E}'_{t,j}$ in~\eqref{equ:theo-2-def-conc-event}, we have
\begin{align*}
    \frac{1}{\hat{\mu}_{I_j(t+\tau),j}(f_j(t+\tau))}-b_{I_j(t+\tau),j}(f_j(t+\tau)) \le & \frac{1}{\mu_{I_j(t+\tau),j}(f_j(t+\tau))}\nonumber\\
    \frac{1}{\hat{\mu}_{I_j(t+\tau),j}(f_j(t+\tau))} + b_{I_j(t+\tau),j}(f_j(t+\tau)) \ge & \frac{1}{\mu_{I_j(t+\tau),j}(f_j(t+\tau))}.
\end{align*}
Therefore, combining the above inequalities and the fact that $\frac{1}{\mu_{I_j(t+\tau),j}(f_j(t+\tau))}\ge 1$, we have
\begin{align*}
    \underline{\mu}_{I_j(t+\tau),j}(f_j(t+\tau)) 
    \le \mu_{I_j(t+\tau),j}(f_j(t+\tau)) 
    \le \bar{\mu}_{I_j(t+\tau),j}(f_j(t+\tau)).
\end{align*}
Then we have
\begin{align}\label{equ:theo-2-q-mu-term-lower-bound}
    & Q_{I_j(t+\tau)}(t+\tau) \mu_{I_j(t+\tau),j}(f_j(t+\tau)) \nonumber\\
    = &
    Q_{I_j(t+\tau)}(t+\tau) \bar{\mu}_{I_j(t+\tau),j}(f_j(t+\tau)) \nonumber\\
    & + Q_{I_j(t+\tau)}(t+\tau) \left( \mu_{I_j(t+\tau),j}(f_j(t+\tau)) - \bar{\mu}_{I_j(t+\tau),j}(f_j(t+\tau))\right)\nonumber\\
    \ge & Q_{I_j(t+\tau)}(t+\tau) \bar{\mu}_{I_j(t+\tau),j}(f_j(t+\tau)) \nonumber\\
    & + Q_{I_j(t+\tau)}(t+\tau) \left( \underline{\mu}_{I_j(t+\tau),j}(f_j(t+\tau)) - \bar{\mu}_{I_j(t+\tau),j}(f_j(t+\tau))\right).
\end{align}
Note that
\begin{align}\label{equ:theo-2-mu-difference-bound}
    & \underline{\mu}_{I_j(t+\tau),j}(f_j(t+\tau)) - \bar{\mu}_{I_j(t+\tau),j}(f_j(t+\tau))\nonumber\\
    = & \frac{1}{\frac{1}{\hat{\mu}_{I_j(t+\tau),j}(f_j(t+\tau))}+b_{I_j(t+\tau),j}(f_j(t+\tau))} - \frac{1}{\max\left\{\frac{1}{\hat{\mu}_{I_j(t+\tau),j}(f_j(t+\tau))}-b_{I_j(t+\tau),j}(f_j(t+\tau)), 1\right\}}\nonumber\\
    = & \frac{\max\left\{\frac{1}{\hat{\mu}_{I_j(t+\tau),j}(f_j(t+\tau))}-b_{I_j(t+\tau),j}(f_j(t+\tau)), 1\right\} - \left(\frac{1}{\hat{\mu}_{I_j(t+\tau),j}(f_j(t+\tau))}+b_{I_j(t+\tau),j}(f_j(t+\tau))\right)}{\left(\frac{1}{\hat{\mu}_{I_j(t+\tau),j}(f_j(t+\tau))}+b_{I_j(t+\tau),j}(f_j(t+\tau))\right) \max\left\{\frac{1}{\hat{\mu}_{I_j(t+\tau),j}(f_j(t+\tau))}-b_{I_j(t+\tau),j}(f_j(t+\tau)), 1\right\}}\nonumber\\
    \ge & \frac{-2b_{I_j(t+\tau),j}(f_j(t+\tau))}{\left(\frac{1}{\hat{\mu}_{I_j(t+\tau),j}(f_j(t+\tau))}+b_{I_j(t+\tau),j}(f_j(t+\tau))\right) \max\left\{\frac{1}{\hat{\mu}_{I_j(t+\tau),j}(f_j(t+\tau))}-b_{I_j(t+\tau),j}(f_j(t+\tau)), 1\right\}}\nonumber\\
    \ge & \frac{-2b_{I_j(t+\tau),j}(f_j(t+\tau))}{\frac{1}{\hat{\mu}_{I_j(t+\tau),j}(f_j(t+\tau))}+b_{I_j(t+\tau),j}(f_j(t+\tau)) }\nonumber\\
    \ge & -2b_{I_j(t+\tau),j}(f_j(t+\tau)),
\end{align}
where the last inequality uses the fact that $\frac{1}{\hat{\mu}_{i,j}(t)} \ge 1$ for any $i,j,t$.
Also note that $\underline{\mu}_{I_j(t+\tau),j}(f_j(t+\tau)) - \bar{\mu}_{I_j(t+\tau),j}(f_j(t+\tau))\ge -1$.
Hence, combining~\eqref{equ:theo-2-mu-difference-bound} and \eqref{equ:theo-2-q-mu-term-lower-bound}, we have
\begin{align}\label{equ:theo-2-q-mu-term-2}
     & Q_{I_j(t+\tau)}(t+\tau) \mu_{I_j(t+\tau),j}(f_j(t+\tau)) \nonumber\\
     \ge & Q_{I_j(t+\tau)}(t+\tau) \bar{\mu}_{I_j(t+\tau),j}(f_j(t+\tau)) 
     - Q_{I_j(t+\tau)}(t+\tau) \min\{2b_{I_j(t+\tau),j}(f_j(t+\tau)), 1\}.
\end{align}
Note that by Lemma~\ref{lemma:bounds-on-diff-queue-lens} and the fact that $t+\tau - f_j(t+\tau)\le U_{\mathrm{S}}$, we have
$Q_{I_j(t+\tau)}(t+\tau) \ge Q_{I_j(t+\tau)}(f_j(t+\tau)) - JU_{\mathrm{S}}.$ Then we have
\begin{align}\label{equ:theo-2-q-mu-term-1}
    Q_{I_j(t+\tau)}(t+\tau) \bar{\mu}_{I_j(t+\tau),j}(f_j(t+\tau)) \ge Q_{I_j(t+\tau)}(f_j(t+\tau)) \bar{\mu}_{I_j(t+\tau),j}(f_j(t+\tau))  - JU_{\mathrm{S}},
\end{align}
where we use the fact that $\bar{\mu}_{I_j(t+\tau),j}(f_j(t+\tau))\le 1$.
By Line~\ref{alg:line:max-weight} in Algorithm~\ref{alg:1} and the definition of $\bar{\mu}_{I_j(t+\tau),j}(f_j(t+\tau))$, we have
\begin{align}\label{equ:theo-2-Q-mu-term-4}
    Q_{I_j(t+\tau)}(f_j(t+\tau)) \bar{\mu}_{I_j(t+\tau),j}(f_j(t+\tau))
    = \max_i  \frac{Q_{i}(f_j(t+\tau))}{\max\left\{\frac{1}{\hat{\mu}_{i,j}(f_j(t+\tau))}-b_{i,j}(f_j(t+\tau)), 1\right\}}.
\end{align}
Combining~\eqref{equ:theo-2-q-mu-term-2},~\eqref{equ:theo-2-q-mu-term-1}, and~\eqref{equ:theo-2-Q-mu-term-4}, we have
\begin{align}\label{equ:theo-2-Q-mu-term-5}
    &Q_{I_j(t+\tau)}(t+\tau) \mu_{I_j(t+\tau),j}(f_j(t+\tau))\nonumber\\  
    \ge &  \max_i  \frac{Q_{i}(f_j(t+\tau))}{\max\left\{\frac{1}{\hat{\mu}_{i,j}(f_j(t+\tau))}-b_{i,j}(f_j(t+\tau)), 1\right\}} - Q_{I_j(t+\tau)}(t+\tau) \min\{2b_{I_j(t+\tau),j}(f_j(t+\tau)), 1\} - JU_{\mathrm{S}} \nonumber\\
    \ge & \max_i  Q_{i}(f_j(t+\tau)) \mu_{i,j}(f_j(t+\tau)) - Q_{I_j(t+\tau)}(t+\tau) \min\{2b_{I_j(t+\tau),j}(f_j(t+\tau)), 1\} - JU_{\mathrm{S}},
\end{align}
where the last inequality uses the fact that $\max\left\{\frac{1}{\hat{\mu}_{i,j}(f_j(t+\tau))}-b_{i,j}(f_j(t+\tau)), 1\right\} \le \frac{1}{\mu_{i,j}(f_j(t+\tau))}$, which is based on the concentration event ${\mathcal E}'_{t,j}$ and the fact that $\frac{1}{\mu_{i,j}(f_j(t+\tau))}\ge 1$.
Substituting~\eqref{equ:theo-2-Q-mu-term-5} into~\eqref{equ:theo-2-service-term-1}, we have
\begin{align}
    & \sum_{\tau=\frac{T}{2}}^{\frac{T}{2}+G_t-1} \hat{E}_{t}\left[\frac{1}{2\|\boldsymbol{Q}(t+\tau)\|_2} 
    \sum_i  2 Q_i(t+\tau) \mathbb{1}_{i,j}(t+\tau) \right]\nonumber\\
    \ge & \sum_{\tau=\frac{T}{2}}^{\frac{T}{2}+G_t-1} \hat{E}_{t}\left[\frac{1}{\|\boldsymbol{Q}(t+\tau)\|_2} 
    Q_{I_j(t+\tau)}(t+\tau) \mu_{I_j(t+\tau),j}(f_j(t+\tau))  \frac{\mathbb{1}_{I_j(t+\tau),j}(t+\tau)}{\mu_{I_j(t+\tau),j}(f_j(t+\tau))}
    \mathbb{1}_{{\mathcal E}'_{t,j}}
    \right]\nonumber\\
    \ge & \hat{E}_{t}\Biggl[ \sum_{\tau=\frac{T}{2}}^{\frac{T}{2}+G_t-1} 
    \frac{1}{\|\boldsymbol{Q}(t+\tau)\|_2}
    \biggl(
    \max_i  Q_{i}(f_j(t+\tau)) \mu_{i,j}(f_j(t+\tau)) \nonumber\\
    &\qquad  \qquad \qquad \quad - Q_{I_j(t+\tau)}(t+\tau) \min\{2b_{I_j(t+\tau),j}(f_j(t+\tau)), 1\} - JU_{\mathrm{S}} \biggr)
    \frac{\mathbb{1}_{I_j(t+\tau),j}(t+\tau)}{\mu_{I_j(t+\tau),j}(f_j(t+\tau))} \mathbb{1}_{{\mathcal E}'_{t,j}}\Biggr]
    \nonumber\\
    \ge & 
    \hat{E}_{t}\left[ \sum_{\tau=\frac{T}{2}}^{\frac{T}{2}+G_t-1} \frac{1}{\|\boldsymbol{Q}(t+\tau)\|_2}  \max_i  Q_{i}(f_j(t+\tau)) \mu_{i,j}(f_j(t+\tau))
    \frac{\mathbb{1}_{I_j(t+\tau),j}(t+\tau)}{\mu_{I_j(t+\tau),j}(f_j(t+\tau))} \mathbb{1}_{{\mathcal E}'_{t,j}}\right]\label{equ:theo-2-sum-service-time-term-1}\\
    & - U_{\mathrm{S}} \hat{E}_{t}\Biggl[ \sum_{\tau=\frac{T}{2}}^{\frac{T}{2}+G_t-1} \frac{1}{\|\boldsymbol{Q}(t+\tau)\|_2} Q_{I_j(t+\tau)}(t+\tau) \min\{2b_{I_j(t+\tau),j}(f_j(t+\tau)), 1\} \mathbb{1}_{I_j(t+\tau),j}(t+\tau) \Biggr]\label{equ:theo-2-ucb-summation-term-1}\\
    & - JU^2_{\mathrm{S}} \hat{E}_{t}\left[ \sum_{\tau=\frac{T}{2}}^{\frac{T}{2}+G_t-1} \frac{1}{\|\boldsymbol{Q}(t+\tau)\|_2} \right],\label{equ:theo-2-service-term-last-term}
\end{align}
where the last inequality is due to the fact that
$\frac{1}{\mu_{I_j(t+\tau),j}(f_j(t+\tau))} \le U_{\mathrm{S}}$.

\subsubsection{Step 3: Bounding the Sum of Queue-Length-Weighted UCB Bonuses}

We first look at the term~\eqref{equ:theo-2-ucb-summation-term-1}.
We will use $\tilde{b}_{i,j}(t)\coloneqq \min\{2b_{i,j}(t), 1\}$ for ease of notation.

Recall the definition of the waiting queue.
Note that if server $j$ is idling in time slot $f_j(t+\tau)$, then $t+\tau = f_j(t+\tau)$ and the waiting queue $\tilde{Q}_{I_j(t+\tau)}(t+\tau)=0$. Hence, we have
$\tilde{Q}_{I_j(t+\tau)}(t+\tau) = \tilde{Q}_{I_j(t+\tau)}(t+\tau)\eta_j(f_j(t+\tau)).$
Also note that
$0\le Q_{I_j(t+\tau)}(t+\tau) - \tilde{Q}_{I_j(t+\tau)}(t+\tau) \le J$
by definition. Hence, we have
\begin{align*}
    Q_{I_j(t+\tau)}(t+\tau) \le & \tilde{Q}_{I_j(t+\tau)}(t+\tau) + J\nonumber\\
    = & \tilde{Q}_{I_j(t+\tau)}(t+\tau)\eta_j(f_j(t+\tau)) + J\nonumber\\
    \le & Q_{I_j(t+\tau)}(t+\tau)\eta_j(f_j(t+\tau)) + J.
\end{align*}
Hence, we have
\begin{align}\label{equ:theo-2-ucb-summation-term-2}
    & \hat{E}_{t}\Biggl[ \sum_{\tau=\frac{T}{2}}^{\frac{T}{2}+G_t-1} \frac{1}{\|\boldsymbol{Q}(t+\tau)\|_2} Q_{I_j(t+\tau)}(t+\tau) \tilde{b}_{I_j(t+\tau),j}(f_j(t+\tau)) \mathbb{1}_{I_j(t+\tau),j}(t+\tau) \Biggr]\nonumber\\
    \le & \hat{E}_{t}\Biggl[ \sum_{\tau=\frac{T}{2}}^{\frac{T}{2}+G_t-1} \frac{Q_{I_j(t+\tau)}(t+\tau)}{\|\boldsymbol{Q}(t+\tau)\|_2}  \tilde{b}_{I_j(t+\tau),j}(f_j(t+\tau)) \eta_j(f_j(t+\tau)) \mathbb{1}_{I_j(t+\tau),j}(t+\tau) \Biggr]\nonumber\\
    & + J \hat{E}_{t}\Biggl[ \sum_{\tau=\frac{T}{2}}^{\frac{T}{2}+G_t-1} \frac{1}{\|\boldsymbol{Q}(t+\tau)\|_2}\Biggr]\nonumber\\
    = & \hat{E}_{t}\Biggl[ \sum_i  \sum_{\tau=\frac{T}{2}}^{\frac{T}{2}+G_t-1} \frac{Q_{i}(t+\tau)}{\|\boldsymbol{Q}(t+\tau)\|_2}  \tilde{b}_{i,j}(f_j(t+\tau)) \eta_j(f_j(t+\tau)) \mathbb{1}_{i,j}(t+\tau) \mathbb{1}_{I_j(t+\tau) = i} \Biggr]\nonumber\\
    & + J \hat{E}_{t}\Biggl[ \sum_{\tau=\frac{T}{2}}^{\frac{T}{2}+G_t-1} \frac{1}{\|\boldsymbol{Q}(t+\tau)\|_2}\Biggr].
\end{align}
Note that
\begin{align}\label{equ:c-s-q}
    \frac{Q_{i}(t+\tau)}{\|\boldsymbol{Q}(t+\tau)\|_2} = \frac{Q_i(t+\tau)}{\sqrt{\sum_{i'} Q^2_{i'}(t+\tau)}} \le \frac{Q_i(t+\tau)}{\sqrt{Q^2_i(t+\tau)}} = 1
\end{align}
Substituting~\eqref{equ:c-s-q} into~\eqref{equ:theo-2-ucb-summation-term-2}, we have
\begin{align}\label{equ:theo-2-ucb-summation-term-3}
    & \hat{E}_{t}\Biggl[ \sum_{\tau=\frac{T}{2}}^{\frac{T}{2}+G_t-1} \frac{1}{\|\boldsymbol{Q}(t+\tau)\|_2} Q_{I_j(t+\tau)}(t+\tau) \tilde{b}_{I_j(t+\tau),j}(f_j(t+\tau)) \mathbb{1}_{I_j(t+\tau),j}(t+\tau) \Biggr]\nonumber\\
    \le & \hat{E}_{t}\Biggl[ \sum_i  \sum_{\tau=\frac{T}{2}}^{\frac{T}{2}+G_t-1}  \tilde{b}_{i,j}(f_j(t+\tau)) \eta_j(f_j(t+\tau)) \mathbb{1}_{i,j}(t+\tau) \mathbb{1}_{I_j(t+\tau) = i} \Biggr] + J \hat{E}_{t}\Biggl[ \sum_{\tau=\frac{T}{2}}^{\frac{T}{2}+G_t-1} \frac{1}{\|\boldsymbol{Q}(t+\tau)\|_2}\Biggr].
\end{align}
Next we bound the sum-of-UCB term $\sum_{\tau=\frac{T}{2}}^{\frac{T}{2}+G_t-1}  \tilde{b}_{i,j}(f_j(t+\tau)) \eta_j(f_j(t+\tau)) \mathbb{1}_{i,j}(t+\tau) \mathbb{1}_{I_j(t+\tau) = i}$.
Considering the event $\{\tilde{b}_{i,j}(f_j(t+\tau)) \le \frac{\delta}{2IJU_{\mathrm{S}}}\}$, we further have
\begin{align}\label{equ:theo-2-ucb-summation-term-4}
    & \sum_{\tau=\frac{T}{2}}^{\frac{T}{2}+G_t-1}  \tilde{b}_{i,j}(f_j(t+\tau)) \eta_j(f_j(t+\tau)) \mathbb{1}_{i,j}(t+\tau) \mathbb{1}_{I_j(t+\tau) = i} \nonumber\\
    = & \sum_{\tau=\frac{T}{2}}^{\frac{T}{2}+G_t-1}  \tilde{b}_{i,j}(f_j(t+\tau)) \eta_j(f_j(t+\tau)) \mathbb{1}_{i,j}(t+\tau) \mathbb{1}_{I_j(t+\tau) = i} \mathbb{1}_{\tilde{b}_{i,j}(f_j(t+\tau)) \le \frac{\delta}{2IJU_{\mathrm{S}}}}\nonumber\\
    & + \sum_{\tau=\frac{T}{2}}^{\frac{T}{2}+G_t-1}  \tilde{b}_{i,j}(f_j(t+\tau)) \eta_j(f_j(t+\tau)) \mathbb{1}_{i,j}(t+\tau) \mathbb{1}_{I_j(t+\tau) = i} \mathbb{1}_{\tilde{b}_{i,j}(f_j(t+\tau)) > \frac{\delta}{2IJU_{\mathrm{S}}}}\nonumber\\
    \le & \frac{\delta G_t }{2IJU_{\mathrm{S}}} 
    + \sum_{\tau=\frac{T}{2}}^{\frac{T}{2}+G_t-1}  \tilde{b}_{i,j}(f_j(t+\tau)) \eta_j(f_j(t+\tau)) \mathbb{1}_{i,j}(t+\tau) \mathbb{1}_{I_j(t+\tau) = i} \mathbb{1}_{\tilde{b}_{i,j}(f_j(t+\tau)) > \frac{\delta}{2IJU_{\mathrm{S}}}}\nonumber\\
    \le & \frac{\delta G_t }{2IJU_{\mathrm{S}}} 
    + \sum_{\tau=\frac{T}{2}}^{\frac{T}{2}+G_t-1}  \tilde{b}_{i,j}(f_j(t+\tau)) \eta_j(f_j(t+\tau)) \mathbb{1}_{i,j}(t+\tau) \mathbb{1}_{I_j(t+\tau) = i} 
    \mathbb{1}_{\hat{N}_{i,j}(f_j(t+\tau)) < \frac{64 I^2 J^2 U_{\mathrm{S}}^4 \log \frac{1}{1-\gamma}}{\delta^2}},
\end{align}
where the last inequality holds since
\begin{align*}
    \tilde{b}_{i,j}(f_j(t+\tau)) = \min\{2b_{i,j}(f_j(t+\tau)), 1\} = & \min\left\{4 U_{\mathrm{S}} \sqrt{\frac{\log \left(\sum_{\tau'=0}^{f_j(t+\tau)-1} \gamma^{\tau'}\right)}{\hat{N}_{i,j}(f_j(t+\tau))}}
    , 1\right\}\nonumber\\
    \le & 4 U_{\mathrm{S}} \sqrt{\frac{\log \frac{1}{1-\gamma}}{\hat{N}_{i,j}(f_j(t+\tau))}}
\end{align*}
according to Line~\ref{alg:line:b} in Algorithm~\ref{alg:1}. Note that by Lemma~\ref{lemma:prop-G}, we have $\frac{T}{2}+G_t\le \frac{(c_2+1)T}{\delta}$. Hence, we can apply Lemma~\ref{lemma:eij} with $\tau_l=\frac{T}{2}$, $\tau_h=\frac{T}{2}+G_t-1$, $L=\frac{(c_2+1)T}{\delta}$, $U_N = \frac{64 I^2 J^2 U_{\mathrm{S}}^4 \log \frac{1}{1-\gamma}}{\delta^2}$ to obtain
\begin{align}\label{equ:theo-2-ucb-summation-term-5}
    & \sum_{\tau=\frac{T}{2}}^{\frac{T}{2}+G_t-1} \tilde{b}_{i,j}(f_j(t+\tau)) \eta_j(f_j(t+\tau))
    \mathbb{1}_{i,j}(t+\tau) \mathbb{1}_{I_j(t+\tau)=i}
    \mathbb{1}_{\hat{N}_{i,j}(f_j(t+\tau)) < \frac{64 I^2 J^2 U_{\mathrm{S}}^4 \log \frac{1}{1-\gamma}}{\delta^2}}\nonumber\\
    \le & \left\lceil \frac{2(c_2+1)T(1-\gamma)}{\delta} \right\rceil \left( 1 + \frac{128 I J U^3_{\mathrm{S}} \log \frac{1}{1-\gamma} }{\delta}\right)\nonumber\\
    \le & \frac{1044 (c_2+1) I J U^3_{\mathrm{S}} \log^2 \frac{1}{1-\gamma}}{\delta^2},
\end{align}
where the last inequality is by the definition of $T$, $\gamma \ge 1-\frac{1}{1+e^{1.5}}$, and $c_2= 5(IU_{\mathrm{A}}+J)\ge 5$. Substituting \eqref{equ:theo-2-ucb-summation-term-5} into \eqref{equ:theo-2-ucb-summation-term-4} and then into \eqref{equ:theo-2-ucb-summation-term-3}, we have
\begin{align*}
    & \hat{E}_{t}\Biggl[ \sum_{\tau=\frac{T}{2}}^{\frac{T}{2}+G_t-1} \frac{1}{\|\boldsymbol{Q}(t+\tau)\|_2} Q_{I_j(t+\tau)}(t+\tau) \tilde{b}_{I_j(t+\tau),j}(f_j(t+\tau)) \mathbb{1}_{I_j(t+\tau),j}(t+\tau) \Biggr]\nonumber\\
    \le & \frac{\delta G_t }{2JU_{\mathrm{S}}} 
    + \frac{1044 (c_2+1) I^2 J U^3_{\mathrm{S}} \log^2 \frac{1}{1-\gamma}}{\delta^2} + J \hat{E}_{t}\Biggl[ \sum_{\tau=\frac{T}{2}}^{\frac{T}{2}+G_t-1} \frac{1}{\|\boldsymbol{Q}(t+\tau)\|_2}\Biggr].
\end{align*}
Substituting the above bound into~\eqref{equ:theo-2-ucb-summation-term-1}, we have
\begin{align}\label{equ:theo-2-ucb-summation-final}
    \eqref{equ:theo-2-ucb-summation-term-1} \ge &  - \frac{\delta G_t }{2J} 
    - \frac{1044 (c_2+1) I^2 J U^4_{\mathrm{S}} \log^2 \frac{1}{1-\gamma}}{\delta^2} - J  U_{\mathrm{S}} \hat{E}_{t}\Biggl[ \sum_{\tau=\frac{T}{2}}^{\frac{T}{2}+G_t-1} \frac{1}{\|\boldsymbol{Q}(t+\tau)\|_2}\Biggr]\nonumber\\
    \ge & - \frac{\delta G_t }{2J} 
    - \frac{6264 (IU_{\mathrm{A}}+J) I^2 J U^4_{\mathrm{S}} \log^2 \frac{1}{1-\gamma}}{\delta^2} - J  U_{\mathrm{S}} \hat{E}_{t}\Biggl[ \sum_{\tau=\frac{T}{2}}^{\frac{T}{2}+G_t-1} \frac{1}{\|\boldsymbol{Q}(t+\tau)\|_2}\Biggr],
\end{align}
where the last inequality is by $c_2=5(IU_{\mathrm{A}}+J)$.

\subsubsection{Step 4: Bounding the Weighted Sum of Job Completion Indicators}

We next look at the term~\eqref{equ:theo-2-sum-service-time-term-1}:
\begin{align*}
    \hat{E}_{t}\left[ \sum_{\tau=\frac{T}{2}}^{\frac{T}{2}+G_t-1} \frac{1}{\|\boldsymbol{Q}(t+\tau)\|_2}  \max_i  Q_{i}(f_j(t+\tau)) \mu_{i,j}(f_j(t+\tau))
    \frac{\mathbb{1}_{I_j(t+\tau),j}(t+\tau)}{\mu_{I_j(t+\tau),j}(f_j(t+\tau))} \mathbb{1}_{{\mathcal E}'_{t,j}}\right].
\end{align*}
Since $\mathbb{1}_{{\mathcal E}'_{t,j}} + \mathbb{1}_{{\mathcal E}^{'\mathrm{c}}_{t,j}} = 1$, we have
\begin{align}\label{equ:job-compl-ind-1}
    & \hat{E}_{t}\left[ \sum_{\tau=\frac{T}{2}}^{\frac{T}{2}+G_t-1} \frac{1}{\|\boldsymbol{Q}(t+\tau)\|_2}  \max_i  Q_{i}(f_j(t+\tau)) \mu_{i,j}(f_j(t+\tau))
    \frac{\mathbb{1}_{I_j(t+\tau),j}(t+\tau)}{\mu_{I_j(t+\tau),j}(f_j(t+\tau))} \mathbb{1}_{{\mathcal E}'_{t,j}}\right]\nonumber\\
    = & \hat{E}_{t}\left[ \sum_{\tau=\frac{T}{2}}^{\frac{T}{2}+G_t-1} \frac{1}{\|\boldsymbol{Q}(t+\tau)\|_2}  \max_i  Q_{i}(f_j(t+\tau)) \mu_{i,j}(f_j(t+\tau))
    \frac{\mathbb{1}_{I_j(t+\tau),j}(t+\tau)}{\mu_{I_j(t+\tau),j}(f_j(t+\tau))} \right]\nonumber\\
    & - \hat{E}_{t}\left[ \sum_{\tau=\frac{T}{2}}^{\frac{T}{2}+G_t-1} \frac{1}{\|\boldsymbol{Q}(t+\tau)\|_2}  \max_i  Q_{i}(f_j(t+\tau)) \mu_{i,j}(f_j(t+\tau))
    \frac{\mathbb{1}_{I_j(t+\tau),j}(t+\tau)}{\mu_{I_j(t+\tau),j}(f_j(t+\tau))} \mathbb{1}_{{\mathcal E}^{'\mathrm{c}}_{t,j}}\right].
\end{align}
Since $\mu_{i,j}(\tau)\le 1$ and $\frac{1}{\mu_{i,j}(\tau)}\le U_{\mathrm{S}}$ for any $i,j,\tau$, we have 
\begin{align}\label{equ:job-compl-ind-2}
    & \sum_{\tau=\frac{T}{2}}^{\frac{T}{2}+G_t-1} \frac{1}{\|\boldsymbol{Q}(t+\tau)\|_2}  \max_i  Q_{i}(f_j(t+\tau)) \mu_{i,j}(f_j(t+\tau))
    \frac{\mathbb{1}_{I_j(t+\tau),j}(t+\tau)}{\mu_{I_j(t+\tau),j}(f_j(t+\tau))}\nonumber\\
    \le & U_{\mathrm{S}}\sum_{\tau=\frac{T}{2}}^{\frac{T}{2}+G_t-1} \frac{1}{\|\boldsymbol{Q}(t+\tau)\|_2}  \max_i  Q_{i}(f_j(t+\tau))\nonumber\\
    \le & U_{\mathrm{S}}\sum_{\tau=\frac{T}{2}}^{\frac{T}{2}+G_t-1} \frac{1}{\|\boldsymbol{Q}(t+\tau)\|_2}  \left(\max_i  Q_{i}(t+\tau) + J U_{\mathrm{S}}\right)\nonumber\\
    = & U_{\mathrm{S}} \sum_{\tau=\frac{T}{2}}^{\frac{T}{2}+G_t-1} \frac{1}{\|\boldsymbol{Q}(t+\tau)\|_2}  \max_i  Q_{i}(t+\tau) + J U^2_{\mathrm{S}} \sum_{\tau=\frac{T}{2}}^{\frac{T}{2}+G_t-1} \frac{1}{\|\boldsymbol{Q}(t+\tau)\|_2}\nonumber\\
    \le & U_{\mathrm{S}} G_t + J U^2_{\mathrm{S}} \sum_{\tau=\frac{T}{2}}^{\frac{T}{2}+G_t-1} \frac{1}{\|\boldsymbol{Q}(t+\tau)\|_2},
\end{align}
where the second inequality uses Lemma~\ref{lemma:bounds-on-diff-queue-lens} and the last inequality is due to the fact that $\max_i Q_i(t+\tau) \le \|\boldsymbol{Q}(t+\tau)\|_2$. Hence, combining \eqref{equ:job-compl-ind-1} and \eqref{equ:job-compl-ind-2} and using the notation $\hat{P}_{t}\left(\cdot\right)$, we obtain
\begin{align}\label{equ:job-compl-ind-3}
    & \hat{E}_{t}\left[ \sum_{\tau=\frac{T}{2}}^{\frac{T}{2}+G_t-1} \frac{1}{\|\boldsymbol{Q}(t+\tau)\|_2}  \max_i  Q_{i}(f_j(t+\tau)) \mu_{i,j}(f_j(t+\tau))
    \frac{\mathbb{1}_{I_j(t+\tau),j}(t+\tau)}{\mu_{I_j(t+\tau),j}(f_j(t+\tau))} \mathbb{1}_{{\mathcal E}'_{t,j}}\right]\nonumber\\
    \ge & \hat{E}_{t}\left[ \sum_{\tau=\frac{T}{2}}^{\frac{T}{2}+G_t-1} \frac{1}{\|\boldsymbol{Q}(t+\tau)\|_2}  \max_i  Q_{i}(f_j(t+\tau)) \mu_{i,j}(f_j(t+\tau))
    \frac{\mathbb{1}_{I_j(t+\tau),j}(t+\tau)}{\mu_{I_j(t+\tau),j}(f_j(t+\tau))} \right]\nonumber\\
    & -  U_{\mathrm{S}} G_t \hat{P}_{t}\left( {\mathcal E}^{'\mathrm{c}}_{t,j} \right) 
    - J U^2_{\mathrm{S}} \hat{E}_{t}\left[ \sum_{\tau=\frac{T}{2}}^{\frac{T}{2}+G_t-1} \frac{1}{\|\boldsymbol{Q}(t+\tau)\|_2}
    \right].
\end{align}
The following lemma shows that the event ${\mathcal E}'_{t,j}$ which is defined in~\eqref{equ:theo-2-def-conc-event} holds with high probability.
\begin{lemma}\label{lemma:concentration-theo-2}
    Let Assumption~\ref{assump:mu-2} holds. Suppose $c_1=2$, $1-\frac{1}{1+e^{1.5}} \le \gamma < 1$, and $T \ge 8U_{\mathrm{S}}$.
    For any $t,j$,
    \[
    \hat{P}_{t}\left( {\mathcal E}^{'\mathrm{c}}_{t,j} \right)\coloneqq \Pr ({\mathcal E}_{t,j}^{'\mathrm c} \left| \mathbf{Q}(t)={\mathbf q}, \mathbf{H}(t)={\mathbf h}  \right. ) \le \frac{516 I(I U_{\mathrm{A}} + J) \left(1-\gamma\right)^{1.5}}{\delta}.
    \]
\end{lemma}
Proof of this Lemma is similar to that of Lemma~\ref{lemma:concentration} and can be found in Section~\ref{app:proof-lemma-concentration-theo-2}. Hence, by Lemma~\ref{lemma:concentration-theo-2}, Lemma~\ref{lemma:prop-G}, and \eqref{equ:job-compl-ind-3}, we have
\begin{align}\label{equ:theo-2-weighted-indicator}
    & \hat{E}_{t}\left[ \sum_{\tau=\frac{T}{2}}^{\frac{T}{2}+G_t-1} \frac{1}{\|\boldsymbol{Q}(t+\tau)\|_2}  \max_i  Q_{i}(f_j(t+\tau)) \mu_{i,j}(f_j(t+\tau))
    \frac{\mathbb{1}_{I_j(t+\tau),j}(t+\tau)}{\mu_{I_j(t+\tau),j}(f_j(t+\tau))} \mathbb{1}_{{\mathcal E}'_{t,j}}\right]\nonumber\\
    \ge & \hat{E}_{t}\left[ \sum_{\tau=\frac{T}{2}}^{\frac{T}{2}+G_t-1} \frac{1}{\|\boldsymbol{Q}(t+\tau)\|_2}  \max_i  Q_{i}(f_j(t+\tau)) \mu_{i,j}(f_j(t+\tau))
    \frac{\mathbb{1}_{I_j(t+\tau),j}(t+\tau)}{\mu_{I_j(t+\tau),j}(f_j(t+\tau))} \right]\nonumber\\
    & -  \frac{516 I(I U_{\mathrm{A}} + J) (c_2+\frac{1}{2})  U_{\mathrm{S}} T(1-\gamma)^{1.5} }{\delta^2} 
    - J U^2_{\mathrm{S}} \hat{E}_{t}\left[ \sum_{\tau=\frac{T}{2}}^{\frac{T}{2}+G_t-1} \frac{1}{\|\boldsymbol{Q}(t+\tau)\|_2}
    \right].
\end{align}
Note that by the definition of $T$ and the fact that $(1-\gamma)^{0.5}\log \frac{1}{1-\gamma} \le \frac{2}{e}$, we have
\begin{align}\label{equ:basic-ineq}
    T (1-\gamma)^{1.5} = 4 (1-\gamma)^{0.5}\log \frac{1}{1-\gamma} \le \frac{8}{e}.
\end{align}
Then from \eqref{equ:theo-2-weighted-indicator} and \eqref{equ:basic-ineq}, we have
\begin{align*}
    & \hat{E}_{t}\left[ \sum_{\tau=\frac{T}{2}}^{\frac{T}{2}+G_t-1} \frac{1}{\|\boldsymbol{Q}(t+\tau)\|_2}  \max_i  Q_{i}(f_j(t+\tau)) \mu_{i,j}(f_j(t+\tau))
    \frac{\mathbb{1}_{I_j(t+\tau),j}(t+\tau)}{\mu_{I_j(t+\tau),j}(f_j(t+\tau))} \mathbb{1}_{{\mathcal E}'_{t,j}}\right]\nonumber\\
    \ge & \hat{E}_{t}\left[ \sum_{\tau=\frac{T}{2}}^{\frac{T}{2}+G_t-1} \frac{1}{\|\boldsymbol{Q}(t+\tau)\|_2}  \max_i  Q_{i}(f_j(t+\tau)) \mu_{i,j}(f_j(t+\tau))
    \frac{\mathbb{1}_{I_j(t+\tau),j}(t+\tau)}{\mu_{I_j(t+\tau),j}(f_j(t+\tau))} \right]\nonumber\\
    & -  \frac{1519 I(I U_{\mathrm{A}} + J) (c_2+\frac{1}{2})  U_{\mathrm{S}} }{\delta^2} 
    - J U^2_{\mathrm{S}} \hat{E}_{t}\left[ \sum_{\tau=\frac{T}{2}}^{\frac{T}{2}+G_t-1} \frac{1}{\|\boldsymbol{Q}(t+\tau)\|_2}
    \right].
\end{align*}
From the condition~\eqref{equ:condition-delta-2} in Theorem~\ref{theo:2}, $\gamma \ge 1-\frac{1}{1+e^{1.5}}$, and $c_2=5(IU_{\mathrm{A}} + J)$, we have $\frac{1519 I(I U_{\mathrm{A}} + J) (c_2+\frac{1}{2})  U_{\mathrm{S}} }{\delta^2} \\\le \frac{(I U_{\mathrm{A}} + J)T}{16 J}$. Then
\begin{align}\label{equ:job-compl-ind-4}
    & \hat{E}_{t}\left[ \sum_{\tau=\frac{T}{2}}^{\frac{T}{2}+G_t-1} \frac{1}{\|\boldsymbol{Q}(t+\tau)\|_2}  \max_i  Q_{i}(f_j(t+\tau)) \mu_{i,j}(f_j(t+\tau))
    \frac{\mathbb{1}_{I_j(t+\tau),j}(t+\tau)}{\mu_{I_j(t+\tau),j}(f_j(t+\tau))} \mathbb{1}_{{\mathcal E}'_{t,j}}\right]\nonumber\\
    \ge & \hat{E}_{t}\left[ \sum_{\tau=\frac{T}{2}}^{\frac{T}{2}+G_t-1} \frac{1}{\|\boldsymbol{Q}(t+\tau)\|_2}  \max_i  Q_{i}(f_j(t+\tau)) \mu_{i,j}(f_j(t+\tau))
    \frac{\mathbb{1}_{I_j(t+\tau),j}(t+\tau)}{\mu_{I_j(t+\tau),j}(f_j(t+\tau))} \right]\nonumber\\
    & -  \frac{(I U_{\mathrm{A}} + J)T}{16J} 
    - J U^2_{\mathrm{S}} \hat{E}_{t}\left[ \sum_{\tau=\frac{T}{2}}^{\frac{T}{2}+G_t-1} \frac{1}{\|\boldsymbol{Q}(t+\tau)\|_2}
    \right].
\end{align}
By Lemma~\ref{lemma:bounds-on-diff-queue-lens}, we have $\|\boldsymbol{Q}(t+\tau)\|_2\le \|\boldsymbol{Q}(f_j(t+\tau))+U_{\mathrm{A}}U_{\mathrm{S}}\boldsymbol{1}\|_2$. Hence, for the first term of \eqref{equ:job-compl-ind-4}, we have
\begin{align*}
    & \hat{E}_{t}\left[ \sum_{\tau=\frac{T}{2}}^{\frac{T}{2}+G_t-1} \frac{1}{\| \boldsymbol{Q}(t+\tau) \|_2}  \max_i  Q_{i}(f_j(t+\tau)) \mu_{i,j}(f_j(t+\tau))
    \frac{\mathbb{1}_{I_j(t+\tau),j}(t+\tau)}{\mu_{I_j(t+\tau),j}(f_j(t+\tau))} \right] \nonumber\\
    \ge & \hat{E}_{t}\left[ \sum_{\tau=\frac{T}{2}}^{\frac{T}{2}+G_t-1} \frac{1}{\|\boldsymbol{Q}(f_j(t+\tau))+U_{\mathrm{A}}U_{\mathrm{S}}\boldsymbol{1}\|_2}  \max_i  Q_{i}(f_j(t+\tau)) \mu_{i,j}(f_j(t+\tau))
    \frac{\mathbb{1}_{I_j(t+\tau),j}(t+\tau)}{\mu_{I_j(t+\tau),j}(f_j(t+\tau))} \right].
\end{align*}
For any $\tau$, define 
\begin{align*}
    v'_j(\tau)\coloneqq \frac{1}{\|\boldsymbol{Q}(\tau)+U_{\mathrm{A}}U_{\mathrm{S}}\boldsymbol{1}\|_2}  \max_i  Q_{i}(\tau) \mu_{i,j}(\tau).
\end{align*}
Then 
\begin{align}\label{equ:job-compl-ind-5}
    & \hat{E}_{t}\left[ \sum_{\tau=\frac{T}{2}}^{\frac{T}{2}+G_t-1} \frac{1}{\| \boldsymbol{Q}(t+\tau) \|_2}  \max_i  Q_{i}(f_j(t+\tau)) \mu_{i,j}(f_j(t+\tau))
    \frac{\mathbb{1}_{I_j(t+\tau),j}(t+\tau)}{\mu_{I_j(t+\tau),j}(f_j(t+\tau))} \right] \nonumber\\
    \ge & \hat{E}_{t}\left[ \sum_{\tau=\frac{T}{2}}^{\frac{T}{2}+G_t-1} 
    v'_j(f_j(t+\tau))
    \frac{\mathbb{1}_{I_j(t+\tau),j}(t+\tau)}{\mu_{I_j(t+\tau),j}(f_j(t+\tau))} \right].
\end{align}
Note that 
\begin{align}\label{equ:theo-2-bound-v-j}
    v'_j(\tau)\le 1
\end{align}
for any $\tau$ since $\max_i  Q_{i}(\tau) \mu_{i,j}(\tau) \le \max_i  Q_{i}(\tau) \le \|\boldsymbol{Q}(\tau)\|_2 \le \|\boldsymbol{Q}(\tau)+U_{\mathrm{A}}U_{\mathrm{S}}\boldsymbol{1}\|_2$.
We can write the term~\eqref{equ:job-compl-ind-5} in a different form by summing over the time slots in which the jobs start, i.e.,
\begin{align}\label{equ:theo-2-sum-service-time-temp1}
    \hat{E}_{t}\left[ \sum_{\tau=\frac{T}{2}}^{\frac{T}{2}+G_t-1} 
    v'_j(f_j(t+\tau))
    \frac{\mathbb{1}_{I_j(t+\tau),j}(t+\tau)}{\mu_{I_j(t+\tau),j}(f_j(t+\tau))} \right]
    \ge \hat{E}_{t}\Biggl[ \sum_{\tau=\frac{T}{2}}^{\frac{T}{2}+G_t-1}  \frac{v'_j(t+\tau)\mathbb{1}_{\hat{i}^*_j(t+\tau)\neq 0}}{\mu_{\hat{i}^*_j(t+\tau),j}(t+\tau)}\Biggr] - U_{\mathrm{S}},
\end{align}
where the inequality holds since the last job starting before $t+\frac{T}{2}+G_t$ may not finish before $t+\frac{T}{2}+G_t$ and the first job finishing at or after $t+\frac{T}{2}$ may not start at or after $t+\frac{T}{2}$, and we also use~\eqref{equ:theo-2-bound-v-j} and the fact that $1/\mu_{i,j}(\tau)\le U_{\mathrm{S}}$ for any $i,j,\tau$.
Next we look at the first term in~\eqref{equ:theo-2-sum-service-time-temp1}.
Let $\mathbb{1}_{\mathrm{idling}}(j, t)\coloneqq 1- \eta_j(t)$, which is equal to $1$ when server $j$ is idling. Dividing the sum into two cases based on whether server $j$ is idling or non-idling,
we have
\begin{align}\label{equ:theo-2-idling-non-idling}
    & \hat{E}_{t}\Biggl[ \sum_{\tau=\frac{T}{2}}^{\frac{T}{2}+G_t-1}   \frac{v'_j(t+\tau)\mathbb{1}_{\hat{i}^*_j(t+\tau)\neq 0}}{\mu_{\hat{i}^*_j(t+\tau),j}(t+\tau)}\Biggr] \nonumber\\
    = &
     \hat{E}_{t}\Biggl[ \sum_{\tau=\frac{T}{2}}^{\frac{T}{2}+G_t-1}  \frac{v'_j(t+\tau)\mathbb{1}_{\hat{i}^*_j(t+\tau)\neq 0}\eta_j(t+\tau)}{\mu_{\hat{i}^*_j(t+\tau),j}(t+\tau)}\Biggr] + \hat{E}_{t}\Biggl[ \sum_{\tau=\frac{T}{2}}^{\frac{T}{2}+G_t-1}  \frac{v'_j(t+\tau)\mathbb{1}_{\hat{i}^*_j(t+\tau)\neq 0} \mathbb{1}_{\mathrm{idling}}(j, t+\tau)}{\mu_{\hat{i}^*_j(t+\tau),j}(t+\tau)}\Biggr] \nonumber\\
    \ge & 
    \hat{E}_{t}\Biggl[ \sum_{\tau=\frac{T}{2}}^{\frac{T}{2}+G_t-1}  \frac{v'_j(t+\tau)\mathbb{1}_{\hat{i}^*_j(t+\tau)\neq 0}\eta_j(t+\tau)}{\mu_{\hat{i}^*_j(t+\tau),j}(t+\tau)}\Biggr] + \hat{E}_{t}\Biggl[ \sum_{\tau=\frac{T}{2}}^{\frac{T}{2}+G_t-1}  v'_j(t+\tau)\mathbb{1}_{\hat{i}^*_j(t+\tau)\neq 0} \mathbb{1}_{\mathrm{idling}}(j, t+\tau) \Biggr],
\end{align}
where the last inequality is due to the fact that $\mu_{i,j}(t)\le 1$ for any $i,j,t$. Note that $E[S_{i,j}(t+\tau)] = 1/\mu_{i,j}(t+\tau)$ and $\hat{E}_{t}[S_{i,j}(t+\tau)] = E[S_{i,j}(t+\tau)]$ since $S_{i,j}(t+\tau)$ is independent of $\boldsymbol{Q}(t)$ and $\boldsymbol{H}(t)$. Hence, we have
\begin{align}\label{equ:theo-2-equalities}
    & \hat{E}_{t}\Biggl[ \sum_{\tau=\frac{T}{2}}^{\frac{T}{2}+G_t-1}  \frac{v'_j(t+\tau)\mathbb{1}_{\hat{i}^*_j(t+\tau)\neq 0}\eta_j(t+\tau)}{\mu_{\hat{i}^*_j(t+\tau),j}(t+\tau)}\Biggr] \nonumber\\
    = & \sum_{i=1}^{I} \hat{E}_{t}\Biggl[ \sum_{\tau=\frac{T}{2}}^{\frac{T}{2}+G_t-1}  \frac{v'_j(t+\tau)\mathbb{1}_{\hat{i}^*_j(t+\tau)=i}\eta_j(t+\tau)}{\mu_{i,j}(t+\tau)}\Biggr]\nonumber\\
    = & \sum_{i=1}^{I} \hat{E}_{t}\Biggl[ \sum_{\tau=\frac{T}{2}}^{\frac{T}{2}+G_t-1}  v'_j(t+\tau)\mathbb{1}_{\hat{i}^*_j(t+\tau)=i}\eta_j(t+\tau)
    \hat{E}_{t}[S_{i,j}(t+\tau)]\Biggr]\nonumber\\
    = & \sum_{i=1}^{I} \hat{E}_{t}\Biggl[ \sum_{\tau=\frac{T}{2}}^{\frac{T}{2}+G_t-1}  v'_j(t+\tau)\mathbb{1}_{\hat{i}^*_j(t+\tau)=i}\eta_j(t+\tau)
    \hat{E}_{t} [S_{i,j}(t+\tau)\left| \boldsymbol{Q}(t+\tau), \boldsymbol{H}(t+\tau), \boldsymbol{A}(t+\tau) \right.] \Biggr]\nonumber\\
    = & \sum_{i=1}^{I} \hat{E}_{t}\Biggl[ \sum_{\tau=\frac{T}{2}}^{\frac{T}{2}+G_t-1}  
    \hat{E}_{t} [v'_j(t+\tau)\mathbb{1}_{\hat{i}^*_j(t+\tau)=i}\eta_j(t+\tau) S_{i,j}(t+\tau)\left| \boldsymbol{Q}(t+\tau), \boldsymbol{H}(t+\tau), \boldsymbol{A}(t+\tau) \right.] \Biggr]\nonumber\\
    = & \sum_{i=1}^{I} \sum_{\tau=\frac{T}{2}}^{\frac{T}{2}+G_t-1}  
    \hat{E}_{t}\left[ \hat{E}_{t} [v'_j(t+\tau)\mathbb{1}_{\hat{i}^*_j(t+\tau)=i}\eta_j(t+\tau) S_{i,j}(t+\tau)\left| \boldsymbol{Q}(t+\tau), \boldsymbol{H}(t+\tau), \boldsymbol{A}(t+\tau) \right.] \right]\nonumber\\
    = &  \sum_{i=1}^{I} \sum_{\tau=\frac{T}{2}}^{\frac{T}{2}+G_t-1} 
    \hat{E}_{t}[v'_j(t+\tau)\mathbb{1}_{\hat{i}^*_j(t+\tau)=i}\eta_j(t+\tau) S_{i,j}(t+\tau)]\nonumber\\
    = & \sum_{\tau=\frac{T}{2}}^{\frac{T}{2}+G_t-1} 
    \hat{E}_{t}[v'_j(t+\tau)\mathbb{1}_{\hat{i}^*_j(t+\tau)\neq 0}\eta_j(t+\tau) S_{\hat{i}^*_j(t+\tau),j}(t+\tau)],
\end{align}
where the third equality is due to the independence between $S_{i,j}(t+\tau)$ and $\boldsymbol{Q}(t+\tau), \boldsymbol{H}(t+\tau), \boldsymbol{A}(t+\tau)$, the fourth equality is due to the fact that $v'_j(t+\tau),\mathbb{1}_{\hat{i}^*_j(t+\tau)=i},\eta_j(t+\tau)$ are fully determined by $\boldsymbol{Q}(t+\tau), \boldsymbol{H}(t+\tau), \boldsymbol{A}(t+\tau)$, and the sixth equality is by the law of iterated expectation. Note that in these derivations we view $\hat{E}_{t}$ as the expectation under the probability measure $\hat{P}_{t}$.
Substituting~\eqref{equ:theo-2-equalities} into~\eqref{equ:theo-2-idling-non-idling}, we have
\begin{align}\label{equ:theo-2-sum-service-time-temp3}
    & \hat{E}_{t}\Biggl[ \sum_{\tau=\frac{T}{2}}^{\frac{T}{2}+G_t-1}   \frac{v'_j(t+\tau)\mathbb{1}_{\hat{i}^*_j(t+\tau)\neq 0}}{\mu_{\hat{i}^*_j(t+\tau),j}(t+\tau)}\Biggr] \nonumber\\
    \ge & \sum_{\tau=\frac{T}{2}}^{\frac{T}{2}+G_t-1} 
    \hat{E}_{t}[v'_j(t+\tau)\mathbb{1}_{\hat{i}^*_j(t+\tau)\neq 0}\eta_j(t+\tau) S_{\hat{i}^*_j(t+\tau),j}(t+\tau)]\nonumber\\
    & + \hat{E}_{t}\Biggl[ \sum_{\tau=\frac{T}{2}}^{\frac{T}{2}+G_t-1} v'_j(t+\tau)\mathbb{1}_{\hat{i}^*_j(t+\tau)\neq 0} \mathbb{1}_{\mathrm{idling}}(j, t+\tau) \Biggr]\nonumber\\
    = & \hat{E}_{t}\Biggl[ \sum_{\tau=\frac{T}{2}}^{\frac{T}{2}+G_t-1}  
    v'_j(t+\tau)\mathbb{1}_{\hat{i}^*_j(t+\tau)\neq 0} \left(\eta_j(t+\tau) S_{\hat{i}^*_j(t+\tau),j}(t+\tau) + \mathbb{1}_{\mathrm{idling}}(j, t+\tau)\right)\Biggr].
\end{align}
Note that the term $\eta_j(t+\tau) S_{\hat{i}^*_j(t+\tau),j}(t+\tau) + \mathbb{1}_{\mathrm{idling}}(j, t+\tau)$ is the actual time that server $j$ spends on the queue $\hat{i}^*_j(t+\tau)$. Hence, \eqref{equ:theo-2-sum-service-time-temp3} can be rewritten using $f_j$ in the following way:
\begin{align}\label{equ:theo-2-sum-service-time-temp4}
    & \hat{E}_{t}\Biggl[ \sum_{\tau=\frac{T}{2}}^{\frac{T}{2}+G_t-1}  
    v'_j(t+\tau)\mathbb{1}_{\hat{i}^*_j(t+\tau)\neq 0} \left(\eta_j(t+\tau) S_{\hat{i}^*_j(t+\tau),j}(t+\tau) + \mathbb{1}_{\mathrm{idling}}(j, t+\tau)\right)\Biggr]\nonumber\\
    = & \hat{E}_{t}\left[ \sum_{\tau=\tau_{\mathrm{start}}}^{\tau_{\mathrm{end}}} v'_j(f_j(t+\tau)) \right],
\end{align}
where $t + \tau_{\mathrm{start}}$ is the starting (or idling) time of the first schedule that starts at or after $t + \frac{T}{2}$ and $t + \tau_{\mathrm{end}}$ is the finishing (or idling) time of the last schedule that starts at or before $t+\frac{T}{2}+G_t-1$.
By~\eqref{equ:theo-2-bound-v-j} and the facts that $t+\tau_{\mathrm{start}}<t + \frac{T}{2} + U_{\mathrm{S}}$ and $t+\tau_{\mathrm{end}}\ge t + \frac{T}{2}+G_t - 1$, we have
\begin{align}\label{equ:theo-2-sum-service-time-temp5}
    & \hat{E}_{t}\left[ \sum_{\tau=t_{\mathrm{start}}}^{t_{\mathrm{end}}} v'_j(f_j(t+\tau)) \right] \nonumber\\
    \ge & \hat{E}_{t}\left[ \sum_{\tau=\frac{T}{2}}^{\frac{T}{2}+G_t-1} v'_j(f_j(t+\tau)) \right] - U_{\mathrm{S}}  \nonumber\\
    \ge & \hat{E}_{t}\left[\sum_{\tau=\frac{T}{2}}^{\frac{T}{2}+G_t-1} 
    \frac{1}{\|\boldsymbol{Q}(f_j(t+\tau))+U_{\mathrm{A}}U_{\mathrm{S}}\boldsymbol{1}\|_2}  \max_i  Q_{i}(f_j(t+\tau)) \left(\mu_{i,j}(t+\tau) - \frac{1}{T^p}\right) \right]  - U_{\mathrm{S}}  \nonumber\\
    = & \hat{E}_{t}\left[\sum_{\tau=\frac{T}{2}}^{\frac{T}{2}+G_t-1} 
    \frac{1}{\|\boldsymbol{Q}(f_j(t+\tau))+U_{\mathrm{A}}U_{\mathrm{S}}\boldsymbol{1}\|_2}
    \max_i \left(Q_i(f_j(t+\tau)) \mu_{i,j}(t+\tau)-\frac{1}{T^p} Q_i(f_j(t+\tau)) \right)\right]  - U_{\mathrm{S}}  \nonumber\\
    \ge & \hat{E}_{t}\left[\sum_{\tau=\frac{T}{2}}^{\frac{T}{2}+G_t-1} 
    \frac{1}{\|\boldsymbol{Q}(f_j(t+\tau))+U_{\mathrm{A}}U_{\mathrm{S}}\boldsymbol{1}\|_2}
    \left( \max_i Q_i(f_j(t+\tau)) \mu_{i,j}(t+\tau)-\frac{1}{T^p} \|\boldsymbol{Q}(f_j(t+\tau))\|_2 \right) \right]  - U_{\mathrm{S}}  \nonumber\\
    = &  \hat{E}_{t}\left[\sum_{\tau=\frac{T}{2}}^{\frac{T}{2}+G_t-1} 
    \frac{1}{\|\boldsymbol{Q}(f_j(t+\tau))+U_{\mathrm{A}}U_{\mathrm{S}}\boldsymbol{1}\|_2}
    \max_i Q_i(f_j(t+\tau)) \mu_{i,j}(t+\tau)\right] \nonumber\\
    & - \frac{1}{T^p} \hat{E}_{t}\left[ \sum_{\tau=\frac{T}{2}}^{\frac{T}{2}+G_t-1} \frac{\|\boldsymbol{Q}(f_j(t+\tau))\|_2}{\|\boldsymbol{Q}(f_j(t+\tau))+U_{\mathrm{A}}U_{\mathrm{S}}\boldsymbol{1}\|_2} \right] - U_{\mathrm{S}}\nonumber\\
    \ge & \hat{E}_{t}\left[\sum_{\tau=\frac{T}{2}}^{\frac{T}{2}+G_t-1} 
    \frac{1}{\|\boldsymbol{Q}(f_j(t+\tau))+U_{\mathrm{A}}U_{\mathrm{S}}\boldsymbol{1}\|_2}
    \max_i Q_i(f_j(t+\tau)) \mu_{i,j}(t+\tau)\right] - \frac{G_t}{T^p} - U_{\mathrm{S}}\nonumber\\
    \ge & \hat{E}_{t}\left[\sum_{\tau=\frac{T}{2}}^{\frac{T}{2}+G_t-1} 
    \frac{1}{\|\boldsymbol{Q}(f_j(t+\tau))+U_{\mathrm{A}}U_{\mathrm{S}}\boldsymbol{1}\|_2}
    \max_i Q_i(f_j(t+\tau)) \mu_{i,j}(t+\tau)\right] - \frac{11(I U_{\mathrm{A}} + J)T^{1-p}}{2\delta} - U_{\mathrm{S}},
\end{align}
where the second inequality uses the fact that $t+\tau-f_j(t+\tau)\le U_{\mathrm{S}}$ and Assumption~\ref{assump:mu-1} (2), the third inequality is due to the fact that $Q_i(\tau) \le \|\boldsymbol{Q}(\tau)\|_2$ for any $\tau,i$, and the last inequality is by Lemma~\ref{lemma:prop-G} and $c_2=5(I U_{\mathrm{A}} + J)$.

Combining~\eqref{equ:theo-2-sum-service-time-temp3},~\eqref{equ:theo-2-sum-service-time-temp4}, and~\eqref{equ:theo-2-sum-service-time-temp5}, we have
\begin{align}\label{equ:theo-2-sum-service-time-temp6}
    & \hat{E}_{t}\Biggl[ \sum_{\tau=\frac{T}{2}}^{\frac{T}{2}+G_t-1}   \frac{v'_j(t+\tau)\mathbb{1}_{\hat{i}^*_j(t+\tau)\neq 0}}{\mu_{\hat{i}^*_j(t+\tau),j}(t+\tau)}\Biggr] \nonumber\\
    \ge & \hat{E}_{t}\left[\sum_{\tau=\frac{T}{2}}^{\frac{T}{2}+G_t-1} 
    \frac{1}{\|\boldsymbol{Q}(f_j(t+\tau))+U_{\mathrm{A}}U_{\mathrm{S}}\boldsymbol{1}\|_2}
    \max_i Q_i(f_j(t+\tau)) \mu_{i,j}(t+\tau)\right] - \frac{11(I U_{\mathrm{A}} + J)T^{1-p}}{2\delta} - U_{\mathrm{S}}.
\end{align}
Substituting~\eqref{equ:theo-2-sum-service-time-temp6} into~\eqref{equ:theo-2-sum-service-time-temp1} and then into~\eqref{equ:job-compl-ind-5} and then into \eqref{equ:job-compl-ind-4}, we have
\begin{align}\label{equ:theo-2-sum-service-time-temp7}
    \eqref{equ:theo-2-sum-service-time-term-1} = & \hat{E}_{t}\left[ \sum_{\tau=\frac{T}{2}}^{\frac{T}{2}+G_t-1} \frac{1}{\|\boldsymbol{Q}(t+\tau)\|_2}  \max_i  Q_{i}(f_j(t+\tau)) \mu_{i,j}(f_j(t+\tau))
    \frac{\mathbb{1}_{I_j(t+\tau),j}(t+\tau)}{\mu_{I_j(t+\tau),j}(f_j(t+\tau))} \mathbb{1}_{{\mathcal E}'_{t,j}}\right]\nonumber\\
    \ge & \hat{E}_{t}\left[\sum_{\tau=\frac{T}{2}}^{\frac{T}{2}+G_t-1} 
    \frac{1}{\|\boldsymbol{Q}(f_j(t+\tau))+U_{\mathrm{A}}U_{\mathrm{S}}\boldsymbol{1}\|_2}
    \max_i Q_i(f_j(t+\tau)) \mu_{i,j}(t+\tau)\right] - \frac{11(I U_{\mathrm{A}} + J)T^{1-p}}{2\delta} 
    \nonumber\\
    & - 2 U_{\mathrm{S}} -  \frac{(I U_{\mathrm{A}} + J)T}{16 J} - J U^2_{\mathrm{S}} \hat{E}_{t}\left[ \sum_{\tau=\frac{T}{2}}^{\frac{T}{2}+G_t-1} \frac{1}{\|\boldsymbol{Q}(t+\tau)\|_2}
     \right].
\end{align}

Combining \eqref{equ:theo-2-sum-service-time-term-1}, \eqref{equ:theo-2-ucb-summation-term-1}, \eqref{equ:theo-2-service-term-last-term}, \eqref{equ:theo-2-ucb-summation-final}, and \eqref{equ:theo-2-sum-service-time-temp7}, we obtain the bound for the per-server service term as follows:
\begin{align*}
    & \sum_{\tau=\frac{T}{2}}^{\frac{T}{2}+G_t-1} \hat{E}_{t}\left[\frac{1}{2\|\boldsymbol{Q}(t+\tau)\|_2} 
    \sum_i  2 Q_i(t+\tau) \mathbb{1}_{i,j}(t+\tau) \right]\nonumber\\
    \ge & \hat{E}_{t}\left[\sum_{\tau=\frac{T}{2}}^{\frac{T}{2}+G_t-1} 
    \frac{1}{\|\boldsymbol{Q}(f_j(t+\tau))+U_{\mathrm{A}}U_{\mathrm{S}}\boldsymbol{1}\|_2}
    \max_i Q_i(f_j(t+\tau)) \mu_{i,j}(t+\tau)\right] - \frac{11(I U_{\mathrm{A}} + J)T^{1-p}}{2\delta} - 2 U_{\mathrm{S}}
    \nonumber\\
    & -  \frac{(I U_{\mathrm{A}} + J)T}{16J} - (2 J U^2_{\mathrm{S}} + J  U_{\mathrm{S}})\hat{E}_{t}\left[ \sum_{\tau=\frac{T}{2}}^{\frac{T}{2}+G_t-1} \frac{1}{\|\boldsymbol{Q}(t+\tau)\|_2}
     \right]
    - \frac{\delta G_t }{2J} 
    - \frac{6264(I U_{\mathrm{A}} + J) I^2 J U^4_{\mathrm{S}} \log^2 \frac{1}{1-\gamma}}{\delta^2}.
\end{align*}
Substituting the above bound into \eqref{equ:theo-2-service-term}, we obtain the bound for the service term as follows:
\begin{align}\label{equ:theo-2-servic-term-final-bound}
    \eqref{equ:theo-2-service-term} \le & - \sum_j \hat{E}_{t}\left[\sum_{\tau=\frac{T}{2}}^{\frac{T}{2}+G_t-1} 
    \frac{1}{\|\boldsymbol{Q}(f_j(t+\tau))+U_{\mathrm{A}}U_{\mathrm{S}}\boldsymbol{1}\|_2}
    \max_i Q_i(f_j(t+\tau)) \mu_{i,j}(t+\tau)\right]\nonumber\\
    & + \frac{11(I U_{\mathrm{A}} + J)JT^{1-p}}{2\delta} + 2 J U_{\mathrm{S}}
    +  \frac{(I U_{\mathrm{A}} + J)T}{16}
    + J^2 U_{\mathrm{S}}(2  U_{\mathrm{S}} +  1 )\hat{E}_{t}\left[ \sum_{\tau=\frac{T}{2}}^{\frac{T}{2}+G_t-1} \frac{1}{\|\boldsymbol{Q}(t+\tau)\|_2}
     \right]\nonumber\\
    & + \frac{\delta G_t }{2}  + \frac{6264(I U_{\mathrm{A}} + J) I^2 J^2 U^4_{\mathrm{S}} \log^2 \frac{1}{1-\gamma}}{\delta^2}.
\end{align}

\subsection{Deriving Negative Lyapunov Drift}

In this subsection, we combine the bounds of the arrival term and the service term to obtain a negative Lyapunov drift.
    
Combining \eqref{equ:theo-2-arrival-term}, \eqref{equ:theo-2-service-term}, \eqref{equ:theo-2-constant-term}, \eqref{equ:theo-2-arrival-term-final}, and \eqref{equ:theo-2-servic-term-final-bound}, we have
\begin{align}\label{equ:theo-2-drift-4}
    & E\left[ L'(t+\frac{T}{2}+G_t) - L'(t+\frac{T}{2}) \left| \boldsymbol{Q}(t)=\boldsymbol{q}, \boldsymbol{H}(t)=\boldsymbol{h}\right.\right]\nonumber\\
    \le & \sum_j \sum_{\tau=\frac{T}{2}}^{\frac{T}{2}+G_t - 1} 
    \hat{E}_{t}\Biggl[
    \Biggl(
    \frac{1}{\bigl\|\bigl(\boldsymbol{Q}(\tau'_{l(t+\tau)}(t))- WJ {\boldsymbol 1} \bigr)_{+} \bigr\|_2}  
    - \frac{1}{\|\boldsymbol{Q}(f_j(t+\tau))+U_{\mathrm{A}}U_{\mathrm{S}}\boldsymbol{1}\|_2}
    \Biggr)\nonumber\\
    & \qquad \qquad \qquad \max_i Q_i(f_j(t+\tau)) \mu_{i,j}(t+\tau)  \Biggr]\nonumber\\
    & + J \max\{U_\mathrm{S} U_{\mathrm{A}}, JW\} \sum_{\tau=\frac{T}{2}}^{\frac{T}{2}+G_t - 1} 
    \hat{E}_{t}\left[\frac{1}{\bigl\|\bigl(\boldsymbol{Q}(\tau'_{l(t+\tau)}(t))- WJ {\boldsymbol 1} \bigr)_{+} \bigr\|_2} \right]\nonumber\\
    & +\left[ J^2 U_{\mathrm{S}}(2  U_{\mathrm{S}} +  1 ) + I W U_{\mathrm{A}} (U_{\mathrm{A}} + 1 ) + \frac{IU_{\mathrm{A}}^2 + J^2 + IJ^2}{2}\right] 
    \sum_{\tau=\frac{T}{2}}^{\frac{T}{2}+G_t - 1} 
    \hat{E}_{t}\left[\frac{1}{\|\boldsymbol{Q}(t + \tau)\|_2}\right] \nonumber\\
    & - \delta G_t
    + \frac{11(I U_{\mathrm{A}} + J)JT^{1-p}}{2\delta} + 2 J U_{\mathrm{S}}
    + \frac{(I U_{\mathrm{A}} + J)T}{16}
    + \frac{\delta G_t }{2} 
    + \frac{6264(I U_{\mathrm{A}} + J) I^2 J^2 U^4_{\mathrm{S}} \log^2 \frac{1}{1-\gamma}}{\delta^2},
\end{align}
which holds when $\sum_i q_i > \frac{(c_2+1)JT}{\delta} + WIJ$.
Next we want to bound the first term in \eqref{equ:theo-2-drift-4}.
Recall from \eqref{equ:theo-2-easy-bound} that 
$
\tau'_{l(t+\tau)}(t) - f_j(t+\tau)\in[-W, U_{\mathrm{S}}].
$
Then by Lemma~\ref{lemma:bounds-on-diff-queue-lens}, we have
\begin{itemize}
    \item If $\tau'_{l(t+\tau)}(t) \ge f_j(t+\tau)$, then 
    \begin{align*}
    Q_i(\tau'_{l(t+\tau)}(t)) \ge Q_i(f_j(t+\tau)) - (\tau'_{l(t+\tau)}(t) - f_j(t+\tau)) J \ge Q_i(f_j(t+\tau)) - J U_\mathrm{S}.
    \end{align*}
    \item If $\tau'_{l(t+\tau)}(t) \le f_j(t+\tau)$, then
    \begin{align*}
    Q_i(\tau'_{l(t+\tau)}(t)) \ge Q_i(f_j(t+\tau)) - (f_j(t+\tau) - \tau'_{l(t+\tau)}(t)) U_{\mathrm{A}} \ge Q_i(f_j(t+\tau)) - U_{\mathrm{A}} W.
    \end{align*}
\end{itemize}
Therefore, we have
\begin{align*}
    Q_i(\tau'_{l(t+\tau)}(t)) \ge Q_i(f_j(t+\tau)) - \max\{J U_\mathrm{S}, U_{\mathrm{A}} W\}.
\end{align*}
Hence, using the above inequality to bound the first term in \eqref{equ:theo-2-drift-4}, we have
\begin{align*}
    & \frac{1}{\bigl\|\bigl(\boldsymbol{Q}(\tau'_{l(t+\tau)}(t))- WJ {\boldsymbol 1} \bigr)_{+} \bigr\|_2}  
    - \frac{1}{\|\boldsymbol{Q}(f_j(t+\tau))+U_{\mathrm{A}}U_{\mathrm{S}}\boldsymbol{1}\|_2}\nonumber\\
    = & \frac{\|\boldsymbol{Q}(f_j(t+\tau))+U_{\mathrm{A}}U_{\mathrm{S}}\boldsymbol{1}\|_2 - \bigl\|\bigl(\boldsymbol{Q}(\tau'_{l(t+\tau)}(t))- WJ {\boldsymbol 1} \bigr)_{+} \bigr\|_2}{\bigl\|\bigl(\boldsymbol{Q}(\tau'_{l(t+\tau)}(t))- WJ {\boldsymbol 1} \bigr)_{+} \bigr\|_2 \|\boldsymbol{Q}(f_j(t+\tau))+U_{\mathrm{A}}U_{\mathrm{S}}\boldsymbol{1}\|_2}
    \nonumber\\
    \le & \frac{\|\boldsymbol{Q}(f_j(t+\tau))+U_{\mathrm{A}}U_{\mathrm{S}}\boldsymbol{1}\|_2 
    - \left\|\left(\boldsymbol{Q}(f_j(t+\tau)) - \max\{J U_\mathrm{S}, U_{\mathrm{A}} W\}\boldsymbol{1} - WJ {\boldsymbol 1} \right)_{+} \right\|_2}{\bigl\|\bigl(\boldsymbol{Q}(\tau'_{l(t+\tau)}(t))- WJ {\boldsymbol 1} \bigr)_{+} \bigr\|_2 \|\boldsymbol{Q}(f_j(t+\tau))+U_{\mathrm{A}}U_{\mathrm{S}}\boldsymbol{1}\|_2}.
\end{align*}
Notice that the bound is positive. Then we have
\begin{align*}
    & \left(\frac{1}{\bigl\|\bigl(\boldsymbol{Q}(\tau'_{l(t+\tau)}(t))- WJ {\boldsymbol 1} \bigr)_{+} \bigr\|_2}  
    - \frac{1}{\|\boldsymbol{Q}(f_j(t+\tau))+U_{\mathrm{A}}U_{\mathrm{S}}\boldsymbol{1}\|_2}\right)
    \max_i Q_i(f_j(t+\tau)) \mu_{i,j}(t+\tau)\nonumber\\
    \le & \frac{\|\boldsymbol{Q}(f_j(t+\tau))+U_{\mathrm{A}}U_{\mathrm{S}}\boldsymbol{1}\|_2 
    - \left\|\left(\boldsymbol{Q}(f_j(t+\tau)) - \max\{J U_\mathrm{S}, U_{\mathrm{A}} W\}\boldsymbol{1} - WJ {\boldsymbol 1} \right)_{+} \right\|_2}{\bigl\|\bigl(\boldsymbol{Q}(\tau'_{l(t+\tau)}(t))- WJ {\boldsymbol 1} \bigr)_{+} \bigr\|_2 \|\boldsymbol{Q}(f_j(t+\tau))+U_{\mathrm{A}}U_{\mathrm{S}}\boldsymbol{1}\|_2}\nonumber\\
    & ~~\cdot\max_i Q_i(f_j(t+\tau)) \mu_{i,j}(t+\tau)\nonumber\\
    \le & \frac{\|\boldsymbol{Q}(f_j(t+\tau))+U_{\mathrm{A}}U_{\mathrm{S}}\boldsymbol{1}\|_2 
    - \left\|\left(\boldsymbol{Q}(f_j(t+\tau)) - \max\{J U_\mathrm{S}, U_{\mathrm{A}} W\}\boldsymbol{1} - WJ {\boldsymbol 1} \right)_{+} \right\|_2}
    {\bigl\|\bigl(\boldsymbol{Q}(\tau'_{l(t+\tau)}(t))- WJ {\boldsymbol 1} \bigr)_{+} \bigr\|_2 }\nonumber\\
    & ~~\cdot\frac{\|\boldsymbol{Q}(f_j(t+\tau))\|_2}{\|\boldsymbol{Q}(f_j(t+\tau))+U_{\mathrm{A}}U_{\mathrm{S}}\boldsymbol{1}\|_2}\nonumber\\
    \le & \frac{\|\boldsymbol{Q}(f_j(t+\tau))+U_{\mathrm{A}}U_{\mathrm{S}}\boldsymbol{1}\|_2 
    - \left\|\left(\boldsymbol{Q}(f_j(t+\tau)) - \max\{J U_\mathrm{S}, U_{\mathrm{A}} W\}\boldsymbol{1} - WJ {\boldsymbol 1} \right)_{+} \right\|_2}
    {\bigl\|\bigl(\boldsymbol{Q}(\tau'_{l(t+\tau)}(t))- WJ {\boldsymbol 1} \bigr)_{+} \bigr\|_2 },
\end{align*}
where the second inequality holds since $\max_i Q_i(f_j(t+\tau)) \mu_{i,j}(t+\tau) \le \max_i Q_i(f_j(t+\tau)) \le \| \boldsymbol{Q}(f_j(t+\tau)) \|_2$. By the triangle inequality, we further have
\begin{align*}
    & \left(\frac{1}{\bigl\|\bigl(\boldsymbol{Q}(\tau'_{l(t+\tau)}(t))- WJ {\boldsymbol 1} \bigr)_{+} \bigr\|_2}  
    - \frac{1}{\|\boldsymbol{Q}(f_j(t+\tau))+U_{\mathrm{A}}U_{\mathrm{S}}\boldsymbol{1}\|_2}\right)
    \max_i Q_i(f_j(t+\tau)) \mu_{i,j}(t+\tau)\nonumber\\
    \le & \frac{\left\|\boldsymbol{Q}(f_j(t+\tau))+U_{\mathrm{A}}U_{\mathrm{S}}\boldsymbol{1}
    - \left(\boldsymbol{Q}(f_j(t+\tau)) - \max\{J U_\mathrm{S}, U_{\mathrm{A}} W\}\boldsymbol{1} - WJ {\boldsymbol 1} \right)_{+} \right\|_2}
    {\bigl\|\bigl(\boldsymbol{Q}(\tau'_{l(t+\tau)}(t))- WJ {\boldsymbol 1} \bigr)_{+} \bigr\|_2 }\nonumber\\
    \le & \frac{\left\|\boldsymbol{Q}(f_j(t+\tau))+U_{\mathrm{A}}U_{\mathrm{S}}\boldsymbol{1}
    - \left(\boldsymbol{Q}(f_j(t+\tau)) - \max\{J U_\mathrm{S}, U_{\mathrm{A}} W\}\boldsymbol{1} - WJ {\boldsymbol 1} \right) \right\|_2}
    {\bigl\|\bigl(\boldsymbol{Q}(\tau'_{l(t+\tau)}(t))- WJ {\boldsymbol 1} \bigr)_{+} \bigr\|_2 }\nonumber\\
    = & \frac{\left\|U_{\mathrm{A}}U_{\mathrm{S}}\boldsymbol{1}
    + \max\{J U_\mathrm{S}, U_{\mathrm{A}} W\}\boldsymbol{1} + WJ {\boldsymbol 1}  \right\|_2}
    {\bigl\|\bigl(\boldsymbol{Q}(\tau'_{l(t+\tau)}(t))- WJ {\boldsymbol 1} \bigr)_{+} \bigr\|_2 }\nonumber\\
    = & \frac{\sqrt{I} \left(U_{\mathrm{A}}U_{\mathrm{S}}
    + \max\{J U_\mathrm{S}, U_{\mathrm{A}} W\} + WJ  \right)}
    {\bigl\|\bigl(\boldsymbol{Q}(\tau'_{l(t+\tau)}(t))- WJ {\boldsymbol 1} \bigr)_{+} \bigr\|_2 }.
\end{align*}
Substituting the above bound into \eqref{equ:theo-2-drift-4}, we have
\begin{align}\label{equ:theo-2-drift-5}
    & E\left[ L'(t+\frac{T}{2}+G_t) - L'(t+\frac{T}{2}) \left| \boldsymbol{Q}(t)=\boldsymbol{q}, \boldsymbol{H}(t)=\boldsymbol{h}\right.\right]\nonumber\\
    \le & \left[\sqrt{I} J \left(U_{\mathrm{A}}U_{\mathrm{S}}
    + \max\{J U_\mathrm{S}, U_{\mathrm{A}} W\} + WJ  \right) + J \max\{U_\mathrm{S} U_{\mathrm{A}}, JW\} \right]\nonumber\\
    & \sum_{\tau=\frac{T}{2}}^{\frac{T}{2}+G_t - 1} 
    \hat{E}_{t}\left[\frac{1}{\bigl\|\bigl(\boldsymbol{Q}(\tau'_{l(t+\tau)}(t))- WJ {\boldsymbol 1} \bigr)_{+} \bigr\|_2} \right]\nonumber\\
    & +\left[ J^2 U_{\mathrm{S}}(2  U_{\mathrm{S}} +  1 ) + I W U_{\mathrm{A}} (U_{\mathrm{A}} + 1 ) + \frac{IU_{\mathrm{A}}^2 + J^2 + IJ^2}{2}\right] 
    \sum_{\tau=\frac{T}{2}}^{\frac{T}{2}+G_t - 1} 
    \hat{E}_{t}\left[\frac{1}{\|\boldsymbol{Q}(t + \tau)\|_2}\right] \nonumber\\
    & - \delta G_t
    + \frac{11(I U_{\mathrm{A}} + J)JT^{1-p}}{2\delta} + 2 J U_{\mathrm{S}}
    + \frac{(I U_{\mathrm{A}} + J)T}{16}
    + \frac{\delta G_t }{2} 
    + \frac{6264(I U_{\mathrm{A}} + J) I^2 J^2 U^4_{\mathrm{S}} \log^2 \frac{1}{1-\gamma}}{\delta^2}.
\end{align}
Recall \eqref{equ:lower-bound-q-t-tau} and \eqref{equ:lower-bound-q-tau-w-j}. We can further bound \eqref{equ:theo-2-drift-5} by
\begin{align*}
    & E\left[ L'(t+\frac{T}{2}+G_t) - L'(t+\frac{T}{2}) \left| \boldsymbol{Q}(t)=\boldsymbol{q}, \boldsymbol{H}(t)=\boldsymbol{h}\right.\right]\nonumber\\
    \le & \left[\sqrt{I} J \left(U_{\mathrm{A}}U_{\mathrm{S}}
    + \max\{J U_\mathrm{S}, U_{\mathrm{A}} W\} + WJ  \right) + J \max\{U_\mathrm{S} U_{\mathrm{A}}, JW\} \right] 
    \frac{G_t \sqrt{I}}{\left(\sum_i q_i - \frac{(c_2+1)JT}{\delta} - WIJ \right)_{+}} \nonumber\\
    & +\left[ J^2 U_{\mathrm{S}}(2  U_{\mathrm{S}} +  1 ) + I W U_{\mathrm{A}} (U_{\mathrm{A}} + 1 ) + \frac{IU_{\mathrm{A}}^2 + J^2 + IJ^2}{2}\right] 
    \frac{G_t \sqrt{I}}{\left(\sum_i q_i - \frac{(c_2+1)JT}{\delta}  \right)_{+}} \nonumber\\
    & - \delta G_t
    + \frac{11(I U_{\mathrm{A}} + J)JT^{1-p}}{2\delta} + 2 J U_{\mathrm{S}}
    + \frac{(I U_{\mathrm{A}} + J)T}{16}
    + \frac{\delta G_t }{2} 
    + \frac{6264(I U_{\mathrm{A}} + J) I^2 J^2 U^4_{\mathrm{S}} \log^2 \frac{1}{1-\gamma}}{\delta^2}\nonumber\\
    \le & \frac{33(I U_{\mathrm{A}} + J) IJ^2 W U_\mathrm{S} U_{\mathrm{A}} T} {\delta \left(\sum_i q_i - \frac{6(I U_{\mathrm{A}} + J)JT}{\delta} - WIJ \right)_{+}} + \frac{31(I U_{\mathrm{A}} + J) I^{\frac{3}{2}} J^2 W U^2_\mathrm{S} U^2_{\mathrm{A}} T }{\delta \left(\sum_i q_i - \frac{6(I U_{\mathrm{A}} + J)JT}{\delta}  \right)_{+}}  -\frac{5(I U_{\mathrm{A}} + J)T}{2}\nonumber\\
    & + \frac{11(I U_{\mathrm{A}} + J)JT^{1-p}}{2\delta} + 2 J U_{\mathrm{S}}
    + \frac{(I U_{\mathrm{A}} + J)T}{16}
    + \frac{6264(I U_{\mathrm{A}} + J) I^2 J^2 U^4_{\mathrm{S}} \log^2 \frac{1}{1-\gamma}}{\delta^2},
\end{align*}
where the last inequality is by Lemma~\ref{lemma:prop-G} and $c_2=5(I U_{\mathrm{A}} + J)$. From the condition \eqref{equ:condition-delta-2} in Theorem~\ref{theo:2} and $\gamma \ge 1-\frac{1}{1+e^{1.5}}$, we have
$\frac{6264(I U_{\mathrm{A}} + J) I^2 J^2 U^4_{\mathrm{S}} \log^2 \frac{1}{1-\gamma}}{\delta^2} \le \frac{(I U_{\mathrm{A}} + J)T}{4}$ and $\frac{11(I U_{\mathrm{A}} + J)JT^{1-p}}{2\delta} \le \frac{(I U_{\mathrm{A}} + J)T}{16}$. Therefore, we have
\begin{align*}
    & E\left[ L'(t+\frac{T}{2}+G_t) - L'(t+\frac{T}{2}) \left| \boldsymbol{Q}(t)=\boldsymbol{q}, \boldsymbol{H}(t)=\boldsymbol{h}\right.\right]\nonumber\\
    \le & \frac{33(I U_{\mathrm{A}} + J) IJ^2 W U_\mathrm{S} U_{\mathrm{A}} T} {\delta \left(\sum_i q_i - \frac{6(I U_{\mathrm{A}} + J)JT}{\delta} - WIJ \right)_{+}} + \frac{31(I U_{\mathrm{A}} + J) I^{\frac{3}{2}} J^2 W U^2_\mathrm{S} U^2_{\mathrm{A}} T }{\delta \left(\sum_i q_i - \frac{6(I U_{\mathrm{A}} + J)JT}{\delta}  \right)_{+}} \nonumber\\
    & + 2JU_{\mathrm{S}} -\left(\frac{5}{2}-\frac{1}{4}-\frac{1}{16}-\frac{1}{16}\right)(I U_{\mathrm{A}} + J)T\nonumber\\
    = & \frac{33(I U_{\mathrm{A}} + J) IJ^2 W U_\mathrm{S} U_{\mathrm{A}} T} {\delta \left(\sum_i q_i - \frac{6(I U_{\mathrm{A}} + J)JT}{\delta} - WIJ \right)_{+}} + \frac{31(I U_{\mathrm{A}} + J) I^{\frac{3}{2}} J^2 W U^2_\mathrm{S} U^2_{\mathrm{A}} T }{\delta \left(\sum_i q_i - \frac{6(I U_{\mathrm{A}} + J)JT}{\delta}  \right)_{+}} \nonumber\\
    & + 2JU_{\mathrm{S}} -\frac{17}{8}(I U_{\mathrm{A}} + J)T..
\end{align*}
Suppose $\sum_i q_i \ge \frac{8(I U_{\mathrm{A}} + J)JT}{\delta}$. Then we have $\sum_i q_i - \frac{6(I U_{\mathrm{A}} + J)JT}{\delta} - WIJ \ge \frac{3(I U_{\mathrm{A}} + J)JT}{2\delta}$ (since $W\le\frac{T}{2}$) and $\sum_i q_i - \frac{6(I U_{\mathrm{A}} + J)JT}{\delta} \ge \frac{2(I U_{\mathrm{A}} + J)JT}{\delta}$. Hence, we have
\begin{align}\label{equ:theo-2-negative-drift}
    & E\left[ L'(t+\frac{T}{2}+G_t) - L'(t+\frac{T}{2}) \left| \boldsymbol{Q}(t)=\boldsymbol{q}, \boldsymbol{H}(t)=\boldsymbol{h}\right.\right]\nonumber\\
    \le & 22 IJ U_\mathrm{S} U_{\mathrm{A}} W + 16 I^{\frac{3}{2}} J U^2_\mathrm{S} U^2_{\mathrm{A}} W + 2 J U_\mathrm{S} -\frac{17}{8}(I U_{\mathrm{A}} + J)T \nonumber\\
    \le & 40 I^{\frac{3}{2}} J U^2_\mathrm{S} U^2_{\mathrm{A}} W -\frac{17}{8}(I U_{\mathrm{A}} + J)T\nonumber\\
    \le & -\left(\frac{17}{8}-\frac{1}{16}\right)(I U_{\mathrm{A}} + J)T\nonumber\\
    \le & -(2+\frac{1}{16}) (I U_{\mathrm{A}} + J)T,
\end{align}
where the third inequality holds since $40 I^{\frac{3}{2}} J U^2_\mathrm{S} U^2_{\mathrm{A}} W \le \frac{40 I^{\frac{3}{2}} J U^2_\mathrm{S} U^2_{\mathrm{A}} W }{\delta^2} \le \frac{(I U_{\mathrm{A}} + J)T}{16}$ by the condition \eqref{equ:condition-delta-2} in Theorem~\ref{theo:2} and $\gamma \ge 1-\frac{1}{1+e^{1.5}}$.
Note that for any $t$ this negative drift \eqref{equ:theo-2-negative-drift} holds when $\sum_i Q_i(t) = \sum_i q_i \ge \frac{8(I U_{\mathrm{A}} + J)JT}{\delta}$.

\subsection{Bounding the Exponential Queue Length Recursively}
Fix any time slot $t$. We will bound the exponential queue length at time $t$ recursively. Let
\begin{align*}
    t'\coloneqq t-\frac{T}{2}-\left\lfloor\frac{c_2 T}{\delta}\right\rfloor.
\end{align*}
Then by Lemma~\ref{lemma:prop-G}, we have
\begin{align}\label{equ:range-of-t-G}
    t=& t-\frac{T}{2}-\left\lfloor\frac{c_2 T}{\delta}\right\rfloor + \frac{T}{2}+\left\lfloor\frac{c_2 T}{\delta}\right\rfloor\nonumber\\
    = &  t' + \frac{T}{2} + \left\lfloor\frac{c_2 T}{\delta}\right\rfloor\nonumber\\
    \le & t' + \frac{T}{2} + G_{t'}\nonumber\\
    \le &  t' + \frac{T}{2} + \frac{c_2 T}{\delta} + W\nonumber\\
    = & t-\frac{T}{2}-\left\lfloor\frac{c_2 T}{\delta}\right\rfloor + \frac{T}{2} + \frac{c_2 T}{\delta} + W\nonumber\\
    \le & t + W + 1,
\end{align}
which means that the length of the interval $[t', t]$ is approximately $\frac{T}{2} + G_{t'}$ with error bounded by $W+1$. 
Define for any nonnegative integers $\tau$ and $n$,
\begin{align*}
    \Delta_n(\tau)\coloneqq \|\boldsymbol{Q}(\tau+n)\|_2 - \|\boldsymbol{Q}(\tau)\|_2.
\end{align*}
By the triangle inequality, we have
\begin{align}\label{equ:Delta-bound}
    |\Delta_n(\tau)| \le & \|\boldsymbol{Q}(\tau+n) - \boldsymbol{Q}(\tau)\|_2\nonumber\\
    \le &  \|\boldsymbol{Q}(\tau+n) - \boldsymbol{Q}(\tau)\|_1\nonumber\\
    = & \sum_i \left| Q_i(\tau+n) -  Q_i(\tau)\right|\nonumber\\
    = & \sum_i \left( Q_i(\tau+n) -  Q_i(\tau)\right) \mathbb{1}_{Q_i(\tau+n) > Q_i(\tau)}  
    + \sum_i \left(Q_i(\tau) - Q_i(\tau+n)\right) \mathbb{1}_{Q_i(\tau+n) < Q_i(\tau)}  \nonumber\\
    \le & (IU_{\mathrm{A}} + J)n,
\end{align}
where the second inequality is due to the fact that $\|\boldsymbol{x}\|_2\le \|\boldsymbol{x}\|_1$ for any $\boldsymbol{x}$, and the last inequality uses Lemma~\ref{lemma:bounds-on-diff-queue-lens}.
From \eqref{equ:range-of-t-G}, \eqref{equ:Delta-bound}, and the definition of $\Delta_n(\tau)$, we have
\begin{align}\label{equ:proof-exp-bound-1}
    \exp\left(\xi \|\boldsymbol{Q}(t)\|_2\right) = & \exp\left(\xi \|\boldsymbol{Q}(t'+\frac{T}{2}+G_{t'})\|_2 \right) 
    \exp\left(\xi \biggl(\|\boldsymbol{Q}(t)\|_2 - \|\boldsymbol{Q}(t'+\frac{T}{2}+G_{t'})\|_2\biggr)\right)\nonumber\\
    = & \exp\left(\xi \|\boldsymbol{Q}(t'+\frac{T}{2}+G_{t'})\|_2 \right) 
    \exp\left( - \xi \Delta_{t'+\frac{T}{2}+G_{t'} - t}(t) \right)\nonumber\\
    \le & \exp\left(\xi \|\boldsymbol{Q}(t'+\frac{T}{2}+G_{t'})\|_2 \right) 
    \exp\left( \xi \left|\Delta_{t'+\frac{T}{2}+G_{t'} - t}(t)\right| \right)\nonumber\\
    \le & \exp\left(\xi \|\boldsymbol{Q}(t'+\frac{T}{2}+G_{t'})\|_2 \right)
    \exp\left(\xi (IU_{\mathrm{A}} + J) (t'+\frac{T}{2}+G_{t'} - t) \right)\nonumber\\
    \le & \exp\left(\xi \|\boldsymbol{Q}(t'+\frac{T}{2}+G_{t'})\|_2 \right)
    \exp\left(\xi (IU_{\mathrm{A}} + J) (W+1) \right),
\end{align}
where the first inequality is by the triangle inequality, the second inequality is due to the fact that $\|\boldsymbol{x}\|_2\le \|\boldsymbol{x}\|_1$ for any $\boldsymbol{x}$. Similarly, we have
\begin{align}\label{equ:proof-exp-bound-2}
    & \exp\left(\xi \|\boldsymbol{Q}(t'+\frac{T}{2}+G_{t'})\|_2 \right)\nonumber\\
    = & \exp\left(\xi \|\boldsymbol{Q}(t'+\frac{T}{2})\|_2 \right) 
    \exp\left(\xi \biggl(\|\boldsymbol{Q}(t'+\frac{T}{2} + G_{t'})\|_2 - \|\boldsymbol{Q}(t'+\frac{T}{2})\|_2\biggr) \right)\nonumber\\
    = & \exp\left(\xi \|\boldsymbol{Q}(t'+\frac{T}{2})\|_2 \right) 
    \exp\left(\xi \Delta_{G_{t'}}(t'+\frac{T}{2}) \right)\nonumber\\
    = & \exp\left(\xi \|\boldsymbol{Q}(t')\|_2 \right)
    \exp\left(\xi \biggl(\|\boldsymbol{Q}(t'+\frac{T}{2})\|_2  - \|\boldsymbol{Q}(t')\|_2 \biggr)\right)
    \exp\left(\xi \Delta_{G_{t'}}(t'+\frac{T}{2}) \right)\nonumber\\
    = & \exp\left(\xi \|\boldsymbol{Q}(t')\|_2 \right)
    \exp\left(\xi \Delta_{\frac{T}{2}}(t')
    \right)
    \exp\left(\xi \Delta_{G_{t'}}(t'+\frac{T}{2}) \right)\nonumber\\
    \le & \exp\left(\xi \|\boldsymbol{Q}(t')\|_2 \right)
    \exp\left(\xi \frac{(IU_{\mathrm{A}} + J)T}{2} \right)
    \exp\left(\xi \Delta_{G_{t'}}(t'+\frac{T}{2}) \right),
\end{align}
where the last inequality is by \eqref{equ:Delta-bound}. Substituting \eqref{equ:proof-exp-bound-2} into \eqref{equ:proof-exp-bound-1}, we have
\begin{align}\label{equ:proof-exp-bound-3}
    & \exp\left(\xi \|\boldsymbol{Q}(t)\|_2\right) \nonumber\\
    \le & \exp\left(\xi \|\boldsymbol{Q}(t')\|_2 \right)
    \exp\left(\xi \Delta_{G_{t'}}(t'+\frac{T}{2}) \right)
    \exp\left(\xi \frac{(IU_{\mathrm{A}} + J)T}{2} \right)
    \exp\left(\xi (IU_{\mathrm{A}} + J) (W+1) \right)\nonumber\\
    = & \exp\left(\xi \|\boldsymbol{Q}(t')\|_2 \right)
    \exp\left(\xi \Delta_{G_{t'}}(t'+\frac{T}{2}) \right)
    \exp\left(\xi \left(\frac{T}{2} + W + 1\right) (IU_{\mathrm{A}} + J) \right)\nonumber\\
    \le & \exp\left(\xi \|\boldsymbol{Q}(t')\|_2 \right)
    \exp\left(\xi \left[\Delta_{G_{t'}}\left(t'+\frac{T}{2}\right)+ (T + 1) (IU_{\mathrm{A}} + J)\right] \right),
\end{align}
where the last inequality is by $W\le T/2$. 
Next, we will use Taylor series to approximate the term $\exp\left(\xi \left[\Delta_{G_{t'}}\left(t'+\frac{T}{2}\right)+ (T + 1) (IU_{\mathrm{A}} + J)\right] \right)$. We first present the following lemma~\cite[Proof of Theorem 3.6]{ChuLu_06}.
\begin{lemma}\label{lemma:taylor}
    For any $y$ such that $|y|\le 3$, we have
    \begin{align*}
        e^y \le 1 + y + \frac{y^2}{2(1-y/3)}\le 1 + y + \frac{y^2}{2(1-|y|/3)}.
    \end{align*}
\end{lemma}
The proof is presented in Section~\ref{app:proof-lemma-taylor} for completeness.
Note that by the triangle inequality and \eqref{equ:Delta-bound}, we have
\begin{align}\label{equ:bound-need-to-used-later}
    \left|\Delta_{G_{t'}}\left(t'+\frac{T}{2}\right)+ (T + 1) (IU_{\mathrm{A}} + J) \right|
    \le & \left|\Delta_{G_{t'}}\left(t'+\frac{T}{2}\right) \right| 
    + (T + 1) (IU_{\mathrm{A}} + J)\nonumber\\
    \le & (IU_{\mathrm{A}} + J) G_{t'} + (T + 1) (IU_{\mathrm{A}} + J) \nonumber\\
    < &  \frac{(c_2+2)(IU_{\mathrm{A}} + J)T}{\delta},
\end{align}
where the last inequality uses Lemma~\ref{lemma:prop-G} and the fact that $T=8 U_{\mathrm{S}}\ge 8$.
Choose $\xi$ such that $0 < \xi \le \frac{3\delta}{(c_2+2)(IU_{\mathrm{A}} + J)T}$. Then we have
\begin{align}\label{equ:bound-xi-delta}
    \xi\left|\Delta_{G_{t'}}\left(t'+\frac{T}{2}\right)+ (T + 1) (IU_{\mathrm{A}} + J)\right| < 3.
\end{align}
By Lemma~\ref{lemma:taylor}, we have
\begin{align}\label{equ:taylor-1}
    & \exp\left(\xi \left[\Delta_{G_{t'}}\left(t'+\frac{T}{2}\right)+ (T + 1) (IU_{\mathrm{A}} + J)\right] \right) \nonumber\\
    \le & 1 +  \xi \left[\Delta_{G_{t'}}\left(t'+\frac{T}{2}\right)+ (T + 1) (IU_{\mathrm{A}} + J)\right]
    + \frac{\xi^2 \left[\Delta_{G_{t'}}\left(t'+\frac{T}{2}\right)+ (T + 1) (IU_{\mathrm{A}} + J)\right]^2}{2\left(1- \frac{\xi\left|\Delta_{G_{t'}}\left(t'+\frac{T}{2}\right)+ (T + 1) (IU_{\mathrm{A}} + J)\right|}{3}\right)}\nonumber\\
    \le & 1 +  \xi \left[\Delta_{G_{t'}}\left(t'+\frac{T}{2}\right)+ (T + 1) (IU_{\mathrm{A}} + J)\right]
    + \frac{\xi^2 (c_2+2)^2(IU_{\mathrm{A}} + J)^2T^2}{2\delta^2\left(1- \frac{\xi(c_2+2)(IU_{\mathrm{A}} + J)T}{3\delta}\right)}
\end{align}
Choose $\xi$ such that
\begin{align}\label{equ:condition-xi}
    0 < \xi \le \frac{3\delta}{ [5(I U_{\mathrm{A}} +J) +2] (I U_{\mathrm{A}} +J) T \left[ 1 + \frac{15(I U_{\mathrm{A}} +J) +6}{\delta} \right] }.
\end{align}
Then by $c_2=5(I U_{\mathrm{A}} +J)$, we can check that
\begin{align}\label{equ:condition-xi-2}
    \xi \le \frac{3\delta}{[5(I U_{\mathrm{A}} +J)+2](I U_{\mathrm{A}} +J)T} = \frac{3\delta}{(c_2+2)(IU_{\mathrm{A}} + J)T}
\end{align}
and
\begin{align}\label{equ:taylor-2}
    \frac{\xi^2 (c_2+2)^2(IU_{\mathrm{A}} + J)^2T^2}{2\delta^2\left(1- \frac{\xi(c_2+2)(IU_{\mathrm{A}} + J)T}{3\delta}\right)} \le \frac{\xi T (IU_{\mathrm{A}} + J) }{2}.
\end{align}
Hence, from \eqref{equ:taylor-1} and \eqref{equ:taylor-2}, we have
\begin{align}\label{equ:taylor-3}
    & \exp\left(\xi \left[\Delta_{G_{t'}}\left(t'+\frac{T}{2}\right)+ (T + 1) (IU_{\mathrm{A}} + J)\right] \right) \nonumber\\
    \le & 1 +  \xi \left[\Delta_{G_{t'}}\left(t'+\frac{T}{2}\right)+ (T + 1) (IU_{\mathrm{A}} + J)\right] + \frac{\xi T (IU_{\mathrm{A}} + J) }{2}\nonumber\\
    \le & 1 + \xi \left[\Delta_{G_{t'}}\left(t'+\frac{T}{2}\right)+ \left(\frac{3T}{2} + 1\right) (IU_{\mathrm{A}} + J)\right],
\end{align}
which holds when the condition \eqref{equ:condition-xi} holds. 
Let $\varphi\coloneqq \frac{8(IU_{\mathrm{A}} + J)JT}{\delta}$. Then by \eqref{equ:proof-exp-bound-3} and the law of total expectation, we have
\begin{align}\label{equ:exp-bound-4}
    & E\left[ \exp\left(\xi \|\boldsymbol{Q}(t)\|_2\right) \right] \nonumber\\
    \le & 
    E\left[
    \exp\left(\xi \|\boldsymbol{Q}(t')\|_2 \right)
    \exp\left(\xi \left[\Delta_{G_{t'}}\left(t'+\frac{T}{2}\right)+ (T + 1) (IU_{\mathrm{A}} + J)\right] \right)
    \right]\nonumber\\
    = & \sum_{\boldsymbol{q}:\|\boldsymbol{q}\|_2>\varphi} 
    \exp\left(\xi \|\boldsymbol{q}\|_2 \right)
    E\left[
    \exp\left(\xi \left[\Delta_{G_{t'}}\left(t'+\frac{T}{2}\right)+ (T + 1) (IU_{\mathrm{A}} + J)\right] \right) \left| \boldsymbol{Q}(t') = \boldsymbol{q} \right.
    \right] \Pr \left( \boldsymbol{Q}(t') = \boldsymbol{q} \right)\nonumber\\
    & + \sum_{\boldsymbol{q}:\|\boldsymbol{q}\|_2 \le \varphi} 
    \exp\left(\xi \|\boldsymbol{q}\|_2 \right)
    E\left[
    \exp\left(\xi \left[\Delta_{G_{t'}}\left(t'+\frac{T}{2}\right)+ (T + 1) (IU_{\mathrm{A}} + J)\right] \right) \left| \boldsymbol{Q}(t') = \boldsymbol{q} \right.
    \right] \Pr \left( \boldsymbol{Q}(t') = \boldsymbol{q} \right)\nonumber\\
    \le & \sum_{\boldsymbol{q}:\|\boldsymbol{q}\|_2>\varphi} 
    \exp\left(\xi \|\boldsymbol{q}\|_2 \right)
    \left(1 + \xi
    E\left[
    \Delta_{G_{t'}}\left(t'+\frac{T}{2}\right)
    \left| \boldsymbol{Q}(t') = \boldsymbol{q} \right.
    \right]
    + \xi \left(\frac{3T}{2} + 1\right) (I U_{\mathrm{A}} + J)
    \right)
    \Pr \left( \boldsymbol{Q}(t') = \boldsymbol{q} \right)\nonumber\\
    & + \sum_{\boldsymbol{q}:\|\boldsymbol{q}\|_2 \le \varphi} 
    \exp\left(\xi \|\boldsymbol{q}\|_2 \right)
    E\left[
    \exp\left(\xi \left[\Delta_{G_{t'}}\left(t'+\frac{T}{2}\right)+ (T + 1) (IU_{\mathrm{A}} + J)\right] \right) \left| \boldsymbol{Q}(t') = \boldsymbol{q} \right.
    \right] \Pr \left( \boldsymbol{Q}(t') = \boldsymbol{q} \right)\nonumber\\
    \le & \sum_{\boldsymbol{q}:\|\boldsymbol{q}\|_2>\varphi} 
    \exp\left(\xi \|\boldsymbol{q}\|_2 \right)
    \left(1 + \xi
    E\left[
    \Delta_{G_{t'}}\left(t'+\frac{T}{2}\right)
    \left| \boldsymbol{Q}(t') = \boldsymbol{q} \right.
    \right]
    + \xi \left(\frac{3T}{2} + 1\right) (I U_{\mathrm{A}} + J)
    \right)
    \Pr \left( \boldsymbol{Q}(t') = \boldsymbol{q} \right)\nonumber\\
    & + 
    \exp\left(\frac{\xi (c_2+2)(I U_{\mathrm{A}} + J)T}{\delta}
    \right)
    \sum_{\boldsymbol{q}:\|\boldsymbol{q}\|_2 \le \varphi} 
    \exp\left(\xi\|\boldsymbol{q}\|_2\right)
    \Pr \left( \boldsymbol{Q}(t') = \boldsymbol{q} \right),
\end{align}
where the second inequality is by \eqref{equ:taylor-3} and the third inequality is by \eqref{equ:bound-need-to-used-later}. 
Note that by the law of iterated expectation and the negative drift \eqref{equ:theo-2-negative-drift} under large queue length, we have
\begin{align}\label{equ:theo-2-negative-drift-cont}
    E\left[
    \Delta_{G_{t'}}\left(t'+\frac{T}{2}\right)
    \left| \boldsymbol{Q}(t') = \boldsymbol{q} \right.
    \right] = & E\left[ 
    E \left[
    \Delta_{G_{t'}}\left(t'+\frac{T}{2}\right)
    \left| \boldsymbol{H}(t'), \boldsymbol{Q}(t')=\boldsymbol{q} \right.
    \right]
    \left| \boldsymbol{Q}(t') = \boldsymbol{q} \right.
    \right]\nonumber\\
    \le & 
    -(2+\frac{1}{16}) (I U_{\mathrm{A}} + J)T
\end{align}
for $\sum_i q_i \ge \varphi$. Noticing the fact that $\sum_i q_i = \|\boldsymbol{q}\|_1 \ge \|\boldsymbol{q}\|_2$, we know that \eqref{equ:theo-2-negative-drift-cont} also holds when $\|\boldsymbol{q}\|_2 \ge \varphi$. Hence, substituting \eqref{equ:theo-2-negative-drift-cont} into \eqref{equ:exp-bound-4}, we obtain
\begin{align}\label{equ:exp-bound-5}
    & E\left[ \exp\left(\xi \|\boldsymbol{Q}(t)\|_2\right) \right] \nonumber\\
    \le & 
    \left(
    1 - \xi T \left( \frac{1}{2} + \frac{1}{16} - \frac{1}{T} \right) (I U_{\mathrm{A}} + J)
    \right)
    \sum_{\boldsymbol{q}:\|\boldsymbol{q}\|_2>\varphi} 
    \exp\left(\xi \|\boldsymbol{q}\|_2 \right)
    \Pr \left( \boldsymbol{Q}(t') = \boldsymbol{q} \right)
    \nonumber\\
    & + 
    \exp\left(\frac{\xi (c_2+2)(I U_{\mathrm{A}} + J)T}{\delta}
    \right)
    \sum_{\boldsymbol{q}:\|\boldsymbol{q}\|_2 \le \varphi} 
    \exp\left(\xi\|\boldsymbol{q}\|_2\right)
    \Pr \left( \boldsymbol{Q}(t') = \boldsymbol{q} \right)\nonumber\\
    \le & 
    \left(
    1 - \frac{\xi T (I U_{\mathrm{A}} + J)}{2}
    \right)
    \sum_{\boldsymbol{q}:\|\boldsymbol{q}\|_2>\varphi} 
    \exp\left(\xi \|\boldsymbol{q}\|_2 \right)
    \Pr \left( \boldsymbol{Q}(t') = \boldsymbol{q} \right)
    \nonumber\\
    & + 
    \exp\left(\frac{\xi (c_2+2)(I U_{\mathrm{A}} + J)T}{\delta}
    \right)
    \sum_{\boldsymbol{q}:\|\boldsymbol{q}\|_2 \le \varphi} 
    \exp\left(\xi\|\boldsymbol{q}\|_2\right)
    \Pr \left( \boldsymbol{Q}(t') = \boldsymbol{q} \right)\nonumber\\
    = & 
    \left(
    1 - \frac{\xi T (I U_{\mathrm{A}} + J)}{2}
    \right)
    \left(\sum_{\boldsymbol{q}:\|\boldsymbol{q}\|_2>\varphi} 
    \exp\left(\xi \|\boldsymbol{q}\|_2 \right)
    \Pr \left( \boldsymbol{Q}(t') = \boldsymbol{q} \right)
    + \sum_{\boldsymbol{q}:\|\boldsymbol{q}\|_2 \le \varphi} 
    \exp\left(\xi\|\boldsymbol{q}\|_2\right)
    \Pr \left( \boldsymbol{Q}(t') = \boldsymbol{q} \right) \right)
    \nonumber\\
    & + 
    \left[\exp\left(\frac{\xi (c_2+2)(I U_{\mathrm{A}} + J)T}{\delta}
    \right)
    - \left(1 - \frac{\xi T (I U_{\mathrm{A}} + J)}{2}\right)
    \right]
    \sum_{\boldsymbol{q}:\|\boldsymbol{q}\|_2 \le \varphi} 
    \exp\left(\xi\|\boldsymbol{q}\|_2\right)
    \Pr \left( \boldsymbol{Q}(t') = \boldsymbol{q} \right)\nonumber\\
    = & 
    \left(
    1 - \frac{\xi T (I U_{\mathrm{A}} + J)}{2}
    \right)
     E \left[ \exp\left(\xi \|\boldsymbol{Q}(t')\|_2 \right) \right]
     \nonumber\\
    & + 
    \left[\exp\left(\frac{\xi (c_2+2)(I U_{\mathrm{A}} + J)T}{\delta}
    \right)
    - \left(1 - \frac{\xi T (I U_{\mathrm{A}} + J)}{2}\right)
    \right]
    \sum_{\boldsymbol{q}:\|\boldsymbol{q}\|_2 \le \varphi} 
    \exp\left(\xi\|\boldsymbol{q}\|_2\right)
    \Pr \left( \boldsymbol{Q}(t') = \boldsymbol{q} \right),
\end{align}
where the second inequality is by $T=\frac{4}{1-\gamma}\log \frac{1}{1-\gamma} > 16$ since $\gamma \ge 1-\frac{1}{1+e^{1.5}}$, and the last equality is by the law of total expectation.
Note that
\begin{align*}
    & \exp\left(\frac{\xi (c_2+2)(I U_{\mathrm{A}} + J)T}{\delta}
    \right)
    - \left(1 - \frac{\xi T (I U_{\mathrm{A}} + J)}{2}\right)\nonumber\\
    \ge & 1 + \frac{\xi (c_2+2)(I U_{\mathrm{A}} + J)T}{\delta}
    - \left(1 - \frac{\xi T (I U_{\mathrm{A}} + J)}{2}\right)
    > 0
\end{align*}
and
\begin{align*}
    \sum_{\boldsymbol{q}:\|\boldsymbol{q}\|_2 \le \varphi} 
    \exp\left(\xi\|\boldsymbol{q}\|_2\right)
    \Pr \left( \boldsymbol{Q}(t') = \boldsymbol{q} \right) 
    \le & \sum_{\boldsymbol{q}:\|\boldsymbol{q}\|_2 \le \varphi} 
    \exp\left(\xi \varphi\right)
    \Pr \left( \boldsymbol{Q}(t') = \boldsymbol{q} \right)\nonumber\\
    \le & \exp\left(\xi \varphi\right) \sum_{\boldsymbol{q}} \Pr \left( \boldsymbol{Q}(t') = \boldsymbol{q} \right)
    = \exp\left(\xi \varphi\right).
\end{align*}
Hence, we can further bound \eqref{equ:exp-bound-5} by
\begin{align}\label{equ:exp-bound-6}
    E\left[ \exp\left(\xi \|\boldsymbol{Q}(t)\|_2\right) \right]    
    \le & \left(
    1 - \frac{\xi T (I U_{\mathrm{A}} + J)}{2}
    \right)
     E \left[ \exp\left(\xi \|\boldsymbol{Q}(t')\|_2 \right) \right]
     \nonumber\\
     & + 
    \left[\exp\left(\frac{\xi (c_2+2)(I U_{\mathrm{A}} + J)T}{\delta}
    \right)
    - \left(1 - \frac{\xi T (I U_{\mathrm{A}} + J)}{2}\right)
    \right] \exp\left(\xi \varphi\right)
\end{align}
Note that by \eqref{equ:condition-xi-2}, we have $0 < \frac{\xi (c_2+2)(I U_{\mathrm{A}} + J)T}{\delta} < 3$. Hence, by Lemma~\ref{lemma:taylor}, we have
\begin{align}\label{equ:taylor-4}
    \exp\left(\frac{\xi (c_2+2)(I U_{\mathrm{A}} + J)T}{\delta}
    \right) \le 1 + \frac{\xi (c_2+2)(I U_{\mathrm{A}} + J)T}{\delta} + \frac{\xi^2 (c_2+2)^2(IU_{\mathrm{A}} + J)^2T^2}{2\delta^2\left(1- \frac{\xi(c_2+2)(IU_{\mathrm{A}} + J)T}{3\delta}\right)}.
\end{align}
From \eqref{equ:exp-bound-6}, \eqref{equ:taylor-4}, and \eqref{equ:taylor-2}, we have
\begin{align}\label{equ:exp-bound-7}
    & E\left[ \exp\left(\xi \|\boldsymbol{Q}(t)\|_2\right) \right]    \nonumber\\
    \le & \left(
    1 - \frac{\xi T (I U_{\mathrm{A}} + J)}{2}
    \right)
     E \left[ \exp\left(\xi \|\boldsymbol{Q}(t')\|_2 \right) \right]
      + 
    \left[\frac{\xi (c_2+2)(I U_{\mathrm{A}} + J)T}{\delta}
    + \xi T (I U_{\mathrm{A}} + J)
    \right] \exp\left(\xi \varphi\right)\nonumber\\
    \le & \left(
    1 - \frac{\xi T (I U_{\mathrm{A}} + J)}{2}
    \right)
     E \left[ \exp\left(\xi \|\boldsymbol{Q}(t')\|_2 \right) \right]
    + 
    \frac{\xi T (c_2+3)(I U_{\mathrm{A}} + J) \exp\left(\xi \varphi\right)}{\delta}.
\end{align}
Let 
\[
    \rho\coloneqq 
    1 - \frac{\xi T (I U_{\mathrm{A}} + J)}{2}
    .
\]
By the condition \eqref{equ:condition-xi} ,we can check that
\begin{align*}
    0 < \rho < 1.
\end{align*}
Recall that $t'= t-\frac{T}{2}-\left\lfloor\frac{c_2 T}{\delta}\right\rfloor.$
Applying \eqref{equ:exp-bound-7} recursively, we obtain
\begin{align}\label{equ:exp-bound-8}
    & E\left[ \exp\left(\xi \|\boldsymbol{Q}(t)\|_2\right) \right] \nonumber\\
    \le & \rho^{\biggl\lfloor 
    \frac{t}{\frac{T}{2}+\left\lfloor\frac{c_2 T}{\delta}\right\rfloor}
    \biggr\rfloor}
    E\left[
    \exp \left(
    \xi \left\| 
    \boldsymbol{Q}\left(
    t - \left(\frac{T}{2}+\left\lfloor\frac{c_2 T}{\delta}\right\rfloor\right) 
    \left\lfloor 
    \frac{t}{\frac{T}{2}+\left\lfloor\frac{c_2 T}{\delta}\right\rfloor}
    \right\rfloor
    \right) 
    \right\|_2
    \right)
    \right]\nonumber\\
    & + \sum_{n=0}^{\biggl\lfloor 
    \frac{t}{\frac{T}{2}+\left\lfloor\frac{c_2 T}{\delta}\right\rfloor}
    \biggr\rfloor-1}
    \rho^n \frac{\xi T (c_2+3)(I U_{\mathrm{A}} + J) \exp\left(\xi \varphi\right)}{\delta}\nonumber\\
    \le &
    E\left[
    \exp \left(
    \xi \left\| 
    \boldsymbol{Q}\left(
    t - \left(\frac{T}{2}+\left\lfloor\frac{c_2 T}{\delta}\right\rfloor\right) 
    \left\lfloor 
    \frac{t}{\frac{T}{2}+\left\lfloor\frac{c_2 T}{\delta}\right\rfloor}
    \right\rfloor
    \right) 
    \right\|_2
    \right)
    \right]
    + \frac{\xi T (c_2+3)(I U_{\mathrm{A}} + J) \exp\left(\xi \varphi\right)}{\delta (1-\rho)},
\end{align}
which holds for any $t$ when $\xi$ satisfies the condition \eqref{equ:condition-xi}. Note that
\[
    t - \left(\frac{T}{2}+\left\lfloor\frac{c_2 T}{\delta}\right\rfloor\right) 
    \left\lfloor 
    \frac{t}{\frac{T}{2}+\left\lfloor\frac{c_2 T}{\delta}\right\rfloor}
    \right\rfloor \le \frac{T}{2}+ \left\lfloor\frac{c_2 T}{\delta}\right\rfloor.
\]
Hence, by Lemma~\ref{lemma:bounds-on-diff-queue-lens} and the initial condition that $Q_i(0)=0$ for all $i$, we have
\begin{align}\label{equ:exp-initial-bound}
    & E\left[
    \exp \left(
    \xi \left\| 
    \boldsymbol{Q}\left(
    t - \left(\frac{T}{2}+\left\lfloor\frac{c_2 T}{\delta}\right\rfloor\right) 
    \left\lfloor 
    \frac{t}{\frac{T}{2}+\left\lfloor\frac{c_2 T}{\delta}\right\rfloor}
    \right\rfloor
    \right) 
    \right\|_2
    \right)
    \right]\nonumber\\
    \le & E\left[
    \exp \left(
    \xi \left\| 
    \boldsymbol{Q}\left(
    t - \left(\frac{T}{2}+\left\lfloor\frac{c_2 T}{\delta}\right\rfloor\right) 
    \left\lfloor 
    \frac{t}{\frac{T}{2}+\left\lfloor\frac{c_2 T}{\delta}\right\rfloor}
    \right\rfloor
    \right) 
    \right\|_1
    \right)
    \right]\nonumber\\
    = & E\left[
    \exp \left(
    \xi 
    \sum_i
    Q_i\left(
    t - \left(\frac{T}{2}+\left\lfloor\frac{c_2 T}{\delta}\right\rfloor\right) 
    \left\lfloor 
    \frac{t}{\frac{T}{2}+\left\lfloor\frac{c_2 T}{\delta}\right\rfloor}
    \right\rfloor
    \right) 
    \right)
    \right]\nonumber\\
    \le & \exp \left(
    \xi \left( \frac{T}{2}+\left\lfloor\frac{c_2 T}{\delta}\right\rfloor \right) I U_{\mathrm{A}}
    \right)\nonumber\\
    \le & e^{3/37},
\end{align}
where the last inequality is by \eqref{equ:condition-xi} and $c_2=5(I U_{\mathrm{A}} + J)$. Substituting \eqref{equ:exp-initial-bound} into \eqref{equ:exp-bound-8}, we have
\begin{align}\label{equ:exp-bound-final}
    E\left[ \exp\left(\xi \|\boldsymbol{Q}(t)\|_2\right) \right] \le & e^{3/37} + \frac{\xi T (c_2+3)(I U_{\mathrm{A}} + J) \exp\left(\xi \varphi\right)}{\delta (1-\rho)}\nonumber\\
    = & e^{3/37} + \frac{\xi T (c_2+3)(I U_{\mathrm{A}} + J) \exp\left(\xi \varphi\right)}{\delta \xi T (I U_{\mathrm{A}} + J)/2} \nonumber\\
    = & e^{3/37} + \frac{ 2 (c_2+3) \exp\left(\xi \varphi\right)}{\delta}\nonumber\\
    \le & \frac{31 I J U_{\mathrm{A}}}{\delta},
\end{align}
where the second line is by $\rho=1 - \frac{\xi T (I U_{\mathrm{A}} + J)}{2}$ and the last inequality is by \eqref{equ:condition-xi}, $c_2=5(I U_{\mathrm{A}} + J)$, and $\varphi = \frac{8(IU_{\mathrm{A}} + J)JT}{\delta}$.
Notice that if $\xi\le 0$, then $E\left[ \exp\left(\xi \|\boldsymbol{Q}(t)\|_2\right) \right] \le 1$.
Hence, the bound \eqref{equ:exp-bound-final} holds when 
$\xi \le \frac{3\delta}{ [5(I U_{\mathrm{A}} +J) +2] (I U_{\mathrm{A}} +J) T \left[ 1 + \frac{15(I U_{\mathrm{A}} +J) +6}{\delta} \right] }$. Note that we choose $t$ arbitrary at the beginning of this subsection. Hence, the proof holds for all $t$.
Theorem~\ref{theo:2} is proved.

\section{Proof of Corollary~\ref{cor:to-theo-2-tail}}
\label{app:sec:proof:cor:to-theo-2-tail}

In this section, we will present the proof of Corollary~\ref{cor:to-theo-2-tail}.

From Theorem~\ref{theo:2}, we have
\begin{align*}
    E\left[e^{\xi \|\boldsymbol{Q}(t)\|_2}\right]
    \le
    \frac{31 I J U_{\mathrm{A}}}{\delta},
\end{align*}
for all $\xi \le \frac{3\delta}{ g(\gamma) [5(I U_{\mathrm{A}} +J) +2] (I U_{\mathrm{A}} +J) \left[ 1 + \frac{15(I U_{\mathrm{A}} +J) +6}{\delta} \right] }$ and all $t$. 

Choose $\xi = \frac{3\delta}{ g(\gamma) [5(I U_{\mathrm{A}} +J) +2] (I U_{\mathrm{A}} +J) \left[ 1 + \frac{15(I U_{\mathrm{A}} +J) +6}{\delta} \right] }$. Let $x$ be any positive real number. Then by Markov's inequality, we have
\begin{align*}
    \Pr \left( \|\boldsymbol{Q}(t)\|_2 \ge x \right)
    = & \Pr \left(\exp\left(\xi\|\boldsymbol{Q}(t)\|_2\right) \ge \exp(\xi x) \right)\nonumber\\
    \le & \frac{E\left[e^{\xi \|\boldsymbol{Q}(t)\|_2}\right]}{e^{\xi x}} \nonumber\\
    \le & \frac{31 I J U_{\mathrm{A}}}{\delta}\exp(-\xi x).
\end{align*}

\section{Proof of Theorem~\ref{theo:3}}
\label{app:sec:proof:theo:3}

In this section, we will present the complete proof of Theorem~\ref{theo:3}.
In the proof, if server $j$ is not available at the beginning of time slot $t$, i.e., $\sum_i M_{i,j}(t) > 0$, we let $\hat{i}^*_{j}(t)=0$.

We present the proof of Theorem~\ref{theo:3} in the following subsections.
\subsection{Dividing the Time Horizon}

Firstly, we want to divide the time horizon into intervals. This step is similar to that in the proof of Theorem~\ref{theo:1}, but the length of each interval is no longer determined by the discount factor.
Since ${\boldsymbol \lambda}+\delta {\boldsymbol 1}\in {\mathcal C}'(W)$,
for any time slot $t$, there exists a $w(t)\in [1,W]$ that satisfies the inequality in the capacity region definition~\eqref{equ:def-capacity-stationary}.
Let $t_{\mathrm{S},0} = 0$ and $t_{\mathrm{S}, k} = t_{\mathrm{S}, k-1} + w(t_{\mathrm{S},k-1})$ for $k\ge 1$. Let $w_k\coloneqq w(t_{\mathrm{S}, k})$ for simplicity. Then the horizon is divided into intervals with length $w_0,w_1,\ldots,w_k,\ldots$, where the $k^{\mathrm{th}}$ interval is $[t_{\mathrm{S}, k}, t_{\mathrm{S}, k+1}]$. Note that $w_k, t_{\mathrm{S}, k}$ are fixed numbers rather than random variables.
In the next subsection, we will analyze the Lyapunov drift with this partition.

\subsection{Decomposing the Lyapunov Drift}
Consider the same Lyapunov function $L(t)\coloneqq \sum_i Q_i^2(t)$ as that in the proof of Theorem~\ref{theo:1}. Fix any time $t$. We first analyze the Lyapunov drift in each interval.
Following the same argument as \eqref{equ:drift-2} and \eqref{equ:drift-variance-term} in the proof of Theorem~\ref{theo:1}, we can obtain
\begin{align}
    E[L(t_{\mathrm{S}, k} + w_k) - L(t_{\mathrm{S}, k})] 
    = & \sum_{\tau=0}^{w_k - 1} E[L(t_{\mathrm{S}, k} + \tau+1) - L(t_{\mathrm{S}, k} +\tau)]\nonumber\\
    = & \sum_{\tau=0}^{w_k - 1} E\left[\sum_i 2Q_i(t_{\mathrm{S}, k} + \tau) A_i(t_{\mathrm{S}, k} + \tau)\right] \label{equ:arrival-term-station}\\
    & - \sum_{\tau=0}^{w_k - 1} E\left[2\sum_i Q_i(t_{\mathrm{S}, k} + \tau) \sum_j \mathbb{1}_{i,j}(t_{\mathrm{S}, k} + \tau)\right]\label{equ:service-term-station}\\
    & + (I U_{\mathrm{A}}^2 + J^2 + IJ^2)w_k.\nonumber
\end{align}

\subsection{Bounding the Arrival Term}
We first analyze the arrival term \eqref{equ:arrival-term-station}.
By law of iterated expectation, we have
\begin{align*}
    \eqref{equ:arrival-term-station} = & 
    \sum_{\tau=0}^{w_k - 1} E\left[ E\left[ \sum_i 2Q_i(t_{\mathrm{S}, k} + \tau) A_i(t_{\mathrm{S}, k} + \tau) \left| \boldsymbol{Q}(t_{\mathrm{S}, k} + \tau) \right.\right] \right]\nonumber\\
    = &
    \sum_{\tau=0}^{w_k - 1} E\left[ \sum_i 2Q_i(t_{\mathrm{S}, k}+\tau) E\left[ A_i(t_{\mathrm{S}, k}+\tau) \left| \boldsymbol{Q}(t_{\mathrm{S}, k}+\tau) \right.\right] \right]\nonumber\\
    = & \sum_{\tau=0}^{w_k - 1} E\left[ \sum_i 2Q_i(t_{\mathrm{S}, k}+\tau) \lambda_i(t_{\mathrm{S}, k}+\tau) \right],
\end{align*}
where the last inequality holds since $A_i(t_{\mathrm{S}, k}+\tau)$ is independent of $\boldsymbol{Q}(t_{\mathrm{S}, k}+\tau)$. By adding and subtracting $\delta$, we have
\begin{align}\label{equ:arrival-term-station-2}
    \eqref{equ:arrival-term-station} = & \sum_{\tau=0}^{w_k - 1} E\left[ \sum_i 2Q_i(t_{\mathrm{S}, k}+\tau) \left(\lambda_i(t_{\mathrm{S}, k}+\tau)+\delta\right) \right] -2\delta \sum_{\tau=0}^{w_k - 1} E\left[ \sum_i Q_i(t_{\mathrm{S}, k}+\tau) \right].
\end{align}
By Lemma~\ref{lemma:bounds-on-diff-queue-lens}, it holds that for any $\tau\in[0,w_k-1]$,
\begin{align*}
    Q_i(t_{\mathrm{S}, k}+\tau) \le Q_i(t_{\mathrm{S}, k}) + \tau U_{\mathrm{A}} \le Q_i(t_{\mathrm{S}, k}) + U_{\mathrm{A}} w_k.
\end{align*}
Hence, we have
\begin{align}\label{equ:bound-queue-lambda}
    & \sum_{\tau=0}^{w_k - 1} \sum_i 2Q_i(t_{\mathrm{S}, k}+\tau) \left(\lambda_i(t_{\mathrm{S}, k}+\tau)+\delta\right)   \nonumber\\
    \le & \sum_{\tau=0}^{w_k - 1} \sum_i 2 (Q_i(t_{\mathrm{S}, k}) + U_{\mathrm{A}} w_k) \left(\lambda_i(t_{\mathrm{S}, k}+\tau)+\delta\right) \nonumber\\
    = &  2 \sum_i Q_i(t_{\mathrm{S}, k}) \sum_{\tau=0}^{w_k - 1} \left(\lambda_i(t_{\mathrm{S}, k}+\tau)+\delta\right) + 2 U_{\mathrm{A}} w_k \sum_i \sum_{\tau=0}^{w_k - 1} \left(\lambda_i(t_{\mathrm{S}, k}+\tau)+\delta\right)\nonumber\\
    \le & 2 \sum_i Q_i(t_{\mathrm{S}, k}) \sum_{\tau=0}^{w_k - 1} \left(\lambda_i(t_{\mathrm{S}, k}+\tau)+\delta\right) + 2 I W U_{\mathrm{A}} (U_{\mathrm{A}} + 1) w_k,
\end{align}
where the last inequality is due to the facts that $\lambda_i(t_{\mathrm{S}, k}+\tau)+\delta \le U_{\mathrm{A}} + 1$ and $w_k\le W$. Since ${\boldsymbol \lambda}+\delta {\boldsymbol 1}\in {\mathcal C}'(W)$, by the definition of ${\cal C}'(W)$, we have
\begin{align}\label{equ:use-def-capacity}
    \sum_{\tau=0}^{w_k - 1} \left(\lambda_i(t_{\mathrm{S}, k}+\tau)+\delta\right) = \sum_{\tau=t_{\mathrm{S}, k}}^{t_{\mathrm{S}, k} + w_k - 1} \left(\lambda_i(\tau)+\delta\right) \le w_k \sum_j \alpha'_{i,j}(t_{\mathrm{S}, k}) \mu_{i,j}
\end{align}
Substituting \eqref{equ:use-def-capacity} into \eqref{equ:bound-queue-lambda}, we have
\begin{align}\label{equ:arrival-term-station-3}
    & \sum_{\tau=0}^{w_k - 1} \sum_i 2Q_i(t_{\mathrm{S}, k}+\tau) \left(\lambda_i(t_{\mathrm{S}, k}+\tau)+\delta\right)   \nonumber\\
    \le & 2 \sum_i Q_i(t_{\mathrm{S}, k}) w_k \sum_j \alpha'_{i,j}(t_{\mathrm{S}, k}) \mu_{i,j} + 2 I W U_{\mathrm{A}} (U_{\mathrm{A}} + 1) w_k\nonumber\\
    \le & 2 w_k \sum_j \max_i Q_i(t_{\mathrm{S}, k}) \mu_{i,j} \sum_{i'} \alpha'_{i',j}(t_{\mathrm{S}, k})  + 2 I W U_{\mathrm{A}} (U_{\mathrm{A}} + 1) w_k\nonumber\\
    \le & 2 w_k \sum_j \max_i Q_i(t_{\mathrm{S}, k}) \mu_{i,j} + 2 I W U_{\mathrm{A}} (U_{\mathrm{A}} + 1) w_k,
\end{align}
where the last inequality uses $\sum_{i'} \alpha'_{i',j}(t_{\mathrm{S}, k})\le 1$ in the definition of ${\cal C}'(W)$.
Recall the definition of $f_j(\cdot)$, which maps a time slot to another
time slot such that if $y=f_j(x)$ then y is the time slot when server j picked the job that was being served at server j in time slot x. If server $j$ is idling in time slot $x$, then $f_j(x)=x$. Note that for $\tau\in[0, w_k-1]$, we have
\begin{align*}
    t_{\mathrm{S}, k} - f_j(t_{\mathrm{S}, k}+\tau) = \left[t_{\mathrm{S}, k} - (t_{\mathrm{S}, k} + \tau)\right] + \left[(t_{\mathrm{S}, k}+\tau) - f_j(t_{\mathrm{S}, k}+\tau)\right]
    \le U_{\mathrm{S}}.
\end{align*}
Similarly,
\begin{align*}
    f_j(t_{\mathrm{S}, k}+\tau) - t_{\mathrm{S}, k} 
    = \left[f_j(t_{\mathrm{S}, k}+\tau) - (t_{\mathrm{S}, k} + \tau) \right] + \left[(t_{\mathrm{S}, k} + \tau) - t_{\mathrm{S}, k} \right] 
    \le w_k \le W.
\end{align*}
Hence, by Lemma~\ref{lemma:bounds-on-diff-queue-lens}, we have
\begin{itemize}
    \item If $t_{\mathrm{S}, k} \ge f_j(t_{\mathrm{S}, k}+\tau)$, then
    \begin{align*}
        Q_i(t_{\mathrm{S}, k}) \le Q_i(f_j(t_{\mathrm{S}, k}+\tau)) + U_{\mathrm{S}} U_{\mathrm{A}}.
    \end{align*}
    \item If $t_{\mathrm{S}, k} < f_j(t_{\mathrm{S}, k}+\tau)$, then
    \begin{align*}
        Q_i(t_{\mathrm{S}, k}) \le Q_i(f_j(t_{\mathrm{S}, k}+\tau)) + J W.
    \end{align*}
\end{itemize}
Hence, we have
\begin{align}\label{equ:queue-bound-station-temp}
    Q_i(t_{\mathrm{S}, k}) \le Q_i(f_j(t_{\mathrm{S}, k}+\tau)) + \max\{U_{\mathrm{S}} U_{\mathrm{A}}, J W\}.
\end{align}
Substituting \eqref{equ:queue-bound-station-temp} into \eqref{equ:arrival-term-station-3} and using the fact that $\mu_{i,j}\le 1$, we have
\begin{align}\label{equ:arrival-term-station-4}
    & \sum_{\tau=0}^{w_k - 1} \sum_i 2Q_i(t_{\mathrm{S}, k}+\tau) \left(\lambda_i(t_{\mathrm{S}, k}+\tau)+\delta\right)   \nonumber\\
    \le & 2 \sum_j \sum_{\tau=0}^{w_k-1} \max_i Q_i(f_j(t_{\mathrm{S}, k}+\tau)) \mu_{i,j} 
    + 2 J \max\{U_{\mathrm{S}} U_{\mathrm{A}}, J W\} w_k
    + 2 I W U_{\mathrm{A}} (U_{\mathrm{A}} + 1) w_k,
\end{align}
Substituting \eqref{equ:arrival-term-station-4} into \eqref{equ:arrival-term-station-2}, we have
\begin{align}\label{equ:arrival-term-station-final}
    \eqref{equ:arrival-term-station} \le &  2 \sum_j E\left[ \sum_{\tau=0}^{w_k-1} \max_i Q_i(f_j(t_{\mathrm{S}, k}+\tau)) \mu_{i,j} \right] -2\delta \sum_{\tau=0}^{w_k - 1} E\left[ \sum_i Q_i(t_{\mathrm{S}, k}+\tau) \right] \nonumber\\
    &+ 2 J \max\{U_{\mathrm{S}} U_{\mathrm{A}}, J W\} w_k + 2 I W U_{\mathrm{A}} (U_{\mathrm{A}} + 1) w_k.
\end{align}

\subsection{Bounding the Service Term}
In this section, we will analyze the service term \eqref{equ:service-term-station}. Let us first fix any server $j$. We first consider the following per-server service term:
\begin{align*}
    \sum_{\tau=0}^{w_k - 1} E\left[\sum_i Q_i(t_{\mathrm{S}, k} + \tau) \mathbb{1}_{i,j}(t_{\mathrm{S}, k} + \tau)\right].
\end{align*}
The process of bounding the service term is the main difference between this proof and the proof of Theorem~\ref{theo:1}. It takes several steps, which are shown in the following.

Firstly, by the definition of $I_j(t_{\mathrm{S}, k} + \tau)$, we have
\begin{align*}
    \sum_{\tau=0}^{w_k - 1} E\left[\sum_i Q_i(t_{\mathrm{S}, k} + \tau) \mathbb{1}_{i,j}(t_{\mathrm{S}, k} + \tau)\right] = \sum_{\tau=0}^{w_k - 1} E\left[Q_{I_j(t_{\mathrm{S}, k} + \tau)}(t_{\mathrm{S}, k} + \tau) \mathbb{1}_{I_j(t_{\mathrm{S}, k} + \tau),j}(t_{\mathrm{S}, k} + \tau)\right].
\end{align*}
By multiplying and dividing the same term, the per-server service term can be further rewritten as
\begin{align*}
    & \sum_{\tau=0}^{w_k - 1} E\left[\sum_i Q_i(t_{\mathrm{S}, k} + \tau) \mathbb{1}_{i,j}(t_{\mathrm{S}, k} + \tau)\right] \nonumber\\
    = & E\left[ \sum_{\tau=0}^{w_k - 1}
    Q_{I_j(t_{\mathrm{S}, k} + \tau)}(t_{\mathrm{S}, k} + \tau)
    \mu_{I_j(t_{\mathrm{S}, k} + \tau),j}
    \frac{\mathbb{1}_{I_j(t_{\mathrm{S}, k} + \tau),j}(t_{\mathrm{S}, k} + \tau)}{ \mu_{I_j(t_{\mathrm{S}, k} + \tau),j} }
    \right].
\end{align*}
By adding and subtracting the same term, we further have
\begin{align}\label{equ:per-server-service-1}
    & \sum_{\tau=0}^{w_k - 1} E\left[\sum_i Q_i(t_{\mathrm{S}, k} + \tau) \mathbb{1}_{i,j}(t_{\mathrm{S}, k} + \tau)\right] \nonumber\\
    = & E\left[\sum_{\tau=0}^{w_k - 1}
    \max_i Q_i(f_j(t_{\mathrm{S}, k} + \tau)) \mu_{i,j}
    \frac{\mathbb{1}_{I_j(t_{\mathrm{S}, k} + \tau),j}(t_{\mathrm{S}, k} + \tau)}{ \mu_{I_j(t_{\mathrm{S}, k} + \tau),j} }\right]\nonumber\\
    & + E\left[ \sum_{\tau=0}^{w_k - 1}
    \left(Q_{I_j(t_{\mathrm{S}, k} + \tau)}(t_{\mathrm{S}, k} + \tau)
    \mu_{I_j(t_{\mathrm{S}, k} + \tau),j} 
     - \max_i Q_i(f_j(t_{\mathrm{S}, k} + \tau)) \mu_{i,j} 
    \right)
    \frac{\mathbb{1}_{I_j(t_{\mathrm{S}, k} + \tau),j}(t_{\mathrm{S}, k} + \tau)}{ \mu_{I_j(t_{\mathrm{S}, k} + \tau),j} }\right].
\end{align}
We now consider the term
\begin{align*}
    Q_{I_j(t_{\mathrm{S}, k} + \tau)}(t_{\mathrm{S}, k} + \tau) \mu_{I_j(t_{\mathrm{S}, k} + \tau),j} 
    - \max_i Q_i(f_j(t_{\mathrm{S}, k} + \tau)) \mu_{i,j}
\end{align*}
in \eqref{equ:per-server-service-1}.
For any time slot $\tau$, any server $j$, we define the event ${\cal E}_{\mathrm{S},\tau,j}$ as follows:
\begin{align}\label{equ:concentration-event-3}
    {\cal E}_{\mathrm{S},\tau,j} \coloneqq \left\{ \mbox{for all } i, 
    \left| \frac{1}{\hat{\mu}_{i,j}(\tau)} - \frac{1}{\mu_{i,j}} \right| \le b_{i,j}(\tau) \right\}.
\end{align}
We first consider the situation that the event ${\cal E}_{\mathrm{S},f_j(t_{\mathrm{S}, k} + \tau),j}$ holds.
Define
\begin{align*}
    \bar{\mu}_{I_j(t_{\mathrm{S}, k}+\tau),j}(f_j(t_{\mathrm{S}, k}+\tau))
    \coloneqq &
    \frac{1}{\max\left\{\frac{1}{\hat{\mu}_{I_j(t_{\mathrm{S}, k}+\tau),j}(f_j(t_{\mathrm{S}, k}+\tau))}-b_{I_j(t_{\mathrm{S}, k}+\tau),j}(f_j(t_{\mathrm{S}, k}+\tau)), 1\right\}}\nonumber\\
    \underline{\mu}_{I_j(t_{\mathrm{S}, k}+\tau),j}(f_j(t_{\mathrm{S}, k}+\tau))
    \coloneqq &
    \frac{1}{\frac{1}{\hat{\mu}_{I_j(t_{\mathrm{S}, k}+\tau),j}(f_j(t_{\mathrm{S}, k}+\tau))}+b_{I_j(t_{\mathrm{S}, k}+\tau),j}(f_j(t_{\mathrm{S}, k}+\tau))}
\end{align*}
By the definition of the event ${\cal E}_{\mathrm{S},f_j(t_{\mathrm{S}, k} + \tau),j}$ in~\eqref{equ:concentration-event-3}, we have
\begin{align*}
    \frac{1}{\hat{\mu}_{I_j(t_{\mathrm{S}, k}+\tau),j}(f_j(t_{\mathrm{S}, k}+\tau))}-b_{I_j(t_{\mathrm{S}, k}+\tau),j}(f_j(t_{\mathrm{S}, k}+\tau)) \le & \frac{1}{\mu_{I_j(t_{\mathrm{S}, k}+\tau),j}}\nonumber\\
    \frac{1}{\hat{\mu}_{I_j(t_{\mathrm{S}, k}+\tau),j}(f_j(t_{\mathrm{S}, k}+\tau))} + b_{I_j(t_{\mathrm{S}, k}+\tau),j}(f_j(t_{\mathrm{S}, k}+\tau)) \ge & \frac{1}{\mu_{I_j(t_{\mathrm{S}, k}+\tau),j}}.
\end{align*}
Therefore, combining the above inequalities with the fact that $\frac{1}{\mu_{I_j(t_{\mathrm{S}, k}+\tau),j}}\ge 1$, we have
\begin{align}\label{equ:high-prob-event}
    \underline{\mu}_{I_j(t_{\mathrm{S}, k}+\tau),j}(f_j(t_{\mathrm{S}, k}+\tau)) 
    \le \mu_{I_j(t_{\mathrm{S}, k}+\tau),j}
    \le \bar{\mu}_{I_j(t_{\mathrm{S}, k}+\tau),j}(f_j(t_{\mathrm{S}, k}+\tau)).
\end{align}
Hence,
when ${\cal E}_{\mathrm{S},f_j(t_{\mathrm{S}, k} + \tau),j}$ holds, we have
\begin{align}\label{equ:q-mu-lower}
    & Q_{I_j(t_{\mathrm{S}, k} + \tau)}(t_{\mathrm{S}, k} + \tau) \mu_{I_j(t_{\mathrm{S}, k} + \tau),j}  \nonumber\\
    = & Q_{I_j(t_{\mathrm{S}, k} + \tau)}(t_{\mathrm{S}, k} + \tau)
    \bar{\mu}_{I_j(t_{\mathrm{S}, k} + \tau),j} (f_j(t_{\mathrm{S}, k} + \tau)) \nonumber\\
    & + Q_{I_j(t_{\mathrm{S}, k} + \tau)}(t_{\mathrm{S}, k} + \tau) \left( \mu_{I_j(t_{\mathrm{S}, k} + \tau),j}
    - \bar{\mu}_{I_j(t_{\mathrm{S}, k} + \tau),j} (f_j(t_{\mathrm{S}, k} + \tau))
    \right)\nonumber\\
    \ge & Q_{I_j(t_{\mathrm{S}, k} + \tau)}(t_{\mathrm{S}, k} + \tau)
    \bar{\mu}_{I_j(t_{\mathrm{S}, k} + \tau),j} (f_j(t_{\mathrm{S}, k} + \tau)) \nonumber\\
    & + 
    Q_{I_j(t_{\mathrm{S}, k} + \tau)}(t_{\mathrm{S}, k} + \tau) \left( \underline{\mu}_{I_j(t_{\mathrm{S}, k} + \tau),j} (f_j(t_{\mathrm{S}, k} + \tau))
    - \bar{\mu}_{I_j(t_{\mathrm{S}, k} + \tau),j} (f_j(t_{\mathrm{S}, k} + \tau))
    \right)
\end{align}
Following the same argument as \eqref{equ:mu-difference-bound} in the proof of Theorem~\ref{theo:1}, we have
\begin{align}\label{equ:mu-diff-bound-station}
    \underline{\mu}_{I_j(t_{\mathrm{S}, k} + \tau),j} (f_j(t_{\mathrm{S}, k} + \tau))
    - \bar{\mu}_{I_j(t_{\mathrm{S}, k} + \tau),j} (f_j(t_{\mathrm{S}, k} + \tau)) \ge -\min \{ 2 b_{I_j(t_{\mathrm{S}, k}+\tau),j}(f_j(t_{\mathrm{S}, k}+\tau)), 1 \}.
\end{align}
Combining \eqref{equ:q-mu-lower} and \eqref{equ:mu-diff-bound-station}, we have
\begin{align}\label{equ:q-mu-lower-2}
    & Q_{I_j(t_{\mathrm{S}, k} + \tau)}(t_{\mathrm{S}, k} + \tau) \mu_{I_j(t_{\mathrm{S}, k} + \tau),j}  \nonumber\\
    \ge & Q_{I_j(t_{\mathrm{S}, k} + \tau)}(t_{\mathrm{S}, k} + \tau)
    \bar{\mu}_{I_j(t_{\mathrm{S}, k} + \tau),j} (f_j(t_{\mathrm{S}, k} + \tau))\nonumber\\
    & - Q_{I_j(t_{\mathrm{S}, k} + \tau)}(t_{\mathrm{S}, k} + \tau)
    \min \{ 2 b_{I_j(t_{\mathrm{S}, k}+\tau),j}(f_j(t_{\mathrm{S}, k}+\tau)), 1 \},
\end{align}
which holds when the event ${\cal E}_{\mathrm{S},f_j(t_{\mathrm{S}, k} + \tau),j}$ holds. By Lemma~\ref{lemma:bounds-on-diff-queue-lens} and the fact that $0\le t_{\mathrm{S}, k} + \tau - f_j(t_{\mathrm{S}, k} + \tau) \le U_{\mathrm{S}}$, \eqref{equ:q-mu-lower-2} can be further bounded by
\begin{align*}
    & Q_{I_j(t_{\mathrm{S}, k} + \tau)}(t_{\mathrm{S}, k} + \tau) \mu_{I_j(t_{\mathrm{S}, k} + \tau),j}  \nonumber\\
    \ge & Q_{I_j(t_{\mathrm{S}, k} + \tau)}(f_j(t_{\mathrm{S}, k} + \tau))
    \bar{\mu}_{I_j(t_{\mathrm{S}, k} + \tau),j} (f_j(t_{\mathrm{S}, k} + \tau)) - J U_{\mathrm{S}}\nonumber\\
    & - Q_{I_j(t_{\mathrm{S}, k} + \tau)}(t_{\mathrm{S}, k} + \tau)
    \min \{ 2 b_{I_j(t_{\mathrm{S}, k}+\tau),j}(f_j(t_{\mathrm{S}, k}+\tau)), 1 \}\nonumber\\
    = & \max_i Q_i(f_j(t_{\mathrm{S}, k}+\tau))
    \bar{\mu}_{i,j}(f_j(t_{\mathrm{S}, k}+\tau))
    - J U_{\mathrm{S}}\nonumber\\
    & - Q_{I_j(t_{\mathrm{S}, k} + \tau)}(t_{\mathrm{S}, k} + \tau)
    \min \{ 2 b_{I_j(t_{\mathrm{S}, k}+\tau),j}(f_j(t_{\mathrm{S}, k}+\tau)), 1 \}\nonumber\\
    \ge & \max_i Q_i(f_j(t_{\mathrm{S}, k}+\tau))
    \mu_{i,j}
    - J U_{\mathrm{S}}
    - Q_{I_j(t_{\mathrm{S}, k} + \tau)}(t_{\mathrm{S}, k} + \tau)
    \min \{ 2 b_{I_j(t_{\mathrm{S}, k}+\tau),j}(f_j(t_{\mathrm{S}, k}+\tau)), 1 \},
\end{align*}
where the equality is due to Line~\ref{alg:line:max-weight} of Algorithm~\ref{alg:1}, and the last inequality uses \eqref{equ:high-prob-event} when the event ${\cal E}_{\mathrm{S},f_j(t_{\mathrm{S}, k} + \tau),j}$ holds. Hence, when the event ${\cal E}_{\mathrm{S},f_j(t_{\mathrm{S}, k} + \tau),j}$ holds, we have
\begin{align}\label{equ:error-event-hold}
    & Q_{I_j(t_{\mathrm{S}, k} + \tau)}(t_{\mathrm{S}, k} + \tau) \mu_{I_j(t_{\mathrm{S}, k} + \tau),j} 
    - \max_i Q_i(f_j(t_{\mathrm{S}, k} + \tau)) \mu_{i,j}\nonumber\\
    \ge & 
    - J U_{\mathrm{S}}
    - Q_{I_j(t_{\mathrm{S}, k} + \tau)}(t_{\mathrm{S}, k} + \tau)
    \min \{ 2 b_{I_j(t_{\mathrm{S}, k}+\tau),j}(f_j(t_{\mathrm{S}, k}+\tau)), 1 \}.
\end{align}
If the event ${\cal E}_{\mathrm{S},f_j(t_{\mathrm{S}, k} + \tau),j}$ does not hold, then we have
\begin{align}\label{equ:error-event-not-hold}
    & Q_{I_j(t_{\mathrm{S}, k} + \tau)}(t_{\mathrm{S}, k} + \tau) \mu_{I_j(t_{\mathrm{S}, k} + \tau),j} 
    - \max_i Q_i(f_j(t_{\mathrm{S}, k} + \tau)) \mu_{i,j}\nonumber\\
    \ge &  - \max_i Q_i(f_j(t_{\mathrm{S}, k} + \tau)) \mu_{i,j}
    \ge - \max_i Q_i(f_j(t_{\mathrm{S}, k} + \tau))
    \ge - \sum_i Q_i(f_j(t_{\mathrm{S}, k} + \tau)).
\end{align}
Combining \eqref{equ:error-event-hold} and \eqref{equ:error-event-not-hold}, we have
\begin{align}\label{equ:error-each-time-slot}
    & Q_{I_j(t_{\mathrm{S}, k} + \tau)}(t_{\mathrm{S}, k} + \tau) \mu_{I_j(t_{\mathrm{S}, k} + \tau),j} 
    - \max_i Q_i(f_j(t_{\mathrm{S}, k} + \tau)) \mu_{i,j}\nonumber\\
    \ge &  - \mathbb{1}_{{\cal E}_{\mathrm{S},f_j(t_{\mathrm{S}, k} + \tau),j}} 
    \left(J U_{\mathrm{S}}
    + Q_{I_j(t_{\mathrm{S}, k} + \tau)}(t_{\mathrm{S}, k} + \tau)
    \min \{ 2 b_{I_j(t_{\mathrm{S}, k}+\tau),j}(f_j(t_{\mathrm{S}, k}+\tau)), 1 \}\right) \nonumber\\
    & - \mathbb{1}_{{\cal E}^{\mathrm{c}}_{\mathrm{S},f_j(t_{\mathrm{S}, k} + \tau),j}}
    \biggl(
    \sum_i Q_i(f_j(t_{\mathrm{S}, k} + \tau))
    \biggr).
\end{align}
Substituting \eqref{equ:error-each-time-slot} into \eqref{equ:per-server-service-1}, we have
\begin{align}\label{equ:per-server-service-2}
    & \sum_{\tau=0}^{w_k - 1} E\left[\sum_i Q_i(t_{\mathrm{S}, k} + \tau) \mathbb{1}_{i,j}(t_{\mathrm{S}, k} + \tau)\right] \nonumber\\
    \ge & E\left[\sum_{\tau=0}^{w_k - 1}
    \max_i Q_i(f_j(t_{\mathrm{S}, k} + \tau)) \mu_{i,j}
    \frac{\mathbb{1}_{I_j(t_{\mathrm{S}, k} + \tau),j}(t_{\mathrm{S}, k} + \tau)}{ \mu_{I_j(t_{\mathrm{S}, k} + \tau),j} }\right]\nonumber\\
    & - E\Biggl[ \sum_{\tau=0}^{w_k - 1}
    \mathbb{1}_{{\cal E}_{\mathrm{S},f_j(t_{\mathrm{S}, k} + \tau),j}} 
    \left(J U_{\mathrm{S}}
    + Q_{I_j(t_{\mathrm{S}, k} + \tau)}(t_{\mathrm{S}, k} + \tau)
    \min \{ 2 b_{I_j(t_{\mathrm{S}, k}+\tau),j}(f_j(t_{\mathrm{S}, k}+\tau)), 1 \} 
    \right)\nonumber\\
    & \qquad \qquad \frac{\mathbb{1}_{I_j(t_{\mathrm{S}, k} + \tau),j}(t_{\mathrm{S}, k} + \tau)}{ \mu_{I_j(t_{\mathrm{S}, k} + \tau),j} }\Biggr]\nonumber\\
    & - E\left[ \sum_{\tau=0}^{w_k - 1}
    \mathbb{1}_{{\cal E}^{\mathrm{c}}_{\mathrm{S},f_j(t_{\mathrm{S}, k} + \tau),j}}
    \sum_i Q_i(f_j(t_{\mathrm{S}, k} + \tau))
    \frac{\mathbb{1}_{I_j(t_{\mathrm{S}, k} + \tau),j}(t_{\mathrm{S}, k} + \tau)}{ \mu_{I_j(t_{\mathrm{S}, k} + \tau),j} }\right]\nonumber\\
    \ge & E\left[\sum_{\tau=0}^{w_k - 1}
    \max_i Q_i(f_j(t_{\mathrm{S}, k} + \tau)) \mu_{i,j}
    \frac{\mathbb{1}_{I_j(t_{\mathrm{S}, k} + \tau),j}(t_{\mathrm{S}, k} + \tau)}{ \mu_{I_j(t_{\mathrm{S}, k} + \tau),j} }\right]\nonumber\\
    & - U_{\mathrm{S}} E\left[ \sum_{\tau=0}^{w_k - 1}
    Q_{I_j(t_{\mathrm{S}, k} + \tau)}(t_{\mathrm{S}, k} + \tau)
    \min \{ 2 b_{I_j(t_{\mathrm{S}, k}+\tau),j}(f_j(t_{\mathrm{S}, k}+\tau)), 1 \}
    \mathbb{1}_{I_j(t_{\mathrm{S}, k} + \tau),j}(t_{\mathrm{S}, k} + \tau)\right]
    \nonumber\\
    & - U_{\mathrm{S}} E\left[ \sum_{\tau=0}^{w_k - 1} \sum_i
    \mathbb{1}_{{\cal E}^{\mathrm{c}}_{\mathrm{S},f_j(t_{\mathrm{S}, k} + \tau),j}}
    Q_i(f_j(t_{\mathrm{S}, k} + \tau))
    \mathbb{1}_{I_j(t_{\mathrm{S}, k} + \tau),j}(t_{\mathrm{S}, k} + \tau) \right]\nonumber\\
    & - J U^2_{\mathrm{S}} w_k,
\end{align}
where the last inequality uses the service time bound $1/ \mu_{I_j(t_{\mathrm{S}, k} + \tau),j}\le U_{\mathrm{S}}$.

\subsection{Telescoping Sum}
Combining \eqref{equ:arrival-term-station}, \eqref{equ:service-term-station}, \eqref{equ:arrival-term-station-final}, and \eqref{equ:per-server-service-2}, we have
\begin{align*}
    & E[L(t_{\mathrm{S}, k} + w_k) - L(t_{\mathrm{S}, k})] \nonumber\\
    \le & 2 \sum_j E\left[ \sum_{\tau=0}^{w_k-1} \max_i Q_i(f_j(t_{\mathrm{S}, k}+\tau)) \mu_{i,j} \right] -2\delta \sum_{\tau=0}^{w_k - 1} E\left[ \sum_i Q_i(t_{\mathrm{S}, k}+\tau) \right] \nonumber\\
    & - 2 \sum_j E\left[\sum_{\tau=0}^{w_k - 1}
    \max_i Q_i(f_j(t_{\mathrm{S}, k} + \tau)) \mu_{i,j}
    \frac{\mathbb{1}_{I_j(t_{\mathrm{S}, k} + \tau),j}(t_{\mathrm{S}, k} + \tau)}{ \mu_{I_j(t_{\mathrm{S}, k} + \tau),j} }\right]\nonumber\\
    & + 2 U_{\mathrm{S}} \sum_j  E\left[ \sum_{\tau=0}^{w_k - 1}
    Q_{I_j(t_{\mathrm{S}, k} + \tau)}(t_{\mathrm{S}, k} + \tau)
    \min \{ 2 b_{I_j(t_{\mathrm{S}, k}+\tau),j}(f_j(t_{\mathrm{S}, k}+\tau)), 1 \}
    \mathbb{1}_{I_j(t_{\mathrm{S}, k} + \tau),j}(t_{\mathrm{S}, k} + \tau)\right]
    \nonumber\\
    & + 2 U_{\mathrm{S}}  \sum_j E\left[ \sum_{\tau=0}^{w_k - 1} \sum_i
    \mathbb{1}_{{\cal E}^{\mathrm{c}}_{\mathrm{S},f_j(t_{\mathrm{S}, k} + \tau),j}}
    Q_i(f_j(t_{\mathrm{S}, k} + \tau))
    \mathbb{1}_{I_j(t_{\mathrm{S}, k} + \tau),j}(t_{\mathrm{S}, k} + \tau) \right]\nonumber\\
    & + 2 J \max\{U_{\mathrm{S}} U_{\mathrm{A}}, J W\} w_k + 2 I W U_{\mathrm{A}} (U_{\mathrm{A}} + 1) w_k  + 2 J^2 U^2_{\mathrm{S}} w_k  + (I U_{\mathrm{A}}^2 + J^2 + IJ^2)w_k.
\end{align*}
Taking summation over $k=0, 1,\ldots, K-1$, we have
\begin{align}
    E[L(t_{\mathrm{S}, K}) - L(0)]
    \le & 2 \sum_j E\left[ \sum_{\tau=0}^{\sum_{k=0}^{K-1} w_k-1} \max_i Q_i(f_j(\tau)) \mu_{i,j} \right] -2\delta \sum_{\tau=0}^{\sum_{k=0}^{K-1}w_k - 1} E\left[ \sum_i Q_i(\tau) \right] \nonumber\\
    & - 2 \sum_j E\left[\sum_{\tau=0}^{\sum_{k=0}^{K-1} w_k - 1}
    \max_i Q_i(f_j(\tau)) \mu_{i,j}
    \frac{\mathbb{1}_{I_j(\tau),j}(\tau)}{ \mu_{I_j(\tau),j} }\right]\label{equ:queue-indicators}\\
    & + 2 U_{\mathrm{S}} \sum_j  E\left[ \sum_{\tau=0}^{\sum_{k=0}^{K-1} w_k - 1}
    Q_{I_j(\tau)}(\tau)
    \min \{ 2 b_{I_j(\tau),j}(f_j(\tau)), 1 \}
    \mathbb{1}_{I_j(\tau),j}(\tau)\right]
    \label{equ:sum-queue-ucb}\\
    & + 2 U_{\mathrm{S}}  \sum_j E\left[ \sum_{\tau=0}^{\sum_{k=0}^{K-1}w_k - 1} \sum_i
    \mathbb{1}_{{\cal E}^{\mathrm{c}}_{\mathrm{S},f_j(\tau),j}}
    Q_i(f_j(\tau))
    \mathbb{1}_{I_j(\tau),j}(\tau) \right]\label{equ:coupling-term}\\
    & + \left(2 J \max\{U_{\mathrm{S}} U_{\mathrm{A}}, J W\} + 
    2 I W U_{\mathrm{A}} (U_{\mathrm{A}} + 1) +
    2 J^2 U^2_{\mathrm{S}} + 
    I U_{\mathrm{A}}^2 + J^2 + IJ^2\right)\sum_{k=0}^{K-1}w_k.\nonumber
\end{align}

\subsection{Decoupling the Queue Length and Concentration}
In this subsection, we consider the term \eqref{equ:coupling-term}.
In fact, due to the indicator $\mathbb{1}_{I_j(\tau),j}(\tau)$ and the definition of $f_j(\tau)$, the sum in \eqref{equ:coupling-term} is taken over only the time slots in which there is a job starting.
Actually we can bound this sum by another sum which is taken over all time slots as follows:
\begin{align}\label{equ:coupling-term-2}
    E\left[ \sum_{\tau=0}^{\sum_{k=0}^{K-1}w_k - 1} \sum_i
    \mathbb{1}_{{\cal E}^{\mathrm{c}}_{\mathrm{S},f_j(\tau),j}}
    Q_i(f_j(\tau))
    \mathbb{1}_{I_j(\tau),j}(\tau) \right]
    \le E\left[ \sum_{\tau=0}^{\sum_{k=0}^{K-1}w_k - 1} \sum_i
    \mathbb{1}_{{\cal E}^{\mathrm{c}}_{\mathrm{S},\tau,j}}
    Q_i(\tau) \right].
\end{align}
First notice that the event ${\cal E}^{\mathrm{c}}_{\mathrm{S},\tau,j}$ is expected to have a small probability by the definition of ${\cal E}_{\mathrm{S},\tau,j}$ in \eqref{equ:concentration-event-3} and concentration inequalities, which is shown in the following lemma:
\begin{lemma}\label{lemma:concentration-3}
    Consider Algorithm~\ref{alg:1} with $c_1=2$ and $\gamma=1$. For any time slot $\tau\ge 1$ and any server $j$, we have
    \begin{align*}
        \Pr\left({\cal E}^{\mathrm{c}}_{\mathrm{S},\tau,j}\right) \le \frac{10 I}{\tau^6},
    \end{align*}
\end{lemma}
Proof of Lemma~\ref{lemma:concentration-3} can be found in Section~\ref{app:proof-lemma-concentration-3}.
Note that the event ${\cal E}^{\mathrm{c}}_{\mathrm{S},\tau,j}$ is correlated with the queue lengths, which is the main difficulty of bounding the above the summation \eqref{equ:coupling-term-2}. Here we borrow the proof idea of Lemma 5.4 and Lemma 5.5 in \cite{FreLykWen_22} to decouple the queue length and the indicator, as shown in the following lemma:
\begin{lemma}\label{lemma:sum-error-queue}
    Consider Algorithm~\ref{alg:1} with $\gamma=1$. For any $t$, we have
    \begin{align*}
        E\left[\sum_{\tau=0}^{t - 1} \sum_i
        \mathbb{1}_{{\cal E}^{\mathrm{c}}_{\mathrm{S},\tau,j}}
        Q_i(\tau) \right] 
        \le \frac{\delta}{6 J U_{\mathrm{S}} } E\left[\sum_{\tau=0}^{t-1} \sum_i Q_i(\tau)\right] + \frac{6IJU_{\mathrm{S}}U_{\mathrm{A}}}{\delta} 
        \sum_{\tau=1}^{t-1} (2\tau - 1) \Pr \left({\cal E}^{\mathrm{c}}_{\mathrm{S},\tau,j}\right).
    \end{align*}
\end{lemma}
Proof of Lemma~\ref{lemma:sum-error-queue} can be found in Section~\ref{app:proof-lemma-sum-error-queue}. Then from Lemma~\ref{lemma:concentration-3} and Lemma~\ref{lemma:sum-error-queue}, we have for any $t$,
\begin{align}\label{equ:bound-error-Q}
    E\left[\sum_{\tau=0}^{t - 1} \sum_i
    \mathbb{1}_{{\cal E}^{\mathrm{c}}_{\mathrm{S},\tau,j}}
    Q_i(\tau) \right]
    \le & \frac{\delta}{6 J U_{\mathrm{S}} } E\left[\sum_{\tau=0}^{t-1} \sum_i Q_i(\tau)\right] + \frac{6IJU_{\mathrm{S}}U_{\mathrm{A}}}{\delta} 
    \sum_{\tau=1}^{t-1} (2\tau - 1) \frac{10 I}{\tau^6}\nonumber\\
    \le & \frac{\delta}{6 J U_{\mathrm{S}} } E\left[\sum_{\tau=0}^{t-1} \sum_i Q_i(\tau)\right] + \frac{120 I^2 JU_{\mathrm{S}}U_{\mathrm{A}}}{\delta} 
    \sum_{\tau=1}^{t-1} \frac{1}{\tau^5}\nonumber\\
    \le & \frac{\delta}{6 J U_{\mathrm{S}} } E\left[\sum_{\tau=0}^{t-1} \sum_i Q_i(\tau)\right] + \frac{150 I^2 JU_{\mathrm{S}}U_{\mathrm{A}}}{\delta},
\end{align}
where the last inequality holds since $\sum_{\tau=1}^{t-1} \frac{1}{\tau^5} \le 1 + \int_{\tau=1}^{t-1} \tau^{-5} d \tau \le \frac{5}{4} - \frac{1}{4(t-1)^4}\le \frac{5}{4}$ by integration.

Note that the proof idea of getting the bound~\eqref{equ:bound-error-Q} cannot be applied to the MaxWeight with discounted UCB algorithm (Algorithm~\ref{alg:1} with $\gamma<1$) to get a similar result because the probability of the complement of the concentration event in the discounting case does not decrease to $0$ as $\tau$ increases to infinity, i.e., Lemma~\ref{lemma:concentration-3} does not hold for $\gamma<1$. In fact, in the discounting case, the probability of error is always lower bounded by a constant no matter how long the horizon is due to the discount factor.

Combining \eqref{equ:coupling-term}, \eqref{equ:coupling-term-2}, and \eqref{equ:bound-error-Q}, we have
\begin{align}\label{equ:bound-coupling-term}
    \eqref{equ:coupling-term} \le & \frac{\delta}{3} E\left[\sum_{\tau=0}^{\sum_{k=0}^{K-1}w_k-1} \sum_i Q_i(\tau)\right] + \frac{300 I^2 J^2 U^2_{\mathrm{S}} U_{\mathrm{A}}}{\delta}.
\end{align}

\subsection{Bounding the Sum of UCB Bonuses}
In this subsection, we consider bounding the sum of UCB bonuses \eqref{equ:sum-queue-ucb}.
Bounding \eqref{equ:sum-queue-ucb} is similar to the process of bounding \eqref{equ:ucb-summation-term-1} in the proof of Theorem~\ref{theo:1}. We will present the proof here for completeness.  
Define for any $i,j,t$,
\begin{align*}
    \tilde{b}_{i,j}(t) \coloneqq \min \{ 2 b_{i,j}(t), 1\}
\end{align*}
for ease of notation.
Recall the definition of the waiting queue $\tilde{Q}$.
Following the same argument as that of proving \eqref{equ:idling-queue-3} in the proof of Theorem~\ref{theo:1}, we have
\begin{align}\label{equ:idling-queue-theo-3}
    Q_{I_j(\tau)}( \tau) \le  Q_{I_j( \tau)}( \tau)\eta_j(f_j( \tau)) + J.
\end{align}
Hence, by \eqref{equ:idling-queue-theo-3} and the fact that $\tilde{b}_{I_j(\tau), j}(f_j(\tau))\le 1$, we have
\begin{align}\label{equ:sum-ucb-1}
    & E\left[ \sum_{\tau=0}^{\sum_{k=0}^{K-1} w_k - 1}
    Q_{I_j(\tau)}(\tau)
    \min \{ 2 b_{I_j(\tau),j}(f_j(\tau)), 1 \}
    \mathbb{1}_{I_j(\tau),j}(\tau)\right]\nonumber\\
     = & E\left[ \sum_{\tau=0}^{\sum_{k=0}^{K-1} w_k - 1}
    Q_{I_j(\tau)}(\tau)
    \tilde{b}_{I_j(\tau), j}(f_j(\tau))
    \mathbb{1}_{I_j(\tau),j}(\tau)\right]\nonumber\\
    \le & E\left[ \sum_{\tau=0}^{\sum_{k=0}^{K-1} w_k - 1}
    Q_{I_j( \tau)}( \tau)
    \tilde{b}_{I_j(\tau), j}(f_j(\tau))
    \eta_j(f_j( \tau))
    \mathbb{1}_{I_j(\tau),j}(\tau)\right] + J  \sum_{k=0}^{K-1} w_k \nonumber\\
    = & 
    \sum_i E\left[ \sum_{\tau=0}^{\sum_{k=0}^{K-1} w_k - 1}
    Q_{i}( \tau)
    \tilde{b}_{i, j}(f_j(\tau))
    \eta_j(f_j( \tau))
    \mathbb{1}_{i,j}(\tau)\mathbb{1}_{I_j(\tau)=i}\right] + J  \sum_{k=0}^{K-1} w_k.
\end{align}
Considering the event $\{\tilde{b}_{i, j}(f_j(\tau)) \le \frac{\delta}{6JU_{\mathrm{S}}}\}$, we further have
\begin{align}\label{equ:sum-ucb-2}
    & E\left[ \sum_{\tau=0}^{\sum_{k=0}^{K-1} w_k - 1}
    Q_{i}( \tau)
    \tilde{b}_{i, j}(f_j(\tau))
    \eta_j(f_j( \tau))
    \mathbb{1}_{i,j}(\tau)\mathbb{1}_{I_j(\tau)=i}\right]\nonumber\\
    = & 
    E\left[ \sum_{\tau=0}^{\sum_{k=0}^{K-1} w_k - 1}
    Q_{i}( \tau)
    \tilde{b}_{i, j}(f_j(\tau))
    \eta_j(f_j( \tau))
    \mathbb{1}_{i,j}(\tau)\mathbb{1}_{I_j(\tau)=i}
    \mathbb{1}_{\tilde{b}_{i, j}(f_j(\tau)) \le \frac{\delta}{6JU_{\mathrm{S}}}}
    \right]\nonumber\\
    & + 
    E\left[ \sum_{\tau=0}^{\sum_{k=0}^{K-1} w_k - 1}
    Q_{i}( \tau)
    \tilde{b}_{i, j}(f_j(\tau))
    \eta_j(f_j( \tau))
    \mathbb{1}_{i,j}(\tau)\mathbb{1}_{I_j(\tau)=i}
    \mathbb{1}_{\tilde{b}_{i, j}(f_j(\tau)) > \frac{\delta}{6JU_{\mathrm{S}}}}
    \right]\nonumber\\
    \le & \frac{\delta}{6JU_{\mathrm{S}}}
    E\left[ \sum_{\tau=0}^{\sum_{k=0}^{K-1} w_k - 1}
    Q_{i}( \tau)\right]
    + E\left[ \sum_{\tau=0}^{\sum_{k=0}^{K-1} w_k - 1}
    Q_{i}( \tau)
    \tilde{b}_{i, j}(f_j(\tau))
    \eta_j(f_j( \tau))
    \mathbb{1}_{i,j}(\tau)\mathbb{1}_{I_j(\tau)=i}
    \mathbb{1}_{\tilde{b}_{i, j}(f_j(\tau)) > \frac{\delta}{6JU_{\mathrm{S}}}}
    \right]\nonumber\\
     \le & \frac{\delta}{6JU_{\mathrm{S}}}
    E\left[ \sum_{\tau=0}^{\sum_{k=0}^{K-1} w_k - 1}
    Q_{i}( \tau)\right]\nonumber\\
    & + E\left[ \sum_{\tau=0}^{\sum_{k=0}^{K-1} w_k - 1}
    Q_{i}( \tau)
    \tilde{b}_{i, j}(f_j(\tau))
    \eta_j(f_j( \tau))
    \mathbb{1}_{i,j}(\tau)\mathbb{1}_{I_j(\tau)=i}
    \mathbb{1}_{\hat{N}_{i, j}(f_j(\tau)) < \frac{576 J^2 U^2_{\mathrm{S}} \log f_j(\tau)}{\delta^2}}
    \right],
\end{align}
where the last inequality holds since
\begin{align}\label{equ:bound-tilde-b-i-j-t}
    \tilde{b}_{i,j}(t) = \min\{2b_{i,j}(t), 1\} = & \min\left\{4 U_{\mathrm{S}} \sqrt{\frac{\log t}{\hat{N}_{i,j}(t)}}
    , 1\right\}
    \le 4 U_{\mathrm{S}} \sqrt{\frac{\log t}{\hat{N}_{i,j}(t)}}
\end{align}
for any $t$
according to Line~\ref{alg:line:b} in Algorithm~\ref{alg:1} with $\gamma=1$.
For the second term in~\eqref{equ:sum-ucb-2}, we will use the same method as Lemma~\ref{lemma:sum-error-queue} to decouple the queue length and the UCB bonus and then the sum of UCB bonuses can be bounded in a way similar to traditional UCB. The result is shown in the following lemma:
\begin{lemma}\label{lemma:decouple-queue-len-ucb}
Consider Algorithm~\ref{alg:1} with $\gamma=1$ and $c_1=2$. For any $K\ge 3$, we have
    \begin{align*}
        & E\left[ \sum_{\tau=0}^{\sum_{k=0}^{K-1} w_k - 1}
        Q_{i}( \tau)
        \tilde{b}_{i, j}(f_j(\tau))
        \eta_j(f_j( \tau))
        \mathbb{1}_{i,j}(\tau)\mathbb{1}_{I_j(\tau)=i}
        \mathbb{1}_{\hat{N}_{i, j}(f_j(\tau)) < \frac{576 J^2 U^2_{\mathrm{S}} \log f_j(\tau)}{\delta^2}}
        \right]\nonumber\\
        \le & \frac{\delta}{6 J U_{\mathrm{S}} } E \left[\sum_{\tau=0}^{\sum_{k=0}^{K-1} w_k -1} Q_i(\tau) \right]
        + \frac{225816 J^3 U^5_{\mathrm{S}} U_{\mathrm{A}} \log^2 \left(\sum_{k=0}^{K-1} w_k - 1\right)}{\delta^3}.
    \end{align*}
\end{lemma}
Proof of Lemma~\ref{lemma:decouple-queue-len-ucb} can be found in Section~\ref{app:proof-lemma-decouple-queue-len-ucb}. By Lemma~\ref{lemma:decouple-queue-len-ucb} and \eqref{equ:sum-ucb-2}, we have
\begin{align*}
    & E\left[ \sum_{\tau=0}^{\sum_{k=0}^{K-1} w_k - 1}
    Q_{i}( \tau)
    \tilde{b}_{i, j}(f_j(\tau))
    \eta_j(f_j( \tau))
    \mathbb{1}_{i,j}(\tau)\mathbb{1}_{I_j(\tau)=i}\right]\nonumber\\
    \le & \frac{\delta}{3 J U_{\mathrm{S}} } E \left[\sum_{\tau=0}^{\sum_{k=0}^{K-1} w_k -1} Q_i(\tau) \right]
    + \frac{225816 J^3 U^5_{\mathrm{S}} U_{\mathrm{A}} \log^2 \left(\sum_{k=0}^{K-1} w_k - 1\right)}{\delta^3}
\end{align*}
for any $K\ge 3$. Substituting the above inequality into \eqref{equ:sum-ucb-1}, we have
\begin{align*}
    & E\left[ \sum_{\tau=0}^{\sum_{k=0}^{K-1} w_k - 1}
    Q_{I_j(\tau)}(\tau)
    \min \{ 2 b_{I_j(\tau),j}(f_j(\tau)), 1 \}
    \mathbb{1}_{I_j(\tau),j}(\tau)\right]\nonumber\\
    \le & \frac{\delta}{3 J U_{\mathrm{S}} } E \left[\sum_{\tau=0}^{\sum_{k=0}^{K-1} w_k -1} \sum_i Q_i(\tau) \right]
    + \frac{225816 I J^3 U^5_{\mathrm{S}} U_{\mathrm{A}} \log^2 \left(\sum_{k=0}^{K-1} w_k - 1\right)}{\delta^3}
    + J  \sum_{k=0}^{K-1} w_k
\end{align*}
for any $K\ge 3$. Substituting the above inequality into \eqref{equ:sum-queue-ucb}, we have
\begin{align}\label{equ:bound-sum-queue-ucb}
    \eqref{equ:sum-queue-ucb} \le \frac{2\delta}{3} E \left[\sum_{\tau=0}^{\sum_{k=0}^{K-1} w_k -1} \sum_i Q_i(\tau) \right]
    + \frac{451632 I J^4 U^6_{\mathrm{S}} U_{\mathrm{A}} \log^2 \left(\sum_{k=0}^{K-1} w_k - 1\right)}{\delta^3}
    + 2 J^2  U_{\mathrm{S}} \sum_{k=0}^{K-1} w_k
\end{align}
for any $K\ge 3$.

\subsection{Bounding the Weighted Sum of Job Completion Indicators}
In this subsection, we consider bounding the weighted sum of job completion indicators \eqref{equ:queue-indicators}. We first look at the term:
\begin{align*}
    E\left[\sum_{\tau=0}^{\sum_{k=0}^{K-1} w_k - 1}
    \max_i Q_i(f_j(\tau)) \mu_{i,j}
    \frac{\mathbb{1}_{I_j(\tau),j}(\tau)}{ \mu_{I_j(\tau),j} }\right].
\end{align*}
We can rewrite the above term in a different form by summing over the time slots in which the jobs start, i.e.,
\begin{align}\label{equ:sum-over-job-start-time}
    & E\left[\sum_{\tau=0}^{\sum_{k=0}^{K-1} w_k - 1}
    \max_i Q_i(f_j(\tau)) \mu_{i,j}
    \frac{\mathbb{1}_{I_j(\tau),j}(\tau)}{ \mu_{I_j(\tau),j} }\right]\nonumber\\
    \ge & E\left[\sum_{\tau=0}^{\sum_{k=0}^{K-1} w_k - 1}
    \max_i Q_i(\tau) \mu_{i,j}
    \frac{\mathbb{1}_{\hat{i}^*_j(\tau)\neq 0}}{ \mu_{\hat{i}^*_j(\tau), j} }\right]
     - E\left[
    \max_i Q_i\left(f_j\biggl(\sum_{k=0}^{K-1} w_k - 1\biggr)\right) \mu_{i,j}
    \frac{1}{ \mu_{I_j(\sum_{k=0}^{K-1} w_k - 1),j} }\right]\nonumber\\
    \ge & E\left[\sum_{\tau=0}^{\sum_{k=0}^{K-1} w_k - 1}
    \max_i Q_i(\tau) \mu_{i,j}
    \frac{\mathbb{1}_{\hat{i}^*_j(\tau)\neq 0}}{ \mu_{\hat{i}^*_j(\tau), j} }\right]
     - U_{\mathrm{S}} E\left[
    \sum_i Q_i\left(f_j\biggl(\sum_{k=0}^{K-1} w_k - 1\biggr)\right) \right],
\end{align}
where the first inequality holds since the last job starting before $\sum_{k=0}^{K-1} w_k$ may not finish before
$\sum_{k=0}^{K-1} w_k$, and the second inequality holds since $\mu_{i,j}\le 1$ and $\frac{1}{\mu_{i,j}}\le U_{\mathrm{S}}$ for any $i,j$.
Note that by Lemma~\ref{lemma:bounds-on-diff-queue-lens} and the bound of service time, we have
\begin{align*}
    \sum_i Q_i\left(f_j\biggl(\sum_{k=0}^{K-1} w_k - 1\biggr)\right)
    \le & \sum_i Q_i\left(\sum_{k=0}^{K-1} w_k - 1\right) + J U_{\mathrm{S}} \nonumber\\
    \le & \sum_i Q_i (\tau) + J U_{\mathrm{S}} + I U_{\mathrm{A}} \left(\sum_{k=0}^{K-1} w_k - 1\right)
\end{align*}
for any $0\le \tau\le \sum_{k=0}^{K-1} w_k - 1$. Then by summing over $\tau$ from $0$ to $\sum_{k=0}^{K-1} w_k - 1$ and dividing both sides by $\sum_{k=0}^{K-1} w_k$, we have
\begin{align*}
    \sum_i Q_i\left(f_j\biggl(\sum_{k=0}^{K-1} w_k - 1\biggr)\right)
    \le \frac{1}{\sum_{k=0}^{K-1} w_k} \sum_{\tau=0}^{\sum_{k=0}^{K-1} w_k - 1}  \sum_i Q_i (\tau) + J U_{\mathrm{S}} + I U_{\mathrm{A}} \left(\sum_{k=0}^{K-1} w_k - 1\right).
\end{align*}
Substituting the above inequality into \eqref{equ:sum-over-job-start-time}, we have
\begin{align}\label{equ:queue-mu-sum}
    & E\left[\sum_{\tau=0}^{\sum_{k=0}^{K-1} w_k - 1}
    \max_i Q_i(f_j(\tau)) \mu_{i,j}
    \frac{\mathbb{1}_{I_j(\tau),j}(\tau)}{ \mu_{I_j(\tau),j} }\right]\nonumber\\
    \ge & E\left[\sum_{\tau=0}^{\sum_{k=0}^{K-1} w_k - 1}
    \max_i Q_i(\tau) \mu_{i,j}
    \frac{\mathbb{1}_{\hat{i}^*_j(\tau)\neq 0}}{ \mu_{\hat{i}^*_j(\tau)j} }\right]\nonumber\\
    &  -  
    \frac{U_{\mathrm{S}}}{\sum_{k=0}^{K-1} w_k} E\left[\sum_{\tau=0}^{\sum_{k=0}^{K-1} w_k - 1}  \sum_i Q_i (\tau) \right] - J U^2_{\mathrm{S}} - I U_{\mathrm{S}}  U_{\mathrm{A}} \left(\sum_{k=0}^{K-1} w_k - 1\right)
\end{align}
Next we look at the term $E\left[\sum_{\tau=0}^{\sum_{k=0}^{K-1} w_k - 1} \max_i Q_i(\tau) \mu_{i,j} \frac{\mathbb{1}_{\hat{i}^*_j(\tau)\neq 0}}{ \mu_{\hat{i}^*_j(\tau)j} }\right]$ in \eqref{equ:queue-mu-sum}.
Let $\mathbb{1}_{\mathrm{idling}}(j,\tau)\coloneqq 1 - \eta_j(\tau)$, which is equal to $1$ when server $j$ is idling. Dividing the sum into two cases based on whether server $j$ is idling or non-idling, we have
\begin{align}\label{equ:sum-queue-mu-idling-or-not}
    & E\left[\sum_{\tau=0}^{\sum_{k=0}^{K-1} w_k - 1}
    \max_i Q_i(\tau) \mu_{i,j}
    \frac{\mathbb{1}_{\hat{i}^*_j(\tau)\neq 0}}{ \mu_{\hat{i}^*_j(\tau)j} }\right]\nonumber\\
    = &  E\left[\sum_{\tau=0}^{\sum_{k=0}^{K-1} w_k - 1}
    \max_i Q_i(\tau) \mu_{i,j} \eta_j(\tau)
    \frac{\mathbb{1}_{\hat{i}^*_j(\tau)\neq 0}}{ \mu_{\hat{i}^*_j(\tau)j} }\right]
    + E\left[\sum_{\tau=0}^{\sum_{k=0}^{K-1} w_k - 1}
    \max_i Q_i(\tau) \mu_{i,j} \mathbb{1}_{\mathrm{idling}}(j,\tau)
    \frac{\mathbb{1}_{\hat{i}^*_j(\tau)\neq 0}}{ \mu_{\hat{i}^*_j(\tau)j} }\right]\nonumber\\
    \ge & E\left[\sum_{\tau=0}^{\sum_{k=0}^{K-1} w_k - 1}
    \max_i Q_i(\tau) \mu_{i,j} \eta_j(\tau)
    \frac{\mathbb{1}_{\hat{i}^*_j(\tau)\neq 0}}{ \mu_{\hat{i}^*_j(\tau)j} }\right]
    + E\left[\sum_{\tau=0}^{\sum_{k=0}^{K-1} w_k - 1}
    \max_i Q_i(\tau) \mu_{i,j} \mathbb{1}_{\mathrm{idling}}(j,\tau)
    \mathbb{1}_{\hat{i}^*_j(\tau)\neq 0}\right],
\end{align}
where the last inequality is due to the fact that $\mu_{i,j}\le 1$ for any $i,j$.
We first look at the first term in \eqref{equ:sum-queue-mu-idling-or-not}. Note that $E[S_{i,j}(\tau)]=1/\mu_{i,j}$ for any $\tau$. Then we have
\begin{align}\label{equ:equalities-station}
    & E\left[\sum_{\tau=0}^{\sum_{k=0}^{K-1} w_k - 1}
    \max_i Q_i(\tau) \mu_{i,j} \eta_j(\tau)
    \frac{\mathbb{1}_{\hat{i}^*_j(\tau)\neq 0}}{ \mu_{\hat{i}^*_j(\tau)j} }\right]\nonumber\\
    = & \sum_{i=1}^I E\left[\sum_{\tau=0}^{\sum_{k=0}^{K-1} w_k - 1}
    \max_{i'} Q_{i'}(\tau) \mu_{i',j} \eta_j(\tau)
    \frac{\mathbb{1}_{\hat{i}^*_j(\tau) = i}}{ \mu_{i,j} }\right] \nonumber\\
    = & \sum_{i=1}^I E\left[\sum_{\tau=0}^{\sum_{k=0}^{K-1} w_k - 1}
    \max_{i'} Q_{i'}(\tau) \mu_{i',j}  \eta_j(\tau)
    \mathbb{1}_{\hat{i}^*_j(\tau) = i} E[S_{i,j}(\tau)] \right]\nonumber\\
    = & \sum_{i=1}^I E\left[\sum_{\tau=0}^{\sum_{k=0}^{K-1} w_k - 1}
    \max_{i'} Q_{i'}(\tau) \mu_{i',j} \eta_j(\tau)
    \mathbb{1}_{\hat{i}^*_j(\tau) = i}
     E[S_{i,j}(\tau)| \boldsymbol{Q}(\tau), \boldsymbol{H}(\tau), \boldsymbol{A}(\tau)]
     \right]\nonumber\\
    = & \sum_{i=1}^I E\left[\sum_{\tau=0}^{\sum_{k=0}^{K-1} w_k - 1}
     E[\max_{i'} Q_{i'}(\tau) \mu_{i',j} \eta_j(\tau) \mathbb{1}_{\hat{i}^*_j(\tau) = i} S_{i,j}(\tau)| \boldsymbol{Q}(\tau), \boldsymbol{H}(\tau), \boldsymbol{A}(\tau)] \right]\nonumber\\
     = & \sum_{i=1}^I \sum_{\tau=0}^{\sum_{k=0}^{K-1} w_k - 1}
     E\left[ E[\max_{i'} Q_{i'}(\tau) \mu_{i',j} \eta_j(\tau) \mathbb{1}_{\hat{i}^*_j(\tau) = i} S_{i,j}(\tau)| \boldsymbol{Q}(\tau), \boldsymbol{H}(\tau), \boldsymbol{A}(\tau)] \right]\nonumber\\
     = & \sum_{i=1}^I \sum_{\tau=0}^{\sum_{k=0}^{K-1} w_k - 1}
     E\left[ \max_{i'} Q_{i'}(\tau) \mu_{i',j} \eta_j(\tau) \mathbb{1}_{\hat{i}^*_j(\tau) = i} S_{i,j}(\tau) \right] \nonumber\\
     = & \sum_{\tau=0}^{\sum_{k=0}^{K-1} w_k - 1}
     E\left[ \max_{i'} Q_{i'}(\tau) \mu_{i',j} \eta_j(\tau) \mathbb{1}_{\hat{i}^*_j(\tau) \neq 0} S_{\hat{i}^*_j(\tau),j}(\tau) \right],
\end{align}
where the third equality is due to the independence between $S_{i,j}(\tau)$ and $\boldsymbol{Q}(\tau), \boldsymbol{H}(\tau), \boldsymbol{A}(\tau)$, the fourth equality is due to the fact that $\max_{i'}Q_{i'}(\tau)\mu_{i',j}$, $\eta_j(\tau)$, $\hat{i}^*_j(\tau)$ are fully determined by $\boldsymbol{Q}(\tau), \boldsymbol{H}(\tau), \boldsymbol{A}(\tau)$, and the sixth equality is by the law of iterated expectation.
Substituting \eqref{equ:equalities-station} into \eqref{equ:sum-queue-mu-idling-or-not}, we have
\begin{align}\label{equ:actual-time-spent}
    & E\left[\sum_{\tau=0}^{\sum_{k=0}^{K-1} w_k - 1}
    \max_i Q_i(\tau) \mu_{i,j}
    \frac{\mathbb{1}_{\hat{i}^*_j(\tau)\neq 0}}{ \mu_{\hat{i}^*_j(\tau)j} }\right]\nonumber\\
    \ge & \sum_{\tau=0}^{\sum_{k=0}^{K-1} w_k - 1}
     E\left[ \max_{i'} Q_{i'}(\tau) \mu_{i',j} \eta_j(\tau) \mathbb{1}_{\hat{i}^*_j(\tau) \neq 0} S_{\hat{i}^*_j(\tau),j}(\tau) \right]\nonumber\\
     & + E\left[\sum_{\tau=0}^{\sum_{k=0}^{K-1} w_k - 1}
    \max_i Q_i(\tau) \mu_{i,j} \mathbb{1}_{\mathrm{idling}}(j,\tau)
    \mathbb{1}_{\hat{i}^*_j(\tau)\neq 0}\right]\nonumber\\
    = & E\left[\sum_{\tau=0}^{\sum_{k=0}^{K-1} w_k - 1}
     \max_{i} Q_{i}(\tau) \mu_{i,j} \mathbb{1}_{\hat{i}^*_j(\tau) \neq 0} 
     \left( \eta_j(\tau) S_{\hat{i}^*_j(\tau),j}(\tau) + \mathbb{1}_{\mathrm{idling}}(j,\tau) \right) \right].
\end{align}
Note that the term $\eta_j(\tau) S_{\hat{i}^*_j(\tau),j}(\tau) + \mathbb{1}_{\mathrm{idling}}(j,\tau)$ is the actual time that server $j$ spends on the queue $\hat{i}^*_j(\tau)$. Therefore, \eqref{equ:actual-time-spent} can be rewritten using $f_j$ in the following way:
\begin{align}\label{equ:actual-time-same-sum}
    & E\left[\sum_{\tau=0}^{\sum_{k=0}^{K-1} w_k - 1}
    \max_{i} Q_{i}(\tau) \mu_{i,j} \mathbb{1}_{\hat{i}^*_j(\tau) \neq 0} 
    \left( \eta_j(\tau) S_{\hat{i}^*_j(\tau),j}(\tau) + \mathbb{1}_{\mathrm{idling}}(j,\tau) \right) \right]\nonumber\\
    \ge & E\left[\sum_{\tau=0}^{\sum_{k=0}^{K-1} w_k - 1}
    \max_{i} Q_{i}(f_j(\tau)) \mu_{i,j}  \right],
\end{align}
where the inequality is due to the fact that the last job starting before time slot $\sum_{k=0}^{K-1} w_k$ at server $j$ may not complete before $\sum_{k=0}^{K-1} w_k$. Combining \eqref{equ:queue-mu-sum}, \eqref{equ:actual-time-spent}, and \eqref{equ:actual-time-same-sum}, we have
\begin{align*}
    & E\left[\sum_{\tau=0}^{\sum_{k=0}^{K-1} w_k - 1}
    \max_i Q_i(f_j(\tau)) \mu_{i,j}
    \frac{\mathbb{1}_{I_j(\tau),j}(\tau)}{ \mu_{I_j(\tau),j} }\right]\nonumber\\
    \ge & E\left[\sum_{\tau=0}^{\sum_{k=0}^{K-1} w_k - 1}
    \max_{i} Q_{i}(f_j(\tau)) \mu_{i,j}  \right]\nonumber\\
    &  -  
    \frac{U_{\mathrm{S}}}{\sum_{k=0}^{K-1} w_k} E\left[\sum_{\tau=0}^{\sum_{k=0}^{K-1} w_k - 1}  \sum_i Q_i (\tau) \right] - J U^2_{\mathrm{S}} - I U_{\mathrm{S}}  U_{\mathrm{A}} \left(\sum_{k=0}^{K-1} w_k - 1\right).
\end{align*}
Substituting the above inequality into \eqref{equ:queue-indicators}, we have
\begin{align}\label{equ:bound-queue-indicators}
    \eqref{equ:queue-indicators} \le & - 2 \sum_j E\left[\sum_{\tau=0}^{\sum_{k=0}^{K-1} w_k - 1}
    \max_{i} Q_{i}(f_j(\tau)) \mu_{i,j}  \right]\nonumber\\
    & + \frac{2 J U_{\mathrm{S}}}{\sum_{k=0}^{K-1} w_k} E\left[\sum_{\tau=0}^{\sum_{k=0}^{K-1} w_k - 1}  \sum_i Q_i (\tau) \right] + 2 J^2 U^2_{\mathrm{S}} + 2 I J U_{\mathrm{S}}  U_{\mathrm{A}} \left(\sum_{k=0}^{K-1} w_k - 1\right).
\end{align}

\subsection{Deriving Negative Lyapunov Drift}

Substituting \eqref{equ:bound-queue-indicators}, \eqref{equ:bound-sum-queue-ucb}, \eqref{equ:bound-coupling-term} into \eqref{equ:queue-indicators}, \eqref{equ:sum-queue-ucb}, \eqref{equ:coupling-term}, respectively, we have
\begin{align}\label{equ:negative-drift-station}
    E[L(t_{\mathrm{S}, K}) - L(0)]
    \le & -\left(\delta - \frac{2 J U_{\mathrm{S}}}{\sum_{k=0}^{K-1} w_k} \right) \sum_{\tau=0}^{\sum_{k=0}^{K-1}w_k - 1} E\left[ \sum_i Q_i(\tau) \right] \nonumber\\
    & + \frac{451632 I J^4 U^6_{\mathrm{S}} U_{\mathrm{A}} \log^2 \left(\sum_{k=0}^{K-1} w_k\right)}{\delta^3}\nonumber\\
    & + 17 I J^2 U^2_{\mathrm{S}} U^2_{\mathrm{A}} W \sum_{k=0}^{K-1} w_k  + \frac{302 I^2 J^2 U^2_{\mathrm{S}} U_{\mathrm{A}}}{\delta}
\end{align}
for all $K\ge 3$.
If $\sum_{k=0}^{K-1} w_k < \frac{4 J U_{\mathrm{S}}}{\delta}$, then by Lemma~\ref{lemma:bounds-on-diff-queue-lens}, we have
\begin{align}\label{equ:finite-time-average-bound-case-1}
    \frac{1}{\sum_{k=0}^{K-1}w_k}\sum_{\tau=0}^{\sum_{k=0}^{K-1}w_k - 1} E\left[ \sum_i Q_i(\tau) \right] \le \frac{4 I J U_{\mathrm{S}} U_{\mathrm{A}}}{\delta}.
\end{align}
If $\sum_{k=0}^{K-1} w_k \ge \frac{4 J U_{\mathrm{S}}}{\delta}$, then $\frac{2 J U_{\mathrm{S}}}{\sum_{k=0}^{K-1} w_k} \le \frac{\delta}{2}$. Hence, from \eqref{equ:negative-drift-station}, we have
\begin{align*}
    E[L(t_{\mathrm{S}, K}) - L(0)]
    \le & -\frac{\delta}{2} \sum_{\tau=0}^{\sum_{k=0}^{K-1}w_k - 1} E\left[ \sum_i Q_i(\tau) \right] \nonumber\\
    & + \frac{451632 I J^4 U^6_{\mathrm{S}} U_{\mathrm{A}} \log^2 \left(\sum_{k=0}^{K-1} w_k\right)}{\delta^3}\nonumber\\
    & + 17 I J^2 U^2_{\mathrm{S}} U^2_{\mathrm{A}} W \sum_{k=0}^{K-1} w_k  + \frac{302 I^2 J^2 U^2_{\mathrm{S}} U_{\mathrm{A}}}{\delta},
\end{align*}
which implies that
\begin{align}\label{equ:finite-time-average-bound-case-2}
    & \frac{1}{\sum_{k=0}^{K-1}w_k} \sum_{\tau=0}^{\sum_{k=0}^{K-1}w_k - 1} E\left[ \sum_i Q_i(\tau) \right]\nonumber\\
    \le & \frac{2E[L(0)]}{\delta \sum_{k=0}^{K-1}w_k}
    + \frac{903264 I J^4 U^6_{\mathrm{S}} U_{\mathrm{A}} \log^2 \left(\sum_{k=0}^{K-1} w_k\right)}{\delta^4 \sum_{k=0}^{K-1}w_k}
    + \frac{34 I J^2 U^2_{\mathrm{S}} U^2_{\mathrm{A}} W }{\delta}
    + \frac{604 I^2 J^2 U^2_{\mathrm{S}} U_{\mathrm{A}}}{\delta^2 \sum_{k=0}^{K-1}w_k}\nonumber\\
    = & \frac{903264 I J^4 U^6_{\mathrm{S}} U_{\mathrm{A}} \log^2 \left(\sum_{k=0}^{K-1} w_k\right)}{\delta^4 \sum_{k=0}^{K-1}w_k}
    + \frac{34 I J^2 U^2_{\mathrm{S}} U^2_{\mathrm{A}} W }{\delta}
    + \frac{604 I^2 J^2 U^2_{\mathrm{S}} U_{\mathrm{A}}}{\delta^2 \sum_{k=0}^{K-1}w_k},
\end{align}
where the last equality is by the initial condition that $Q_i(0)=0$ for all $i$.
From \eqref{equ:finite-time-average-bound-case-1} and \eqref{equ:finite-time-average-bound-case-2}, we have
\begin{align}\label{equ:final-bound-station-1}
    & \frac{1}{\sum_{k=0}^{K-1}w_k} \sum_{\tau=0}^{\sum_{k=0}^{K-1}w_k - 1} E\left[ \sum_i Q_i(\tau) \right]\nonumber\\
    \le & \frac{903264 I J^4 U^6_{\mathrm{S}} U_{\mathrm{A}} \log^2 \left(\sum_{k=0}^{K-1} w_k\right)}{\delta^4 \sum_{k=0}^{K-1}w_k}
    + \frac{34 I J^2 U^2_{\mathrm{S}} U^2_{\mathrm{A}} W }{\delta}
    + \frac{604 I^2 J^2 U^2_{\mathrm{S}} U_{\mathrm{A}}}{\delta^2 \sum_{k=0}^{K-1}w_k}
\end{align}
for all $K\ge 3$.
Note that for any $t$ there exists an integer $K$ such that $t\le \sum_{k=0}^{K-1}w_k -1 < t + W$ since $w_k\le W$ by the definition of the capacity region~\eqref{equ:def-capacity-stationary}. Hence, we have
\begin{align}\label{equ:range-t-station}
    \sum_{k=0}^{K-1}w_k -1 \ge t \ge \sum_{k=0}^{K-1}w_k - W.
\end{align}
Therefore, we have
\begin{align}\label{equ:transform-average-queue-station}
    \frac{1}{t} \sum_{\tau=1}^{t} E\left[ \sum_i Q_i(\tau) \right] 
    = & \frac{1}{t} \sum_{\tau=0}^{t} E\left[ \sum_i Q_i(\tau) \right] \nonumber\\
    = & \frac{\sum_{k=0}^{K-1}w_k}{t} \frac{1}{\sum_{k=0}^{K-1}w_k} \sum_{\tau=0}^{t} E\left[ \sum_i Q_i(\tau) \right] \nonumber\\
    \le & \frac{t+W}{t} \frac{1}{\sum_{k=0}^{K-1}w_k} \sum_{\tau=0}^{\sum_{k=0}^{K-1}w_k -1} E\left[ \sum_i Q_i(\tau) \right].
\end{align}
From \eqref{equ:range-t-station} and $w_k\le W$, we have $t \le KW -1$. Hence, a sufficient condition for $K\ge 3$ is $t \ge 3W - 1$. Therefore, from \eqref{equ:final-bound-station-1} and \eqref{equ:transform-average-queue-station}, for all $t\ge \max\{3W -1,e^2-1\}$, we have
\begin{align}\label{equ:final-bound-station-2}
    & \frac{1}{t} \sum_{\tau=1}^{t} E\left[ \sum_i Q_i(\tau) \right]\nonumber\\
    \le & \left(1 + \frac{W}{t}\right) \left(\frac{903264 I J^4 U^6_{\mathrm{S}} U_{\mathrm{A}} \log^2 \left(\sum_{k=0}^{K-1} w_k\right)}{\delta^4 \sum_{k=0}^{K-1}w_k}
    + \frac{34 I J^2 U^2_{\mathrm{S}} U^2_{\mathrm{A}} W }{\delta}
    + \frac{604 I^2 J^2 U^2_{\mathrm{S}} U_{\mathrm{A}}}{\delta^2 \sum_{k=0}^{K-1}w_k}\right)\nonumber\\
    \le & \left(1 + \frac{W}{t}\right) \left(\frac{903264 I J^4 U^6_{\mathrm{S}} U_{\mathrm{A}} \log^2 (t+1)}{\delta^4 (t+1)}
    + \frac{34 I J^2 U^2_{\mathrm{S}} U^2_{\mathrm{A}} W }{\delta}
    + \frac{604 I^2 J^2 U^2_{\mathrm{S}} U_{\mathrm{A}}}{\delta^2 (t+1)}\right),
\end{align}
where the last inequality uses \eqref{equ:range-t-station} and the fact that $\frac{\log^2 x}{x}$ is decreasing when $x\ge e^2$. For $t<\max\{3W -1,e^2-1\}$, by Lemma~\ref{lemma:bounds-on-diff-queue-lens}, we have
\begin{align}\label{equ:final-bound-station-3}
    \frac{1}{t} \sum_{\tau=1}^{t} E\left[ \sum_i Q_i(\tau) \right] \le I U_{\mathrm{A}} \max\{3W -1,e^2-1\} \le (3+e^2) I U_{\mathrm{A}} W.
\end{align}
From \eqref{equ:final-bound-station-2} and \eqref{equ:final-bound-station-3}, we have for all $t$,
\begin{align*}
    \frac{1}{t} \sum_{\tau=1}^{t} E\left[ \sum_i Q_i(\tau) \right]
    \le \left(1 + \frac{W}{t}\right) \left(\frac{903264 I J^4 U^6_{\mathrm{S}} U_{\mathrm{A}} \log^2 (t+1)}{\delta^4 (t+1)}
    + \frac{34 I J^2 U^2_{\mathrm{S}} U^2_{\mathrm{A}} W }{\delta}
    + \frac{604 I^2 J^2 U^2_{\mathrm{S}} U_{\mathrm{A}}}{\delta^2 (t+1)}\right).
\end{align*}
The finite-time bound in Theorem~\ref{theo:3} is proved.

Letting $t\rightarrow \infty$, we have
\begin{align*}
    \limsup_{t\rightarrow \infty} \frac{1}{t} \sum_{\tau=1}^{t} E\left[ \sum_i Q_i(\tau) \right]
    \le \frac{34 I J^2 U^2_{\mathrm{S}} U^2_{\mathrm{A}} W }{\delta}.
\end{align*}
The asymptotic bound in Theorem~\ref{theo:3} is proved.

\section{Proofs of Auxiliary Lemmas}
\label{app:sec:proof:lemmas}

In this section, we present the proofs of all the lemmas that appeared in the paper. In the proofs, if server $j$ is not available at the beginning of time slot $t$, i.e., $\sum_i M_{i,j}(t) > 0$, we let $\hat{i}^*_{j}(t)=0$. Let $T\coloneqq g(\gamma)$ for ease of notation.

\subsection{Proof of Lemma~\ref{lemma:concentration}}
\label{app:proof-lemma-concentration}

\proof
Recall that
\[
b_{i,j}(t_k+\tau) = 2\sqrt{\frac{U_{\mathrm{S}}^2\log \left(\sum_{\tau'=0}^{t_k+\tau-1} \gamma^{\tau'} \right)}{\hat{N}_{i,j}(t_k+\tau)}} 
= 2\sqrt{\frac{U_{\mathrm{S}}^2\log \left(
\frac{1-\gamma^{t_k+\tau}}{1-\gamma} \right)}{\hat{N}_{i,j}(t_k+\tau)}}.
\]
Consider the event
\begin{align}\label{equ:def-conc-single-event}
    {\mathcal E}_{t_k,i,j,\tau}\coloneqq \left\{ \left|\frac{1}{\hat{\mu}_{i,j}(t_k+\tau)}-\frac{1}{\mu_{i,j}(t_k+\tau)}\right|\le 2\sqrt{\frac{U_{\mathrm{S}}^2\log \left(
\frac{1-\gamma^{t_k+\tau}}{1-\gamma} \right)}{\hat{N}_{i,j}(t_k+\tau)}} \right\}.
\end{align}
We have
\begin{align}\label{equ:lm1-log-term}
    \hat{P}_{t_k} \left( {\mathcal E}^{ {\mathrm c}}_{t_k,i,j,\tau} \right)
    = \hat{P}_{t_k} \left(  \left|\frac{1}{\hat{\mu}_{i,j}(t_k+\tau)}-\frac{1}{\mu_{i,j}(t_k+\tau)}\right|> 2\sqrt{\frac{ U_{\mathrm{S}}^2\log \left(
    \frac{1-\gamma^{t_k+\tau}}{1-\gamma} \right)}{\hat{N}_{i,j}(t_k+\tau)}} \right).
\end{align}
Note that $t_k+\tau \ge \tau \ge D_k-\frac{T}{8}\ge \frac{3T}{8}$ by Lemma~\ref{lemma:prop-D}. Hence,
\begin{align}\label{equ:lm1-log-term-2}
    \gamma^{t_k+\tau} \le \gamma^{\frac{3T}{8}} = \gamma^{\frac{3}{2(1-\gamma)}\log \frac{1}{1-\gamma}} 
    \le \left(\gamma^{\frac{1}{1-\gamma}}\right)^{\log \frac{1}{1-\gamma}} \le \exp\left(-\log \frac{1}{1-\gamma}\right) = 1-\gamma,
\end{align}
where the last inequality is due to the fact that
$\gamma^{\frac{1}{1-\gamma}} \le \lim_{\gamma \rightarrow 1} \gamma^{\frac{1}{1-\gamma}} =  \lim_{x\rightarrow \infty} (1-\frac{1}{x})^x = e^{-1}$ since $\gamma^{\frac{1}{1-\gamma}}$ is increasing in $\gamma$.
Hence, from~\eqref{equ:lm1-log-term} and~\eqref{equ:lm1-log-term-2}, we have
\begin{align*}
    \hat{P}_{t_k} \left( {\mathcal E}^{ {\mathrm c}}_{t_k,i,j,\tau} \right)
    \le & \hat{P}_{t_k} \left(  \left|\frac{1}{\hat{\mu}_{i,j}(t_k+\tau)}-\frac{1}{\mu_{i,j}(t_k+\tau)}\right|> 2\sqrt{\frac{ U_{\mathrm{S}}^2
    \log \frac{\gamma}{1-\gamma} }{\hat{N}_{i,j}(t_k+\tau)}} \right).
\end{align*}
Consider the events $\{\hat{N}_{i,j}(t_k+\tau)>4\log \frac{\gamma}{1-\gamma}\}$ and $\{\hat{N}_{i,j}(t_k+\tau)\le 4\log \frac{\gamma}{1-\gamma}\}$. Then we have
\begin{align*}
    \hat{P}_{t_k} \left( {\mathcal E}^{ {\mathrm c}}_{t_k,i,j,\tau} \right)
    \le & \hat{P}_{t_k} \left(  \left|\frac{1}{\hat{\mu}_{i,j}(t_k+\tau)}-\frac{1}{\mu_{i,j}(t_k+\tau)}\right|> 2\sqrt{\frac{U_{\mathrm{S}}^2\log \frac{\gamma}{1-\gamma} }{\hat{N}_{i,j}(t_k+\tau)}},  \hat{N}_{i,j}(t_k+\tau)>4\log \frac{\gamma}{1-\gamma} \right) \nonumber\\
     &  + \hat{P}_{t_k} \left(  \left|\frac{1}{\hat{\mu}_{i,j}(t_k+\tau)}-\frac{1}{\mu_{i,j}(t_k+\tau)}\right|> 2\sqrt{\frac{U_{\mathrm{S}}^2\log \frac{\gamma}{1-\gamma}}{\hat{N}_{i,j}(t_k+\tau)}},
     \hat{N}_{i,j}(t_k+\tau)\le 4\log  \frac{\gamma}{1-\gamma}\right).
\end{align*}
Since
\begin{align*}
    & \hat{P}_{t_k} \left(  \left|\frac{1}{\hat{\mu}_{i,j}(t_k+\tau)}-\frac{1}{\mu_{i,j}(t_k+\tau)}\right|> 2\sqrt{\frac{ U_{\mathrm{S}}^2\log \frac{\gamma}{1-\gamma}}{\hat{N}_{i,j}(t_k+\tau)}},
     \hat{N}_{i,j}(t_k+\tau)\le 4\log \frac{\gamma}{1-\gamma} \right)\\
     \le & \hat{P}_{t_k} \left(  \left|\frac{1}{\hat{\mu}_{i,j}(t_k+\tau)}-\frac{1}{\mu_{i,j}(t_k+\tau)}\right|> U_{\mathrm{S}} \right) = 0,
\end{align*}
we then have
\begin{align}\label{equ:conc-prob-1}
    & \hat{P}_{t_k} \left( {\mathcal E}^{  {\mathrm c}}_{t_k,i,j,\tau} \right)\nonumber\\
    \le & \hat{P}_{t_k} \left(  \left|\frac{1}{\hat{\mu}_{i,j}(t_k+\tau)}-\frac{1}{\mu_{i,j}(t_k+\tau)}\right|> 2\sqrt{\frac{U_{\mathrm{S}}^2\log \frac{\gamma}{1-\gamma}}{\hat{N}_{i,j}(t_k+\tau)}},  \hat{N}_{i,j}(t_k+\tau)>4\log \frac{\gamma}{1-\gamma}\right) \nonumber\\
     = & \hat{P}_{t_k} \Biggl(  \left|\hat{\phi}_{i,j}(t_k+\tau)-\frac{\hat{N}_{i,j}(t_k+\tau)}{\mu_{i,j}(t_k+\tau)}\right|> 2\sqrt{\hat{N}_{i,j}(t_k+\tau) U_{\mathrm{S}}^2\log \frac{\gamma}{1-\gamma}}, \hat{N}_{i,j}(t_k+\tau)>4\log \frac{\gamma}{1-\gamma}\Biggr).
\end{align}
Define a random mapping $f_j$ that maps a time slot to another time slot such that if $y = f_j(x)$ then $y$ is the time slot when server $j$ picked the job that was being served at server $j$ in time slot $x$. If server $j$ was idling in time slot $x$, then let $f_j(x)=x$. That is,
\begin{align*}
    f_j(x)\coloneqq\max\left\{ t: t\le x, \hat{i}^*_{j}(t) = I_j(x) \right\}.
\end{align*}
Let $M_1$ be such that $t_k+M_1$ is the first time slot when server $j$ picked queue $i$ at or after $t_k$, i,e.,
\begin{align*}
    M_1 \coloneqq \min\left\{m: m\ge 0, \hat{i}^*_{j}(t_k+m) = i \right\}.
\end{align*}
Let $M_2\coloneqq M_1 + (S_{i,j}(t_k+M_1) - 1)\eta_j(t_k+M_1)$. Then $t_k+M_2$ is the time slot when the job picked by server $j$ at time $t_k+M_1$ was completed or if server $j$ was idling because the selected waiting queue is empty at $t_k+M_1$, $M_2=M_1$. Hence, $t_k+M_1=f_j(t_k+M_2)$.
Note that $M_1,M_2$ are random variables. Then according to the algorithm, for $\tau>M_2$, we have
\begin{align}\label{equ:phi-equal-sum}
     & \hat{\phi}_{i,j}(t_k+\tau)\nonumber\\
    = & \gamma^{\tau-M_2}\hat{\phi}_{i,j}(t_k+M_2) \nonumber\\
    & + \sum_{m=M_2+1}^{\tau} \gamma^{\tau-m+M_{i,j}(t_k+m-1)} \mathbb{1}_{i,j}(t_k+m-1) \eta_j(t_k+m-1)  [M_{i,j}(t_k+m-1) + 1]\nonumber\\
    =& \gamma^{\tau-M_2}\hat{\phi}_{i,j}(t_k+M_2) \nonumber\\
    & + \sum_{m=M_2+1}^{\tau} \gamma^{\tau-m+S_{i,j}(f_j(t_k+m-1)) - 1} \mathbb{1}_{i,j}(t_k+m-1) \eta_j(f_j(t_k+m-1))  S_{i,j}(f_j(t_k+m-1)).
\end{align}
Note that the above equalities hold even when $\tau \le M_2$ if we define $\sum_{m=a}^{b} g(m)\coloneqq 0$ if $a>b$ for any function $g$. 
To see this, we first prove the following claim:
\begin{claim}\label{claim:no-job-completion}
There is no job completion of type $i$ at server $j$ in the time interval $[t_k+\frac{T}{8}, t_k+M_2-1]$ if $M_2-1 \ge \frac{T}{8}$.    
\end{claim}
\proof
    This can be proved by contradiction. Suppose there is a job of type $i$ that was completed at server $j$ in $[t_k+\frac{T}{8}, t_k+M_2-1]$. Then the job must start at or after $t_k$ since $U_{\mathrm{S}} \le \frac{T}{8}$. Also, the job must start before $t_k+M_1$ since there is another job at server $j$ starting at $t_k+M_1$ and finishes at $t_k+M_2$ by the definition of $M_1$ and $M_2$. Therefore, the job that was completed at server $j$ in $[t_k+\frac{T}{8}, t_k+M_2-1]$ should start in the time interval $[t_k, t_k+M_1-1]$.
    However, by the definition of $M_1$, there should not be any job of type $i$ starting at server $j$ in $[t_k, t_k+M_1-1]$, which is a contradiction.
\endproof

If $M_2 > \tau$, then $M_2 > \tau \ge D_k - \frac{T}{8} \ge \frac{T}{8}$, and hence $M_2-1 \ge \frac{T}{8}$. By Claim~\ref{claim:no-job-completion}, there is no job completion of type $i$ at server $j$ in the time interval $[t_k+\frac{T}{8}, t_k+M_2-1]$. Hence, there is no job completion of type $i$ at server $j$ in the time interval $[t_k+\tau, t_k+M_2-1]$ since $\tau \ge \frac{T}{8}$. Therefore, we have $\hat{\phi}_{i,j}(t_k+M_2) = \gamma^{M_2 - \tau} \hat{\phi}_{i,j}(t_k+\tau)$ if $M_2>\tau$. Note that this also holds if $M_2=\tau$. Therefore, \eqref{equ:phi-equal-sum} holds for both $\tau\le M_2$ and $\tau > M_2$.

Note that the summation in~\eqref{equ:phi-equal-sum} only includes the time slots when there is job completion of queue $i$ at server $j$. This can be transformed into summing over the time slots when server $j$ is available and picks queue $i$, i.e.,
\begin{align*}
    \hat{\phi}_{i,j}(t_k+\tau)
    =  \gamma^{\tau-M_2}\hat{\phi}_{i,j}(t_k+M_2) + \sum_{m=M_1+1}^{M_3} \gamma^{\tau-m} \mathbb{1}_{\hat{i}^*_j(t_k+m-1)=i}~ \eta_j(t_k+m-1) S_{i,j}(t_k+m-1)
\end{align*}
where $M_3$ is a random variable such that $t_k+M_3=f_j(t_k+\tau)$. Since there is no job of type $i$ starting at server $j$ in the time interval $[t_k, t_k+M_1-1]$, we have
\begin{align}\label{equ:phi-eq-sum}
    \hat{\phi}_{i,j}(t_k+\tau)
    =  \gamma^{\tau-M_2}\hat{\phi}_{i,j}(t_k+M_2) + \sum_{m=1}^{M_3} \gamma^{\tau-m} \mathbb{1}_{\hat{i}^*_j(t_k+m-1)=i}~ \eta_j(t_k+m-1) S_{i,j}(t_k+m-1).
\end{align}
Consider the case where $M_2-1 \ge \frac{T}{8}$.
Based on Claim~\ref{claim:no-job-completion} we know that there is no job completion of type $i$ at server $j$ in the time interval $[t_k+\frac{T}{8}, t_k+M_2-1]$ if $M_2-1 \ge \frac{T}{8}$. Then from the algorithm we have
\begin{align*}
    \hat{\phi}_{i,j}(t_k+M_2) = \gamma^{M_2-\frac{T}{8}}\hat{\phi}_{i,j}\left(t_k+\frac{T}{8}\right) = \gamma^{M_2-\min\left\{\frac{T}{8}, M_2\right\}}\hat{\phi}_{i,j}\left(t_k+\min\left\{\frac{T}{8}, M_2\right\}\right),
\end{align*}
where the last equality holds since $\min\left\{\frac{T}{8}, M_2\right\} = \frac{T}{8}$. If $M_2-1 < \frac{T}{8}$, we have
\begin{align*}
    \hat{\phi}_{i,j}(t_k+M_2)  = \gamma^{M_2-\min\left\{\frac{T}{8}, M_2\right\}}\hat{\phi}_{i,j}\left(t_k+\min\left\{\frac{T}{8}, M_2\right\}\right)
\end{align*}
since $\min\left\{\frac{T}{8}, M_2\right\} = M_2$. Combining the above two cases, we have
\begin{align}\label{equ:gamma-phi-eq}
    \gamma^{\tau-M_2}\hat{\phi}_{i,j}(t_k+M_2) = \gamma^{\tau-\min\left\{\frac{T}{8},M_2\right\}}\hat{\phi}_{i,j}\left(t_k+\min\left\{\frac{T}{8},M_2\right\}\right).
\end{align}
We want to upper bound this term. Since 
\begin{align*}
    \hat{\phi}_{i,j}\left(t_k+\min\left\{\frac{T}{8},M_2\right\}\right) \le U_{\mathrm{S}}\sum_{t=0}^{\infty} \gamma^t = \frac{U_{\mathrm{S}}}{1-\gamma},
\end{align*}
we have
\begin{align}\label{equ:gamma-phi-ineq}
    \gamma^{\tau-\min\left\{\frac{T}{8},M_2\right\}}\hat{\phi}_{i,j}\left(t_k+\min\left\{\frac{T}{8},M_2\right\}\right) \le \gamma^{\tau-\min\left\{\frac{T}{8},M_2\right\}}  \frac{U_{\mathrm{S}}}{1-\gamma}
    \le  \gamma^{D_k-\frac{T}{4}}  \frac{U_{\mathrm{S}}}{1-\gamma}
    \le  \gamma^{\frac{T}{4}}  \frac{U_{\mathrm{S}}}{1-\gamma},
\end{align}
where the second inequality holds since $\tau \ge D_k-\frac{T}{8}$ and the third inequality uses the bound on $D_k$ in Lemma~\ref{lemma:prop-D}.
Note that 
\begin{align}\label{equ:lm1-gamma-power-T}
    \gamma^{\frac{T}{4}} = \left(\gamma^{\frac{1}{1-\gamma}}\right)^{\log \frac{1}{1-\gamma}} \le \exp\left(-\log \frac{1}{1-\gamma}\right) = 1-\gamma,
\end{align}
where the inequality is due to the fact that
$\gamma^{\frac{1}{1-\gamma}} \le \lim_{\gamma \rightarrow 1} \gamma^{\frac{1}{1-\gamma}} =  \lim_{x\rightarrow \infty} (1-\frac{1}{x})^x = e^{-1}$ since $\gamma^{\frac{1}{1-\gamma}}$ is increasing in $\gamma$. From \eqref{equ:gamma-phi-ineq} and \eqref{equ:lm1-gamma-power-T}, we have
\begin{align}\label{equ:gamma-phi-ineq-2}
    \gamma^{\tau-\min\left\{\frac{T}{8},M_2\right\}}\hat{\phi}_{i,j}\left(t_k+\min\left\{\frac{T}{8},M_2\right\}\right)
    \le  U_{\mathrm{S}}.
\end{align}
Substituting~\eqref{equ:gamma-phi-ineq-2} into~\eqref{equ:gamma-phi-eq} and then into~\eqref{equ:phi-eq-sum}, we have
\begin{align}\label{equ:lemma-phi}
    \hat{\phi}_{i,j}(t_k+\tau) \le U_{\mathrm{S}} + \sum_{m=1}^{M_3} \gamma^{\tau-m} \mathbb{1}_{\hat{i}^*_j(t_k+m-1)=i}~ \eta_j(t_k+m-1) S_{i,j}(t_k+m-1).
\end{align}
Similarly, we have
\begin{align}\label{equ:lemma-N}
    \hat{N}_{i,j}(t_k+\tau) \le 1 + \sum_{m=1}^{M_3} \gamma^{\tau-m} \mathbb{1}_{\hat{i}^*_j(t_k+m-1)=i}~ \eta_j(t_k+m-1).
\end{align}
From~\eqref{equ:phi-eq-sum}, we also have
\begin{align}\label{equ:lemma-phi-2}
    \hat{\phi}_{i,j}(t_k+\tau) \ge \sum_{m=1}^{M_3} \gamma^{\tau-m} \mathbb{1}_{\hat{i}^*_j(t_k+m-1)=i}~ \eta_j(t_k+m-1) S_{i,j}(t_k+m-1).
\end{align}
Similarly, we have
\begin{align}\label{equ:lemma-N-2}
    \hat{N}_{i,j}(t_k+\tau) \ge \sum_{m=1}^{M_3} \gamma^{\tau-m} \mathbb{1}_{\hat{i}^*_j(t_k+m-1)=i}~ \eta_j(t_k+m-1).
\end{align}
Note that $E[S_{i,j}(t_k+\tau)]=\frac{1}{\mu_{i,j}(t_k+\tau)}$. Let $\epsilon_m \coloneqq \mathbb{1}_{\hat{i}^*_j(t_k+m-1)=i} \eta_j(t_k+m-1)$. Substituting \eqref{equ:lemma-phi}, \eqref{equ:lemma-N}, \eqref{equ:lemma-phi-2}, and \eqref{equ:lemma-N-2} into~\eqref{equ:conc-prob-1}, we have
\begin{align}\label{equ:conclusion-reused-in-theo-2}
    & \hat{P}_{t_k} \left( {\mathcal E}^{  {\mathrm c}}_{t_k,i,j,\tau} \right)\nonumber\\
     = & \hat{P}_{t_k} \Biggl(  \left|\hat{\phi}_{i,j}(t_k+\tau)-E[S_{i,j}(t_k+\tau)] \hat{N}_{i,j}(t_k+\tau)\right|> 2\sqrt{\hat{N}_{i,j}(t_k+\tau) U_{\mathrm{S}}^2\log \frac{\gamma}{1-\gamma}},  \nonumber\\
     & \qquad \hat{N}_{i,j}(t_k+\tau)>4\log \frac{\gamma}{1-\gamma}\Biggr)\nonumber\\
     \le & \hat{P}_{t_k} \Biggl(  \left|\sum_{m=1}^{M_3} \gamma^{\tau-m} \epsilon_m S_{i,j}(t_k+m-1) - \sum_{m=1}^{M_3} \gamma^{\tau-m} \epsilon_m E[S_{i,j}(t_k+\tau)] \right| \Biggr.\nonumber\\
     & \quad \quad \Biggl.> 2\sqrt{\hat{N}_{i,j}(t_k+\tau) U_{\mathrm{S}}^2\log \frac{\gamma}{1-\gamma}} - U_{\mathrm{S}}, ~\hat{N}_{i,j}(t_k+\tau)>4\log \frac{\gamma}{1-\gamma}\Biggr)\nonumber\\
     \le & \hat{P}_{t_k} \Biggl(  \left|\sum_{m=1}^{M_3} \gamma^{\tau-m} \epsilon_m \left(S_{i,j}(t_k+m-1) - E[S_{i,j}(t_k+m-1)]\right)\right| \Biggr. \nonumber\\
     & \quad \quad + \sum_{m=1}^{M_3} \gamma^{\tau-m} \epsilon_m \left| E[S_{i,j}(t_k+m-1)] - E[S_{i,j}(t_k+\tau)]\right| > 2\sqrt{\hat{N}_{i,j}(t_k+\tau) U_{\mathrm{S}}^2\log \frac{\gamma}{1-\gamma}} - U_{\mathrm{S}},\nonumber\\
     & \quad \quad \Biggl.\hat{N}_{i,j}(t_k+\tau)>4\log \frac{\gamma}{1-\gamma}\Biggr),
\end{align}
where in the last inequality we add and subtract the term $E[S_{i,j}(t_k+m-1)]$ and use the triangle inequality. We note that in the stationary setting, $E[S_{i,j}(t)]$ does not depend on $t$ and the last step is not needed. However, in the nonstationary setting, we need some assumptions on the variability of mean service times. Recall Assumption~\ref{assump:mu-1} (1) on the time-varying service times. We have
\begin{align*}
    \left\lvert E[S_{i,j}(t_k+m-1)]-E[S_{i,j}(t_k+\tau)]\right\rvert \le \frac{1}{T}\left(\frac{1}{\gamma}\right)^{\tau - m}.
\end{align*}
Hence, we have
\begin{align}\label{equ:conc-prob-2}
    \hat{P}_{t_k} \left( {\mathcal E}^{  {\mathrm c}}_{t_k,i,j,\tau} \right)
     \le & \hat{P}_{t_k} \Biggl(  \left|\sum_{m=1}^{M_3} \gamma^{\tau-m} \epsilon_m \left(S_{i,j}(t_k+m-1) - E[S_{i,j}(t_k+m-1)]\right)\right| 
     \Biggr.\nonumber\\
     & \quad \quad \Biggl.> 2\sqrt{\hat{N}_{i,j}(t_k+\tau) U_{\mathrm{S}}^2\log \frac{\gamma}{1-\gamma}} - U_{\mathrm{S}} - \frac{M_3}{T}, \hat{N}_{i,j}(t_k+\tau)>4\log \frac{\gamma}{1-\gamma}\Biggr).
\end{align}
Recall that $t_k+M_3=f_j(t_k+\tau)$.
Hence, $t_k+ M_3\in [t_k+\tau- U_{\mathrm{S}} + 1, t_k+\tau]$. Since $U_{\mathrm{S}} \le \frac{T}{8}$ and $\tau \in [D_k - \frac{T}{8}, D_k+D_{k+1}-1]$, we have
$
    D_k - \frac{T}{4} + 1\le M_3\le D_k+D_{k+1}-1.
$
By the bound on $D_k$ and $D_{k+1}$ in Lemma~\ref{lemma:prop-D}, we further have
\begin{align}\label{equ:bound-M_3}
    \frac{T}{4} + 1 \le M_3 \le 2T - 1.
\end{align}
Hence, we have
\begin{align}\label{equ:conc-prob-aux}
    U_{\mathrm{S}} + \frac{M_3}{T} \le U_{\mathrm{S}} + 2 \le \sqrt{4U_{\mathrm{S}}^2 \left(\log \frac{\gamma}{1-\gamma}\right)^2}\le \sqrt{U_{\mathrm{S}}^2\hat{N}_{i,j}(t_k+\tau)\log \frac{\gamma}{1-\gamma}},
\end{align}
where the second inequality holds since $\gamma \ge 1- \frac{1}{1+ e^{1.5}}$ and $U_{\mathrm{S}}\ge 1$, and the last inequality holds when $\hat{N}_{i,j}(t_k+\tau)>4\log \frac{\gamma}{1-\gamma}$. Based on~\eqref{equ:conc-prob-aux}, we can continue to bound~\eqref{equ:conc-prob-2} and obtain
\begin{align}\label{equ:conc-prob-3}
     &\hat{P}_{t_k} \left( {\mathcal E}^{  {\mathrm c}}_{t_k,i,j,\tau} \right)\nonumber\\
     \le & \hat{P}_{t_k} \Biggl(  \left|\sum_{m=1}^{M_3} \gamma^{\tau-m} \epsilon_m \left(S_{i,j}(t_k+m-1) - E[S_{i,j}(t_k+m-1)]\right)\right| 
     > \sqrt{\hat{N}_{i,j}(t_k+\tau) U_{\mathrm{S}}^2\log \frac{\gamma}{1-\gamma}}\Biggr).
\end{align}
From~\eqref{equ:lemma-N-2} we know
$
    \hat{N}_{i,j}(t_k+\tau) \ge \sum_{m=1}^{M_3} \gamma^{\tau-m} \mathbb{1}_{\hat{i}^*_j(t_k+m-1)=i}~ \eta_j(t_k+m-1) = \sum_{m=1}^{M_3} \gamma^{\tau-m} \epsilon_m.
$
Hence, we can further bound~\eqref{equ:conc-prob-3} as
\begin{align*}
    \hat{P}_{t_k} \left( {\mathcal E}^{  {\mathrm c}}_{t_k,i,j,\tau} \right)
     \le & \hat{P}_{t_k} \Biggl(  \frac{\left|\sum_{m=1}^{M_3} \gamma^{\tau-m} \epsilon_m \left(S_{i,j}(t_k+m-1) - E[S_{i,j}(t_k+m-1)]\right)\right|}{\sqrt{\sum_{m=1}^{M_3} \gamma^{\tau-m} \epsilon_m}}
     > \sqrt{U_{\mathrm{S}}^2\log \frac{\gamma}{1-\gamma}}\Biggr).
\end{align*}
Since $M_3\le \tau$ and $\gamma < 1$, we have $\sqrt{\gamma^{M_3-\tau}} \ge 1$. Hence, we have
\begin{align}\label{equ:conc-prob-4}
    & \hat{P}_{t_k} \left( {\mathcal E}^{  {\mathrm c}}_{t_k,i,j,\tau} \right)\nonumber\\
    \le & \hat{P}_{t_k} \Biggl(  \frac{ \gamma^{M_3-\tau} \left|\sum_{m=1}^{M_3} \gamma^{\tau-m} \epsilon_m \left(S_{i,j}(t_k+m-1) - E[S_{i,j}(t_k+m-1)]\right)\right|}{ \sqrt{\gamma^{M_3-\tau}} \sqrt{\sum_{m=1}^{M_3} \gamma^{\tau-m} \epsilon_m}}
     > \sqrt{ U_{\mathrm{S}}^2\log \frac{\gamma}{1-\gamma}}\Biggr)\nonumber\\
     = & \hat{P}_{t_k} \Biggl(  \frac{\left|\sum_{m=1}^{M_3} \gamma^{M_3-m} \epsilon_m \left(S_{i,j}(t_k+m-1) - E[S_{i,j}(t_k+m-1)]\right)\right|}{\sqrt{\sum_{m=1}^{M_3} \gamma^{M_3-m} \epsilon_m}}
     > \sqrt{ U_{\mathrm{S}}^2\log \frac{\gamma}{1-\gamma}}\Biggr)\nonumber\\
     \le & \hat{P}_{t_k} \Biggl(  \frac{\left|\sum_{m=1}^{M_3} \gamma^{M_3-m} \epsilon_m \left(S_{i,j}(t_k+m-1) - E[S_{i,j}(t_k+m-1)]\right)\right|}{\sqrt{\sum_{m=1}^{M_3} \gamma^{2(M_3-m)} \epsilon_m}}
     > \sqrt{ U_{\mathrm{S}}^2\log \frac{\gamma}{1-\gamma}}\Biggr),
\end{align}
where we added ``2'' in the last inequality because we want to use the Hoeffding-type inequality for self-normalized means~\cite[Theorem 22]{GarMou_08} later in the proof.

Consider the event
$
    {\mathcal E}_{t_k,i,j}\coloneqq \bigcap_{\tau=D_k-\frac{T}{8}}^{D_k+D_{k+1}-1} {\mathcal E}_{t_k,i,j,\tau}.
$
Then from the result~\eqref{equ:conc-prob-4}, we have
\begin{align}\label{equ:conc-prob-5}
    &\hat{P}_{t_k} \left( {\mathcal E}^{  {\mathrm c}}_{t_k,i,j} \right)\nonumber\\
    \le &  \hat{P}_{t_k} \Biggl( \mbox{there exists } \tau\in \left[D_k-\frac{T}{8}, D_k+D_{k+1} - 1\right],\Biggr.\nonumber\\
    &\quad \quad \Biggl.\frac{\left|\sum_{m=1}^{M_3} \gamma^{M_3-m} \epsilon_m \left(S_{i,j}(t_k+m-1) - E[S_{i,j}(t_k+m-1)]\right)\right|}{\sqrt{\sum_{m=1}^{M_3} \gamma^{2(M_3-m)} \epsilon_m}}
     > \sqrt{ U_{\mathrm{S}}^2\log \frac{\gamma}{1-\gamma}}\Biggr)\nonumber\\
     \le & \hat{P}_{t_k} \Biggl( \mbox{there exists } M\in \left[\frac{T}{4}+1, 2T - 1\right],\Biggr.\nonumber\\
    &\quad \quad \Biggl.\frac{\left|\sum_{m=1}^{M} \gamma^{M-m} \epsilon_m \left(S_{i,j}(t_k+m-1) - E[S_{i,j}(t_k+m-1)]\right)\right|}{\sqrt{\sum_{m=1}^{M} \gamma^{2(M-m)} \epsilon_m}}
     > \sqrt{ U_{\mathrm{S}}^2\log \frac{\gamma}{1-\gamma}}\Biggr)\nonumber\\
     \le & \hat{P}_{t_k} \Biggl( \sup_{1\le M\le 2T - 1}\frac{\left|\sum_{m=1}^{M} \gamma^{M-m} \epsilon_m \left(S_{i,j}(t_k+m-1) - E[S_{i,j}(t_k+m-1)]\right)\right|}{\sqrt{\sum_{m=1}^{M} \gamma^{2(M-m)} \epsilon_m}}
     > \sqrt{U_{\mathrm{S}}^2\log \frac{\gamma}{1-\gamma}}\Biggr).
\end{align}
where the second inequality uses the bound~\eqref{equ:bound-M_3} on $M_3$.

We restate the Hoeffding-type inequality for self-normalized means in~\cite[Theorem 22]{GarMou_08},~\cite{GarMou_11} as follows:
\begin{theorem}[Hoeffding-type inequality for self-normalized means]\label{theo:a-restate}
\textbf{\emph{(Theorem 22 in~\cite{GarMou_08})}}
~\\
Let $(X_m)_{m\ge 1}$ be a sequence of nonnegative independent bounded random variables defined on a probability space $(\Omega, {\cal A}, \hat{P})$ with $X_m\in[0, B]$. Let $\hat{E}$ be the expectation under the probability measure $\hat{P}$. Let ${\mathcal F}_{m}$ be an increasing sequence of $\sigma$-algebras of ${\cal A}$ such that $\sigma\left(X_1,\ldots,X_m\right)\subset {\mathcal F}_{m}$ and for $n>m$, $X_n$ is independent of ${\mathcal F}_{m}$. Consider a previsible sequence $(\epsilon_m)_{m\ge 1}$ of Bernoulli random variables, i.e., $\epsilon_m$ is ${\mathcal F}_{m-1}$-measurable. 
For all positive integers $N$ and all $\beta>0$, 
\begin{align*}
    & \hat{P} \Biggl(  \sup_{1\le M \le N}\frac{\sum_{m=1}^{M} \gamma^{M-m}  X_m \epsilon_m - \sum_{m=1}^{M} \gamma^{M-m}  \hat{E}[X_m] \epsilon_m}{\sqrt{\sum_{m=1}^{M} \gamma^{2(M-m)} \epsilon_m}}
     > \beta \Biggr) \nonumber\\
     \le & \left\lceil \frac{\log \left(\gamma^{-2N}\sum_{m=1}^{N} (\gamma^2)^{N-m} \right)}{\log (1+\zeta)} \right\rceil \exp \left( -\frac{2\beta^2}{B^2} \left(1-\frac{\zeta^2}{16}\right)\right)
\end{align*}
for all $\zeta>0$.
\end{theorem}
Let us view the conditional probability $\hat{P}_{t_k}$ as a new probability measure. Then $\hat{E}_{t_k}$ is the expectation under this measure.
Note that $\left(S_{i,j}(t_k+m-1)\right)_{m=1}^{\infty}$ is a sequence of independent bounded random variables under this new measure since they are independent of $\boldsymbol{Q}(t_k)$ and $\boldsymbol{H}(t_k)$, which also implies that
\begin{align}\label{equ:expt-expt-hat}
    E[S_{i,j}(t_k+m-1)] = \hat{E}_{t_k}[S_{i,j}(t_k+m-1)].
\end{align}
Let ${\mathcal F}_m$ defined as
\begin{align}\label{equ:def-filtration}
    {\mathcal F}_m \coloneqq \sigma \left( (\boldsymbol{S}(t_k+n-1))_{n=1}^m, (\boldsymbol{A}(t_k+n-1))_{n=1}^{m+1}, (\boldsymbol{Q}(t_k+n-1))_{n=1}^{m+1}, (\boldsymbol{H}(t_k+n-1))_{n=1}^{m+1}\right),
\end{align}
where $\sigma(\cdot)$ denotes the $\sigma$-algebra generated by the random variables.
Note that 
\begin{align*}
    \sigma(S_{i,j}(t_k),...,S_{i,j}(t_k+m-1))\subset {\mathcal F}_m
\end{align*}
and for any $n>m$, $S_{i,j}(t_k+n-1)$ is independent of ${\mathcal F}_m$.
Recall that $\epsilon_m \coloneqq \mathbb{1}_{\hat{i}^*_j(t_k+m-1)=i} \eta_j(t_k+m-1)$. Since the scheduling decision at time $t_k+m-1$ is determined by $\boldsymbol{Q}(t_k+m-1)$ and $\boldsymbol{H}(t_k+m-1)$, $\mathbb{1}_{\hat{i}^*_j(t_k+m-1)=i}$ is ${\mathcal F}_{m-1}$-measurable. Since $\eta_j(t_k+m-1)$ is determined by $\boldsymbol{A}(t_k+m-1)$, $\boldsymbol{Q}(t_k+m-1)$, and $\boldsymbol{H}(t_k+m-1)$, $\eta_j(t_k+m-1)$ is also ${\mathcal F}_{m-1}$-measurable. Therefore, $\epsilon_m$ is ${\mathcal F}_{m-1}$-measurable, i.e., $\left(\epsilon_m\right)_{m=1}^{\infty}$ is a previsible (or predictable) sequence of Bernoulli random variables.
Therefore,
Applying Theorem~\ref{theo:a-restate} with $\hat{P}=\hat{P}_{t_k}$, $X_m=S_{i,j}(t_k+m-1)$, $\beta=\sqrt{U_{\mathrm{S}}^2\log \frac{\gamma}{1-\gamma}}$, $B=U_{\mathrm{S}}$, and $N=2T-1$, we have
\begin{align}\label{equ:bound-single-side-1}
    & \hat{P}_{t_k} \Biggl( \sup_{1\le M \le 2T-1} \frac{\sum_{m=1}^{M} \gamma^{M-m} \epsilon_m \left(S_{i,j}(t_k+m-1) - \hat{E}_{t_k}[S_{i,j}(t_k+m-1)]\right)}{\sqrt{\sum_{m=1}^{M} \gamma^{2(M-m)} \epsilon_m}}
     > \sqrt{U_{\mathrm{S}}^2\log \frac{\gamma}{1-\gamma}}\Biggr)\nonumber\\
     \le & \left(\frac{\log \left(\gamma^{-4T+2}\sum_{m=1}^{2T-1} (\gamma^2)^{2T-1-m} \right)}{\log (1+\zeta)}+1\right)\exp\left(-2\left(1-\frac{\zeta^2}{16}\right)\log \frac{\gamma}{1-\gamma}\right)
\end{align}
for all $\zeta>0$.
Note that
\begin{align}\label{equ:bound-single-side-part}
    \log \left(\gamma^{-4T+2}\sum_{m=1}^{2T-1} (\gamma^2)^{2T-1-m} \right) 
    \le & \log \left( \gamma^{-4T+2} \frac{1}{1-\gamma^2}\right) = (4T-2)\log \left( \frac{1}{\gamma} \right) + \log \left(\frac{1}{1-\gamma^2}\right)\nonumber\\
    \le & (4T-2)\left(\frac{1-\gamma}{\gamma}\right) + \log \left(\frac{1}{1-\gamma^2}\right)\nonumber\\
    \le & 4T\left(\frac{1-\gamma}{\gamma}\right) + \log \left(\frac{1}{1-\gamma}\right)\nonumber\\
    = & \left(1+\frac{16}{\gamma}\right)\log \frac{1}{1-\gamma},
\end{align}
where the second inequality is due to the fact that $\log x \le x-1$ for any $x>0$, and the last equality is by the definition of $T$.
Substituting~\eqref{equ:bound-single-side-part} into~\eqref{equ:bound-single-side-1}, we have
\begin{align}\label{equ:overestimation}
    & \hat{P}_{t_k} \Biggl( \sup_{1\le M \le 2T-1} \frac{\sum_{m=1}^{M} \gamma^{M-m} \epsilon_m \left(S_{i,j}(t_k+m-1) - \hat{E}_{t_k}[S_{i,j}(t_k+m-1)]\right)}{\sqrt{\sum_{m=1}^{M} \gamma^{2(M-m)} \epsilon_m}}
     > \sqrt{U_{\mathrm{S}}^2\log \frac{\gamma}{1-\gamma}}\Biggr)\nonumber\\
     \le & \left(\frac{\left(1+\frac{16}{\gamma}\right)\log \frac{1}{1-\gamma}}{\log (1+\zeta)}+1\right)\exp\left(-2\left(1-\frac{\zeta^2}{16}\right)\log \frac{\gamma}{1-\gamma}\right)
\end{align}
for all $\zeta>0$.
Although in~\cite{GarMou_08} the bound is only proved for overestimation, the proof can be extended to show that the bound also holds for underestimation. Specifically, note that
\begin{align*}
    \hat{E}_{t_k} \left[\hat{E}_{t_k}[S_{i,j}(t_k+m-1)] - S_{i,j}(t_k+m-1)\right]=0
\end{align*}
and
\begin{align*}
    \hat{E}_{t_k}[S_{i,j}(t_k+m-1)]-U_{\mathrm{S}}\le \hat{E}_{t_k}[S_{i,j}(t_k+m-1)]-S_{i,j}(t_k+m-1)\le \hat{E}_{t_k}[S_{i,j}(t_k+m-1)].
\end{align*}
Hence, considering the random variable $\hat{E}_{t_k}[S_{i,j}(t_k+m-1)] - S_{i,j}(t_k+m-1)$, from~\cite[Lemma 8.1]{devroye1996probabilistic}, for any $\lambda>0$, we have
\begin{align*}
    \log \hat{E}_{t_k} [\exp (-\lambda S_{i,j}(t_k+m-1))] \le \frac{\lambda^2U^2_{\mathrm{S}}}{8}-\lambda \hat{E}_{t_k}[S_{i,j}(t_k+m-1)].
\end{align*}
Hence, we can apply the same proof in~\cite[Theorem 22]{GarMou_08} by replacing $\lambda$ in the proof with $-\lambda$. Then for underestimation, we also have the same bound, i.e.,
\begin{align}\label{equ:underestimation}
    & \hat{P}_{t_k} \Biggl( \sup_{1\le M \le 2T-1} \frac{\sum_{m=1}^{M} \gamma^{M-m} \epsilon_m \left(\hat{E}_{t_k}[S_{i,j}(t_k+m-1)] - S_{i,j}(t_k+m-1) \right)}{\sqrt{\sum_{m=1}^{M} \gamma^{2(M-m)} \epsilon_m}}
     > \sqrt{U_{\mathrm{S}}^2\log \frac{\gamma}{1-\gamma}}\Biggr)\nonumber\\
     \le & \left(\frac{\left(1+\frac{16}{\gamma}\right)\log \frac{1}{1-\gamma}}{\log (1+\zeta)}+1\right)\exp\left(-2\left(1-\frac{\zeta^2}{16}\right)\log \frac{\gamma}{1-\gamma}\right)
\end{align}
for all $\zeta>0$. Taking the union bound over underestimation~\eqref{equ:underestimation} and overestimation~\eqref{equ:overestimation}, we have
\begin{align*}
    & \hat{P}_{t_k} \Biggl( \sup_{1\le M\le 2T-1} \frac{\left|\sum_{m=1}^{M} \gamma^{M-m} \epsilon_m \left(S_{i,j}(t_k+m-1) - \hat{E}_{t_k}[S_{i,j}(t_k+m-1)]\right)\right|}{\sqrt{\sum_{m=1}^{M} \gamma^{2(M-m)} \epsilon_m}}
     > \sqrt {U_{\mathrm{S}}^2\log \frac{\gamma}{1-\gamma}}\Biggr)\nonumber\\
    \le & 2\left(\frac{\left(1+\frac{16}{\gamma}\right)\log \frac{1}{1-\gamma}}{\log (1+\zeta)}+1\right)\exp\left(-2\left(1-\frac{\zeta^2}{16}\right)\log \frac{\gamma}{1-\gamma}\right)
\end{align*}
for all $\zeta>0$.
Setting $\zeta=0.3$, we have
\begin{align*}
    & \hat{P}_{t_k} \Biggl( \sup_{1\le M\le 2T-1} \frac{\left|\sum_{m=1}^{M} \gamma^{M-m} \epsilon_m \left(S_{i,j}(t_k+m-1) - \hat{E}_{t_k}[S_{i,j}(t_k+m-1)]\right)\right|}{\sqrt{\sum_{m=1}^{M} \gamma^{2(M-m)} \epsilon_m}}
     > \sqrt {U_{\mathrm{S}}^2\log \frac{\gamma}{1-\gamma}}\Biggr)\nonumber\\
    \le & \left[\left(8+\frac{122}{\gamma}\right)\log \frac{1}{1-\gamma} + 2\right]\left(\frac{1-\gamma}{\gamma}\right)^{1.99}.
\end{align*}
Since $\gamma \ge 1-\frac{1}{1+e^{1.5}}$ and $(1-\gamma)^{0.49} \log \frac{1}{1-\gamma} \le \frac{1}{0.49e}$, we have
\begin{align}\label{equ:conc-prob-6}
    & \hat{P}_{t_k} \Biggl( \sup_{1\le M\le 2T-1} \frac{\left|\sum_{m=1}^{M} \gamma^{M-m} \epsilon_m \left(S_{i,j}(t_k+m-1) - \hat{E}_{t_k}[S_{i,j}(t_k+m-1)]\right)\right|}{\sqrt{\sum_{m=1}^{M} \gamma^{2(M-m)} \epsilon_m}}
     > \sqrt {U_{\mathrm{S}}^2\log \frac{\gamma}{1-\gamma}}\Biggr)\nonumber\\
    \le & 186(1-\gamma)^{1.5}.
\end{align}

Combining~\eqref{equ:conc-prob-5},~\eqref{equ:expt-expt-hat}, and~\eqref{equ:conc-prob-6}, we have
\begin{align*}
    \hat{P}_{t_k} \left( {\mathcal E}^{  {\mathrm c}}_{t_k,i,j} \right) \le 186(1-\gamma)^{1.5}
\end{align*}
for all $1-\frac{1}{1+e^{1.5}}\le \gamma < 1 $. Taking the union bound over $i$, we have
\begin{align*}
    \hat{P}_{t_k} ({\mathcal E}_{t_k,j}^{ {\mathrm c}}) \le \sum_{i=1}^{I} \hat{P}_{t_k} \left( {\mathcal E}^{ {\mathrm c}}_{t_k,i,j} \right) \le 186I(1-\gamma)^{1.5}.
\end{align*}

\endproof

\subsection{Proof of Lemma~\ref{lemma:prop-D}}
\label{app:proof-lemma-prop-D}

\proof
Recall the definition of $D(t)$:
\begin{align*}
    D(t)=& \min_{n} \sum_{l=0}^n w(\tau_l(t))\\
    & \mbox{s.t. } \sum_{l=0}^{n} w(\tau_l(t)) \ge \frac{T}{2}.
\end{align*}
Recall that $n^*(t)$ is the optimal solution to the above optimization problem. Note that
$D(t)=\sum_{l=0}^{n^*(t)} w(\tau_l(t)) \ge \frac{T}{2}$
and
$\sum_{l=0}^{n^*(t)-1} w(\tau_l(t)) <\frac{T}{2}$.
Hence, we have
\begin{align*}
    D(t)=\sum_{l=0}^{n^*(t)}w(\tau_l(t))=\sum_{l=0}^{n^*(t)-1}w(\tau_l(t)) + w(\tau_{n^*(t)}(t))\le \frac{T}{2} + W,
\end{align*}
where the last inequality is due to the bound $w(\tau)\le W$ for any $\tau$. Therefore, for any $t$, we have
\begin{align*}
    \frac{T}{2}\le D(t) \le \frac{T}{2}+W \le T,
\end{align*}
where the last inequality is by $W\le \frac{T}{2}$.
 
\endproof

\subsection{Proof of Lemma~\ref{lemma:queue-bound}}
\label{app:proof-lemma-queue-bound}

\proof
Fix $i$ and $t$. Consider two cases. The first case is that there exists $j$ such that $\eta_j(t)=0$ (server $j$ is idling) and $I_j(t)=i$. The second case is that for all servers $j$, $\eta_j(t)=1$ or $I_j(t)\neq i$.

Notice that for the first case we must have
\begin{align*}
    \tilde{Q}_{i}(t) + A_{i}(t) = 0
\end{align*}
since server $j$ is scheduled to $i$ and is idling. Hence, we have
\begin{align*}
    Q_{i}(t) + A_{i}(t) \le \tilde{Q}_{i}(t) + J + A_{i}(t) = J.
\end{align*}
Hence, by the queue dynamics~\eqref{equ:queue-dynamics} and the above inequality, we have
\begin{align*}
    Q_i(t+1) \le Q_i(t)+A_i(t) \le J
\end{align*}
for the first case.
For the second case, we have
\begin{align*}
    Q_i(t+1) =& Q_i(t)+A_i(t)-\sum_j \mathbb{1}_{i,j}(t)\eta_j(t)\nonumber\\
    =& Q_i(t)+A_i(t)-\sum_j \mathbb{1}_{i,j}(t),
\end{align*}
where the second inequality holds since for any server $j$, either $\eta_j(t)=1$ or $\mathbb{1}_{i,j}(t)=0$.

Combining the two cases, we obtain that for any $i,t$,
\begin{align*}
    Q_i(t+1) \le \max \left\{ J, Q_i(t)+A_i(t)-\sum_j \mathbb{1}_{i,j}(t)\right\}.
\end{align*}
 
\endproof

\subsection{Proof of Lemma~\ref{lemma:bounds-on-diff-queue-lens}}
\label{app:proof-lemma-bounds-on-diff-queue-lens}

\proof
(1) holds since $Q_{i}(t)$ can increase at most $U_{\mathrm{A}}$ and can decrease at most $J$ for each time slot by the queue dynamics~\eqref{equ:queue-dynamics}.

(2) holds since the total queue length can decrease by at most $J$ for each time slot. This is because there are $J$ servers in total and each server can serve at most one job at a time.

\endproof

\subsection{Proof of Lemma~\ref{lemma:decouple-queue-and-ucb-bonus}}
\label{app:proof-lemma-decouple-queue-and-ucb-bonus}

\proof
    Let $h\coloneqq \lceil e_{i,j}z \rceil$, where we will choose $z$ later.
    Let $\tau\in[D_k, D_k+D_{k+1}-1]$. Then
    \begin{align}\label{equ:lemma-decouple-q-bound-1}
        Q_i(t_k+\tau)  = & \frac{1}{h}\sum_{\tau'=\max\{\tau- h + 1, D_k\}}^{\max\{\tau - h + 1, D_k\} + h - 1} Q_i(t_k + \tau)
    \end{align}
    We want to bound the difference $|\tau - \tau'|$. Consider the following cases.
    \begin{enumerate}
        \item $\tau-h+1 \ge D_k$:
        
        We have $\tau'\in[\tau-h+1, \tau]$. Hence,
        \begin{align*}
            |\tau - \tau'| \le h-1.
        \end{align*}

        \item $\tau-h+1 < D_k$:
        
            We have $\tau'\in[D_k, D_k+h-1]$. We further discuss the following two cases.
            \begin{enumerate}
                \item[(1)] $\tau \le \tau'$: 

                We have
                \begin{align*}
                    |\tau - \tau'| = \tau' - \tau \le (D_k+h-1) - D_k = h-1,
                \end{align*}
                since $\tau'\le D_k+h-1$ and $\tau\ge D_k$.

                \item[(2)] $\tau > \tau'$:

                We have
                \begin{align*}
                    |\tau - \tau'| = \tau - \tau' \le (D_k+h-1) - D_k = h - 1,
                \end{align*}
                since $\tau -h + 1 \le D_k$ and $\tau' \ge D_k$.
            \end{enumerate}
    \end{enumerate}
    Combining these cases, we have
    \begin{align}\label{equ:a-simple-bound}
        |\tau - \tau'| \le h-1.
    \end{align}
    By Lemma~\ref{lemma:bounds-on-diff-queue-lens} and~\eqref{equ:a-simple-bound}, we have
    \begin{align*}
        Q_i(t_k+\tau) \le Q_i(t_k+\tau') + \max\{U_{\mathrm{A}}, J\}(h-1).
    \end{align*}
    Substituting the above inequality into~\eqref{equ:lemma-decouple-q-bound-1}, we have
    \begin{align}\label{equ:lemma-decouple-q-bound-2}
        Q_i(t_k+\tau) \le & \frac{1}{h} \sum_{\tau'=\max\{\tau-h+1, D_k\}}^{\max\{\tau - h+1, D_k\} + h - 1} \left(Q_i(t_k+\tau')+\max\{U_{\mathrm{A}}, J\}(h-1)\right)\nonumber\\
        = & \frac{1}{h} \sum_{\tau'=\max\{\tau-h+1, D_k\}}^{\max\{\tau - h+1, D_k\} + h - 1} Q_i(t_k+\tau')+\max\{U_{\mathrm{A}}, J\}(h-1)
    \end{align}
     Suppose $h\le D_{k+1}$. Then we have
     \begin{align}\label{equ:upper-bound-on-upper-limit}
         \max\{\tau-h+1, D_k\}+h-1 = & \max\{\tau, D_k + h - 1\}\nonumber\\
         \le & D_k+D_{k+1}-1,
     \end{align}
     where the last inequality is by $\tau \le D_k+D_{k+1}-1$ and $h\le D_{k+1}$.
     Also note that
     \begin{align}\label{equ:lower-bound-on-lower-limit}
         \max\{\tau - h + 1, D_k\} \ge D_k.
     \end{align}
     Combining~\eqref{equ:lemma-decouple-q-bound-2},~\eqref{equ:upper-bound-on-upper-limit}, and~\eqref{equ:lower-bound-on-lower-limit}, we have
     \begin{align*}
         Q_i(t_k+\tau) \le \frac{1}{h} \sum_{\tau'=D_k}^{D_k+D_{k+1}-1} Q_i(t_k+\tau')+\max\{U_{\mathrm{A}}, J\} (h-1),
     \end{align*}
     which holds as long as $h\le D_{k+1}$.
     Therefore, if $h\le D_{k+1}$, we have
     \begin{align*}
         & \sum_{\tau=D_k}^{D_k+D_{k+1}-1} Q_{i}(t_k+\tau) \tilde{b}_{i,j}(f_j(t_k+\tau)) \eta_j(f_j(t_k+\tau))
        \mathbb{1}_{i,j}(t_k+\tau) \mathbb{1}_{I_j(t_k+\tau)=i}
        \mathbb{1}_{\hat{N}_{i,j}(f_j(t_k+\tau)) < \frac{64 J^2 U_{\mathrm{S}}^4 \log \frac{1}{1-\gamma}}{\delta^2}} \nonumber\\
        \le & \sum_{\tau=D_k}^{D_k+D_{k+1}-1} 
        \left(\frac{1}{h} \sum_{\tau'=D_k}^{D_k+D_{k+1}-1} Q_i(t_k+\tau')+\max\{U_{\mathrm{A}}, J\}(h-1)\right)\nonumber\\
        & \qquad \qquad \quad \tilde{b}_{i,j}(f_j(t_k+\tau)) \eta_j(f_j(t_k+\tau))
        \mathbb{1}_{i,j}(t_k+\tau) \mathbb{1}_{I_j(t_k+\tau)=i}
        \mathbb{1}_{\hat{N}_{i,j}(f_j(t_k+\tau)) < \frac{64 J^2 U_{\mathrm{S}}^4 \log \frac{1}{1-\gamma}}{\delta^2}}\nonumber\\
        = & \left(\frac{1}{h} \sum_{\tau'=D_k}^{D_k+D_{k+1}-1} Q_i(t_k+\tau')+\max\{U_{\mathrm{A}}, J\}(h-1)\right)\nonumber\\
        & \sum_{\tau=D_k}^{D_k+D_{k+1}-1} \tilde{b}_{i,j}(f_j(t_k+\tau)) \eta_j(f_j(t_k+\tau))
        \mathbb{1}_{i,j}(t_k+\tau) \mathbb{1}_{I_j(t_k+\tau)=i}
        \mathbb{1}_{\hat{N}_{i,j}(f_j(t_k+\tau)) < \frac{64 J^2 U_{\mathrm{S}}^4 \log \frac{1}{1-\gamma}}{\delta^2}}\nonumber\\
        = & \left(\frac{1}{h} \sum_{\tau'=D_k}^{D_k+D_{k+1}-1} Q_i(t_k+\tau')+\max\{U_{\mathrm{A}}, J\}(h-1)\right) e_{i,j}\nonumber\\
        \le & \frac{1}{z} \sum_{\tau'=D_k}^{D_k+D_{k+1}-1} Q_i(t_k+\tau') + \max\{U_{\mathrm{A}}, J\} e^2_{i,j} z,
     \end{align*}
     where the last equality holds since $ e_{i,j} z \le h \le 1 + e_{i,j} z$. let $z=\frac{4JU_{\mathrm{S}}}{\delta}$. Then
     \begin{align}\label{equ:lemma-decouple-bound}
         & \sum_{\tau=D_k}^{D_k+D_{k+1}-1} Q_{i}(t_k+\tau) \tilde{b}_{i,j}(f_j(t_k+\tau)) \eta_j(f_j(t_k+\tau))
        \mathbb{1}_{i,j}(t_k+\tau) \mathbb{1}_{I_j(t_k+\tau)=i}
        \mathbb{1}_{\hat{N}_{i,j}(f_j(t_k+\tau)) < \frac{64 J^2 U_{\mathrm{S}}^4 \log \frac{1}{1-\gamma}}{\delta^2}} \nonumber\\
        \le & \frac{\delta}{4JU_{\mathrm{S}}} \sum_{\tau'=D_k}^{D_k+D_{k+1}-1} Q_i(t_k+\tau') +  \frac{4JU_{\mathrm{S}} \max\{U_{\mathrm{A}}, J\} e^2_{i,j}}{\delta},
     \end{align}
     which holds as long as $h=\left \lceil \frac{4JU_{\mathrm{S}}e_{i,j}}{\delta}\right \rceil \le D_{k+1}$. Since $D_{k+1}\ge \frac{T}{2}$ by Lemma~\ref{lemma:prop-D}, a sufficient condition for~\eqref{equ:lemma-decouple-bound} to hold is that $2 \left \lceil  \frac{4 J U_{\mathrm{S}} e_{i,j}}{\delta} \right \rceil\le T$.
 
\endproof

\subsection{Proof of Lemma~\ref{lemma:eij}}
\label{app:proof-lemma-eij}

\proof
    Since $\tau\in [\tau_l, \tau_h]$ and $\tau_l \ge U_{\mathrm{S}}, \tau_h \le L - 1$, we have $t \le t+ \tau - U_{\mathrm{S}} \le f_j(t+\tau) \le t+\tau \le t+ L - 1$, i.e.,
    $f_j(t+\tau)\in[t, t+L-1]$.    
    Divide the time interval $[t, t+L-1]$ into parts where each part contains $\left\lceil \frac{1}{2(1-\gamma)} \right\rceil$ samples (except the last part). Then there are at most $\left\lceil 2L (1-\gamma) \right\rceil$ parts. let $P$ denote the number of parts and $l_p$ denote the length of part $p$, $p=0,\ldots,P-1$. Then $l_p\le \left\lceil \frac{1}{2(1-\gamma)} \right\rceil$ for all $p$ and $P\le \left\lceil 2L (1-\gamma) \right\rceil$. Then, $S_\mathrm{UCB}$ can be bounded by
    \begin{align}\label{equ:eij-bound}
        S_\mathrm{UCB} \le \sum_{p=0}^{P-1} \sum_{m=0}^{l_p} \tilde{b}_{i,j}(t_{p,m}) \eta_j(t_{p,m})
        \mathbb{1}_{\hat{i}^*_j(t_{p,m})=i}
        \mathbb{1}_{\hat{N}_{i,j}(t_{p,m}) < U_N},
    \end{align}
    where $t_{p,m}$ denotes the $m^{\mathrm{th}}$ time slot of part $p$.
    The summation includes only the UCB bonuses when server $j$ picks queue $i$ and server $j$ is not idling.
    Let $S_{\mathrm{UCB},p}$ denote the sum of part $p$, i.e.,
    \begin{align*}
        S_{\mathrm{UCB},p} \coloneqq \sum_{m=0}^{l_p} \tilde{b}_{i,j}(t_{p,m}) \eta_j(t_{p,m})
        \mathbb{1}_{\hat{i}^*_j(t_{p,m})=i}
        \mathbb{1}_{\hat{N}_{i,j}(t_{p,m}) < U_N}.
    \end{align*}
    For part $p$ such that $\sum_{m=0}^{l_p} \eta_j(t_{p,m}) \mathbb{1}_{\hat{i}^*_j(t_{p,m})=i} = 0$, we have
    \begin{align*}
        S_{\mathrm{UCB},p} = 0.
    \end{align*}
    Consider part $p$ such that $\sum_{m=0}^{l_p} \eta_j(t_{p,m}) \mathbb{1}_{\hat{i}^*_j(t_{p,m})=i} > 0$. Then
    \begin{align}\label{equ:eijp-bound}
        S_{\mathrm{UCB},p} = \sum_{n=1}^{N_p} \tilde{b}_{i,j}(\tau_{p,n}) \mathbb{1}_{\hat{N}_{i,j}(\tau_{p,n}) < U_N},
    \end{align}
    where $N_p\coloneqq \sum_{m=0}^{l_p} \eta_j(t_{p,m}) \mathbb{1}_{\hat{i}^*_j(t_{p,m})=i} > 0$ and $\tau_{p,n}$ is the time slot for the $n^{\mathrm{th}}$ time such that $\eta_j(t_{p,m}) \mathbb{1}_{\hat{i}^*_j(t_{p,m})=i} = 1$ counting from $m=0$. For $n=1$, we have
    \begin{align}\label{equ:bound-bij-1}
        \hat{N}_{i,j}(\tau_{p,1}) \ge 0 \qquad \tilde{b}_{i,j}(\tau_{p,1}) \le 1.
    \end{align}
    Consider the contribution of the completion of the job starting at $\tau_{p,1}$ to $\hat{N}_{i,j}(\tau_{p,2})$. From the update rule~\eqref{equ:alg:update} in the algorithm, we know that the discounting process starts from the service starting time of the job.
    Since $\tau_{p,2}-\tau_{p,1}\le \left\lceil \frac{1}{2(1-\gamma)} \right\rceil$ and $\frac{1}{2(1-\gamma)}\ge 1$ (by $\gamma\ge \frac{1}{2}$), we have
    \begin{align*}
        \hat{N}_{i,j}(\tau_{p,2}) \ge  \gamma^{\tau_{p,2}-\tau_{p,1}-1}\ge \gamma^{\frac{1}{2(1-\gamma)}} = \left(1-\frac{1}{\frac{1}{1-\gamma}}\right)^{\frac{1}{2(1-\gamma)}} \ge \frac{1}{2},
    \end{align*}
    where the last inequality follows from the fact that $(1-x)^y\ge 1-xy$ for any $x\in[0,1]$ and $y\ge 1$. This can be easily verified by taking the first and second derivatives of $f(x)=(1-x)^y-1+xy$ with respect to $x$. For a general $n$, we have
    \begin{align*}
        \hat{N}_{i,j}(\tau_{p,n+1})\ge \sum_{s=1}^{n} \gamma^{\tau_{p,n+1}-\tau_{p,s}-1}.
    \end{align*}
    Since $\tau_{p,n+1}-\tau_{p,s} \le  \left\lceil\frac{1}{2(1-\gamma)}\right\rceil$ for any $s,$ we have
    \begin{align*}
        \gamma^{\tau_{p,n+1}-\tau_{p,s}-1}\geq \frac{1}{2},
    \end{align*}
    and thus
    \begin{align}\label{equ:bound-Nij-n}
        \hat{N}_{i,j}(\tau_{p,n+1})\ge \sum_{s=1}^{n} \gamma^{\tau_{p,n+1}-\tau_{p,s}-1} \ge \frac{n}{2}.
    \end{align}
    From \eqref{equ:bound-b-tilde} and \eqref{equ:bound-Nij-n}, we have
    \begin{align}\label{equ:bound-bij-n}
        \tilde{b}_{i,j}(\tau_{p,n+1})\le 
        4 U_{\mathrm{S}} \sqrt{\frac{\log \frac{1}{1-\gamma}}{\hat{N}_{i,j}(\tau_{p,n+1})}}
        \le 4 U_{\mathrm{S}} \sqrt{\frac{2\log \frac{1}{1-\gamma}}{n}},
    \end{align}
    for $n\ge 1$.
    Dividing the sum in~\eqref{equ:eijp-bound} into two parts and noticing the fact that $\left\lfloor2U_N\right\rfloor + 1$ could possibly be greater than $N_p$, we have
    \begin{align}\label{equ:boung-eijp-2-term}
        S_{\mathrm{UCB},p} \le & \sum_{n=1}^{\left\lfloor2U_N\right\rfloor + 1}
        \tilde{b}_{i,j}(\tau_{p,n}) \mathbb{1}_{\hat{N}_{i,j}(\tau_{p,n}) < U_N}
        + \sum_{n=\left\lfloor2U_N\right\rfloor + 2}^{N_p} 
        \tilde{b}_{i,j}(\tau_{p,n}) \mathbb{1}_{\hat{N}_{i,j}(\tau_{p,n}) < U_N}.
    \end{align}
    Note that from~\eqref{equ:bound-Nij-n} we have $\hat{N}_{i,j}(\tau_{p,n}) \ge \frac{n-1}{2}$. In the second term of the right-hand side of~\eqref{equ:boung-eijp-2-term}, we have $n\ge \left\lfloor2U_N\right\rfloor + 2$. Hence, $\hat{N}_{i,j}(\tau_{p,n}) \ge \frac{n-1}{2} \ge (\left\lfloor2U_N\right\rfloor + 1) / 2 \ge U_N$. Therefore, the indicator function in the second term of the right-hand side of~\eqref{equ:boung-eijp-2-term} is equal to $0$. Then
    \begin{align*}
        S_{\mathrm{UCB},p} \le & \sum_{n=1}^{\left\lfloor2U_N\right\rfloor + 1}
        \tilde{b}_{i,j}(\tau_{p,n}) \mathbb{1}_{\hat{N}_{i,j}(\tau_{p,n}) < U_N}\nonumber\\
        \le & 1 + \sum_{n=1}^{\left\lfloor2U_N\right\rfloor} 4 U_{\mathrm{S}} \sqrt{\frac{2\log \frac{1}{1-\gamma}}{n}}\nonumber\\
        \le & 1 + 16 U_{\mathrm{S}} \sqrt{U_N \log \frac{1}{1-\gamma}},
    \end{align*}
    where the second inequality uses~\eqref{equ:bound-bij-1} and~\eqref{equ:bound-bij-n}, and the last inequality is by integration. Hence, by~\eqref{equ:eij-bound} and the definition of $S_{\mathrm{UCB},p}$, we have
    \begin{align*}
        S_{\mathrm{UCB}} \le \sum_{p=0}^{P-1} S_{\mathrm{UCB},p} \le P \left(1 + 16 U_{\mathrm{S}} \sqrt{U_N \log \frac{1}{1-\gamma}}\right)
        \le \left\lceil 2L (1-\gamma) \right\rceil \left( 1 + 16 U_{\mathrm{S}} \sqrt{U_N \log \frac{1}{1-\gamma}} \right),
    \end{align*}
    where the last inequality is due to the fact that there are at most $\left\lceil 2L (1-\gamma) \right\rceil$ parts.
 
\endproof

\subsection{Proof of Lemma~\ref{lemma:bound-on-sum-queue}}
\label{app:proof-lemma-bound-on-sum-queue}

\proof
By Lemma~\ref{lemma:bounds-on-diff-queue-lens}, we have
\begin{align}\label{equ:lower-bound-Q_i_t_k_plus_tau}
    \sum_i Q_i(t_k+\tau) \ge \sum_i Q_i(t_k) - J\tau \ge \sum_i Q_i(t_k) - 2JT,
\end{align}
where the last inequality holds since $\tau\le D_k+D_{k+1}-1 \le 2T$ by Lemma~\ref{lemma:prop-D}. Based on~\eqref{equ:lower-bound-Q_i_t_k_plus_tau}, we have
\begin{align*}
    \sum_i  q_i \le &  \frac{1}{D_{k+1}}\sum_{\tau=D_k}^{D_k+D_{k+1}-1} \hat{E}_{t_k}\left[ \sum_i Q_i(t_k+\tau) + 2JT \right].
\end{align*}
 
\endproof

\subsection{Proof of Lemma~\ref{lemma:prop-G}}
\label{app:proof-lemma-prop-G}

\proof
    Recall the definition of $G_t$:
    \begin{align*}
        G_t=& \min_{n} \sum_{l=0}^n w(\tau'_l(t))\\
        & \mbox{s.t. } \sum_{l=0}^{n} w(\tau'_l(t)) \ge \frac{c_2T}{\delta}.
    \end{align*}
    Let $n_{G}^*(t)$ denote the optimal solution to the above optimization problem. Note that
    $G_t=\sum_{l=0}^{n_{G}^*(t)} w(\tau'_l(t)) \ge \frac{c_2T}{\delta}$
    and
    $\sum_{l=0}^{n_{G}^*(t)-1} w(\tau'_l(t)) < \frac{c_2T}{\delta}$.
    Hence, we have
    \begin{align*}
        G_t=\sum_{l=0}^{n_{G}^*(t)}w(\tau'_l(t))=\sum_{l=0}^{n_{G}^*(t)-1}w(\tau'_l(t)) + w(\tau'_{n_{G}^*(t)}(t))\le \frac{c_2T}{\delta} + W,
    \end{align*}
    where the last inequality is due to the bound $w(\tau)\le W$ for any $\tau$. Therefore, for any $t$, we have
    \begin{align*}
        \frac{c_2T}{\delta} \le G_t \le \frac{c_2T}{\delta} +W \le \frac{(c_2+\frac{1}{2})T}{\delta},
    \end{align*}
    where the last inequality is by $W\le \frac{T}{2} \le \frac{T}{2\delta}$.
 
\endproof

\subsection{Proof of Lemma~\ref{lemma:concentration-theo-2}}
\label{app:proof-lemma-concentration-theo-2}

\proof
    Recall that
    \[
    b_{i,j}(t+\tau) = 2\sqrt{\frac{U_{\mathrm{S}}^2\log \left(\sum_{\tau'=0}^{t+\tau - 1} \gamma^{\tau'}\right)}{\hat{N}_{i,j}(t+\tau)}} = 2\sqrt{\frac{U_{\mathrm{S}}^2\log \left(\frac{1-\gamma^{t+\tau}}{1-\gamma}\right)}{\hat{N}_{i,j}(t+\tau)}}.
    \]
    Consider the following event ${\mathcal E}_{t,i,j,\tau}$:
    \begin{align*}
        {\mathcal E}_{t,i,j,\tau}\coloneqq \left\{ \left|\frac{1}{\hat{\mu}_{i,j}(t+\tau)}-\frac{1}{\mu_{i,j}(t+\tau)}\right|\le 2 \sqrt{\frac{U_{\mathrm{S}}^2\log \left(\frac{1-\gamma^{t+\tau}}{1-\gamma}\right)}{\hat{N}_{i,j}(t+\tau)}} \right\}.
    \end{align*}
    Let $\epsilon'_m \coloneqq \mathbb{1}_{\hat{i}^*_j(t+m-1)=i} \eta_j(t+m-1)$.
    Let $M'_3$ be a random variable such that $t+M'_3=f_j(t+\tau)$. Following the same proof as the proof of Eq.~\eqref{equ:conclusion-reused-in-theo-2} in the proof of Lemma~\ref{lemma:concentration} (Section~\ref{app:proof-lemma-concentration}), we can obtain that if $\tau \ge \frac{3T}{8}$,
    \begin{align}
         & \hat{P}_{t} \left( {\mathcal E}^{  {\mathrm c}}_{t,i,j,\tau} \right)\nonumber\\
         \le & \hat{P}_{t} \Biggl(  \left|\sum_{m=1}^{M'_3} \gamma^{\tau-m} \epsilon'_m \left(S_{i,j}(t+m-1) - E[S_{i,j}(t+m-1)]\right)\right| \Biggr. \nonumber\\
         & ~~ + \sum_{m=1}^{M'_3} \gamma^{\tau-m} \epsilon'_m \left| E[S_{i,j}(t+m-1)] - E[S_{i,j}(t+\tau)]\right| > 2\sqrt{\hat{N}_{i,j}(t+\tau) U_{\mathrm{S}}^2\log \frac{\gamma}{1-\gamma}} - U_{\mathrm{S}},\nonumber\\
         & \quad \quad \Biggl.\hat{N}_{i,j}(t+\tau)>4\log \frac{\gamma}{1-\gamma}\Biggr).
    \end{align}
    We note that in the stationary setting, $E[S_{i,j}(t)]$
    does not depend on $t$ and so the term 
    \[\sum_{m=1}^{M'_3} \gamma^{\tau-m} \epsilon'_m \left| E[S_{i,j}(t+m-1)] - E[S_{i,j}(t+\tau)]\right|\]
    is zero. However, in the nonstationary setting, we need some assumptions on the variability of mean service times. Note that 
    \begin{align}\label{equ:bound-on-m3-tau}
        1\le m\le M'_3\le \tau \le \frac{T}{2} + G_t - 1 \le \frac{(c_2+1) T}{\delta},
    \end{align}
    where the last inequality is by Lemma~\ref{lemma:prop-G}.
    Hence, different from Lemma~\ref{lemma:concentration}, we need a stronger assumption here because $\tau$ and $M'_3$ could be as large as $\frac{(c_2+1) T}{\delta}$.
    From~\eqref{equ:bound-on-m3-tau}, we know $(t + \tau) - (t + m - 1)\le \frac{(c_2+1) T}{\delta}$. By Assumption~\ref{assump:mu-2} on the time-varying service times, We have
    \begin{align*}
        \left\lvert E[S_{i,j}(t+m-1)]-E[S_{i,j}(t+\tau)]\right\rvert \le \frac{\delta}{(c_2+1)T}\left(\frac{1}{\gamma}\right)^{\tau - m}.
    \end{align*}
    Hence, we have
    \begin{align}\label{equ:theo-2-conc-prob-2}
        \hat{P}_{t} \left( {\mathcal E}^{  {\mathrm c}}_{t,i,j,\tau} \right)
         \le & \hat{P}_{t} \Biggl(  \left|\sum_{m=1}^{M'_3} \gamma^{\tau-m} \epsilon'_m \left(S_{i,j}(t+m-1) - E[S_{i,j}(t+m-1)]\right)\right| 
         \Biggr.\nonumber\\
         & \quad \quad \Biggl.> 2\sqrt{\hat{N}_{i,j}(t+\tau) U_{\mathrm{S}}^2\log \frac{\gamma}{1-\gamma}} - U_{\mathrm{S}} - \frac{\delta M'_3}{(c_2+1)T}, \hat{N}_{i,j}(t+\tau)>4\log \frac{\gamma}{1-\gamma}\Biggr).
    \end{align}
    By~\eqref{equ:bound-on-m3-tau}, we have
    \begin{align}\label{equ:theo-2-conc-prob-aux}
        U_{\mathrm{S}} + \frac{\delta M'_3}{(c_2+1)T} \le U_{\mathrm{S}} + 1 \le \sqrt{4 U_{\mathrm{S}}^2 \left(\log \frac{\gamma}{1-\gamma}\right)^2}\le \sqrt{U_{\mathrm{S}}^2\hat{N}_{i,j}(t+\tau)\log \frac{\gamma}{1-\gamma}},
    \end{align}
    where the second inequality holds since $\gamma \ge 1-\frac{1}{1+e^{1.5}}$ and $U_{\mathrm{S}}\ge 1$, and the last inequality holds when $\hat{N}_{i,j}(t+\tau)>4\log \frac{\gamma}{1-\gamma}$. Based on~\eqref{equ:theo-2-conc-prob-aux}, we can continue to bound~\eqref{equ:theo-2-conc-prob-2} and obtain
    \begin{align}\label{equ:theo-2-conc-prob-3}
         \hat{P}_{t} \left( {\mathcal E}^{  {\mathrm c}}_{t,i,j,\tau} \right)
         \le  \hat{P}_{t} \Biggl(  \left|\sum_{m=1}^{M'_3} \gamma^{\tau-m} \epsilon'_m \left(S_{i,j}(t+m-1) - E[S_{i,j}(t+m-1)]\right)\right| 
         > \sqrt{\hat{N}_{i,j}(t+\tau) U_{\mathrm{S}}^2\log \frac{\gamma}{1-\gamma}}\Biggr).
    \end{align}
    Following the same proof as the proof of Eq.~\eqref{equ:lemma-N-2} in the proof of Lemma~\ref{lemma:concentration} (Section~\ref{app:proof-lemma-concentration})
    we can obtain that
    $
        \hat{N}_{i,j}(t+\tau) \ge \sum_{m=1}^{M'_3} \gamma^{\tau-m} \mathbb{1}_{\hat{i}^*_j(t+m-1)=i}~ \eta_j(t+m-1) = \sum_{m=1}^{M'_3} \gamma^{\tau-m} \epsilon'_m.
    $
    Hence, we can further bound~\eqref{equ:theo-2-conc-prob-3} as
    \begin{align*}
        \hat{P}_{t} \left( {\mathcal E}^{  {\mathrm c}}_{t,i,j,\tau} \right)
         \le & \hat{P}_{t} \Biggl(  \frac{\left|\sum_{m=1}^{M'_3} \gamma^{\tau-m} \epsilon'_m \left(S_{i,j}(t+m-1) - E[S_{i,j}(t+m-1)]\right)\right|}{\sqrt{\sum_{m=1}^{M'_3} \gamma^{\tau-m} \epsilon'_m}}
         > \sqrt{U_{\mathrm{S}}^2\log \frac{\gamma}{1-\gamma}}\Biggr).
    \end{align*}
    Since $M'_3\le \tau$ and $\gamma < 1$, we have $\sqrt{\gamma^{M'_3-\tau}} \ge 1$. Hence, we have
    \begin{align}\label{equ:theo-2-conc-prob-4}
        \hat{P}_{t} \left( {\mathcal E}^{  {\mathrm c}}_{t,i,j,\tau} \right)
        \le & \hat{P}_{t} \Biggl(  \frac{ \gamma^{M'_3-\tau} \left|\sum_{m=1}^{M'_3} \gamma^{\tau-m} \epsilon'_m \left(S_{i,j}(t+m-1) - E[S_{i,j}(t+m-1)]\right)\right|}{ \sqrt{\gamma^{M'_3-\tau}} \sqrt{\sum_{m=1}^{M'_3} \gamma^{\tau-m} \epsilon'_m}}
         > \sqrt{ U_{\mathrm{S}}^2\log \frac{\gamma}{1-\gamma}}\Biggr)\nonumber\\
         = & \hat{P}_{t} \Biggl(  \frac{\left|\sum_{m=1}^{M'_3} \gamma^{M'_3-m} \epsilon'_m \left(S_{i,j}(t+m-1) - E[S_{i,j}(t+m-1)]\right)\right|}{\sqrt{\sum_{m=1}^{M'_3} \gamma^{M'_3-m} \epsilon'_m}}
         > \sqrt{ U_{\mathrm{S}}^2\log \frac{\gamma}{1-\gamma}}\Biggr)\nonumber\\
         \le & \hat{P}_{t} \Biggl(  \frac{\left|\sum_{m=1}^{M'_3} \gamma^{M'_3-m} \epsilon'_m \left(S_{i,j}(t+m-1) - E[S_{i,j}(t+m-1)]\right)\right|}{\sqrt{\sum_{m=1}^{M'_3} \gamma^{2(M'_3-m)} \epsilon'_m}}
         > \sqrt{ U_{\mathrm{S}}^2\log \frac{\gamma}{1-\gamma}}\Biggr),
    \end{align}
    where we added ``2'' in the last inequality because we want to use Theorem~\ref{theo:a-restate} (the Hoeffding-type inequality for self-normalized means~\cite[Theorem 22]{GarMou_08}) later in the proof.

    Consider the event
    $
        {\mathcal E}'_{t,i,j}\coloneqq \bigcap_{\tau=\frac{3T}{8}}^{\frac{T}{2}+G_t-1} {\mathcal E}_{t,i,j,\tau}.
    $
    Then from the result~\eqref{equ:theo-2-conc-prob-4}, we have
    \begin{align*}
        &\hat{P}_{t} \left( {\mathcal E}^{'\mathrm{c}}_{t,i,j} \right)\nonumber\\
        \le &  \hat{P}_{t} \Biggl( \mbox{there exists } \tau\in \left[\frac{3T}{8}, \frac{T}{2}+G_t- 1\right],\Biggr.\nonumber\\
        &\quad \quad \Biggl.\frac{\left|\sum_{m=1}^{M'_3} \gamma^{M'_3-m} \epsilon'_m \left(S_{i,j}(t+m-1) - E[S_{i,j}(t+m-1)]\right)\right|}{\sqrt{\sum_{m=1}^{M'_3} \gamma^{2(M'_3-m)} \epsilon'_m}}
         > \sqrt{ U_{\mathrm{S}}^2\log \frac{\gamma}{1-\gamma}}\Biggr).
    \end{align*}
    Recall that $M'_3$ is defined by $t+M'_3=f_j(t+\tau)$. Hence, $M'_3\ge \tau - U_{\mathrm{S}}\ge \frac{3T}{8} - \frac{T}{8} = \frac{T}{4}$. Also, $M'_3\le \frac{(c_2+1)T}{\delta}$ by~\eqref{equ:bound-on-m3-tau}. Hence,
    \begin{align}\label{equ:theo-2-conc-prob-5}
        &\hat{P}_{t} \left( {\mathcal E}^{'\mathrm{c}}_{t,i,j} \right) \nonumber\\
        \le & \hat{P}_{t} \Biggl( \mbox{there exists } M\in \left[\frac{T}{4}, \frac{(c_2+1)T}{\delta}\right],\Biggr.\Biggl.\frac{\left|\sum_{m=1}^{M} \gamma^{M-m} \epsilon'_m \left(S_{i,j}(t+m-1) - E[S_{i,j}(t+m-1)]\right)\right|}{\sqrt{\sum_{m=1}^{M} \gamma^{2(M-m)} \epsilon'_m}}\nonumber\\
         & \quad > \sqrt{ U_{\mathrm{S}}^2\log \frac{\gamma}{1-\gamma}}\Biggr)\nonumber\\
         \le & \hat{P}_{t} \Biggl( \sup_{1\le M\le \frac{(c_2+1)T}{\delta}}\frac{\left|\sum_{m=1}^{M} \gamma^{M-m} \epsilon'_m \left(S_{i,j}(t+m-1) - E[S_{i,j}(t+m-1)]\right)\right|}{\sqrt{\sum_{m=1}^{M} \gamma^{2(M-m)} \epsilon'_m}}
         > \sqrt{U_{\mathrm{S}}^2\log \frac{\gamma}{1-\gamma}}\Biggr).
    \end{align}
    
    Let us view the conditional probability $\hat{P}_{t}$ as a new probability measure. Then $\hat{E}_{t}$ is the expectation under this measure.
    Note that $\left(S_{i,j}(t+m-1)\right)_{m=1}^{\infty}$ is a sequence of independent bounded random variables under this new measure since they are independent of $\boldsymbol{Q}(t)$ and $\boldsymbol{H}(t)$, which also implies that
    \begin{align}\label{equ:theo-2-expt-expt-hat}
        E[S_{i,j}(t+m-1)] = \hat{E}_{t}[S_{i,j}(t+m-1)].
    \end{align}
    Define ${\mathcal F}'_m$ as follows:
    \begin{align*}
        {\mathcal F}'_m \coloneqq \sigma \left( (\boldsymbol{S}(t+n-1))_{n=1}^m, (\boldsymbol{A}(t+n-1))_{n=1}^{m+1}, (\boldsymbol{Q}(t+n-1))_{n=1}^{m+1}, (\boldsymbol{H}(t+n-1))_{n=1}^{m+1}\right),
    \end{align*}
    where $\sigma(\cdot)$ denotes the $\sigma$-algebra generated by the random variables.
    Note that 
    \begin{align*}
        \sigma(S_{i,j}(t),...,S_{i,j}(t+m-1))\subset {\mathcal F}'_m
    \end{align*}
    and for any $n>m$, $S_{i,j}(t+n-1)$ is independent of ${\mathcal F}'_m$.
    Recall that $\epsilon'_m \coloneqq \mathbb{1}_{\hat{i}^*_j(t+m-1)=i} \eta_j(t+m-1)$. Since the scheduling decision at time $t+m-1$ is determined by $\boldsymbol{Q}(t+m-1)$ and $\boldsymbol{H}(t+m-1)$, $\mathbb{1}_{\hat{i}^*_j(t+m-1)=i}$ is ${\mathcal F}_{m-1}$-measurable. Since $\eta_j(t+m-1)$ is determined by $\boldsymbol{A}(t+m-1)$, $\boldsymbol{Q}(t+m-1)$, and $\boldsymbol{H}(t+m-1)$, $\eta_j(t+m-1)$ is also ${\mathcal F}_{m-1}$-measurable. Therefore, $\epsilon'_m$ is ${\mathcal F}_{m-1}$-measurable, i.e., $\left(\epsilon'_m\right)_{m=1}^{\infty}$ is a previsible (or predictable) sequence of Bernoulli random variables.
    Therefore,
    Applying Theorem~\ref{theo:a-restate} with $\hat{P}=\hat{P}_{t}$, ${\cal F}_m={\cal F}'_m$, $X_m=S_{i,j}(t+m-1)$, $\epsilon_m=\epsilon'_m$, $\beta=\sqrt{U_{\mathrm{S}}^2\log \frac{\gamma}{1-\gamma}}$, $B=U_{\mathrm{S}}$, and $N=\lfloor\frac{(c_2+1)T}{\delta}\rfloor$, we have
    \begin{align}\label{equ:theo-2-bound-single-side-1}
        & \hat{P}_{t} \Biggl( \sup_{1\le M \le \lfloor\frac{(c_2+1)T}{\delta}\rfloor} \frac{\sum_{m=1}^{M} \gamma^{M-m} \epsilon'_m \left(S_{i,j}(t+m-1) - \hat{E}_{t}[S_{i,j}(t+m-1)]\right)}{\sqrt{\sum_{m=1}^{M} \gamma^{2(M-m)} \epsilon'_m}}
         > \sqrt{U_{\mathrm{S}}^2\log \frac{\gamma}{1-\gamma}}\Biggr)\nonumber\\
         \le & \left(\frac{\log \left(\gamma^{-2\left\lfloor\frac{(c_2+1)T}{\delta}\right\rfloor}\sum_{m=1}^{\left\lfloor\frac{(c_2+1)T}{\delta}\right\rfloor} (\gamma^2)^{\left\lfloor\frac{(c_2+1)T}{\delta}\right\rfloor-m} \right)}{\log (1+\zeta)}+1\right)\exp\left(-2\left(1-\frac{\zeta^2}{16}\right)\log \frac{\gamma}{1-\gamma}\right)
    \end{align}
    for all $\zeta>0$.
    Note that
    \begin{align}\label{equ:theo-2-bound-single-side-part}
        \log \left(\gamma^{-2\left\lfloor\frac{(c_2+1)T}{\delta}\right\rfloor}\sum_{m=1}^{\left\lfloor\frac{(c_2+1)T}{\delta}\right\rfloor} (\gamma^2)^{\left\lfloor\frac{(c_2+1)T}{\delta}\right\rfloor-m} \right)
        \le & \log \left( \gamma^{-2\left\lfloor\frac{(c_2+1)T}{\delta}\right\rfloor} \frac{1}{1-\gamma^2}\right) \nonumber\\
        \le & \frac{2(c_2+1)T}{\delta} \log \left( \frac{1}{\gamma} \right) + \log \left(\frac{1}{1-\gamma^2}\right)\nonumber\\
        \le & \frac{2(c_2+1)T}{\delta}\left(\frac{1-\gamma}{\gamma}\right) + \log \left(\frac{1}{1-\gamma^2}\right)\nonumber\\
        \le & \frac{2(c_2+1)T}{\delta} \left(\frac{1-\gamma}{\gamma}\right) + \log \left(\frac{1}{1-\gamma}\right)\nonumber\\
        = & \left(1+\frac{8(c_2+1)}{\gamma \delta}\right) \log \frac{1}{1-\gamma},
    \end{align}
    where the third inequality is due to the fact that $\log x \le x-1$ for any $x>0$, and the last equality is by the definition of $T$.
    Substituting~\eqref{equ:theo-2-bound-single-side-part} into~\eqref{equ:theo-2-bound-single-side-1}, we have
    \begin{align}\label{equ:theo-2-overestimation}
        & \hat{P}_{t} \Biggl( \sup_{1\le M \le \lfloor\frac{(c_2+1)T}{\delta}\rfloor} \frac{\sum_{m=1}^{M} \gamma^{M-m} \epsilon'_m \left(S_{i,j}(t+m-1) - \hat{E}_{t}[S_{i,j}(t+m-1)]\right)}{\sqrt{\sum_{m=1}^{M} \gamma^{2(M-m)} \epsilon'_m}}
         > \sqrt{U_{\mathrm{S}}^2\log \frac{\gamma}{1-\gamma}}\Biggr)\nonumber\\
         \le & \left(\frac{\left(1+\frac{8(c_2+1)}{\gamma \delta}\right) \log \frac{1}{1-\gamma}}{\log (1+\zeta)}+1\right)\exp\left(-2\left(1-\frac{\zeta^2}{16}\right)\log \frac{\gamma}{1-\gamma}\right)
    \end{align}
    for all $\zeta>0$. Recall that we showed in Section~\ref{app:proof-lemma-concentration} that the bound in Theorem~\ref{theo:a-restate} also holds for underestimation. Hence, we have
    \begin{align}\label{equ:theo-2-underestimation}
        & \hat{P}_{t} \Biggl( \sup_{1\le M \le \lfloor\frac{(c_2+1)T}{\delta}\rfloor} \frac{\sum_{m=1}^{M} \gamma^{M-m} \epsilon'_m \left(\hat{E}_{t}[S_{i,j}(t+m-1)] - S_{i,j}(t+m-1) \right)}{\sqrt{\sum_{m=1}^{M} \gamma^{2(M-m)} \epsilon'_m}}
         > \sqrt{U_{\mathrm{S}}^2\log \frac{\gamma}{1-\gamma}}\Biggr)\nonumber\\
         \le & \left(\frac{\left(1+\frac{8(c_2+1)}{\gamma \delta}\right) \log \frac{1}{1-\gamma}}{\log (1+\zeta)}+1\right)\exp\left(-2\left(1-\frac{\zeta^2}{16}\right)\log \frac{\gamma}{1-\gamma}\right)
    \end{align}
    for all $\zeta>0$. Taking the union bound over underestimation~\eqref{equ:theo-2-underestimation} and overestimation~\eqref{equ:theo-2-overestimation}, we have
    \begin{align*}
        & \hat{P}_{t} \Biggl( \sup_{1\le M\le \lfloor\frac{(c_2+1)T}{\delta}\rfloor} \frac{\left|\sum_{m=1}^{M} \gamma^{M-m} \epsilon'_m \left(S_{i,j}(t+m-1) - \hat{E}_{t}[S_{i,j}(t+m-1)]\right)\right|}{\sqrt{\sum_{m=1}^{M} \gamma^{2(M-m)} \epsilon'_m}}
         > \sqrt {U_{\mathrm{S}}^2\log \frac{\gamma}{1-\gamma}}\Biggr)\nonumber\\
        \le & 2\left(\frac{\left(1+\frac{8(c_2+1)}{\gamma \delta}\right) \log \frac{1}{1-\gamma}}{\log (1+\zeta)}+1\right)\exp\left(-2\left(1-\frac{\zeta^2}{16}\right)\log \frac{\gamma}{1-\gamma}\right)
    \end{align*}
    for all $\zeta>0$.
    Setting $\zeta=0.3$, we have
    \begin{align*}
        & \hat{P}_{t} \Biggl( \sup_{1\le M\le \lfloor\frac{(c_2+1)T}{\delta}\rfloor} \frac{\left|\sum_{m=1}^{M} \gamma^{M-m} \epsilon'_m \left(S_{i,j}(t+m-1) - \hat{E}_{t}[S_{i,j}(t+m-1)]\right)\right|}{\sqrt{\sum_{m=1}^{M} \gamma^{2(M-m)} \epsilon'_m}}
         > 
         \sqrt {U_{\mathrm{S}}^2\log \frac{\gamma}{1-\gamma}}\Biggr)\nonumber\\
        \le & 
        \left[\left(8+\frac{61(c_2+1)}{\gamma \delta}\right) \log \frac{1}{1-\gamma}+2\right]
        \left(\frac{1-\gamma}{\gamma}\right)^{1.99}
    \end{align*}
    Since $\gamma \ge 1-\frac{1}{1+e^{1.5}}$, $(1-\gamma)^{0.49} \log \frac{1}{1-\gamma} \le \frac{1}{0.49e}$, and $c_2= 5(IU_{\mathrm{A}}+J)\ge 5$, we have
    \begin{align}\label{equ:theo-2-conc-prob-6}
        & \hat{P}_{t} \Biggl( \sup_{1\le M\le \lfloor\frac{(c_2+1)T}{\delta}\rfloor} \frac{\left|\sum_{m=1}^{M} \gamma^{M-m} \epsilon'_m \left(S_{i,j}(t+m-1) - \hat{E}_{t}[S_{i,j}(t+m-1)]\right)\right|}{\sqrt{\sum_{m=1}^{M} \gamma^{2(M-m)} \epsilon'_m}}
         > 
         \sqrt {U_{\mathrm{S}}^2\log \frac{\gamma}{1-\gamma}}\Biggr)\nonumber\\
        \le & \frac{516(IU_{\mathrm{A}}+J)\left(1-\gamma\right)^{1.5}}{\delta} .
    \end{align}
    
    Combining~\eqref{equ:theo-2-conc-prob-5},~\eqref{equ:theo-2-expt-expt-hat},~\eqref{equ:theo-2-conc-prob-6} and noticing that $M$ is an integer, we have
    \begin{align*}
        \hat{P}_{t} \left( {\mathcal E}^{'{\mathrm c}}_{t,i,j} \right) \le \frac{516(IU_{\mathrm{A}}+J)\left(1-\gamma\right)^{1.5}}{\delta}.
    \end{align*}
    Taking the union bound over $i$, we have
    \begin{align*}
        \hat{P}_{t} ({\mathcal E}_{t,j}^{'{\mathrm c}}) \le \sum_{i=1}^{I} \hat{P}_{t} \left( {\mathcal E}^{'{\mathrm c}}_{t,i,j} \right) \le \frac{516 I(I U_{\mathrm{A}} + J) \left(1-\gamma\right)^{1.5}}{\delta}.
    \end{align*}
 
\endproof

\subsection{Proof of Lemma~\ref{lemma:taylor}}
\label{app:proof-lemma-taylor}

\proof
    The proof can be found in~\cite[Proof of Theorem 3.6]{ChuLu_06}. We present the proof in the following for completeness.
    The proof is based on Taylor series of $e^y$.
    Note that
    \begin{align*}
        e^y=\sum_{k=0}^{\infty} \frac{y^k}{k!},
    \end{align*}
    which converges for all $y$. Hence, we have
    \begin{align*}
        \frac{2(e^y-1-y)}{y^2} = 2 \sum_{k=2}^{\infty} \frac{y^{k-2}}{k!} \le \sum_{k=2}^{\infty} \frac{y^{k-2}}{3^{k-2}} = \frac{1}{1-y/3}, 
    \end{align*}
    where the inequality is by the fact that $k!\ge 2\times 3^{k-2}$, and the last equality holds when $|y|\le 3$.
    Then we have
    \begin{align*}
        e^y \le 1 + y + \frac{y^2}{2(1-y/3)} \le 1 + y + \frac{y^2}{2(1-|y|/3)}.
    \end{align*}
 
\endproof

\subsection{Proof of Lemma~\ref{lemma:concentration-3}}
\label{app:proof-lemma-concentration-3}

\proof
    Recall that when $\gamma=1$,
    \[
    b_{i,j}(\tau) = 2\sqrt{\frac{U_{\mathrm{S}}^2\log \tau }{\hat{N}_{i,j}(\tau)}}.
    \]
    Consider the event
    \begin{align}\label{equ:def-conc-single-event-station}
        {\cal E}_{\mathrm{S},\tau,i,j} \coloneqq \left\{
        \left| \frac{1}{\hat{\mu}_{i,j}(\tau)} - \frac{1}{\mu_{i,j}} \right| \le 2\sqrt{\frac{U_{\mathrm{S}}^2\log \tau }{\hat{N}_{i,j}(\tau)}} \right\}.
    \end{align}
    We have
    \begin{align}\label{equ:log-term-station}
        \Pr \left( {\cal E}^{ {\mathrm c}}_{\mathrm{S},\tau,i,j} \right)
        = & \Pr \left( \left| \frac{1}{\hat{\mu}_{i,j}(\tau)} - \frac{1}{\mu_{i,j}} \right| > 2\sqrt{\frac{U_{\mathrm{S}}^2\log \tau }{\hat{N}_{i,j}(\tau)}} \right)\nonumber\\
        = & \Pr \Biggl(  \left|\hat{\phi}_{i,j}(\tau)-\frac{\hat{N}_{i,j}(\tau)}{\mu_{i,j}}\right|> 2\sqrt{\hat{N}_{i,j}(\tau) U_{\mathrm{S}}^2\log \tau}\Biggr).
    \end{align}
    Recall the update rule of $\hat{\phi}_{i,j}(\tau)$ and $\hat{N}_{i,j}(\tau)$ in \eqref{equ:alg:update}. Since $\gamma=1$, we have
    \begin{align}\label{equ:sum-job-completion}
        \hat{\phi}_{i,j}(\tau) = & \sum_{\tau'=0}^{\tau - 1} \mathbb{1}_{i,j}(\tau')\eta_j(\tau')[M_{i,j}(\tau')+1]\nonumber\\
        = & \sum_{\tau'=0}^{\tau - 1} \mathbb{1}_{i,j}(\tau')\eta_j(\tau')S_{i,j}(f_j(\tau')),
    \end{align}
    where $f_j(\tau')$ is the service starting time of the job that is being served at server $j$ in time slot $\tau'$. Note that the summation in \eqref{equ:sum-job-completion} only includes the time slots when there is job completion of queue $i$ at server $j$. This can be transformed into summing over the time slots when server $j$ is available and picks queue $i$, i.e.,
    \begin{align}\label{equ:sum-phi-start}
        \hat{\phi}_{i,j}(\tau) = \sum_{\tau'=0}^{f_j(\tau)-1} \mathbb{1}_{\hat{i}^*_j(\tau')=i}\eta_j(\tau')S_{i,j}(\tau'),
    \end{align}
    where we note that $f_j(\tau)$ is a random variable.
    Similarly, we have
    \begin{align}\label{equ:sum-N-start}
        \hat{N}_{i,j}(\tau) =  \sum_{\tau'=0}^{\tau - 1} \mathbb{1}_{i,j}(\tau')\eta_j(\tau')
        =  \sum_{\tau'=0}^{f_j(\tau)-1} \mathbb{1}_{\hat{i}^*_j(\tau')=i}\eta_j(\tau').
    \end{align}
    Let $\epsilon_{\mathrm{S},\tau'}\coloneqq \mathbb{1}_{\hat{i}^*_j(\tau'-1)=i}\eta_j(\tau'-1)$.
    Substituting \eqref{equ:sum-phi-start} and \eqref{equ:sum-N-start} into \eqref{equ:log-term-station}, we have
    \begin{align*}
        \Pr \left( {\cal E}^{ {\mathrm c}}_{\mathrm{S},\tau,i,j} \right)
        = & \Pr \left(  \left| \sum_{\tau'=0}^{f_j(\tau)-1} \epsilon_{\mathrm{S},\tau'+1}S_{i,j}(\tau')
        -
        \frac{\sum_{\tau'=0}^{f_j(\tau)-1} \epsilon_{\mathrm{S},\tau'+1}}{\mu_{i,j}}\right|> 2\sqrt{\left(\sum_{\tau'=0}^{f_j(\tau)-1} \epsilon_{\mathrm{S},\tau'+1} \right) U_{\mathrm{S}}^2\log \tau}\right)\nonumber\\
        = & \Pr \left(  \left| \sum_{\tau'=0}^{f_j(\tau)-1} \epsilon_{\mathrm{S},\tau'+1}
        \bigl(S_{i,j}(\tau') - E[S_{i,j}(\tau')]\bigr)
        \right|
        > 2\sqrt{\left(\sum_{\tau'=0}^{f_j(\tau)-1} \epsilon_{\mathrm{S},\tau'+1} \right) U_{\mathrm{S}}^2\log \tau}\right)\nonumber\\
        = & \Pr \left(  \frac{\left|\sum_{\tau'=0}^{f_j(\tau)-1} \epsilon_{\mathrm{S},\tau'+1}
        \bigl(S_{i,j}(\tau') - E[S_{i,j}(\tau')]\bigr)
        \right|}{\sqrt{
        \sum_{\tau'=0}^{f_j(\tau)-1} \epsilon_{\mathrm{S},\tau'+1}
        }}
        > 2\sqrt{ U_{\mathrm{S}}^2\log \tau}\right),
    \end{align*}
    where the second equality holds since $E[S_{i,j}(\tau')]=\mu_{i,j}$, and the last equality holds since $        \sum_{\tau'=0}^{f_j(\tau)-1} \epsilon_{\mathrm{S},\tau'+1}\\\neq 0$ due to the strict inequality. Since $f_j(\tau)$ is a random variable taking values in $[1,\tau]$, we further have
    \begin{align}\label{equ:event-tau-i-j}
        \Pr \left( {\cal E}^{ {\mathrm c}}_{\mathrm{S},\tau,i,j} \right)
        \le & \Pr \left( \mbox{there exists } M\in[1, \tau],   \frac{\left|\sum_{\tau'=0}^{M-1} \epsilon_{\mathrm{S},\tau'+1}
        \bigl(S_{i,j}(\tau') - E[S_{i,j}(\tau')]\bigr)
        \right|}{\sqrt{
        \sum_{\tau'=0}^{M-1} \epsilon_{\mathrm{S},\tau'+1}
        }}
        > 2\sqrt{ U_{\mathrm{S}}^2\log \tau}\right)\nonumber\\
        = & \Pr \left( \sup_{1\le M\le \tau}  \frac{\left|\sum_{\tau'=0}^{M-1} \epsilon_{\mathrm{S},\tau'+1}
        \bigl(S_{i,j}(\tau') - E[S_{i,j}(\tau')]\bigr)
        \right|}{\sqrt{
        \sum_{\tau'=0}^{M-1} \epsilon_{\mathrm{S},\tau'+1}
        }}
        > 2\sqrt{ U_{\mathrm{S}}^2\log \tau}\right)\nonumber\\
        = & \Pr \left( \sup_{1\le M\le \tau}  \frac{\left|\sum_{m=1}^{M} \epsilon_{\mathrm{S},m}
        \bigl(S_{i,j}(m-1) - E[S_{i,j}(m-1)]\bigr)
        \right|}{\sqrt{
        \sum_{m=1}^{M} \epsilon_{\mathrm{S},m}
        }}
        > 2\sqrt{ U_{\mathrm{S}}^2\log \tau}\right).
    \end{align}
    Note that $(S_{i,j}(m-1))_{m=1}^{\infty}$ is a sequence of independent bounded random variables. Let ${\cal F}_{\mathrm{S}, m}$ defined as
    \begin{align*}
        {\cal F}_{\mathrm{S}, m} \coloneqq \sigma \left( (\boldsymbol{S}(n-1))_{n=1}^{m}, (\boldsymbol{A}(n-1))_{n=1}^{m+1}, (\boldsymbol{Q}(n-1))_{n=1}^{m+1}, (\boldsymbol{H}(n-1))_{n=1}^{m+1}\right),
    \end{align*}
    where $\sigma(\cdot)$ denotes the $\sigma$-algebra generated by the random variables.
    Note that 
    \begin{align*}
        \sigma(S_{i,j}(0),...,S_{i,j}(m-1))\subset {\cal F}_{\mathrm{S}, m}
    \end{align*}
    and for any $n>m$, $S_{i,j}(n-1)$ is independent of ${\cal F}_{\mathrm{S}, m}$.
    Recall that $\epsilon_{\mathrm{S},m} \coloneqq \mathbb{1}_{\hat{i}^*_j(m-1)=i} \eta_j(m-1)$. Since the scheduling decision at time $m-1$ is determined by $\boldsymbol{Q}(m-1)$ and $\boldsymbol{H}(m-1)$, $\mathbb{1}_{\hat{i}^*_j(m-1)=i}$ is ${\mathcal F}_{\mathrm{S}, m-1}$-measurable. Since $\eta_j(m-1)$ is determined by $\boldsymbol{A}(m-1)$, $\boldsymbol{Q}(m-1)$, and $\boldsymbol{H}(m-1)$, $\eta_j(m-1)$ is also ${\mathcal F}_{\mathrm{S},m-1}$-measurable. Therefore, $\epsilon_{\mathrm{S},m}$ is ${\mathcal F}_{\mathrm{S}, m-1}$-measurable, i.e., $\left(\epsilon_{\mathrm{S},m}\right)_{m=1}^{\infty}$ is a previsible (or predictable) sequence of Bernoulli random variables.
    Therefore,
    Applying Theorem~\ref{theo:a-restate} (Hoeffding-type inequality for self-normalized means~\cite{GarMou_08}) with $\gamma=1$, $X_m=S_{i,j}(m-1)$, $\beta=2\sqrt{U_{\mathrm{S}}^2\log \tau}$, $B=U_{\mathrm{S}}$, and $N=\tau$, we have
    \begin{align}\label{equ:bound-conc-station-over}
        & \Pr \left( \sup_{1\le M\le \tau}  \frac{\sum_{m=1}^{M} \epsilon_{\mathrm{S},m}
        \bigl(S_{i,j}(m-1) - E[S_{i,j}(m-1)]\bigr)
        }{\sqrt{
        \sum_{m=1}^{M} \epsilon_{\mathrm{S},m}
        }}
        > 2\sqrt{ U_{\mathrm{S}}^2\log \tau}\right)\nonumber\\
        \le & \left(\frac{\log \tau}{\log (1+\zeta)}+1\right)\exp\left(-8\left(1-\frac{\zeta^2}{16}\right)\log \tau\right)
    \end{align}
    for all $\zeta>0$.
    Recall that in the proof of Lemma~\ref{lemma:concentration} (Section~\ref{app:proof-lemma-concentration}) we showed that the bound in Theorem~\ref{theo:a-restate} holds not only for overestimation but also for underestimation. Hence, we have
    \begin{align}\label{equ:bound-conc-station-under}
        & \Pr \left( \sup_{1\le M\le \tau}  \frac{\sum_{m=1}^{M} \epsilon_{\mathrm{S},m}
        \bigl(E[S_{i,j}(m-1)] - S_{i,j}(m-1)\bigr)
        }{\sqrt{
        \sum_{m=1}^{M} \epsilon_{\mathrm{S},m}
        }}
        > 2\sqrt{ U_{\mathrm{S}}^2\log \tau}\right)\nonumber\\
        \le & \left(\frac{\log \tau}{\log (1+\zeta)}+1\right)\exp\left(-8\left(1-\frac{\zeta^2}{16}\right)\log \tau\right)
    \end{align}
    for all $\zeta>0$. Taking the union bound over underestimation~\eqref{equ:bound-conc-station-under} and overestimation~\eqref{equ:bound-conc-station-over}, we have
    \begin{align*}
        & \Pr \left( \sup_{1\le M\le \tau}  \frac{ \left| \sum_{m=1}^{M} \epsilon_{\mathrm{S},m}
        \bigl(S_{i,j}(m-1) - E[S_{i,j}(m-1)]\bigr) \right|
        }{\sqrt{
        \sum_{m=1}^{M} \epsilon_{\mathrm{S},m}
        }}
        > 2\sqrt{ U_{\mathrm{S}}^2\log \tau}\right)\nonumber\\
        \le & 2 \left(\frac{\log \tau}{\log (1+\zeta)}+1\right)\exp\left(-8\left(1-\frac{\zeta^2}{16}\right)\log \tau\right)
    \end{align*}
    for all $\zeta>0$.
    Setting $\zeta=0.3$, we have
    \begin{align}\label{equ:concen-bound-station}
        \Pr \left( \sup_{1\le M\le \tau}  \frac{ \left| \sum_{m=1}^{M} \epsilon_{\mathrm{S},m}
        \bigl(S_{i,j}(m-1) - E[S_{i,j}(m-1)]\bigr) \right|
        }{\sqrt{
        \sum_{m=1}^{M} \epsilon_{\mathrm{S},m}
        }}
        > 2\sqrt{ U_{\mathrm{S}}^2\log \tau}\right)
        \le \frac{8 \log \tau + 2}{\tau^7} \le \frac{10}{\tau^6},
    \end{align}
    where the last inequality holds since $\log \tau \le \tau - 1$ and $\tau\ge 1$.
    Substituting \eqref{equ:concen-bound-station} into \eqref{equ:event-tau-i-j} and taking the union bound over $i$, we have
    \begin{align*}
        \Pr \left( {\cal E}^{ {\mathrm c}}_{\mathrm{S},\tau,j} \right) \le \sum_{i=1}^{I} \Pr \left( {\cal E}^{ {\mathrm c}}_{\mathrm{S},\tau,i,j} \right) \le \frac{10 I}{\tau^6}.
    \end{align*}
     
\endproof

\subsection{Proof of Lemma~\ref{lemma:sum-error-queue}}
\label{app:proof-lemma-sum-error-queue}

\proof

Let
\begin{align*}
    e_{j}\coloneqq \sum_{\tau=0}^{t-1} \mathbb{1}_{{\cal E}^{\mathrm{c}}_{\mathrm{S},\tau,j}}.
\end{align*}
Let
\begin{align*}
    h\coloneqq \left\lceil\frac{6J U_{\mathrm{S}} e_j }{\delta}  \right\rceil.
\end{align*}
Fix any $\tau\in[0, t-1]$. 
We consider two cases, $\tau\le h -1$ and $\tau\ge h$.
If $\tau\le h -1$, then by Lemma~\ref{lemma:bounds-on-diff-queue-lens} and the fact that $Q_i(0)=0$, we have
\begin{align*}
    \sum_i Q_i(\tau) \le I U_{\mathrm{A}} \tau \le I U_{\mathrm{A}} (h - 1).
\end{align*}
If $\tau\ge h$, then we have
\begin{align*}
    \sum_i Q_i(\tau) = & \frac{1}{h} \sum_{\tau'=\tau - h +1}^{\tau} \sum_i Q_i(\tau)\nonumber\\
    \le & \frac{1}{h} \sum_{\tau'=\tau - h +1}^{\tau} \left( \sum_i Q_i(\tau') + I U_{\mathrm{A}} (h-1)\right)\nonumber\\
    = & \frac{1}{h} \sum_{\tau'=\tau - h +1}^{\tau} \sum_i Q_i(\tau') + I U_{\mathrm{A}} (h-1)\nonumber\\
    \le &\frac{1}{h} \sum_{\tau'=0}^{t-1} \sum_i Q_i(\tau') + I U_{\mathrm{A}} (h-1),
\end{align*}
where the first inequality is by Lemma~\ref{lemma:bounds-on-diff-queue-lens} and the last inequality holds since $\tau\le t-1$ and $\tau\ge h$. Combining these two cases, we have
\begin{align*}
    \sum_i Q_i(\tau) \le \frac{1}{h} \sum_{\tau'=0}^{t-1} \sum_i Q_i(\tau') + I U_{\mathrm{A}} (h-1)
\end{align*}
for any $\tau\in[0,t-1]$.
Hence, we have
\begin{align*}
    \sum_{\tau=0}^{t - 1} \sum_i
    \mathbb{1}_{{\cal E}^{\mathrm{c}}_{\mathrm{S},\tau,j}}
    Q_i(\tau) 
    \le &\sum_{\tau=0}^{t - 1}
    \mathbb{1}_{{\cal E}^{\mathrm{c}}_{\mathrm{S},\tau,j}}
    \left(\frac{1}{h} \sum_{\tau'=0}^{t-1} \sum_i Q_i(\tau') + I U_{\mathrm{A}} (h-1)\right)\nonumber\\
    = & \frac{e_j}{h} \sum_{\tau'=0}^{t-1} \sum_i Q_i(\tau') + I U_{\mathrm{A}} (h-1)e_j\nonumber\\
    \le &\frac{\delta}{6 J U_{\mathrm{S}} } \sum_{\tau=0}^{t-1} \sum_i Q_i(\tau) + \frac{6IJU_{\mathrm{S}}U_{\mathrm{A}}}{\delta} e^2_{j},
\end{align*}
where the last inequality holds since $\frac{6J U_{\mathrm{S}} e_j }{\delta} \le h \le 1 + \frac{6J U_{\mathrm{S}} e_j }{\delta}$.
Therefore, we have
\begin{align}\label{equ:sum-tau-queue-indicator}
    E\left[\sum_{\tau=0}^{t - 1} \sum_i
    \mathbb{1}_{{\cal E}^{\mathrm{c}}_{\mathrm{S},\tau,j}}
    Q_i(\tau) \right] 
    \le \frac{\delta}{6 J U_{\mathrm{S}} } E\left[\sum_{\tau=0}^{t-1} \sum_i Q_i(\tau)\right] + \frac{6IJU_{\mathrm{S}}U_{\mathrm{A}}}{\delta} E \left[e^2_{j}\right].
\end{align}
Next we need to bound the term $E[e_j^2]$ in \eqref{equ:sum-tau-queue-indicator}.
By the definition of $e_j$, we have
\begin{align}\label{equ:e-square}
    E[e_j^2] = & E\left[ \left( \sum_{\tau=0}^{t-1} \mathbb{1}_{{\cal E}^{\mathrm{c}}_{\mathrm{S},\tau,j}} \right)^2 \right]\nonumber\\
    = & E\left[\sum_{\tau=0}^{t-1} \mathbb{1}_{{\cal E}^{\mathrm{c}}_{\mathrm{S},\tau,j}} 
    + 2\sum_{\tau=0}^{t-1}\sum_{\tau'=\tau + 1}^{t-1} \mathbb{1}_{{\cal E}^{\mathrm{c}}_{\mathrm{S},\tau,j}} \mathbb{1}_{{\cal E}^{\mathrm{c}}_{\mathrm{S},\tau',j}}\right]\nonumber\\
    = & E\left[\sum_{\tau=0}^{t-1} \mathbb{1}_{{\cal E}^{\mathrm{c}}_{\mathrm{S},\tau,j}} 
    + 2\sum_{\tau'=1}^{t-1}\sum_{\tau=0}^{\tau'-1} \mathbb{1}_{{\cal E}^{\mathrm{c}}_{\mathrm{S},\tau,j}} \mathbb{1}_{{\cal E}^{\mathrm{c}}_{\mathrm{S},\tau',j}}\right]\nonumber\\
    \le & E\left[\sum_{\tau=0}^{t-1} \mathbb{1}_{{\cal E}^{\mathrm{c}}_{\mathrm{S},\tau,j}} 
    + 2\sum_{\tau'=1}^{t-1} (\tau' - 1)  \mathbb{1}_{{\cal E}^{\mathrm{c}}_{\mathrm{S},\tau',j}}\right]\nonumber\\
    = & \sum_{\tau=1}^{t-1} \Pr \left( {\cal E}^{\mathrm{c}}_{\mathrm{S},\tau,j} \right)
    + 2 \sum_{\tau=1}^{t-1} (\tau - 1) \Pr \left({\cal E}^{\mathrm{c}}_{\mathrm{S},\tau,j}\right)\nonumber\\
    = & \sum_{\tau=1}^{t-1} (2\tau - 1) \Pr \left({\cal E}^{\mathrm{c}}_{\mathrm{S},\tau,j}\right).
\end{align}
Substituting \eqref{equ:e-square} into \eqref{equ:sum-tau-queue-indicator}, we have
\begin{align*}
    E\left[\sum_{\tau=0}^{t - 1} \sum_i
    \mathbb{1}_{{\cal E}^{\mathrm{c}}_{\mathrm{S},\tau,j}}
    Q_i(\tau) \right] 
    \le \frac{\delta}{6 J U_{\mathrm{S}} } E\left[\sum_{\tau=0}^{t-1} \sum_i Q_i(\tau)\right] + \frac{6IJU_{\mathrm{S}}U_{\mathrm{A}}}{\delta} 
    \sum_{\tau=1}^{t-1} (2\tau - 1) \Pr \left({\cal E}^{\mathrm{c}}_{\mathrm{S},\tau,j}\right).
\end{align*}
 
\endproof

\subsection{Proof of Lemma~\ref{lemma:decouple-queue-len-ucb}}
\label{app:proof-lemma-decouple-queue-len-ucb}

\proof

The first step of proving this lemma is to decouple the queue length and the UCB bonus, which is the same as the proof of Lemma~\ref{lemma:sum-error-queue} in Section~\ref{app:proof-lemma-sum-error-queue}. We present the proof here for completeness.
Let
\begin{align*}
    e_{i,j}\coloneqq \sum_{\tau=0}^{\sum_{k=0}^{K-1} w_k - 1}
        \tilde{b}_{i, j}(f_j(\tau))
        \eta_j(f_j( \tau))
        \mathbb{1}_{i,j}(\tau)\mathbb{1}_{I_j(\tau)=i}
        \mathbb{1}_{\hat{N}_{i, j}(f_j(\tau)) < \frac{576 J^2 U^2_{\mathrm{S}} \log f_j(\tau)}{\delta^2}}.
\end{align*}
Let
\begin{align*}
    h\coloneqq \left\lceil\frac{6J U_{\mathrm{S}} e_{i,j} }{\delta}  \right\rceil.
\end{align*}
Fix any $\tau\in[0, \sum_{k=0}^{K-1} w_k-1]$. 
We consider two cases, $\tau\le h -1$ and $\tau\ge h$.
If $\tau\le h -1$, then by Lemma~\ref{lemma:bounds-on-diff-queue-lens} and the fact that $Q_i(0)=0$, we have
\begin{align*}
    Q_i(\tau) \le U_{\mathrm{A}} \tau \le U_{\mathrm{A}} (h - 1).
\end{align*}
If $\tau\ge h$, then we have
\begin{align*}
      Q_i(\tau) = & \frac{1}{h} \sum_{\tau'=\tau - h +1}^{\tau}   Q_i(\tau)\nonumber\\
    \le & \frac{1}{h} \sum_{\tau'=\tau - h +1}^{\tau} \left(   Q_i(\tau') + U_{\mathrm{A}} (h-1)\right)\nonumber\\
    = & \frac{1}{h} \sum_{\tau'=\tau - h +1}^{\tau}   Q_i(\tau') + U_{\mathrm{A}} (h-1)\nonumber\\
    \le &\frac{1}{h} \sum_{\tau'=0}^{\sum_{k=0}^{K-1} w_k-1}   Q_i(\tau') + U_{\mathrm{A}} (h-1),
\end{align*}
where the first inequality is by Lemma~\ref{lemma:bounds-on-diff-queue-lens} and the last inequality holds since $\tau\le \sum_{k=0}^{K-1} w_k -1$ and $\tau\ge h$. Combining these two cases, we have
\begin{align*}
    Q_i(\tau) \le \frac{1}{h} \sum_{\tau'=0}^{\sum_{k=0}^{K-1} w_k-1} Q_i(\tau') + U_{\mathrm{A}} (h-1)
\end{align*}
for any $\tau\in[0,\sum_{k=0}^{K-1} w_k-1]$.
Hence, we have
\begin{align}\label{equ:sum-queue-eij}
    & \sum_{\tau=0}^{\sum_{k=0}^{K-1} w_k - 1}
    Q_{i}( \tau)
    \tilde{b}_{i, j}(f_j(\tau))
    \eta_j(f_j( \tau))
    \mathbb{1}_{i,j}(\tau)\mathbb{1}_{I_j(\tau)=i}
    \mathbb{1}_{\hat{N}_{i, j}(f_j(\tau)) < \frac{576 J^2 U^2_{\mathrm{S}} \log f_j(\tau)}{\delta^2}}\nonumber\\
    \le & \sum_{\tau=0}^{\sum_{k=0}^{K-1} w_k - 1}
    \left(\frac{1}{h} \sum_{\tau'=0}^{\sum_{k=0}^{K-1} w_k-1} Q_i(\tau') + U_{\mathrm{A}} (h-1)\right)\nonumber\\
    & \qquad \qquad \tilde{b}_{i, j}(f_j(\tau))
    \eta_j(f_j( \tau))
    \mathbb{1}_{i,j}(\tau)\mathbb{1}_{I_j(\tau)=i}
    \mathbb{1}_{\hat{N}_{i, j}(f_j(\tau)) < \frac{576 J^2 U^2_{\mathrm{S}} \log f_j(\tau)}{\delta^2}}\nonumber\\
    = & \frac{e_{i,j}}{h} \sum_{\tau'=0}^{\sum_{k=0}^{K-1} w_k-1}
    Q_i(\tau') + U_{\mathrm{A}} (h-1)e_{i,j}\nonumber\\
    \le &\frac{\delta}{6 J U_{\mathrm{S}} } \sum_{\tau=0}^{\sum_{k=0}^{K-1} w_k -1} Q_i(\tau) + \frac{6JU_{\mathrm{S}}U_{\mathrm{A}}}{\delta} e_{i,j}^2,
\end{align}
where the last inequality holds since $\frac{6J U_{\mathrm{S}} e_{i,j} }{\delta} \le h \le 1 + \frac{6J U_{\mathrm{S}} e_{i,j} }{\delta}$.
The next step is to bound $e_{i,j}^2$ in \eqref{equ:sum-queue-eij}. Let us first bound $e_{i,j}$:
\begin{align*}
    e_{i,j} = \sum_{\tau=0}^{\sum_{k=0}^{K-1} w_k - 1}
    \tilde{b}_{i, j}(f_j(\tau))
    \eta_j(f_j( \tau))
    \mathbb{1}_{i,j}(\tau)\mathbb{1}_{I_j(\tau)=i}
    \mathbb{1}_{\hat{N}_{i, j}(f_j(\tau)) < \frac{576 J^2 U^2_{\mathrm{S}} \log f_j(\tau)}{\delta^2}},
\end{align*}
where the sum is actually taken over the starting time slots of services that finish within $[0, \sum_{k=0}^{K-1} w_k - 1]$. Hence, the sum can be bounded as follows:
\begin{align*}
    e_{i,j} \le \sum_{\tau=0}^{\sum_{k=0}^{K-1} w_k - 1}
    \tilde{b}_{i, j}(\tau)
    \eta_j(\tau)
    \mathbb{1}_{\hat{i}^*_j(\tau)=i}
    \mathbb{1}_{\hat{N}_{i, j}(\tau) < \frac{576 J^2 U^2_{\mathrm{S}} \log \tau}{\delta^2}},
\end{align*}
where the summation includes only the UCB bonuses when server $j$ picks queue $i$ and server $j$ is not idling. 
If $\sum_{\tau=0}^{\sum_{k=0}^{K-1} w_k - 1} \eta_j(\tau)\mathbb{1}_{\hat{i}^*_j(\tau)=i} = 0$, then $e_{i,j}=0$. Hence we only need to consider the case where $\sum_{\tau=0}^{\sum_{k=0}^{K-1} w_k - 1} \eta_j(\tau)\mathbb{1}_{\hat{i}^*_j(\tau)=i} > 0$. Then
\begin{align*}
    e_{i,j} \le \sum_{n=1}^{N}
    \tilde{b}_{i, j}(\tau_n)
    \mathbb{1}_{\hat{N}_{i, j}(\tau_n) < \frac{576 J^2 U^2_{\mathrm{S}} \log \tau_n}{\delta^2}},
\end{align*}
where $N\coloneqq \sum_{\tau=0}^{\sum_{k=0}^{K-1} w_k - 1} \eta_j(\tau)\mathbb{1}_{\hat{i}^*_j(\tau)=i}$ and $\tau_n$ is the time slot for the $n^{\mathrm{th}}$ time such that $\eta_j(\tau)\mathbb{1}_{\hat{i}^*_j(\tau)=i} = 1$ counting from $\tau=0$.
By the update rule \eqref{equ:alg:update} of $\hat{N}_{i, j}(\tau_n)$ and the definition of $\tau_n$, we have $\hat{N}_{i,j}(\tau_n)= n-1$. Hence, we have
\begin{align}\label{equ:sum-eij-upper}
    e_{i,j} \le & \sum_{n=1}^{N}
    \tilde{b}_{i, j}(\tau_n)
    \mathbb{1}_{n < 1 + \frac{576 J^2 U^2_{\mathrm{S}} \log \tau_n}{\delta^2}}\nonumber\\
    \le & \sum_{n=1}^{\left\lceil\frac{576 J^2 U^2_{\mathrm{S}} \log \left(\sum_{k=0}^{K-1} w_k - 1\right)}{\delta^2}\right\rceil}
    \tilde{b}_{i, j}(\tau_n),
\end{align}
where the last inequality holds since $\tau_n \le \sum_{k=0}^{K-1} w_k - 1$ by the definition of $\tau_n$.
Note that by the definition of $\tilde{b}_{i, j}(\tau_n)$, for $n=1$, we have
\begin{align}\label{equ:bound-tilde-bij-1}
    \tilde{b}_{i, j}(\tau_1) \le 1,
\end{align}
and since $\hat{N}_{i,j}(\tau_n)= n-1$, for $n\ge 2$ we have
\begin{align}\label{equ:bound-tilde-bij-2}
    \tilde{b}_{i, j}(\tau_n) \le 4 U_\mathrm{S} \sqrt{ \frac{\log \tau_n}{n-1}} \le 4 U_\mathrm{S} \sqrt{ \frac{\log \left(\sum_{k=0}^{K-1} w_k - 1\right)}{n-1}},
\end{align}
where the last inequality holds since $\tau_n \le \sum_{k=0}^{K-1} w_k - 1$.
Combining \eqref{equ:sum-eij-upper}, \eqref{equ:bound-tilde-bij-1}, and \eqref{equ:bound-tilde-bij-2}, we have
\begin{align}\label{equ:bound-eij-station}
    e_{i,j} \le & \tilde{b}_{i, j}(\tau_1) +  \sum_{n=2}^{\left\lceil\frac{576 J^2 U^2_{\mathrm{S}} \log \left(\sum_{k=0}^{K-1} w_k - 1\right)}{\delta^2}\right\rceil}
    \tilde{b}_{i, j}(\tau_n)\nonumber\\
    \le & 1 + \sum_{n=2}^{\left\lceil\frac{576 J^2 U^2_{\mathrm{S}} \log \left(\sum_{k=0}^{K-1} w_k - 1\right)}{\delta^2}\right\rceil} 4 U_\mathrm{S} \sqrt{ \frac{\log \left(\sum_{k=0}^{K-1} w_k - 1\right)}{n-1}}\nonumber\\
    \le & 1 + \frac{192 J U^2_\mathrm{S} \log \left(\sum_{k=0}^{K-1} w_k - 1\right) }{\delta}\nonumber\\
    \le & \frac{194 J U^2_\mathrm{S} \log \left(\sum_{k=0}^{K-1} w_k - 1\right) }{\delta},
\end{align}
where the last inequality holds since $\frac{192 J U^2_\mathrm{S} \log \left(\sum_{k=0}^{K-1} w_k - 1\right) }{\delta} \ge 192 \log 2$ for any $K\ge 3$ because $w_k\ge 1$.
Therefore, from \eqref{equ:sum-queue-eij} and \eqref{equ:bound-eij-station}, we have
\begin{align}
    & E\left[ \sum_{\tau=0}^{\sum_{k=0}^{K-1} w_k - 1}
    Q_{i}( \tau)
    \tilde{b}_{i, j}(f_j(\tau))
    \eta_j(f_j( \tau))
    \mathbb{1}_{i,j}(\tau)\mathbb{1}_{I_j(\tau)=i}
    \mathbb{1}_{\hat{N}_{i, j}(f_j(\tau)) < \frac{576 J^2 U^2_{\mathrm{S}} \log f_j(\tau)}{\delta^2}}
    \right] \nonumber\\
    \le & \frac{\delta}{6 J U_{\mathrm{S}} } E \left[\sum_{\tau=0}^{\sum_{k=0}^{K-1} w_k -1} Q_i(\tau) \right]+ \frac{225816 J^3 U^5_{\mathrm{S}} U_{\mathrm{A}} \log^2 \left(\sum_{k=0}^{K-1} w_k - 1\right)}{\delta^3}.
\end{align}
 
\endproof

\section{Additional Details of the Simulations}
\label{app:simu}

In this section, we present more details of the simulations.

\subsection{Settings}

The arrival rates follow the Bernoulli distribution.
The service times follow the truncated heavy-tail Weibull distribution, which is defined as follows. Let $\iota$ denote the success probability. A discrete random variable $X$ is said to follow a Weibull distribution with success probability $\iota$ and parameter $\beta$ if
\begin{align*}
    \sum_{j=k}^{\infty} \Pr(X=j) = \iota^{k^{\beta}}.
\end{align*}
Let $\Pr(X=j;\iota,\beta)$ denote the probability that the random variable $X$ is equal to $j$ under the Weibull distribution with success probability $\iota$ and parameter $\beta$. Define $X_{\mathrm{T}}$ be the truncated version of $X$ such that
\begin{align*}
    \Pr(X_{\mathrm{T}}=j) = \Pr(X_{\mathrm{T}}=j;\iota,\beta) \coloneqq \frac{\Pr(X=j;\iota,\beta)}{1-\iota^{\sqrt{U_{\mathrm{S}}}}}.
\end{align*}
We call $X_{\mathrm{T}}$ follows the truncated heavy-tail Weibull distribution if $0< \beta\le 1$. 
In the simulation, we set $\beta=0.5$ and $U_{\mathrm{S}}=100$.
Let $\iota_{i,j}$ denote the success probability of the distribution of the service time for job-type $i$ at server $j$.

\textbf{For the stationary setting}, we set the arrival rates $\lambda_i=0.15$ for all $i\in\{1,2,\ldots,10\}$.
We set the success probabilities $\iota_{i,j}$ of the service times as follows:
$$\iota_{2k+1,2l+1} = 0.5, \quad \iota_{2k+1,2l+2} = 0.7, \quad \iota_{2k+2, 2l+1}=0.8, \quad \iota_{2k+2,2l+2} = 0.4,$$
for all $k,l\in\{0, 1, 2, 3, 4\}$.

\textbf{For the nonstationary setting}, we set the arrival rates $\lambda_i(t)=0.15$ for all $i\in\{1,2,\ldots,10\}$ and all $t$.
When $t<150k$, we set the success probabilities $\iota_{i,j}(t)$ of the service times as follows:
\begin{align*}
    \iota_{2k+1,2l+1}(t)=& (0.5, 0.5, \ldots, 0.5),\\
    \iota_{2k+1,2l+2}(t)=& (0,7, 0.7, \ldots, 0.7),\\
    \iota_{2k+2,2l+1}(t)=& (0.8, 0.8, \ldots, 0.8),\\
    \iota_{2k+2,2l+2}(t)=& (0.4, 0.4, \ldots, 0.4),
\end{align*}
for all $k,l\in\{0, 1, 2, 3, 4\}$.
When $t\ge 150k$, we set the success probabilities $\iota_{i,j}(t)$ of the service times as follows:
\begin{align*}
    \iota_{2k+1,2l+1}(t)=& (0.8, 0.8, \ldots, 0.8),\\
    \iota_{2k+1,2l+2}(t)=& (0,4, 0.4, \ldots, 0.4),\\
    \iota_{2k+2,2l+1}(t)=& (0.5, 0.5, \ldots, 0.5),\\
    \iota_{2k+2,2l+2}(t)=& (0.7, 0.7, \ldots, 0.7),
\end{align*}
for all $k,l\in\{0, 1, 2, 3, 4\}$.

\subsection{Parameters}
\begin{figure}[htb]
    \centering
    \subfloat[Stationary Setting.]{%
    \includegraphics[width=0.45\textwidth]{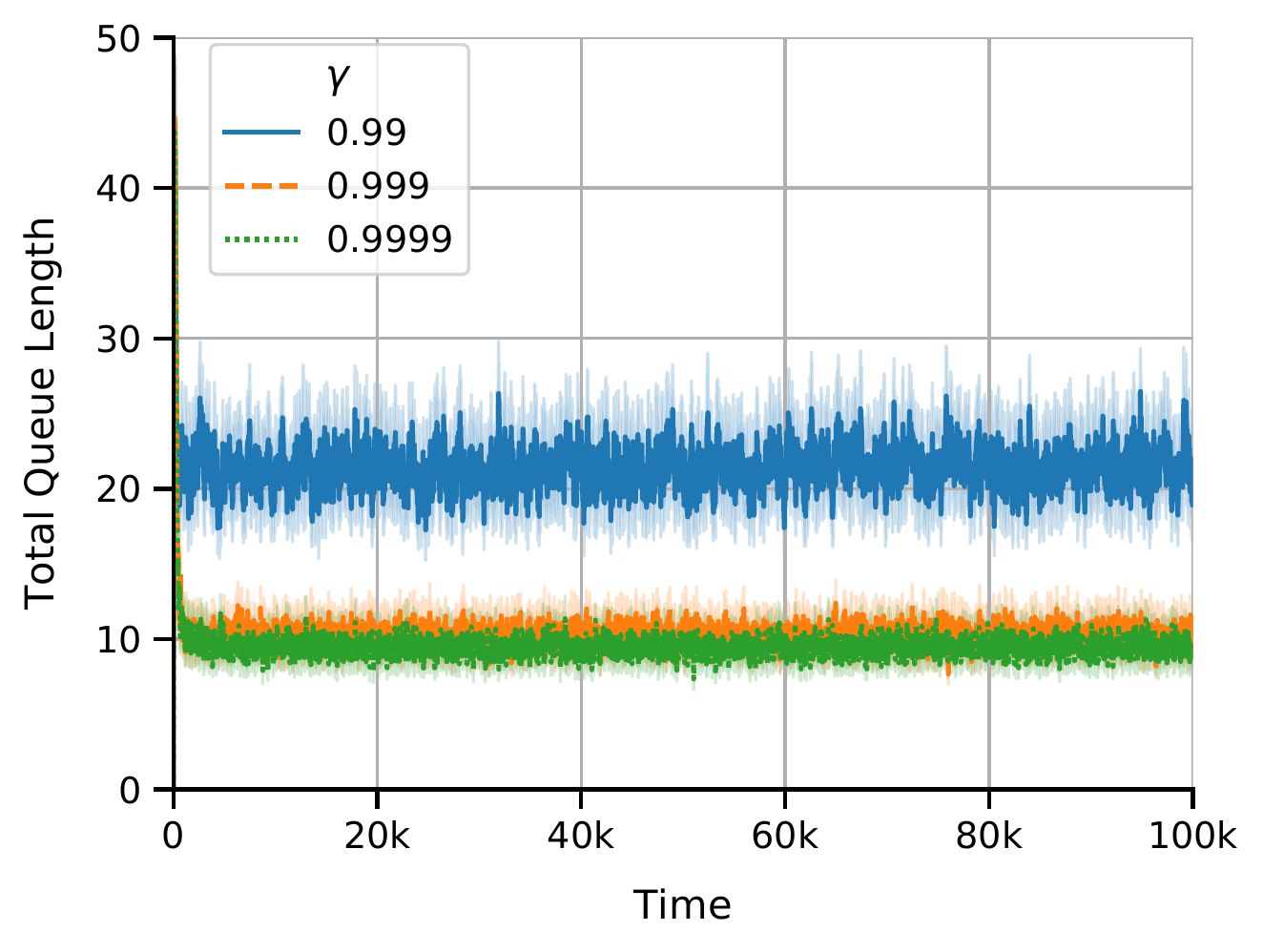}%
    }%
    \hspace{4ex}
    \subfloat[Nonstationary Setting.]{%
    \includegraphics[width=0.43\textwidth]{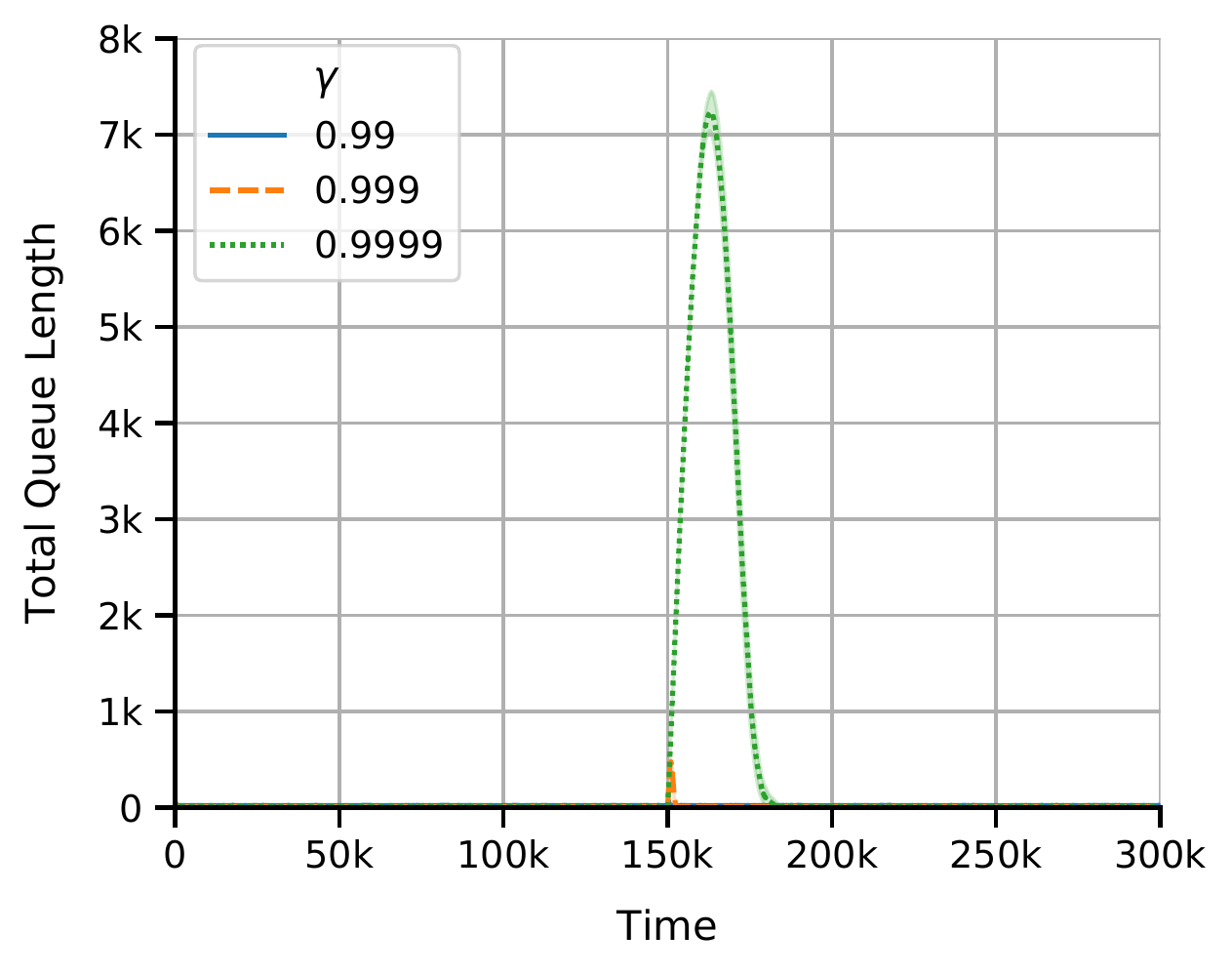}%
    }%
    \caption{Simulation Results of \emph{MaxWeigh With Discounted UCB} With Different $\gamma$.}
    \label{fig:app:simu-diff-gamma}
\end{figure}

\begin{figure}[htb]
    \centering
    \subfloat[Stationary Setting.]{%
    \includegraphics[width=0.442\textwidth]{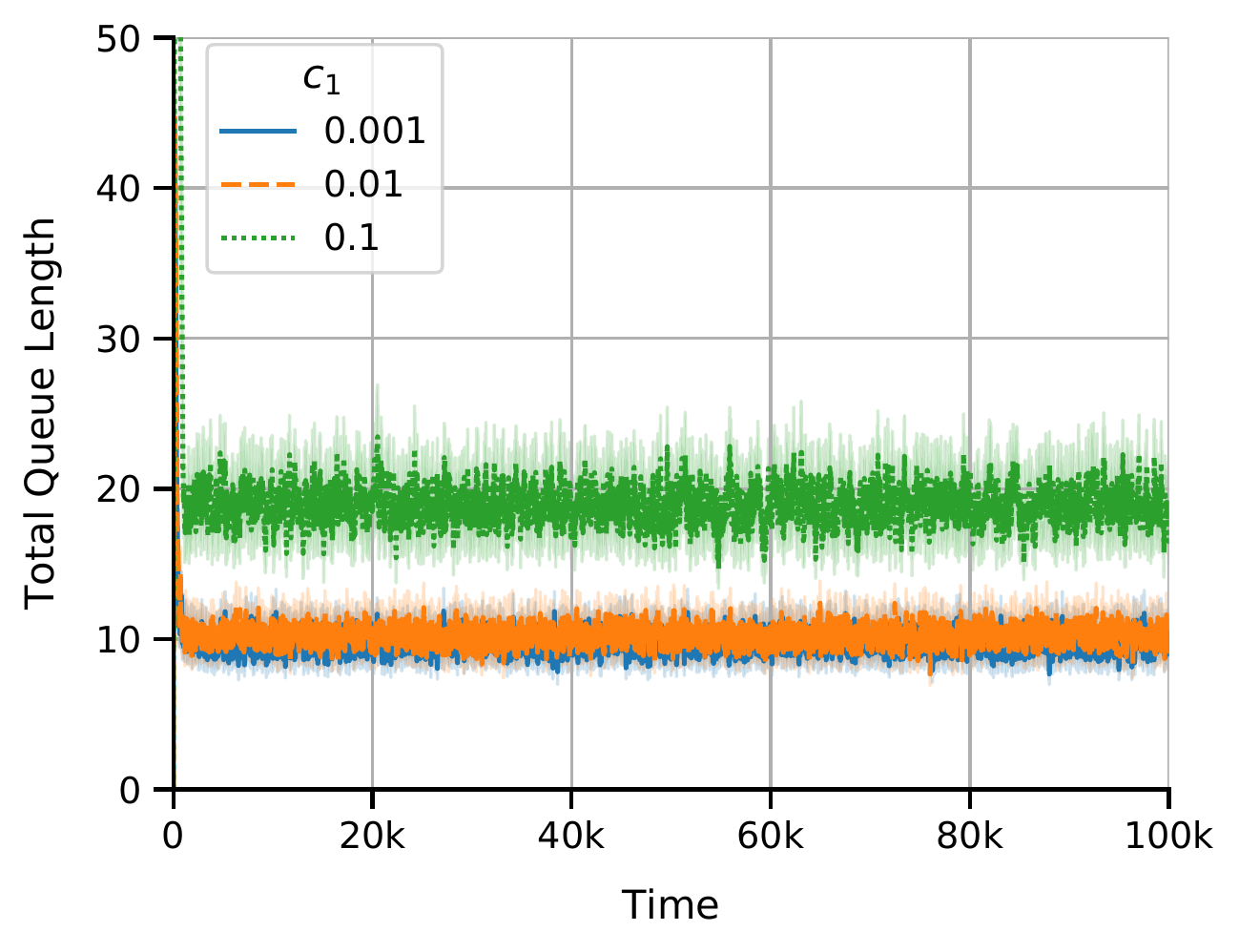}%
    }%
    \hspace{4ex}
    \subfloat[Nonstationary Setting.]{%
    \includegraphics[width=0.45\textwidth]{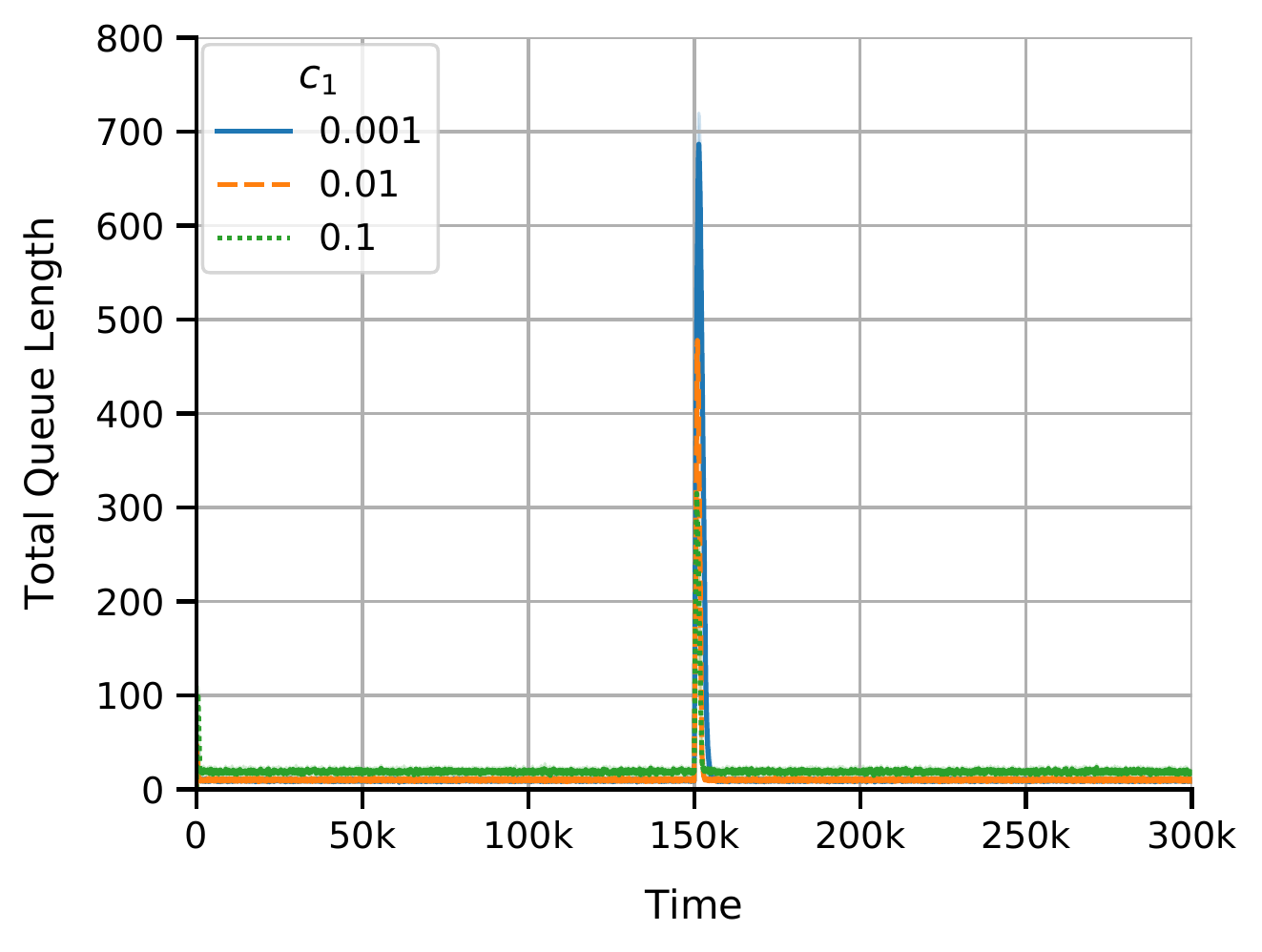}%
    }%
    \caption{Simulation Results of \emph{MaxWeigh With Discounted UCB} With Different $c_1$.}
    \label{fig:app:simu-diff-c1}
\end{figure}

\textbf{For the proposed \emph{MaxWeight with discounted UCB}}, we set $\gamma=0.999$ and $c_1=0.01$ for both stationary and nonstationary setting. Simulation results of \emph{MaxWeight with discounted UCB} with different $\gamma$ are shown in Figure~\ref{fig:app:simu-diff-gamma}. As shown in the figures, the proposed algorithm works well under different values of $\gamma$. 
Note that in the nonstationary setting, although the performance of $\gamma=0.9999$ is worse than that of $\gamma=0.99$ and $\gamma=0.999$, it is still significantly better than the other algorithms whose queue length can be over 20k as shown in Figure~\ref{fig:simu-nonstationary-10-users}.
Simulation results of \emph{MaxWeight with discounted UCB} with different $c_1$ are shown in Figure~\ref{fig:app:simu-diff-c1}. As shown in the figures, the proposed algorithm also works well under different values of $c_1$. Therefore, the algorithm is robust to the value of $\gamma$ and $c_1$. 

\textbf{For \emph{MaxWeight with UCB}}, we use the same $c_1$ as \emph{MaxWeight with discounted UCB}. 

\begin{figure}[htb]
    \centering
    \subfloat[Stationary Setting.]{%
    \includegraphics[width=0.45\textwidth]{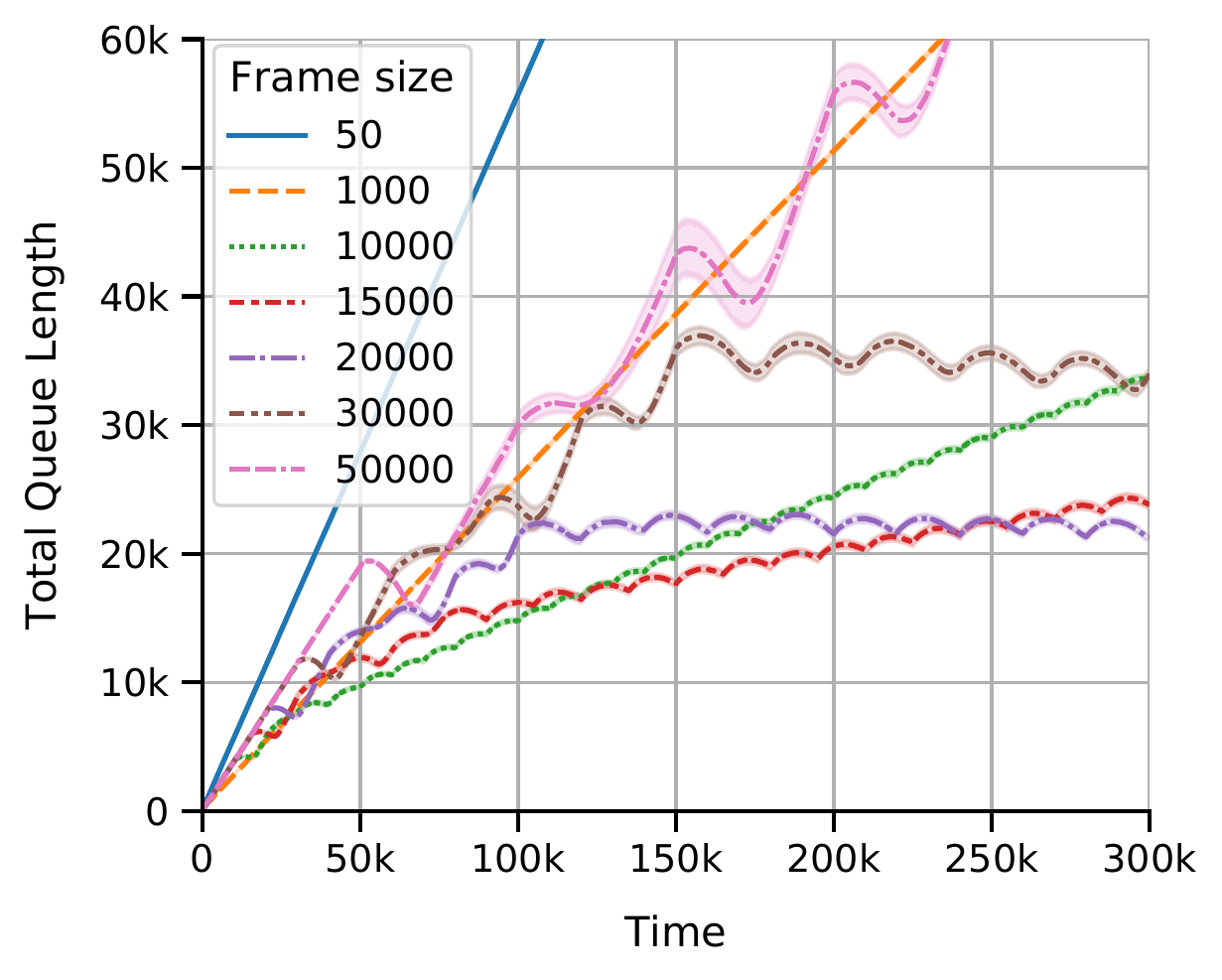}%
    \label{fig:app:simu-diff-framesize-stationary}%
    }%
    \hspace{4ex}
    \subfloat[Nonstationary Setting.]{%
    \includegraphics[width=0.45\textwidth]{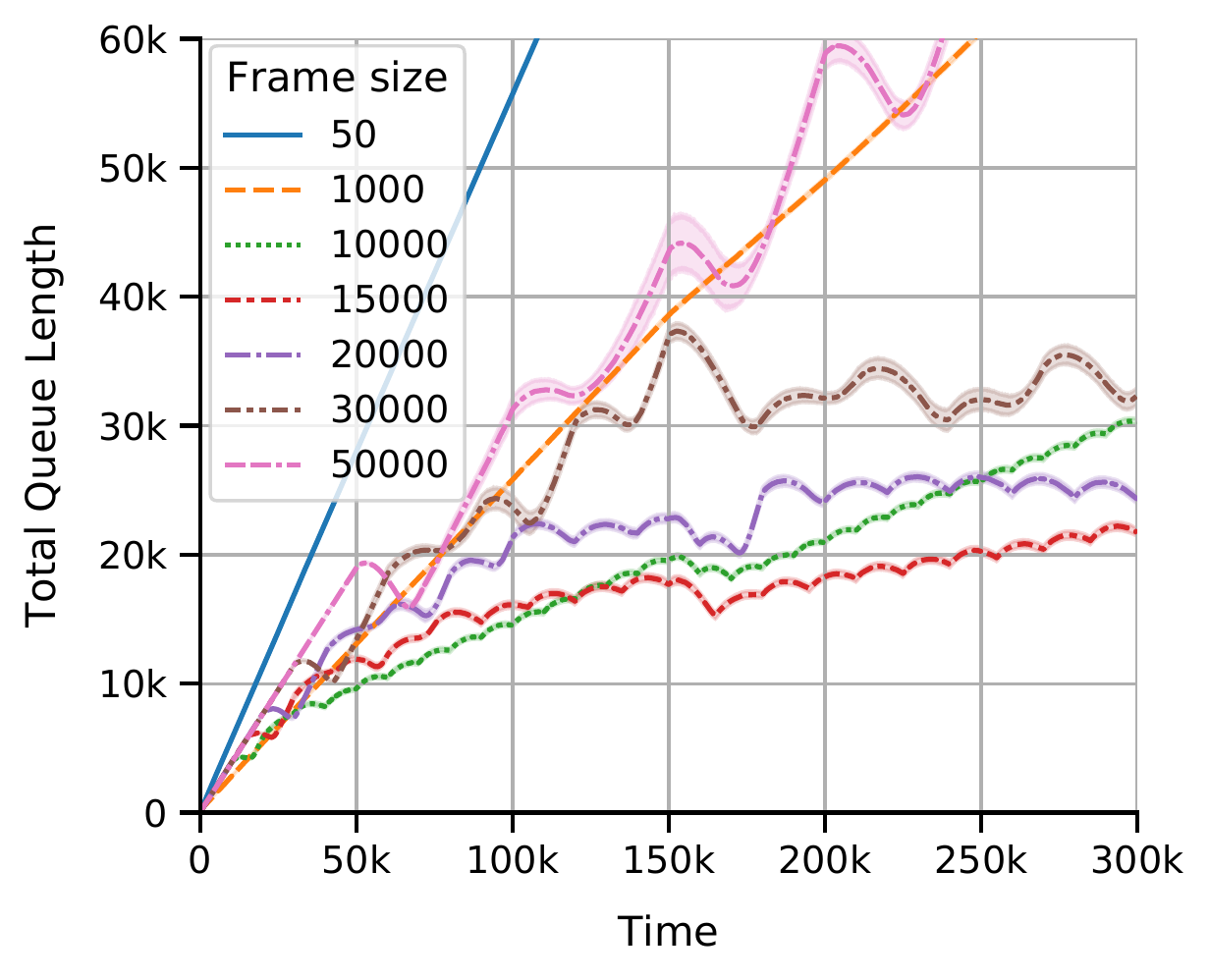}%
    \label{fig:app:simu-diff-framesize-nonstationary}%
    }%
    \caption{Simulation Results of \emph{frame-based MaxWeight} With Different Frame Sizes.}
    \label{fig:app:simu-diff-framesize}
\end{figure}

\textbf{For \emph{frame-based MaxWeight}}, we try different frame sizes and then choose the one that has the best performance. As shown in Figure~\ref{fig:app:simu-diff-framesize-stationary}, the best frame size in the stationary setting is 20000. As shown in Figure~\ref{fig:app:simu-diff-framesize-nonstationary}, the best frame size in the nonstationary setting is 20000.

\begin{figure}[htb]
    \centering
    \subfloat[Stationary Setting.]{%
    \includegraphics[width=0.455\textwidth]{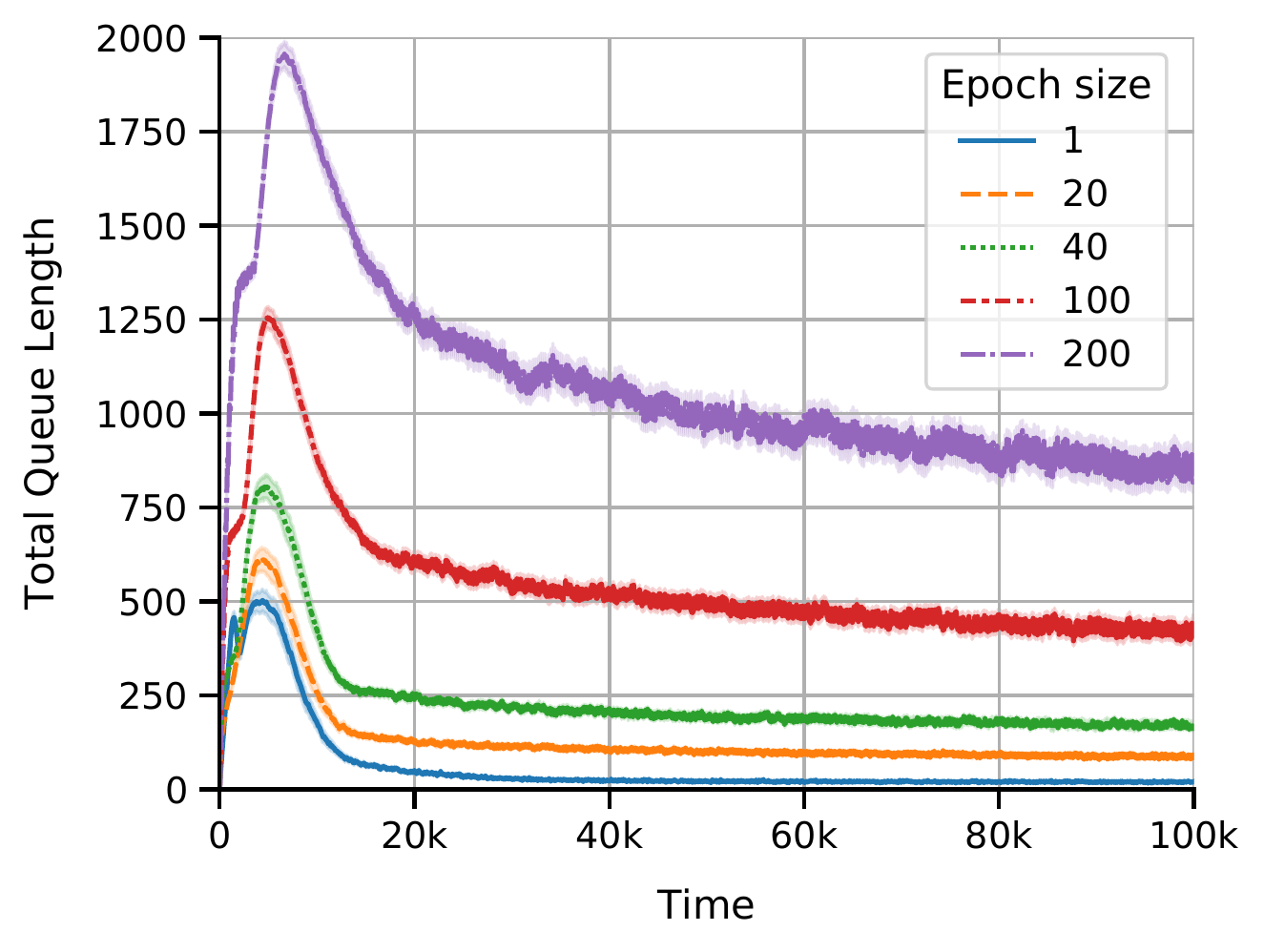}%
    \label{fig:app:simu-diff-epochsize-stationary}%
    }%
    \hspace{4ex}
    \subfloat[Nonstationary Setting.]{%
    \includegraphics[width=0.435\textwidth]{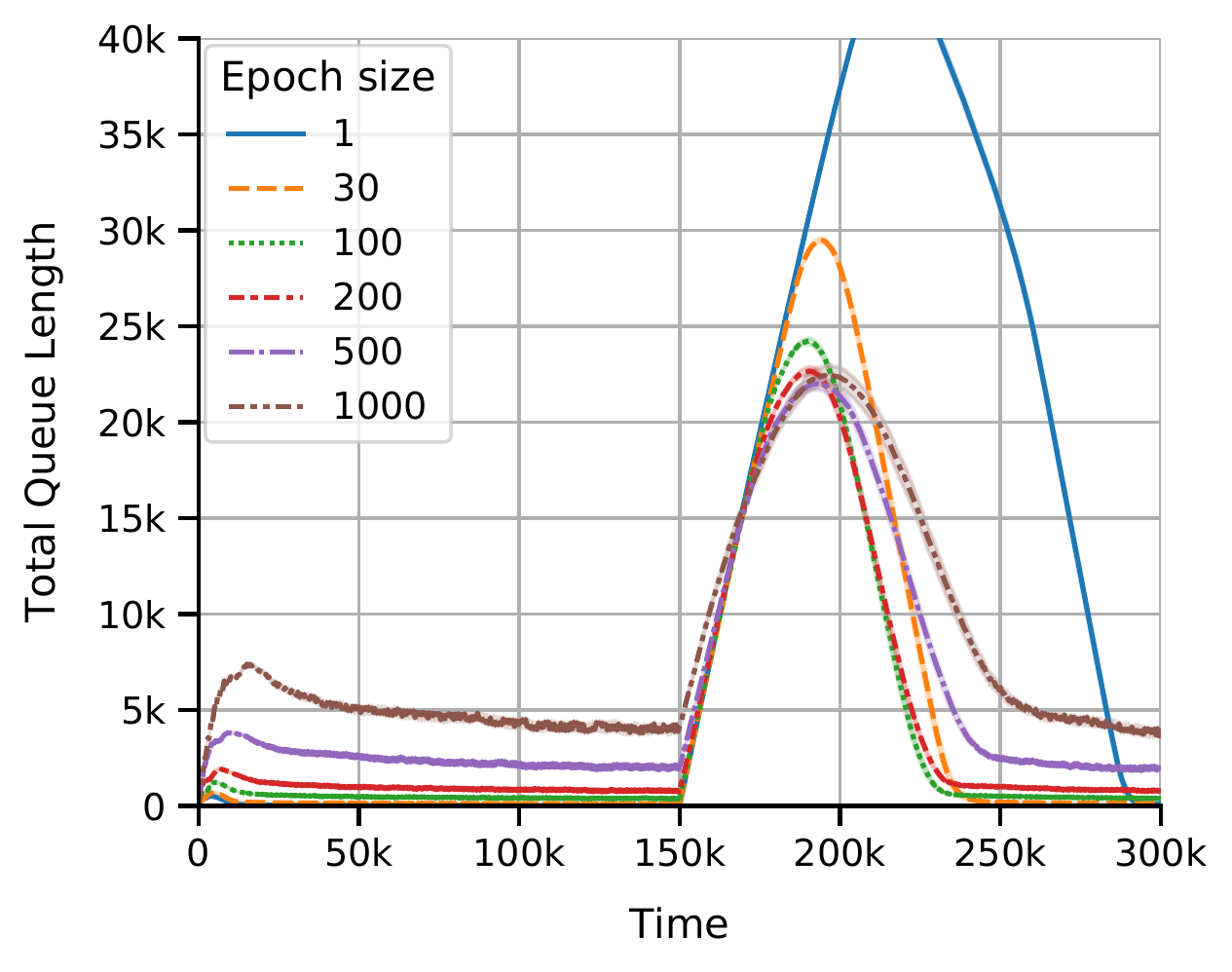}%
    \label{fig:app:simu-diff-epochsize-nonstationary}%
    }%
    \caption{Simulation Results of \emph{DAM.UCB} With Different Epoch Sizes.}
    \label{fig:app:simu-diff-epochsize}
\end{figure}

\textbf{For \emph{DAM.UCB}}, we try different epoch sizes and then choose the one that has the best performance. As shown in Figure~\ref{fig:app:simu-diff-epochsize-stationary}, the best epoch size in the stationary setting is 1. As shown in Figure~\ref{fig:app:simu-diff-epochsize-nonstationary}, the best epoch size in the nonstationary setting is 200.

The total queue length $\sum_i Q_i(t)$ of all the curves in this paper is averaged over 100 runs. The shaded area in all the figures is the $95\%$ confidence interval. For all the curves, we plot one point every 10 time slots.

\vfill

\end{document}